\pdfoutput=1
\documentclass[10pt,twocolumn,letterpaper]{article}
\usepackage{iccv}
\usepackage{times}
\usepackage{epsfig}
\usepackage{graphicx}
\usepackage{amsmath}
\usepackage{amssymb}
\usepackage{tabularx}
\usepackage[dvipsnames]{xcolor}

\usepackage[pagebackref=true,breaklinks=true,letterpaper=true,colorlinks,bookmarks=false]{hyperref}
\iccvfinalcopy 

\def\iccvPaperID{***} 

\begin{document}

\title{Egocentric Pose Estimation from Human Vision Span}

\author{Hao Jiang and Vamsi Krishna Ithapu \\
Facebook Reality Labs Research, Redmond WA, USA
}

\maketitle

\begin{abstract}

Estimating camera wearer's body pose from an egocentric view (egopose) is a vital task in 
augmented and virtual reality. Existing approaches either use a narrow field of view front 
facing camera that barely captures the wearer, or an extruded head-mounted top-down camera for 
maximal wearer visibility. In this paper, we tackle the egopose estimation from a more natural 
human vision span, where camera wearer can be seen in the peripheral view and depending on the head pose the 
wearer may become invisible or has a limited partial view. This is a realistic 
visual field for user-centric wearable devices like glasses which have front facing wide angle 
cameras. Existing solutions are not appropriate for this setting, and so, we propose a novel deep learning 
system taking advantage of both the dynamic features from camera SLAM and the body shape imagery. 
We compute 3D head pose, 3D body pose, the figure/ground separation, all at the same time while 
explicitly enforcing a certain geometric consistency across pose attributes. We further show that this system can be 
trained robustly with lots of existing mocap data so we do not have to collect and annotate large new datasets. 
Lastly, our system estimates egopose in real time and on the fly while maintaining high accuracy.
\end{abstract}


\vspace{-16pt}
\section{Introduction \label{sec:intro}}
\vspace{-4pt}
Truly immersive experiences in augmented and virtual reality (AR and VR) are driven by explicit characterization of user's (i.e., the device wearer) pose. 
In particular, this user's pose needs to be estimated from the perspective of the device, which implicitly corresponds to their egocentric perspective.  
Typically referred to as the {\it egopose}, this corresponds to the 3D head and body pose of the camera wearer. 
Egopose drives the necessary inputs required for constructing naturalistic experiences in AR and VR. 
For instance, world locked egopose representations provide the necessary inputs for user's interacting with the audio and visual objects in a virtual scene. 
In particular, for conversations involving a combination of real people and virtual entities (like avatars or holograms), 
a precise characterization of egopose is necessary to enable seamless switching between multiple speakers while retaining immersion. 

\begin{figure}[!t]
	\centering
	\scalebox{1}{	
\setlength\tabcolsep{1pt}
\begin{tabularx}{\linewidth}{c X }
        \rotatebox{90}{\hspace{0pt}{\tiny }} 
	& \colorbox{CadetBlue}{\includegraphics[width=0.185\linewidth]{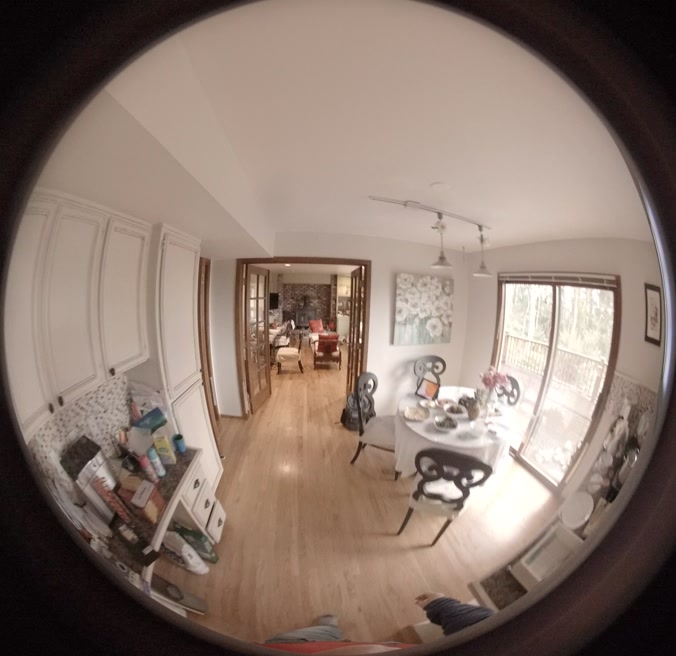}}%
\colorbox{CadetBlue}{\includegraphics[width=0.18\linewidth]{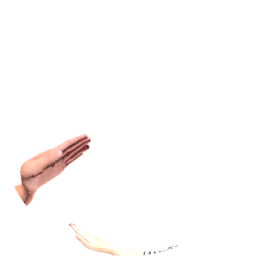}}%
\colorbox{CadetBlue}{\includegraphics[width=0.18\linewidth]{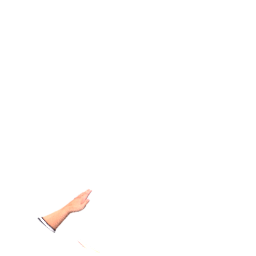}}%
\colorbox{CadetBlue}{\includegraphics[width=0.18\linewidth]{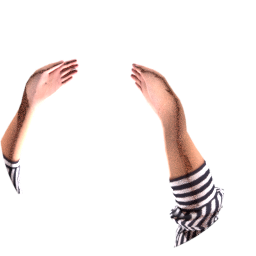}}%
\colorbox{CadetBlue}{\includegraphics[width=0.18\linewidth]{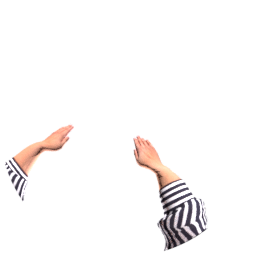}}%
\\
\rotatebox{90}{\hspace{0pt}{\tiny}}	&\includegraphics[width=0.205\linewidth]{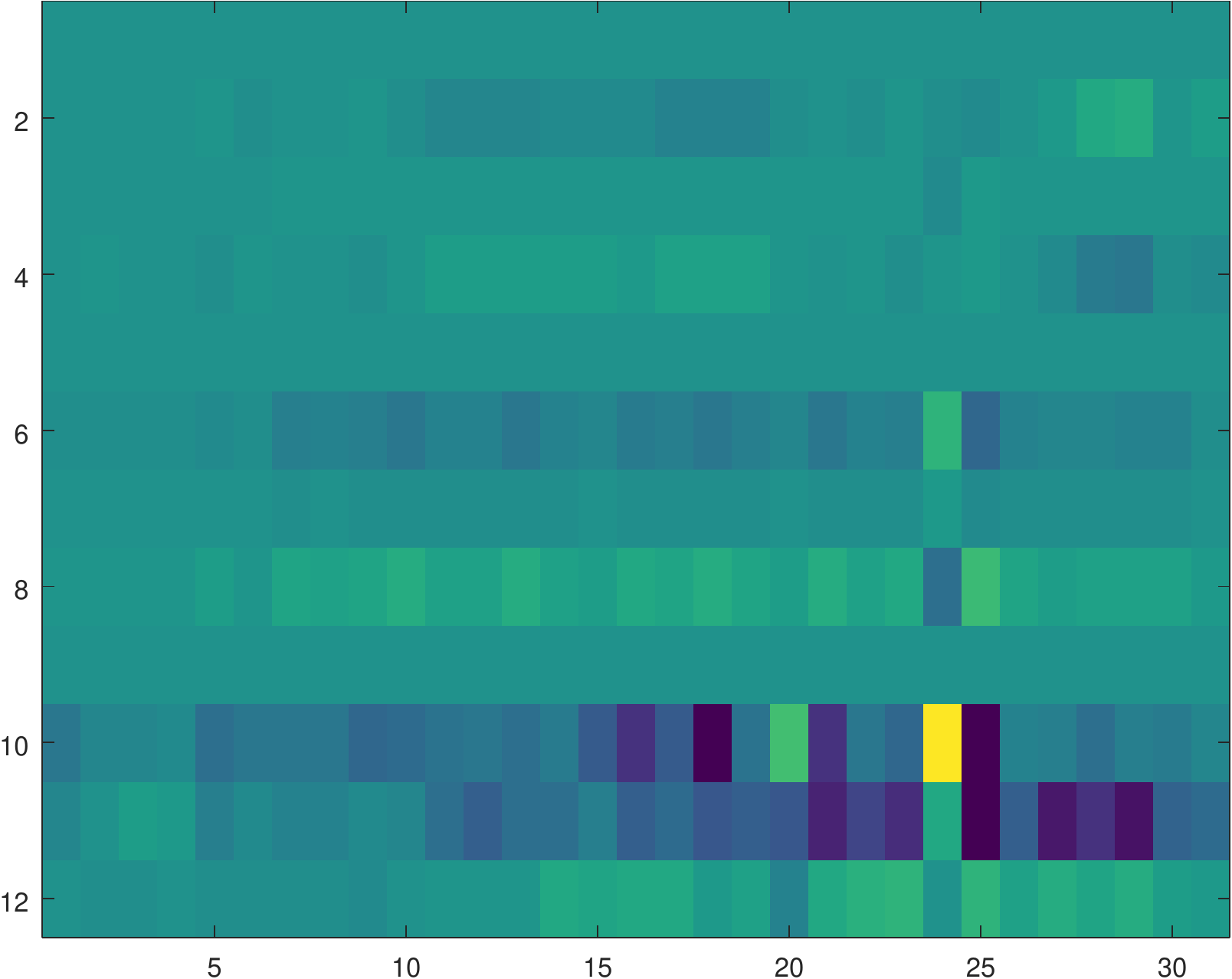}%
\includegraphics[width=0.205\linewidth]{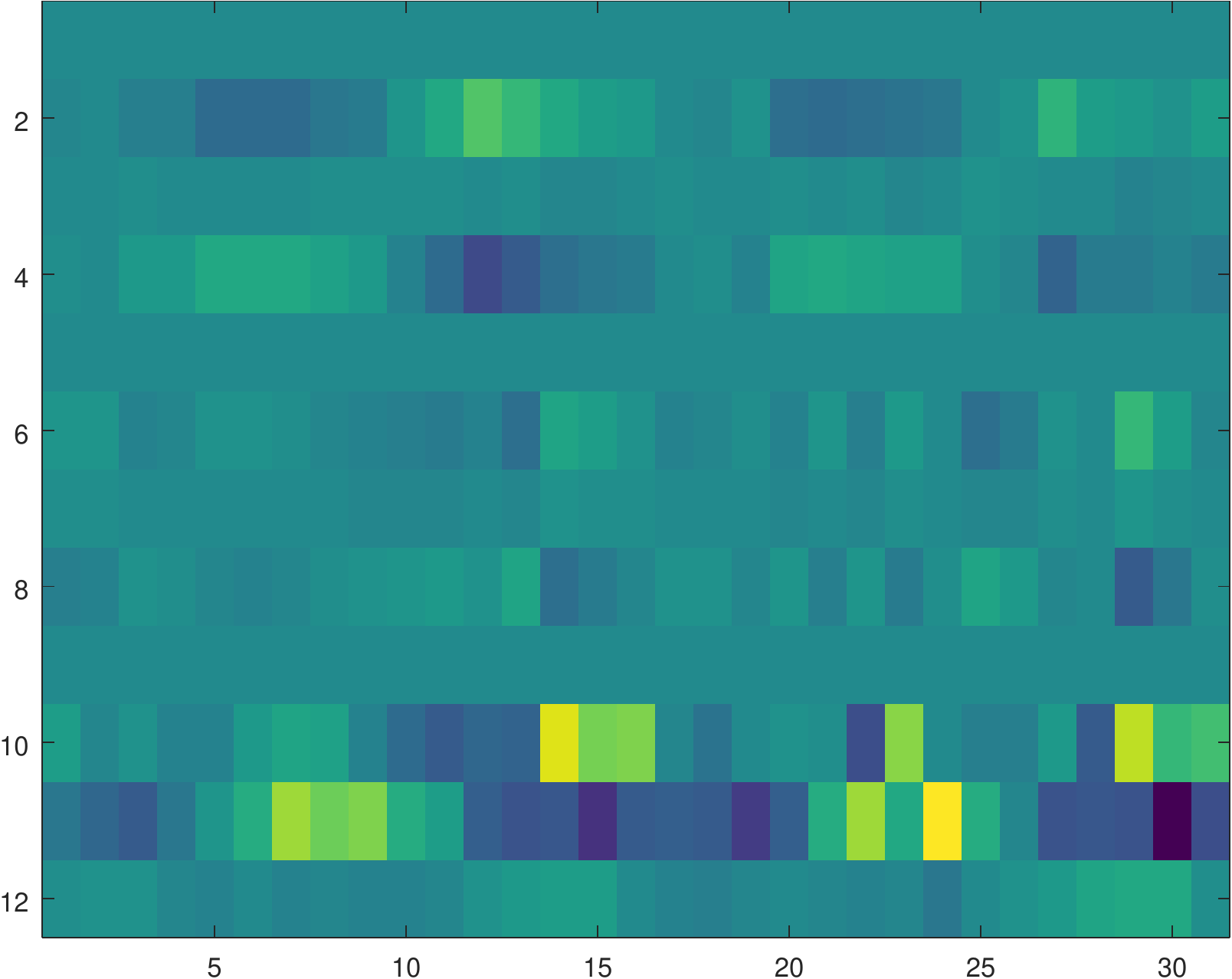}%
\includegraphics[width=0.205\linewidth]{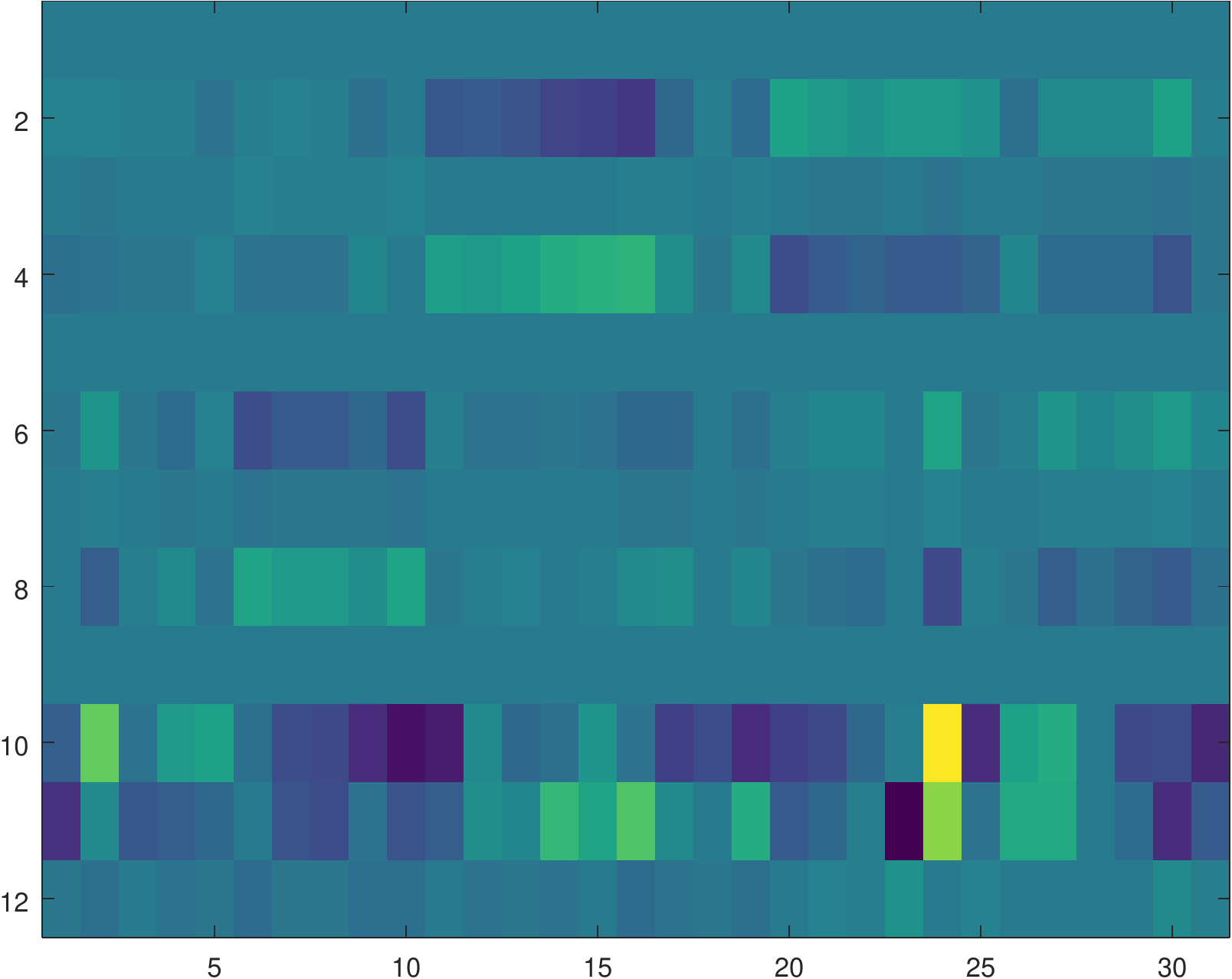}%
\includegraphics[width=0.205\linewidth]{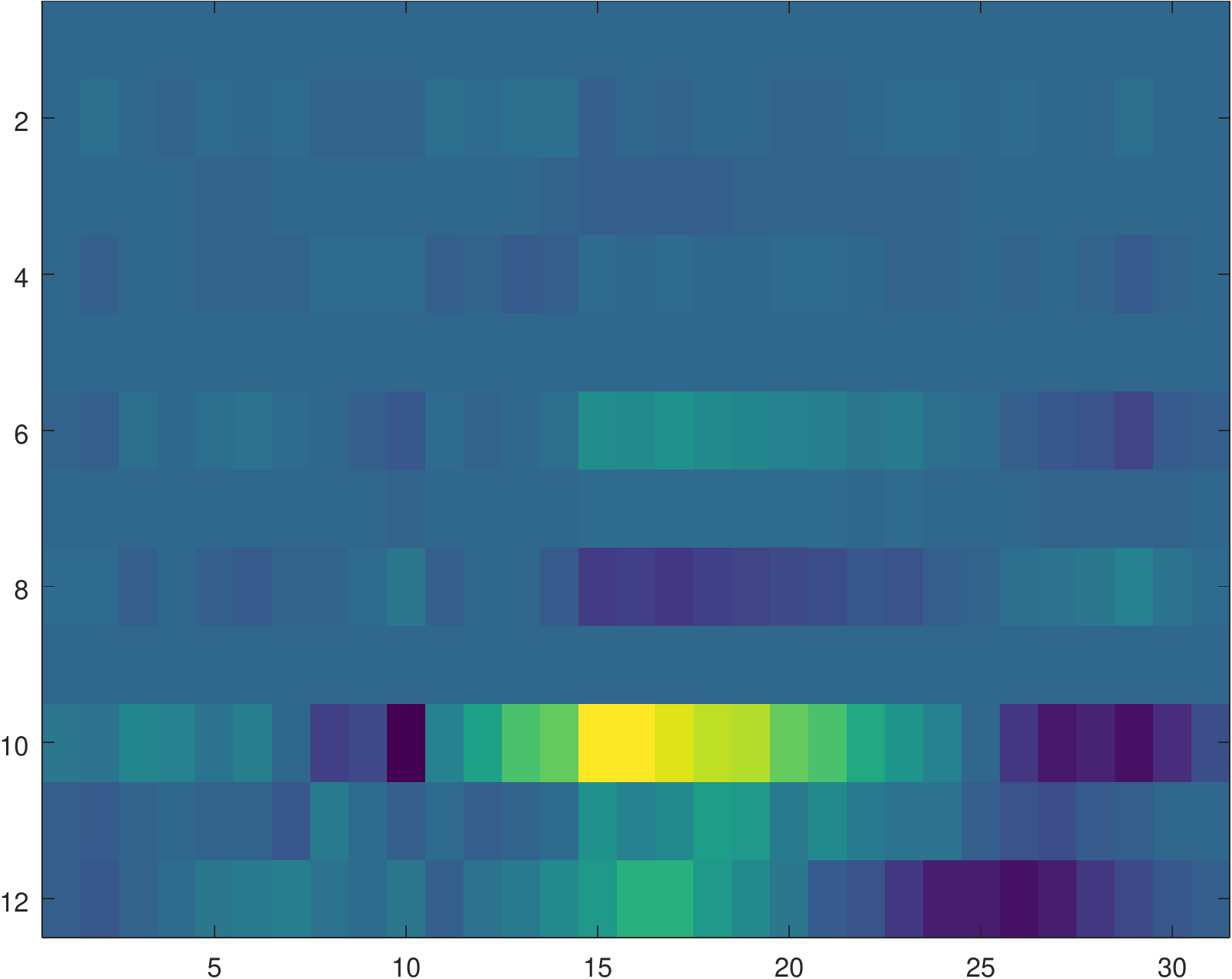}%
\includegraphics[width=0.205\linewidth]{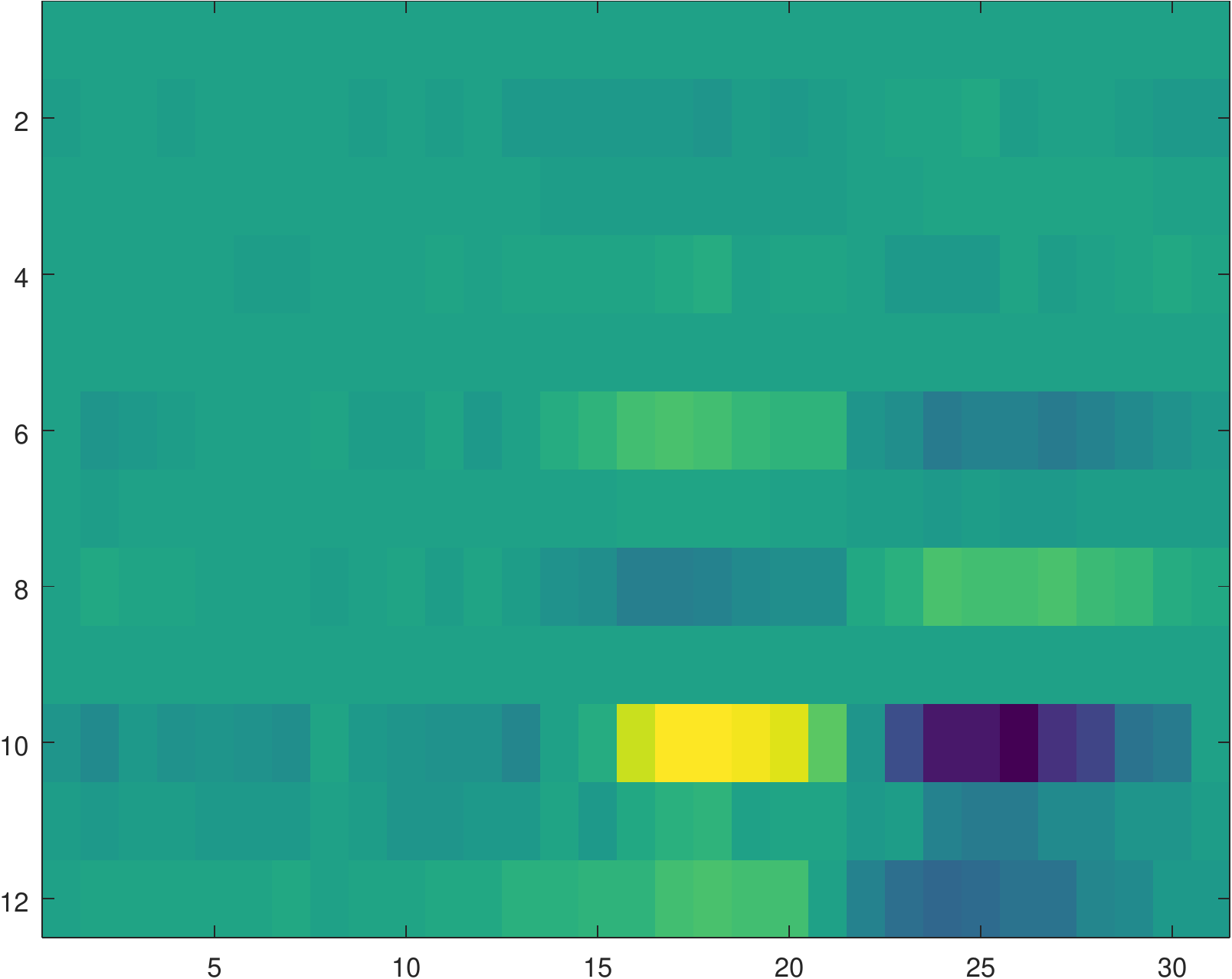}%
\\
	\rotatebox{90}{\hspace{10pt}{\scriptsize Our Results}} &
\includegraphics[width=0.2\linewidth, height=0.275\linewidth]{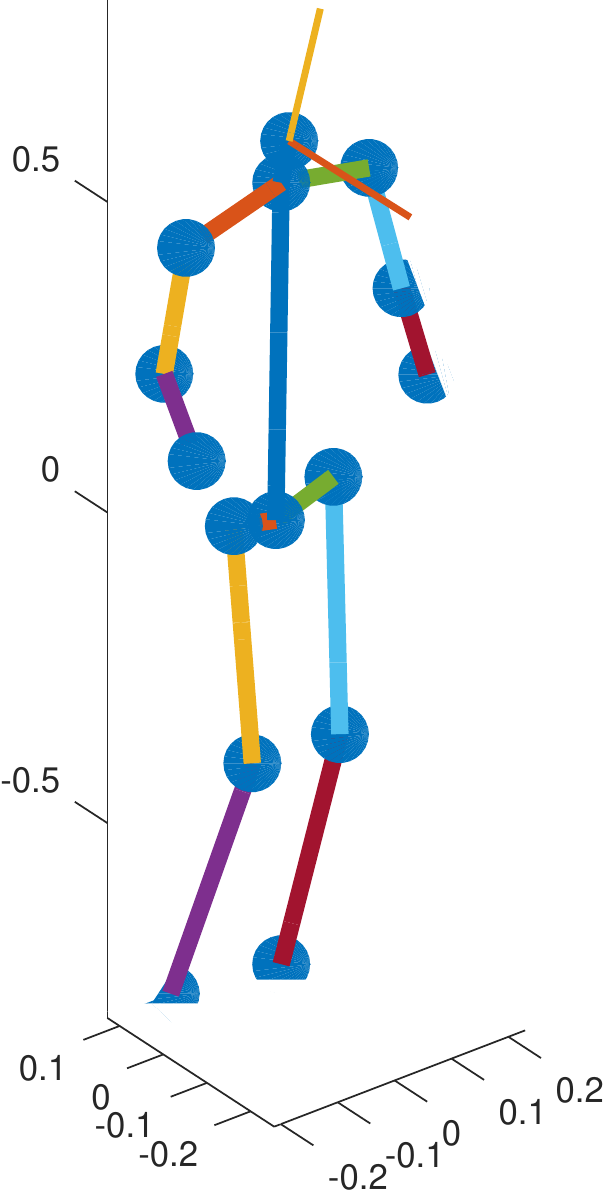}%
\includegraphics[width=0.2\linewidth, height=0.275\linewidth]{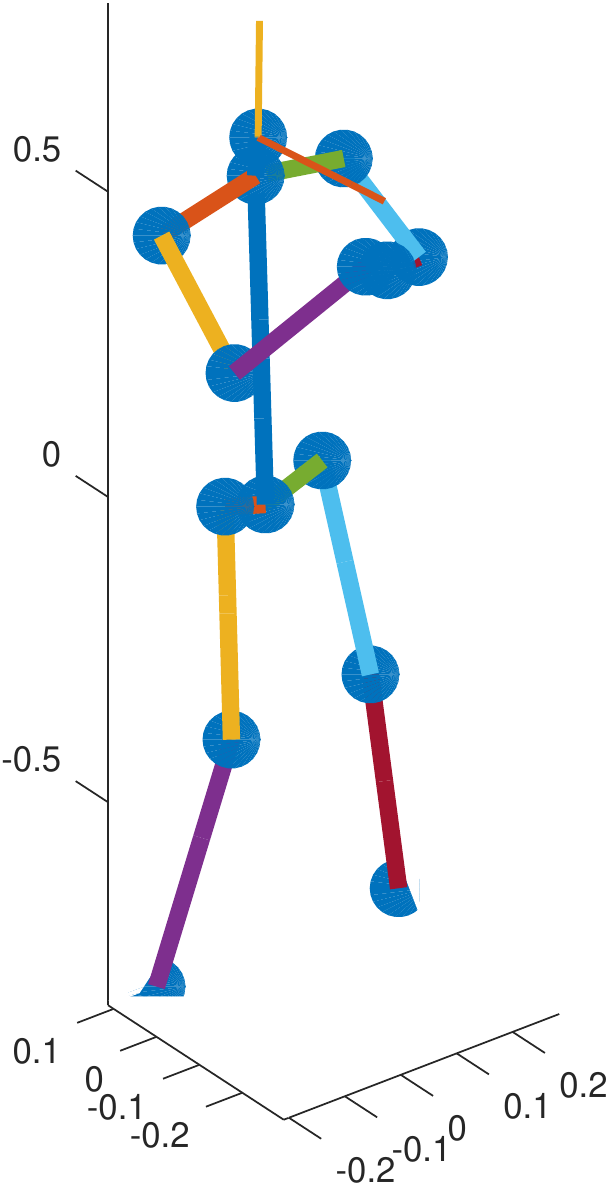}%
\includegraphics[width=0.2\linewidth, height=0.275\linewidth]{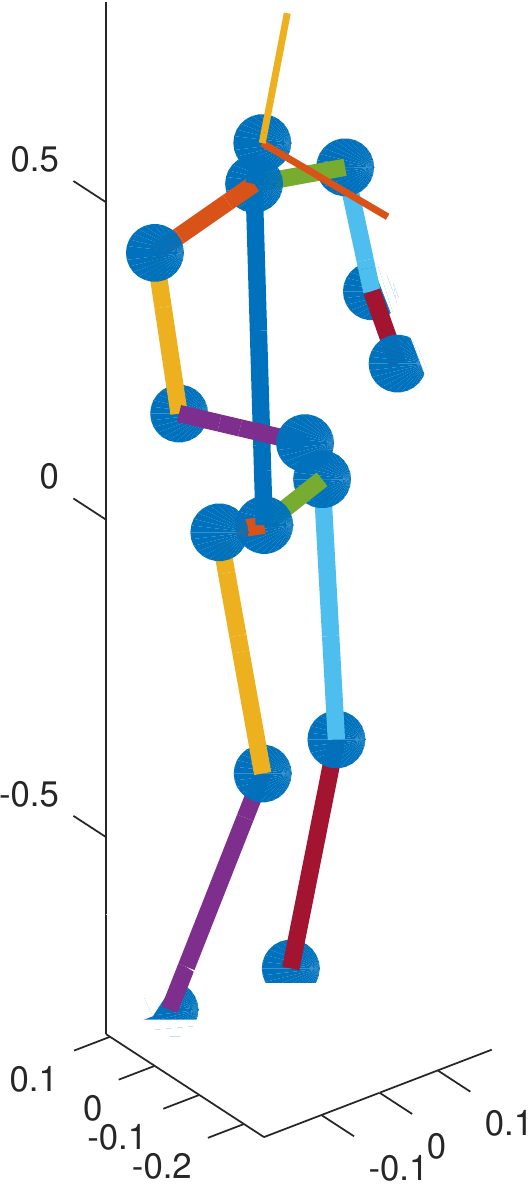}%
\includegraphics[width=0.2\linewidth, height=0.275\linewidth]{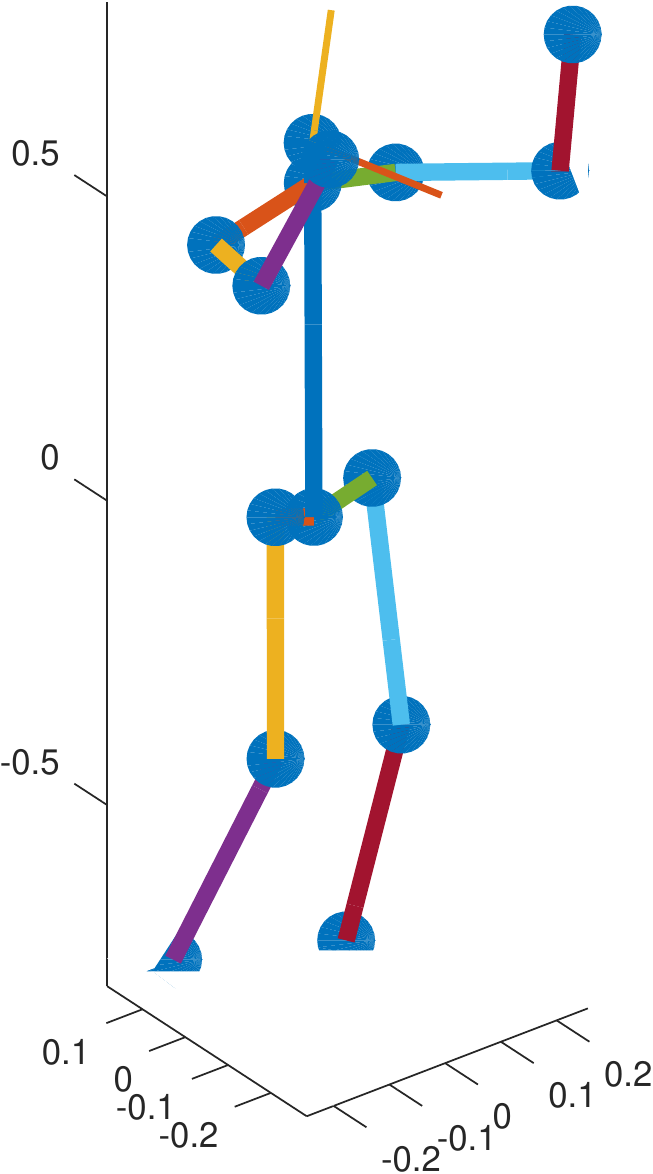}%
\includegraphics[width=0.2\linewidth, height=0.275\linewidth]{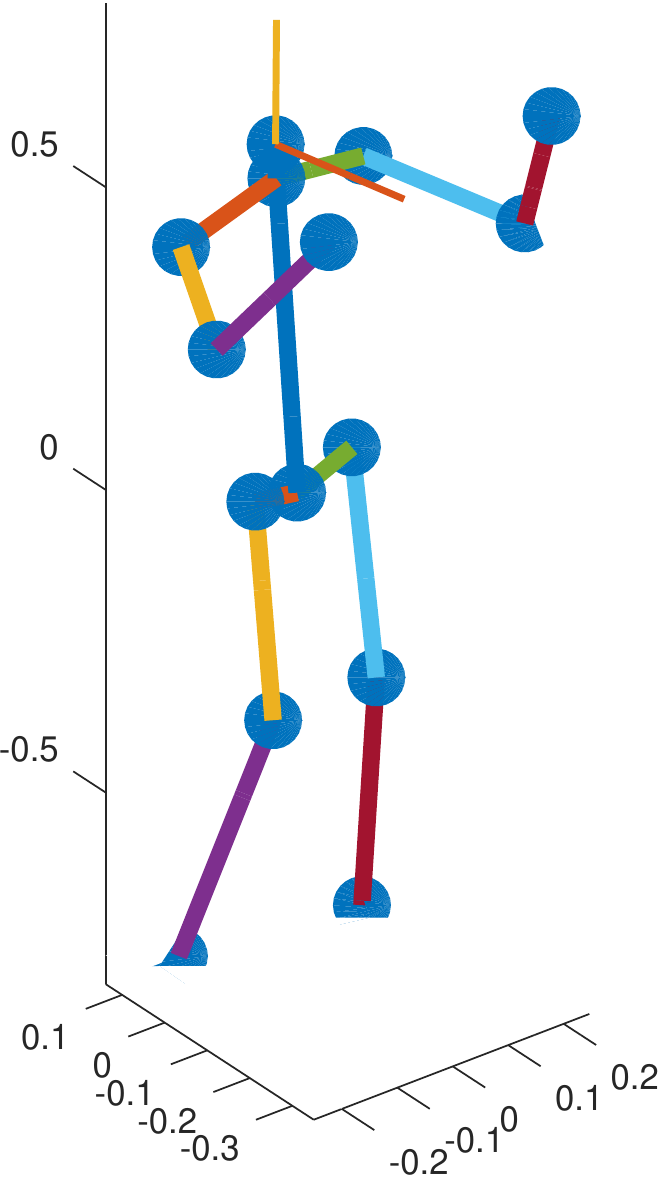}%
\\
	\rotatebox{90}{\hspace{10pt}{\scriptsize Ground Truth}} &
\includegraphics[width=0.2\linewidth, height=0.275\linewidth]{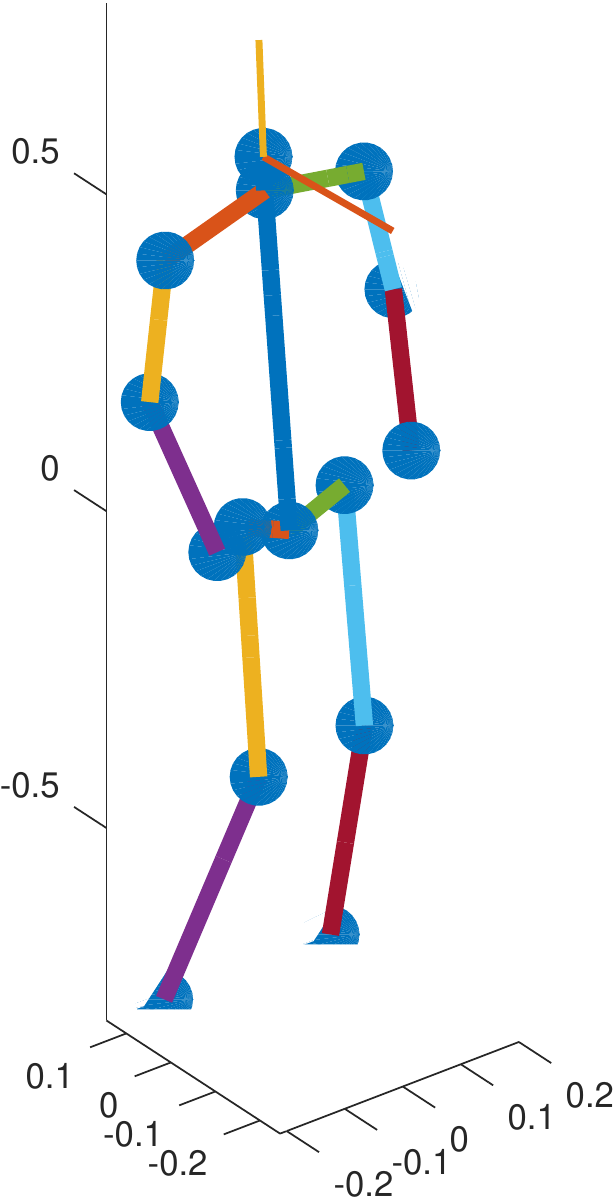}%
\includegraphics[width=0.2\linewidth, height=0.275\linewidth]{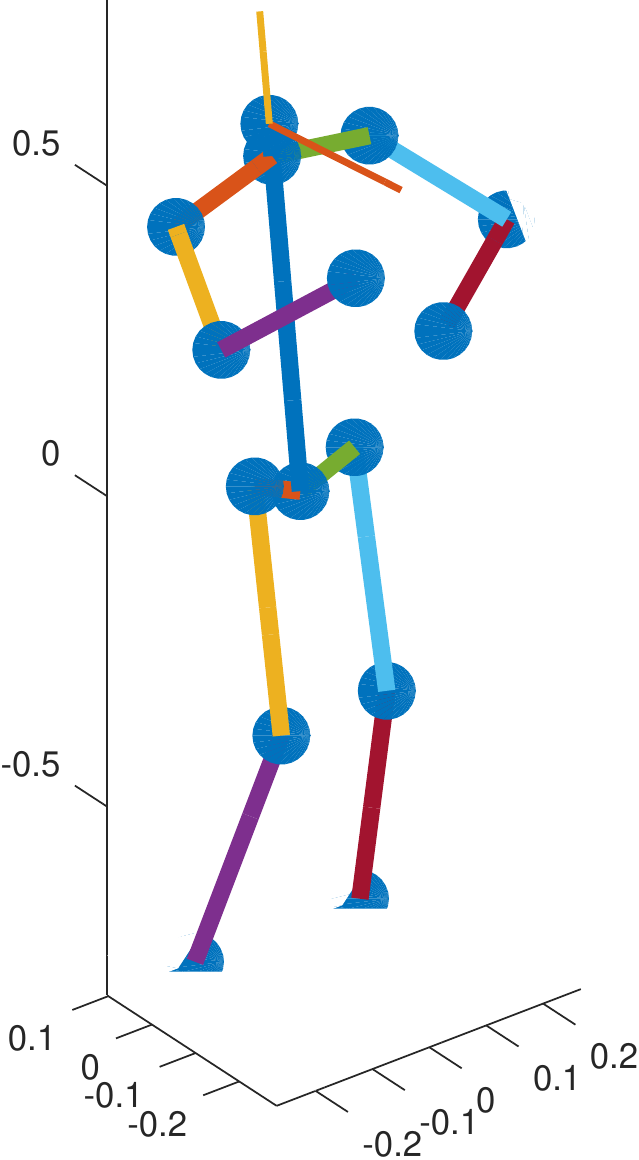}%
\includegraphics[width=0.2\linewidth, height=0.275\linewidth]{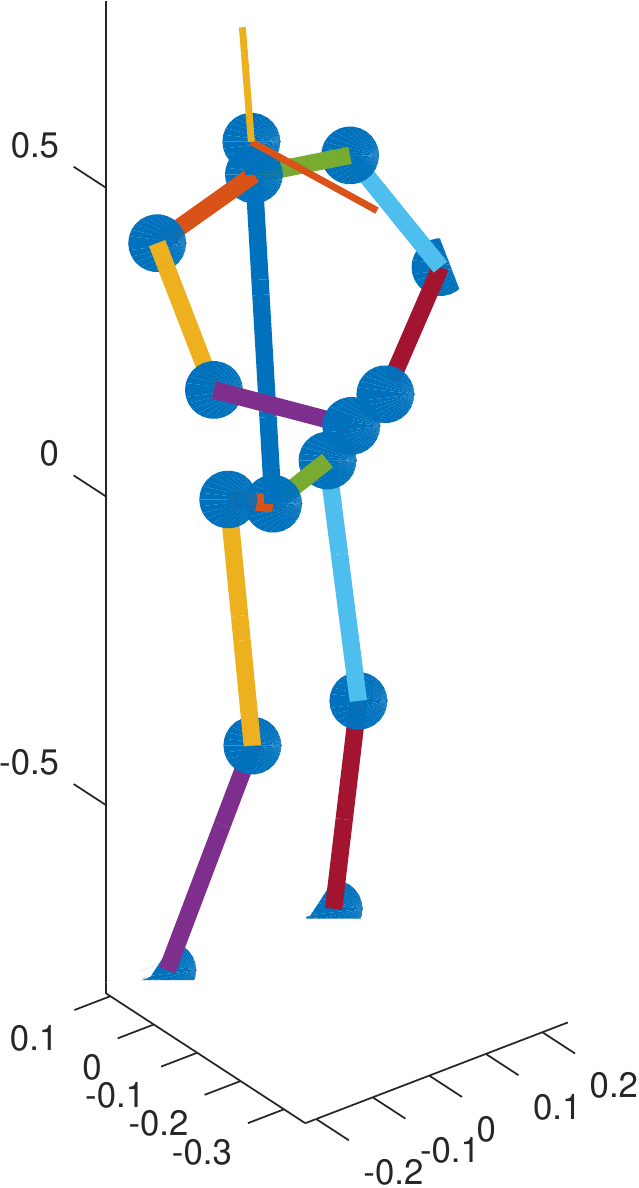}%
\includegraphics[width=0.2\linewidth, height=0.275\linewidth]{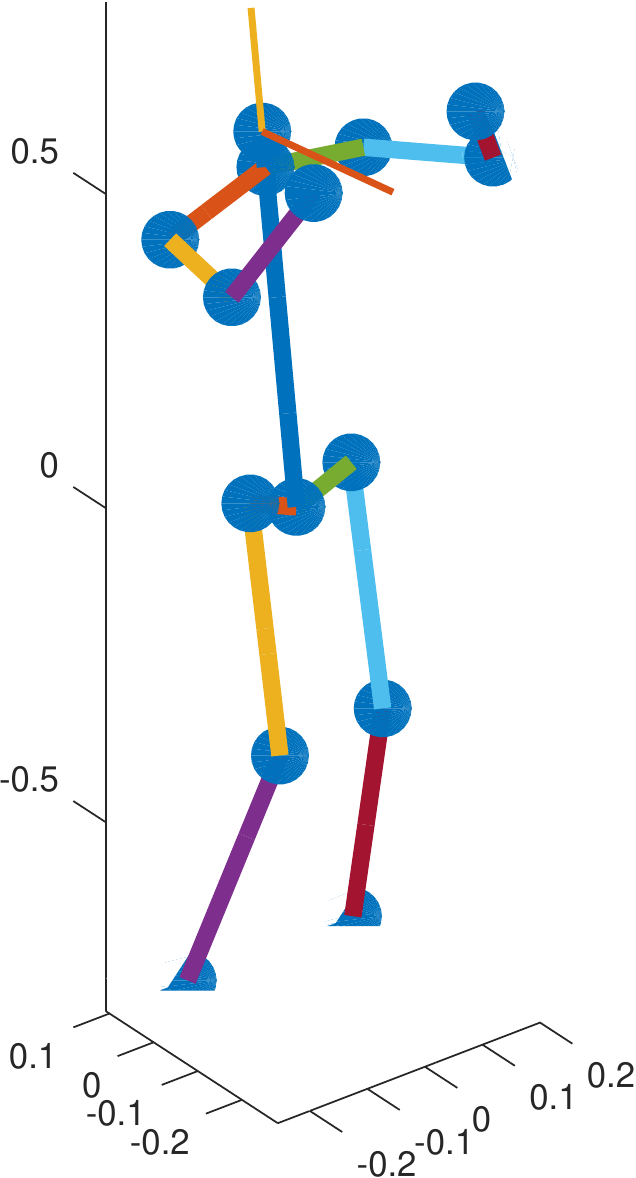}%
\includegraphics[width=0.2\linewidth, height=0.275\linewidth]{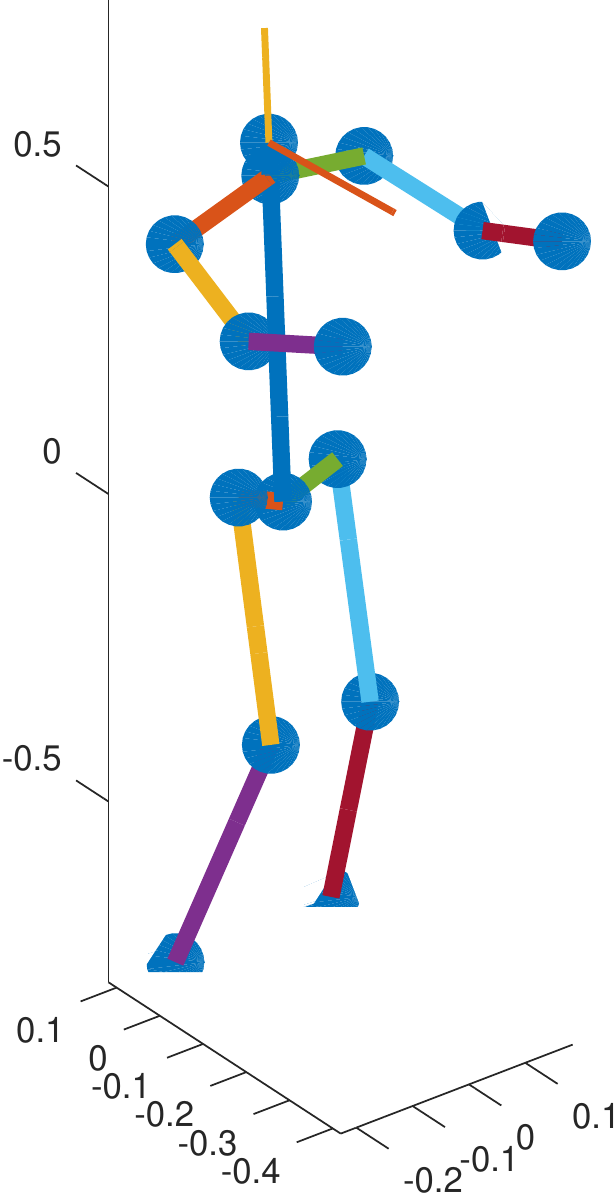}%
\end{tabularx}}
	\vspace{-2pt}
\caption{Egopose estimation from a human vision span. 
The head-mounted front facing fisheye camera rarely sees the wearer. 
When the wearer is visible in the peripheral view, the visible body parts are often limited.
We extract the body part segmentation (row one), motion history image (row two) and estimate the body and head pose of the wearer in real time (row three).  
Row four shows the ground truth egoposes.}
\label{fig:teaser}
	\vspace{-15pt}
\end{figure}

Egopose estimation is a challenging task. Existing approaches generally fall under two categories: 
non-optical sensor based methods, and camera based approaches. 
Sensors based approaches relying on magnetic and inertial attributes give robust estimate of the egopose \cite{imu, rokoko}.
However, they need specially designed equipment, are usually harder to set up, and reasonably intrusive, inhibiting the user's general movement.
Camera based methods are less intrusive and can work in different environments. 
One category of these approaches relies on top-down head-mounted camera to have
the best view of the wearer \cite{egocap, mo2cap2, fb1, fb2}, 
while the other uses the narrow field of view (FOV) front facing cameras in which camera wearer is mostly invisible \cite{Jiang, austin, cmu1, cmu2}. 
The former setting leads to reliable results as long as they can `see' body parts clearly.
However, the head-mounted downward cameras need to extrude to the front to avoid the occlusion of the nose and cheek. 
When the wearer is missing from the field of view (FOV), the pose estimation would completely fail. 
The later setting has the advantage of estimating egopose without seeing the wearer, 
although it cannot resolve some ambiguous body poses, especially the arm poses. 

In AR and VR devices, it is natural to have cameras close to wearer's face and have a visual field similar to human eyes:
for the most part, the camera can see the wearer's hands and some other parts of the body only in the peripheral view, 
and for a significant portion of the time it cannot see the wearer at all, for instance when the camera wearer looks up.
This presents a new setting for egopose -- a human-eye-like vision span, which we believe has not been studied. 
Our solution framework, as shown in Fig.~\ref{fig:teaser}, 
takes advantage of both the camera motion and visible body parts to give robust egopose
estimation not matter the wearer is visible or not in the camera's FOV. 
We propose a deep learning approach to tackle this problem. 

Firstly, our proposal uses both the dynamic motion information obtained from camera SLAM, and the occasionally visible body parts for predictions.
In addition to predicting the egopose, the model computes 3D head pose and the figure-ground segmentation of camera wearer in the egocentric view. 
Because of this joint estimation of head and body pose, we can enforce certain geometrical consistency during the inference, 
which can further improve results and enable us to reposition the egopose into a global coordinate system with camera SLAM information.
Secondly, the proposed method allows wearer to be invisible in the field of view;
and in cases where the camera wearer is partially visible, 
our method can take advantage of both motion and visible shape features to further improve the results.

Third, one of the biggest challenges in egopose estimation is the availability of good datasets. 
It takes a lot of effort to capture synchronized egocentric video and body/head poses for hundreds of subjects. 
In this work, we instead utilize existing datasets to the best extent possible, 
specifically leveraging mocap data collected over the past decades. 
These mocap data usually only capture the body joints movement and they do not include the egocentric video. 
Building on \cite{fb1}, in this work, we also propose an approach to synthesize not only the virtual view egocentric images, 
but also the dynamic information associated with the pose changes. 
We show that such synthetically generated datasets already have superior generalization power on real videos. 
Lastly, our main application is in AR and VR setting, and hence, 
we propose the model with low latency design constrains so it can be deployed in real-time applications. 

The contributions of this paper are: 
\begin{itemize}  \setlength{\itemsep}{0pt}	\setlength{\parskip}{0pt}
	\item An egopose estimation model from a novel perspective of the human vision span, critical to small factor AR/VR glasses, 
		where the FOV covers very limited and sometimes 
		no view of the wearer; 
    \item A joint estimation procedure for ego-head and ego-body poses;		
	\item An approach for synthesizing data for egopose from existing mocap data, which is generalizable to real scenarios; and
	\item A pose estimation model that is real-time, thereby enabling real-world AR and VR applications.  
\end{itemize}


\section{Related works} \label{sec:related}

Human pose estimation is a critical task in computer vision.
Both 2D and 3D human pose estimation techniques have been extensively studied from a {\it third person} perspective 
\cite{openpose, maskrcnn, 3da1, 3da2, 3da3, 3da4, 3db1, 3db2, 3db3}. 
More recently, egocentric pose estimation has also received interest because of its relevance to immersive motion capture and AR/VR applications.
By attaching multiple cameras to a person's body, 3D egopose can be optimized by using camera SLAM and body structure constraints \cite{mslam}.
In \cite{rgbd-pose1, rgbd-pose2}, a chest or head-mounted rgbd camera is used to estimate the camera wearer's hand, arm and torso motions.
Jiang and Grauman \cite{Jiang} use a chest-mounted rgb camera to estimate the wearer's full body 3D poses.
They use a random forest to estimate the pose classes on global motion features,
followed by a convolutional network to classify the sitting vs. standing pose based on the scene context.
The resulting estimates are fused and jointly optimized over a long video to extract the human pose sequence.
This is not real-time, and it does not explicitly use the visible body parts to disambiguate upper body pose.
Interaction with other people in the chest-mounted camera's FOV has also been used to improve egopose estimation \cite{austin} with a deep learning approach.
Yuan and Kitani \cite{cmu1, cmu2} propose deep learning and control-based approaches for egopose estimation using a narrow FOV head-mounted rgb camera.
These methods use optical flow as the input and camera wearer is mostly invisible in the camera's FOV.
It is hard to use these methods to reconstruct poses that cannot be disambiguated by head motion alone.
Experiments show that these previous motion based methods are not suitable for our new setting; our proposed method gives 
much better results.


Egocentric pose estimation using body cameras that look at the camera wearer has also been studied.
In \cite{monoeye1, monoeye2}, a chest mounted fisheye camera is used and they rely on the partial imagery of the wearer.
However, this is intrusive and the camera is hard to be mounted rigidly.
A more often used setting is a head-mounted downward looking camera, which can always see most of the camera wearer.
Rhodin et. al. \cite{egocap} use two head-mounted downward fisheye cameras to capture egoposes.
A single downward fisheye camera \cite{fb1, fb2, mo2cap2}
has also been used to give accurate 2D keypoints and 3D egocentric poses with deep learning approaches. 
In \cite{mo2cap2, tvcg18}, with a downward camera setting, the rotation of the torso is also estimated so that the 3D pose can be transformed to a global system.
The downward camera setup may be suitable for a large VR or AR headset which can position the cameras to have the best view of the wearer.
However, for small factor AR/VR glasses and for hardware configurations that are less {\it bulky},
the cameras have to be very close to the face, thereby the downward facing camera would
have a bad view of the wearer due to the occlusion of nose, mouth and cheek.
The existing approaches cannot handle such small factor AR/VR setups, a vital aspect of the proposed approach.
We tackle the egopose from a new perspective where the camera has a human-eye-like field of view,
which can see the wearer using the peripheral view and depending on the head pose the camera may have very limited or no view of the camera wearer.
Since existing approaches are not suitable for this more naturalistic setting, we propose a new model.



\section{Method} \label{sec:method}

\subsection{Overview} \label{sec:overview}

\noindent {\bf Problem definition:}
Given a sequence of video frames $\{I_{t}\}$ of a front facing head-mounted fisheye camera at each time instant $t$,
we estimate the 3D ego-body-pose $\mathcal{B}_t$ and ego-head-pose $\mathcal{H}_t$.
$\mathcal{B}_t$ is an $N\times3$ body keypoint matrix and $\mathcal{H}_{t}$ is a $2\times3$ head orientation matrix. 
The ego-body-pose is defined in a local coordinate system in which the hip line is rotated horizontally so that it is parallel to the $xz$ plane,
and the hip line center is at the origin as shown in Fig.~\ref{fig:teaser}.
The ego-head-pose comprises of two vectors: a facing direction $\mathbf{f}$ and the top of the head's pointing direction $\mathbf{u}$.
Estimating the head and body pose together allows us to transform the body pose to a global coordinate system using camera SLAM.
We target at real-time egopose estimation so the deep models should be efficient and accurate.  

\begin{figure}[tb]
\includegraphics[width=\linewidth, height=0.3\linewidth]{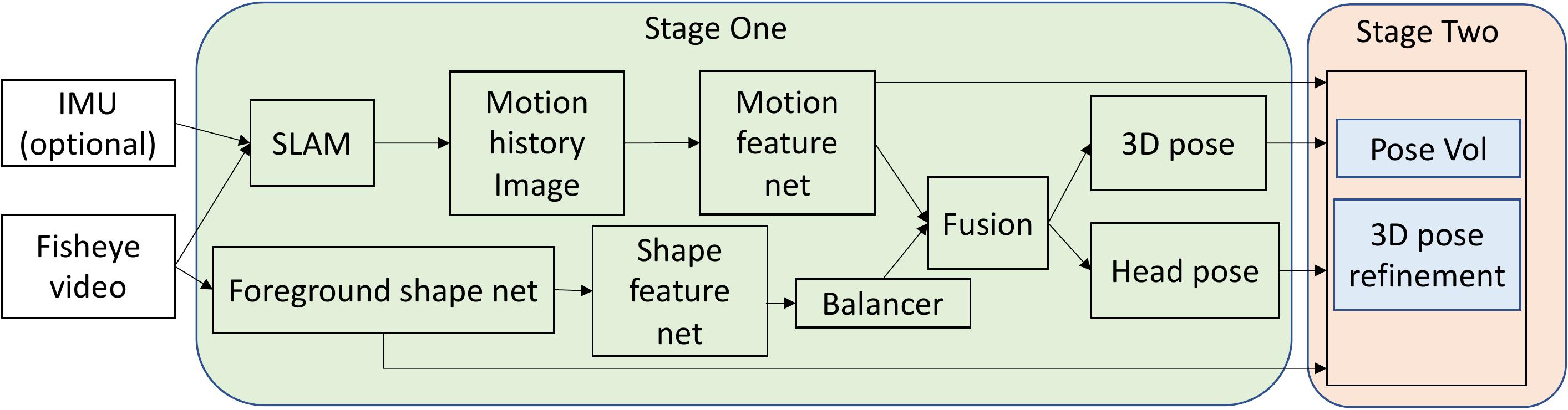}
\caption{The proposed system architecture} \label{fig:sysarch}
	\vspace{-16pt}
\end{figure}

Our proposed system is driven by a head-mounted front facing fisheye camera with an around 180-degree FOV. 
As motivated, and similar to a human-vision span, the camera mostly focuses on the
scene in the front with minimal visual of wearer's body parts via peripheral view.
In such a setting, egopose estimation using only the head motion or the visible parts imagery is not reliable. 
Our proposed method takes advantage of both these information streams and optimizes for the combination efficiently. 
The overall {\bf system architecture} is shown in Figure \ref{fig:sysarch}. 
The sequence of blocks and operations is as follows: 
In one branch, the fisheye video and optional IMU are used to extract the camera pose and position in a global coordinate system. 
We convert the camera motion and position to a compact representation denoted as the {motion history image}. 
The {motion feature net} processes the motion history image to extract dynamic features.
Separately, in a parallel branch, the fisheye image is also sent to the shape net to extract the wearer's foreground shape.
We further extract shape features from the foreground shape presentation. 
The fusion network balances and combines the two branch outputs (dynamic features and shape features) and 
gives the egopose estimates -- the initial body keypoints and head pose estimations.
Once this is done, we further refine the body keypoints using a 3D approach, leading to the final egopose estimate.
We address each of these components one at a time. 

\subsection{Stage 1: Egopose initial estimate} \label{sec:stage1}

We propose a new method using both dynamic features and shape features for robust egopose estimation.

\vspace{-8pt}
\subsubsection{Motion history image and motion feature net} \label{sec:motion}
\vspace{-4pt}

We propose the \textit{motion history image}, a representation which is invariant to scene structures and can characterize the rotation, 
translation and height evolution in each time interval. 
At each time instant $t$, we compute the incremental camera rotation $R_t$ and translation $d_t$ 
from the previous time instant $t-1$ using camera poses and positions from SLAM \cite{fisheyeslam}.
We incorporate $R_t-\mathcal{I}_{3\times3}$, where $\mathcal{I}$ is an indentity matrix, into the motion representation.
$d_t$ needs to be converted to the camera local system at each time instant $t$ so that it is invariant to the wearer's facing orientation.
To remove unknown scaling factor, we further scale it with the wearer's height estimate. 
The transformed and normalized $d_t$ is $\hat{{d}_{t}}$.
Based on SLAM, a simple calibration procedure in which the wearer stands and then squats can be used 
to extract the person's height and ground plane's rough position.

\begin{figure}[tb]
\includegraphics[width=0.125\linewidth, height=0.2\linewidth]{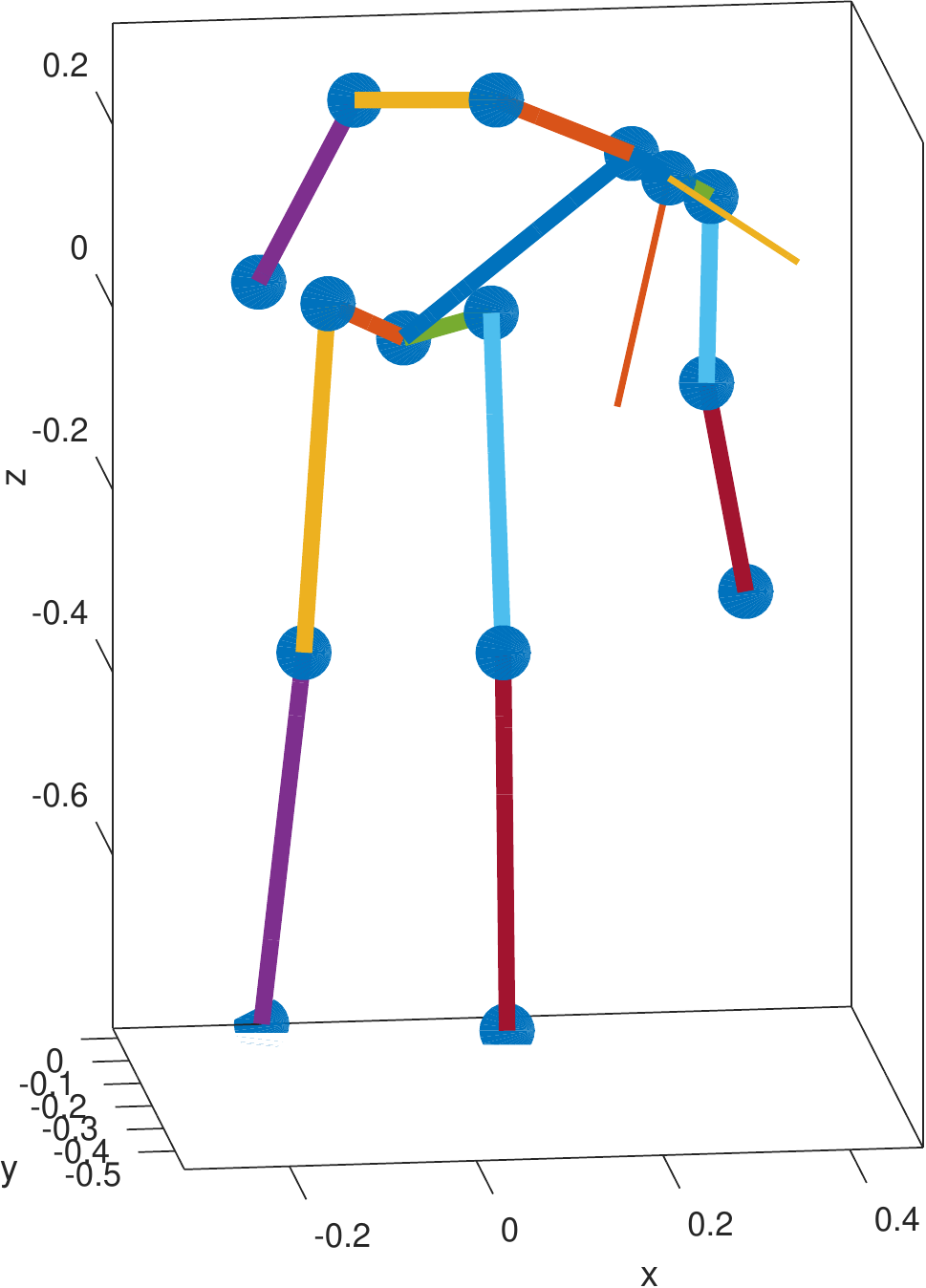}%
\includegraphics[width=0.125\linewidth, height=0.2\linewidth]{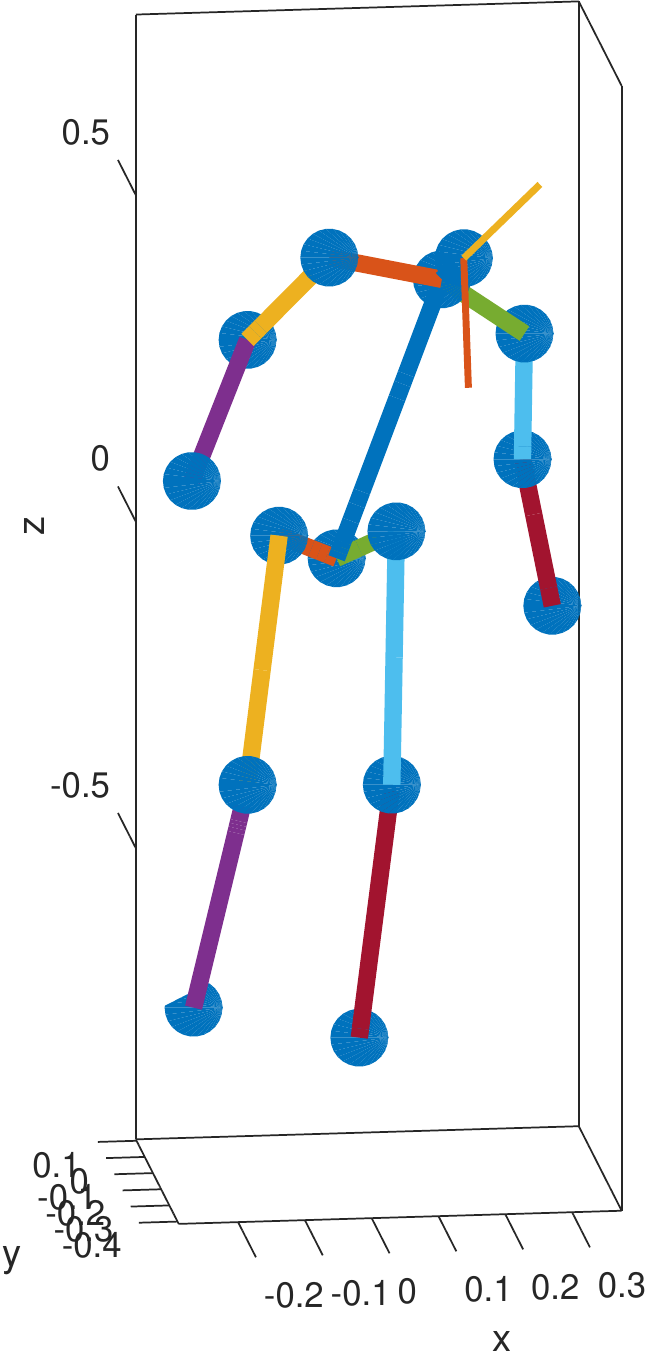}%
\includegraphics[width=0.125\linewidth, height=0.2\linewidth]{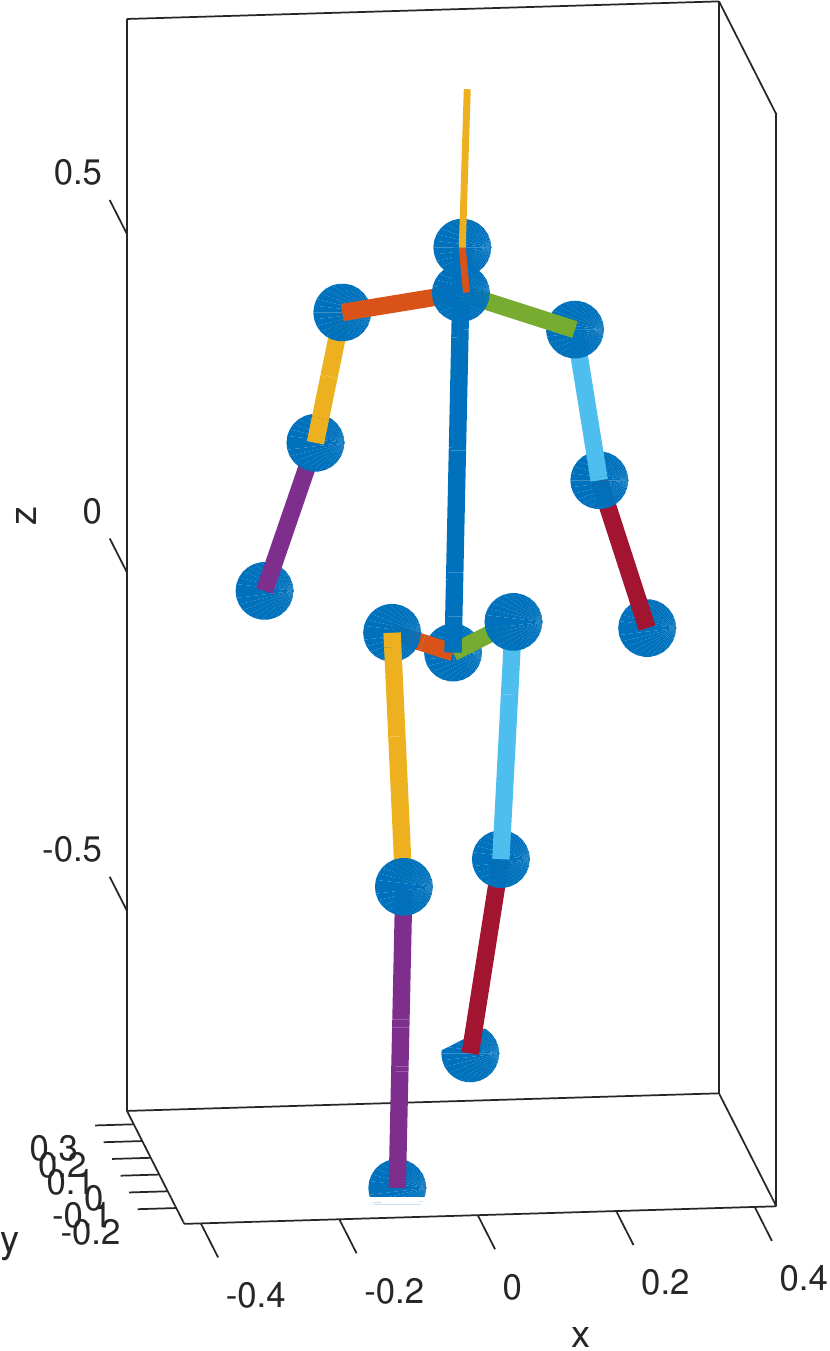}%
\includegraphics[width=0.125\linewidth, height=0.2\linewidth]{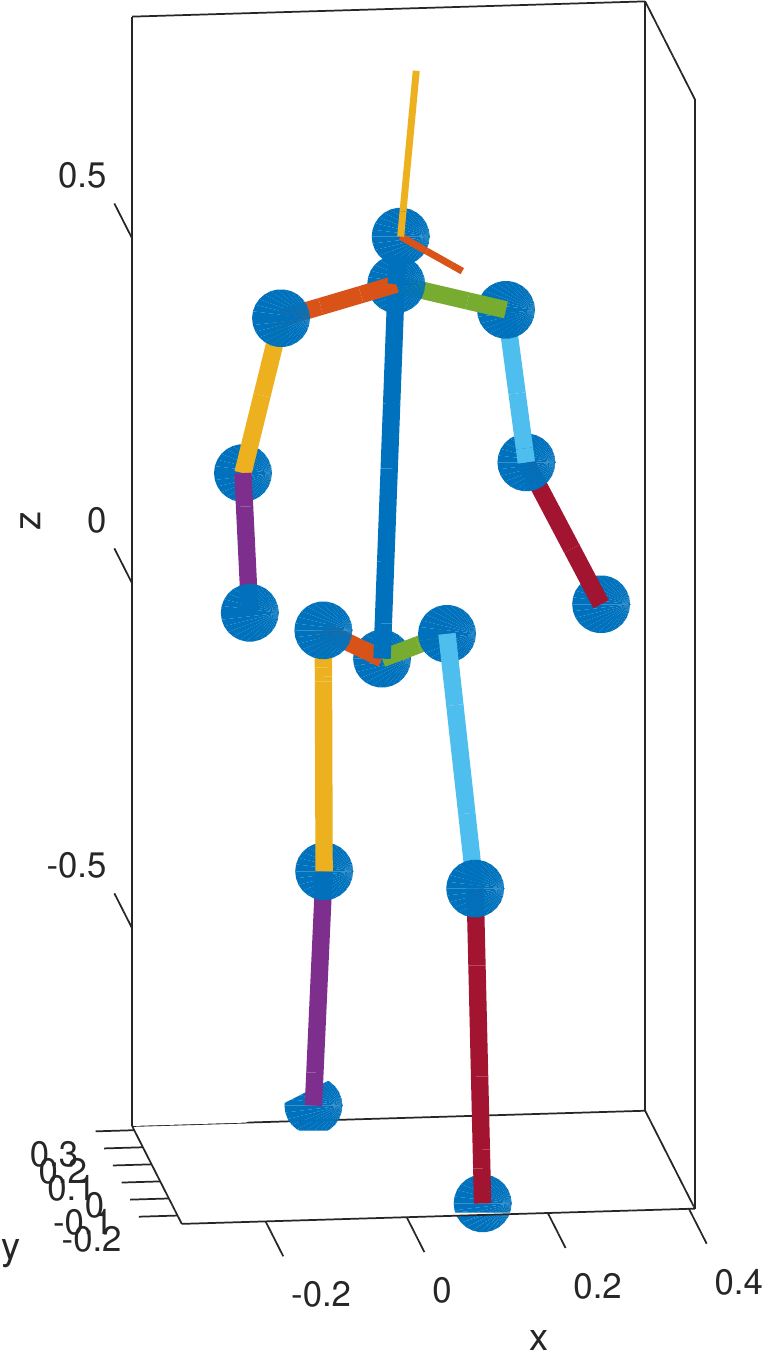}%
\includegraphics[width=0.125\linewidth, height=0.2\linewidth]{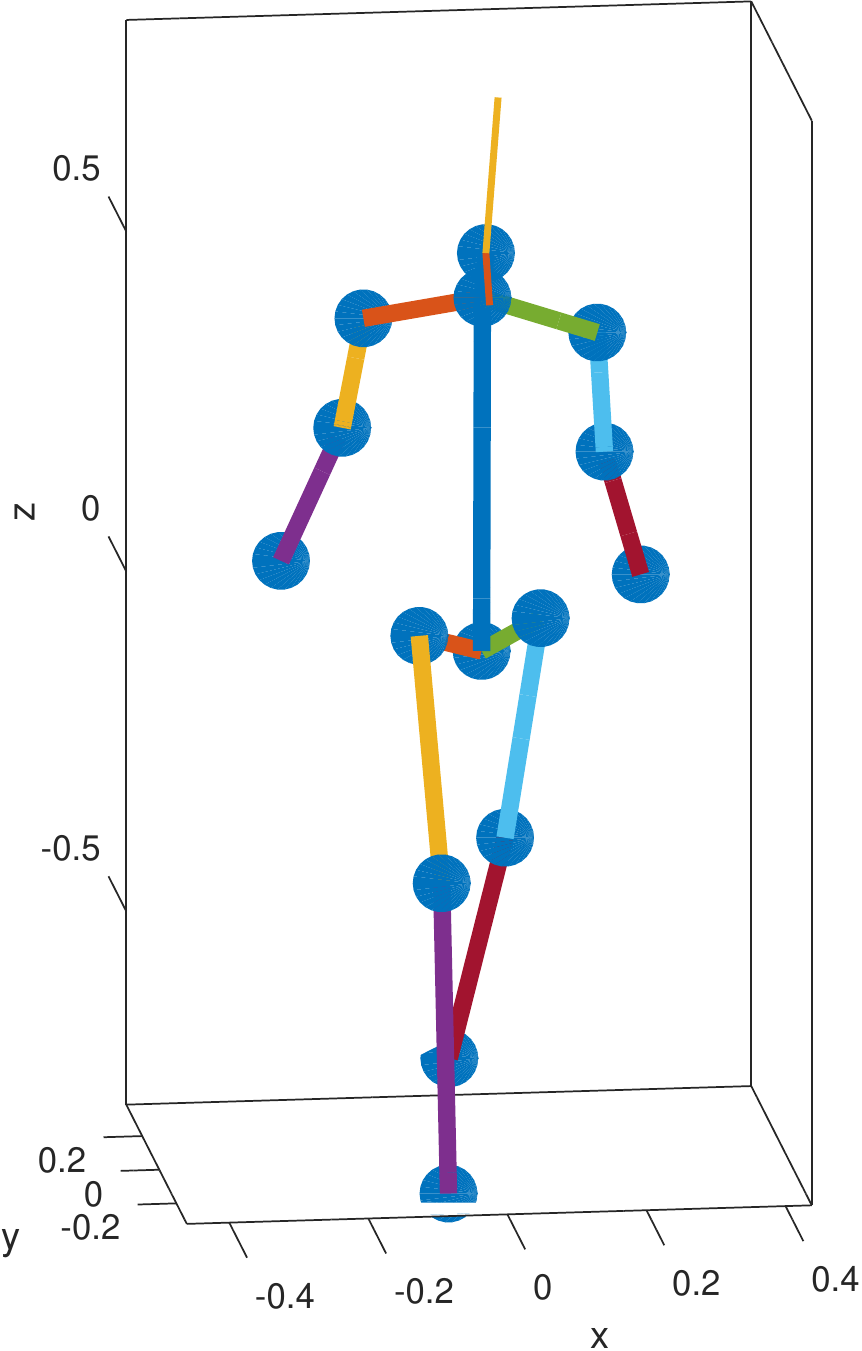}%
\includegraphics[width=0.125\linewidth, height=0.2\linewidth]{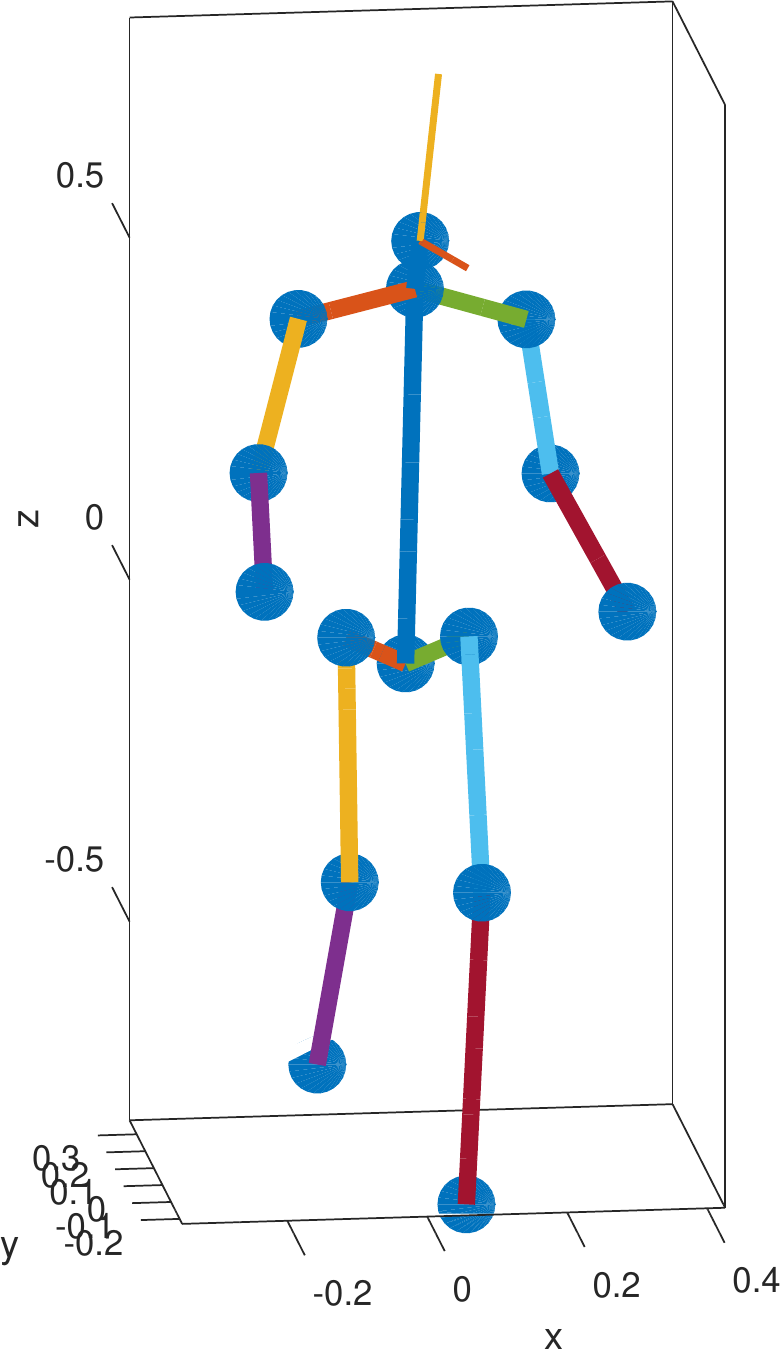}%
\includegraphics[width=0.125\linewidth, height=0.2\linewidth]{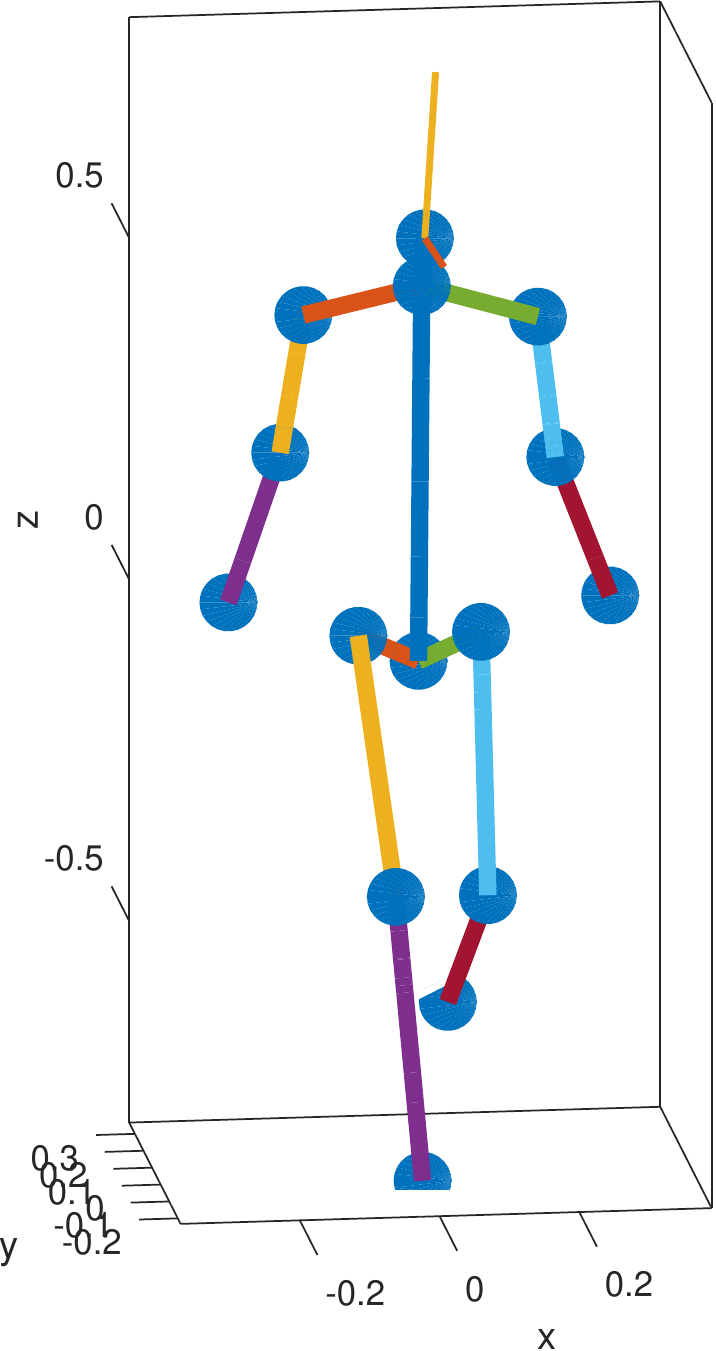}%
\includegraphics[width=0.125\linewidth, height=0.2\linewidth]{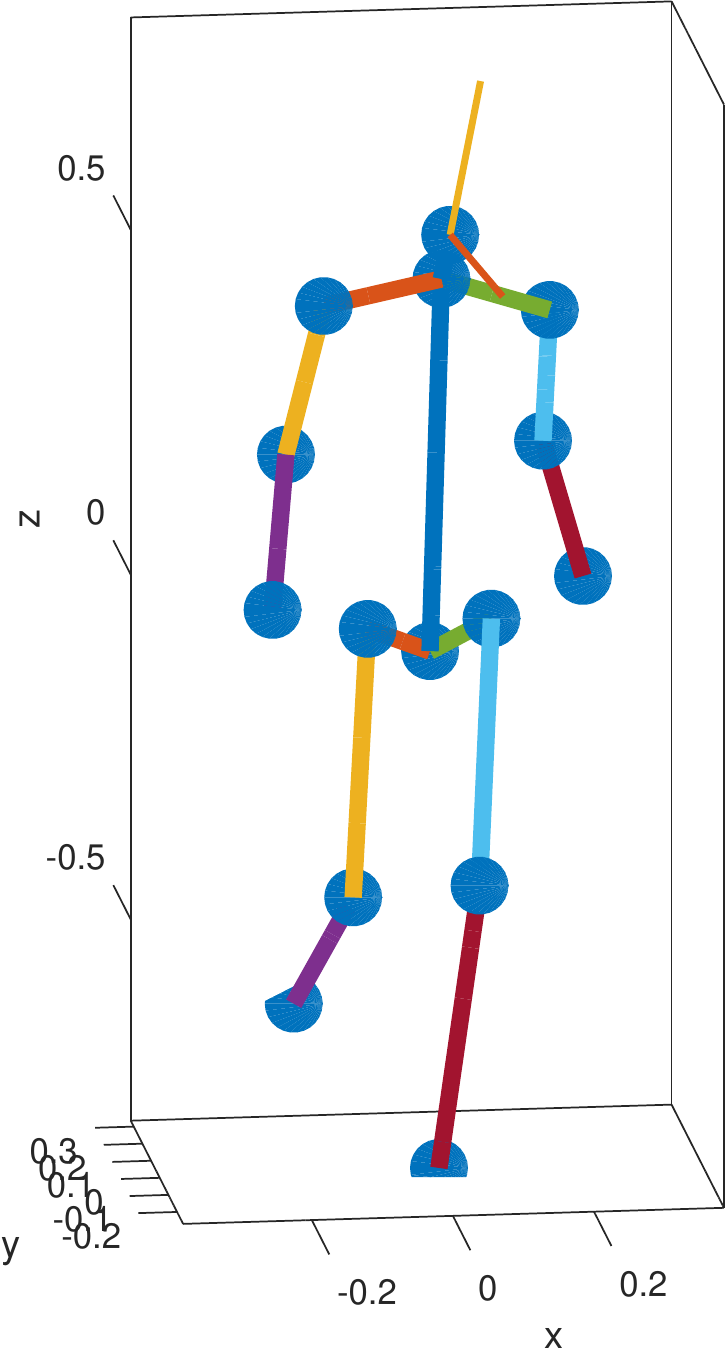}%
\linebreak
\includegraphics[width=0.125\linewidth, height=0.125\linewidth]{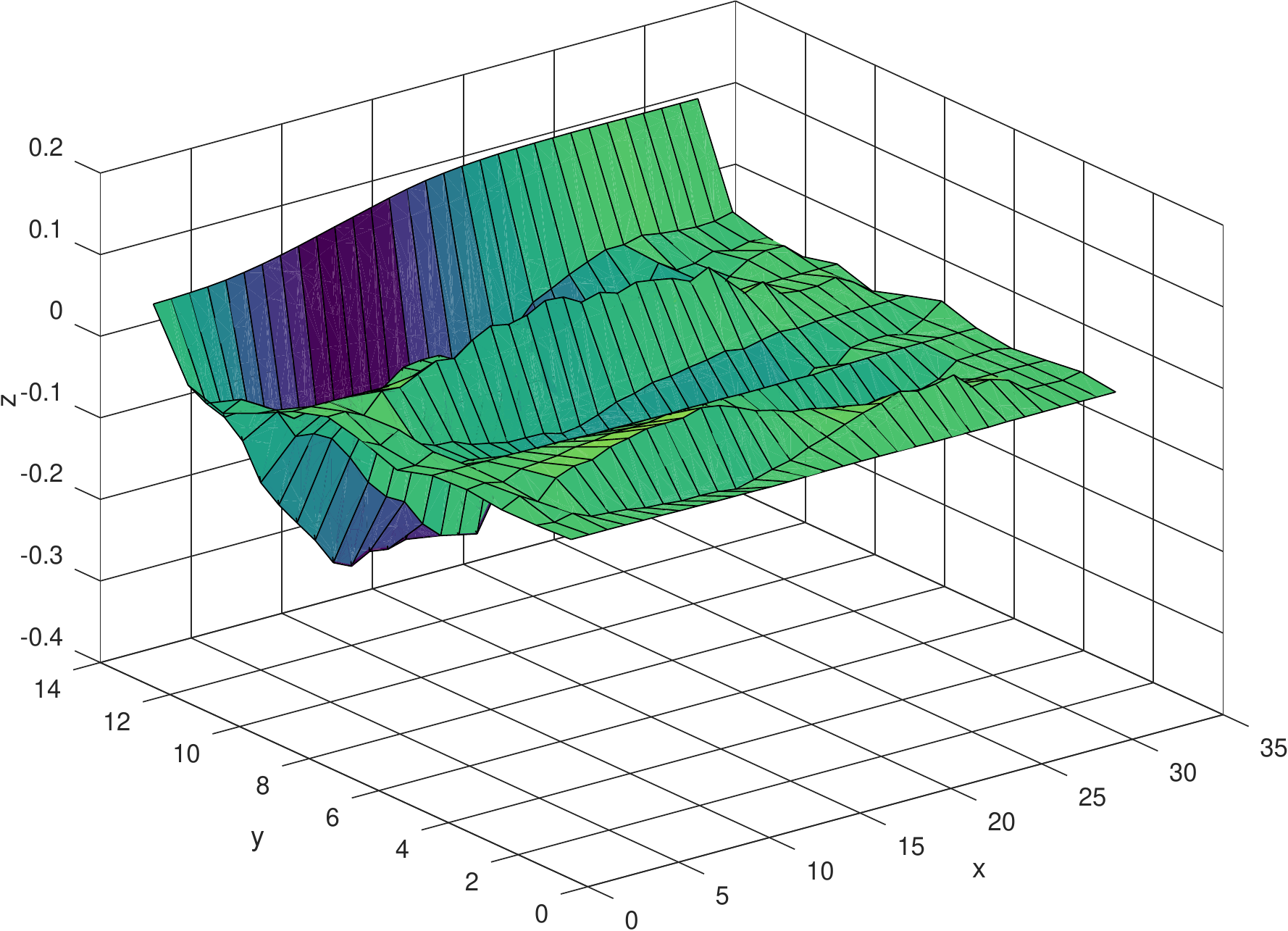}%
\includegraphics[width=0.125\linewidth, height=0.125\linewidth]{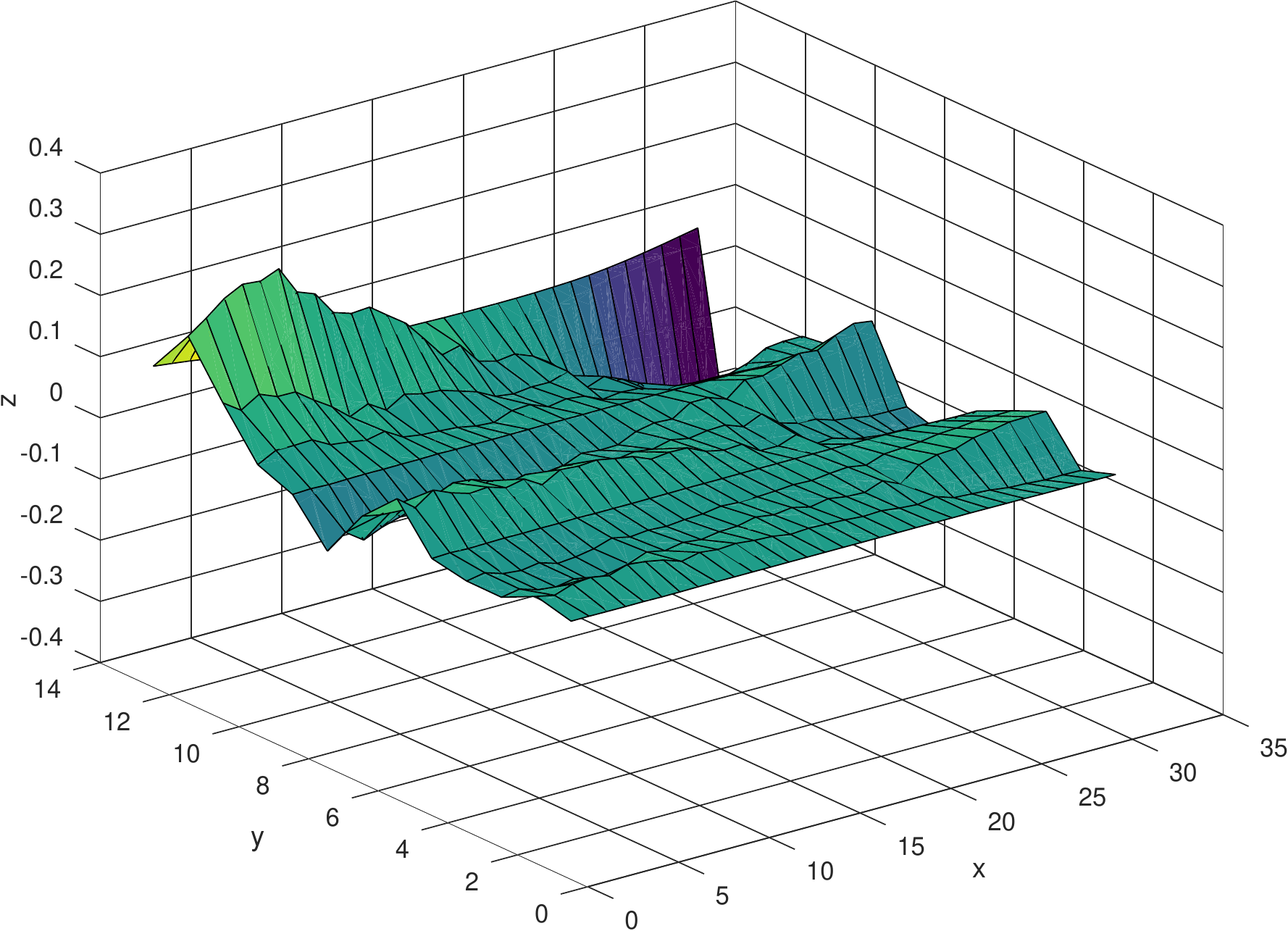}%
\includegraphics[width=0.125\linewidth, height=0.125\linewidth]{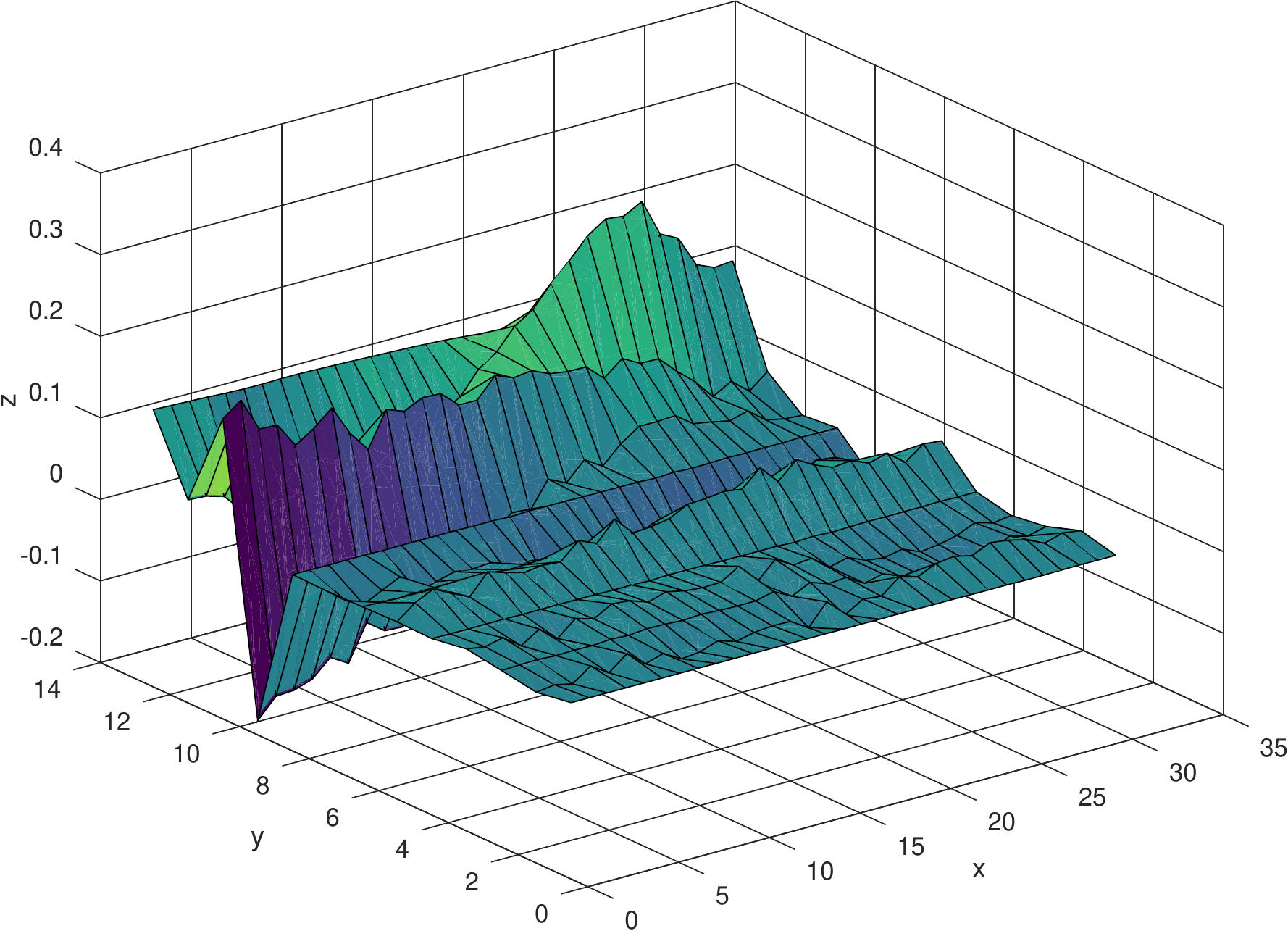}%
\includegraphics[width=0.125\linewidth, height=0.125\linewidth]{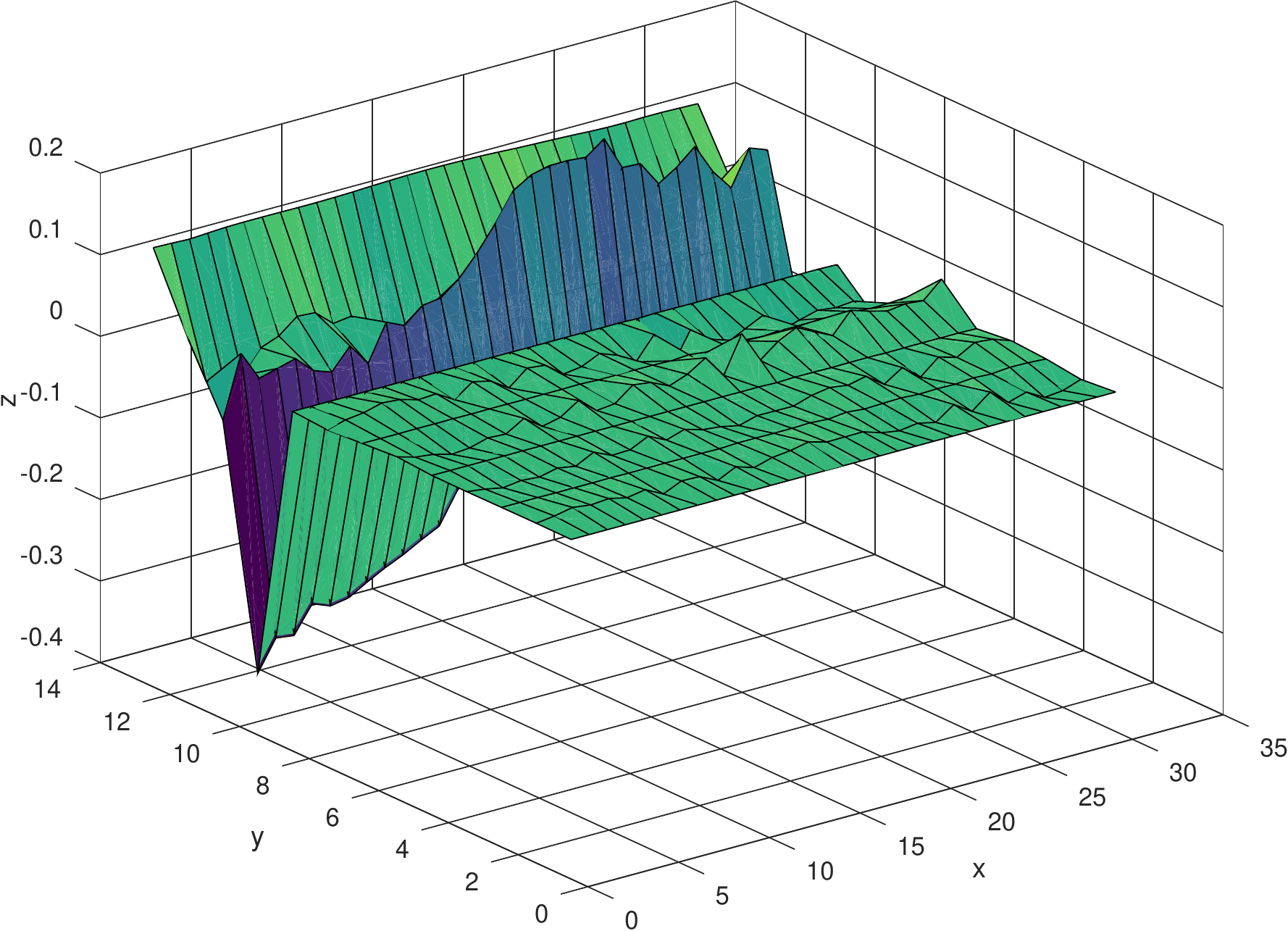}%
\includegraphics[width=0.125\linewidth, height=0.125\linewidth]{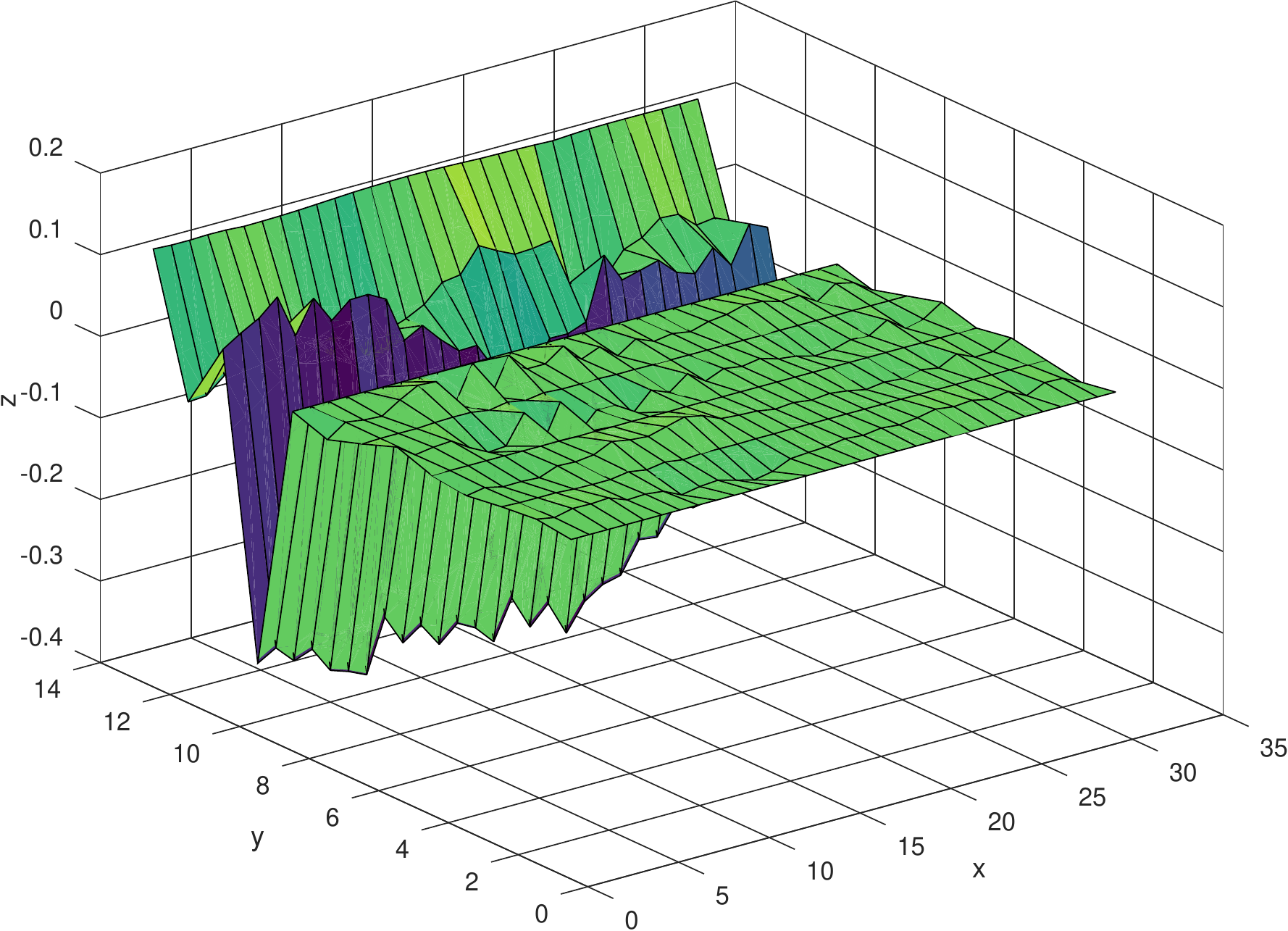}%
\includegraphics[width=0.125\linewidth, height=0.125\linewidth]{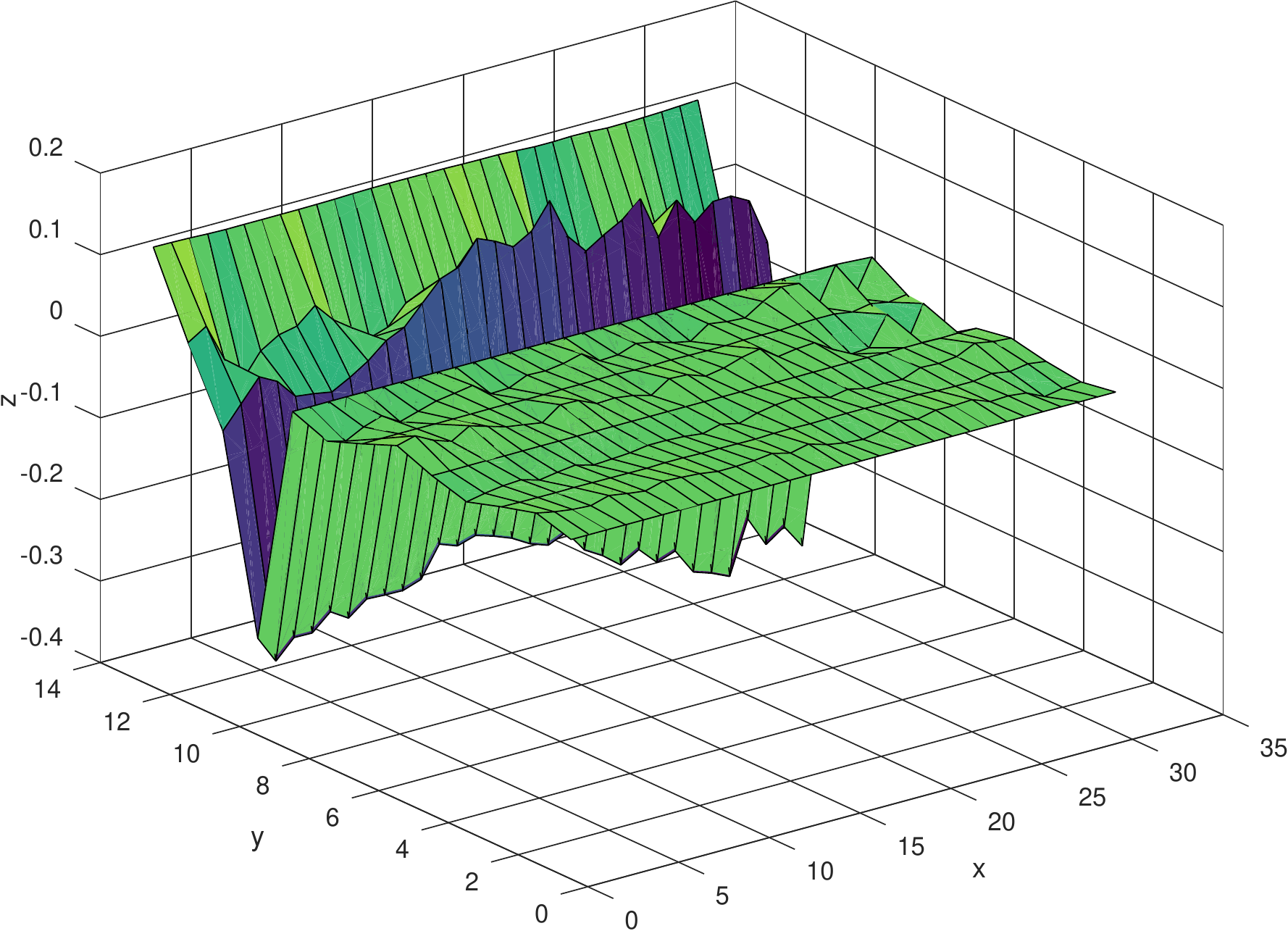}%
\includegraphics[width=0.125\linewidth, height=0.125\linewidth]{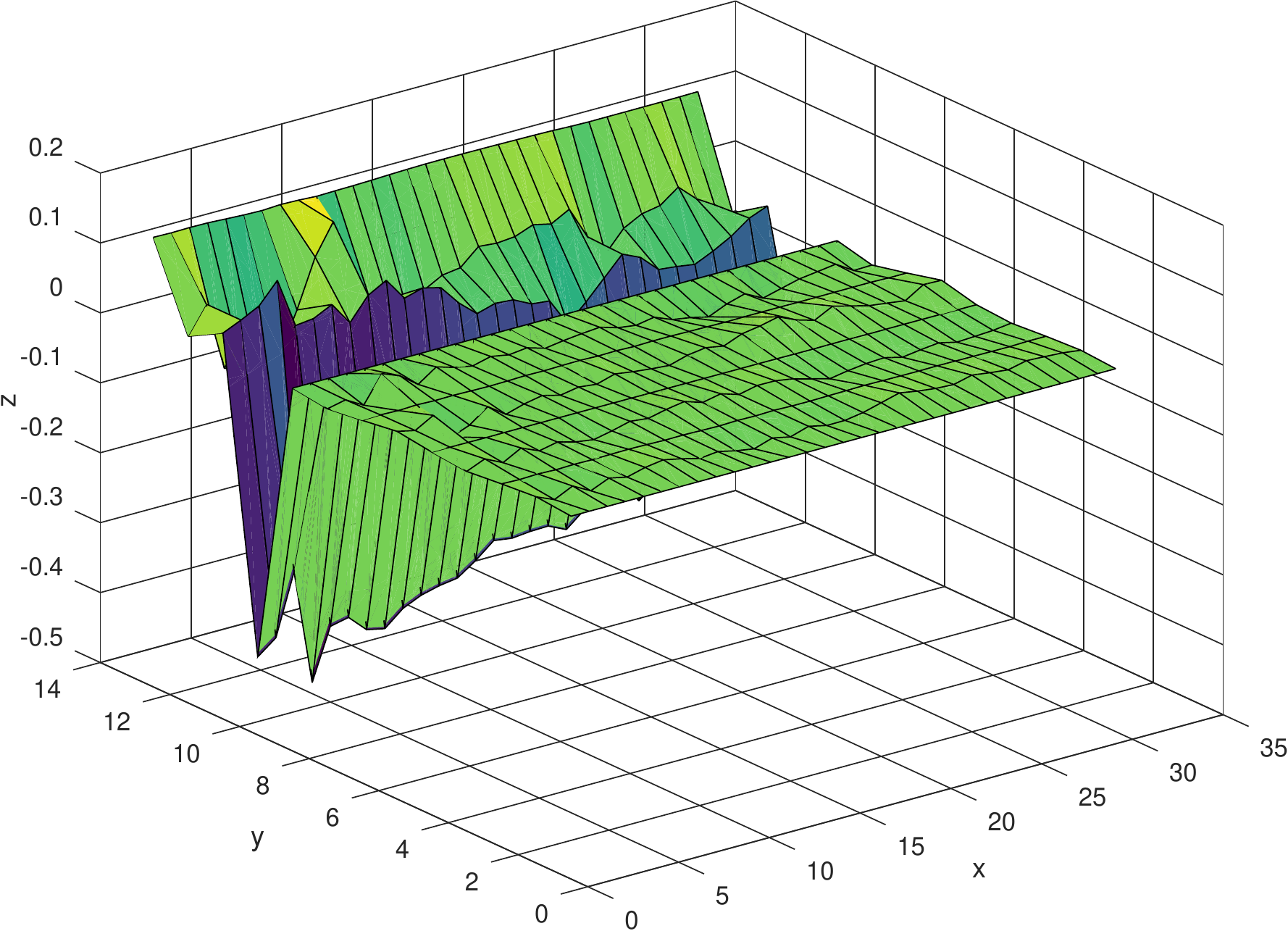}%
\includegraphics[width=0.125\linewidth, height=0.125\linewidth]{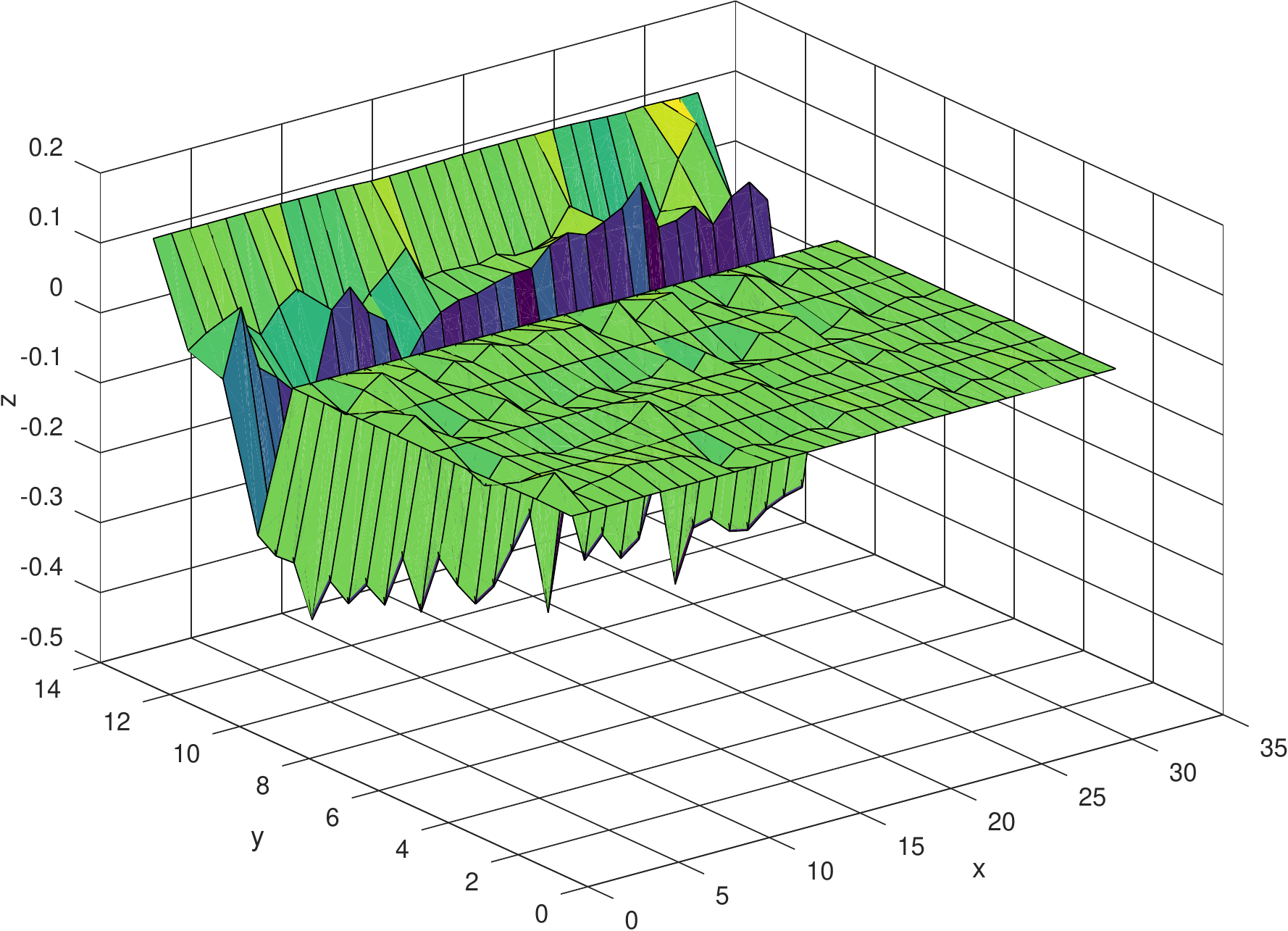}%
\caption{The motion history image representation. Row one: head and body pose samples in a sequence. 
	Row two: corresponding motion history images. 
	This dynamic feature characterizes the head and body poses and movements. (Figures in this paper are best viewed in color).}
	\label{fig:motion-history}	
	\vspace{-6pt}
\end{figure}

\begin{figure}
\centering
\includegraphics[width=0.5\textwidth]{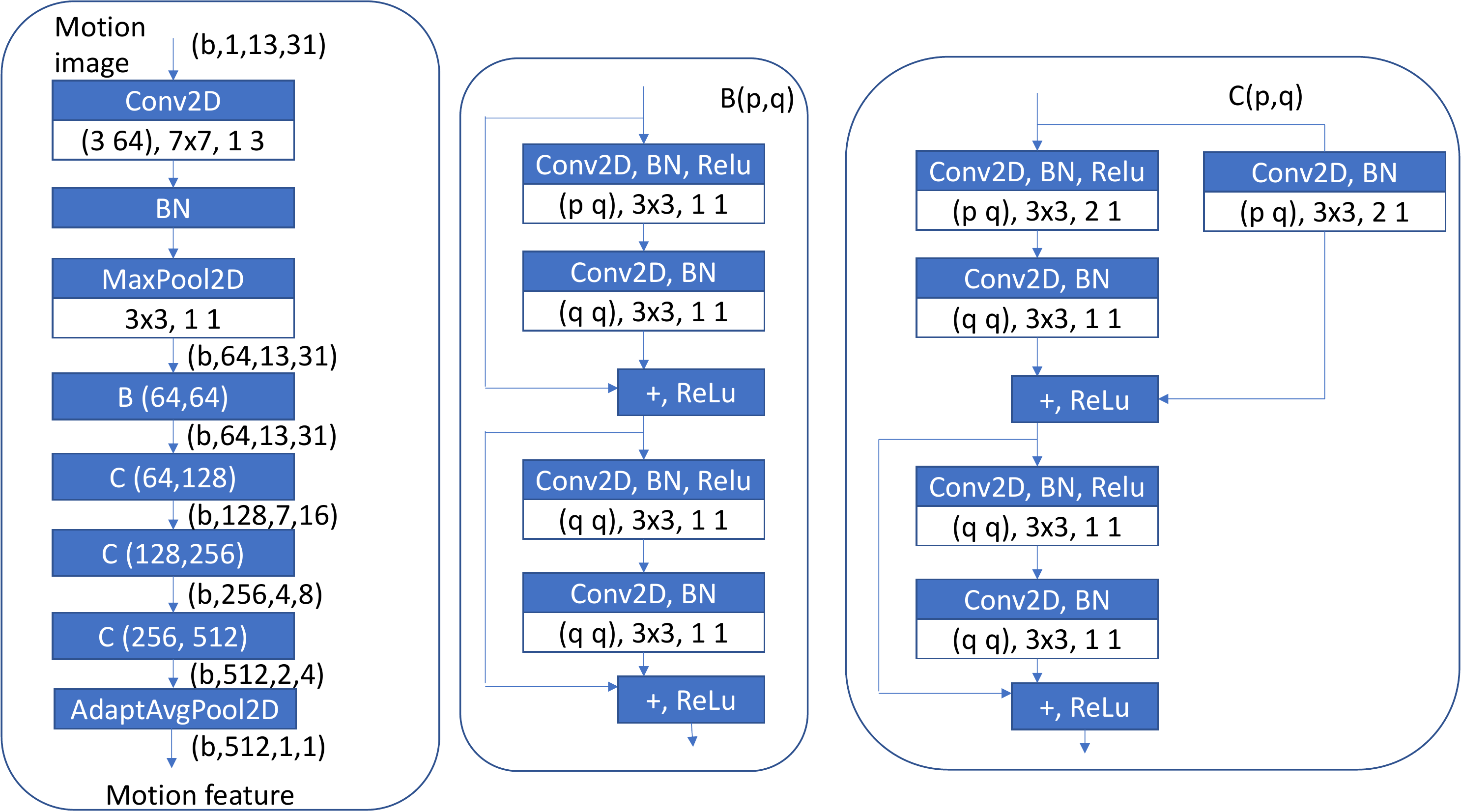}
\caption{Motion feature network. Parameters for convolution layers are input/output channels, kernel size, stride and padding. For maxpooling layers,
	the parameters are kernel size, stride and padding. b: batch size.}
\label{fig:motionnet}
	\vspace{-15pt}
\end{figure}

$R_t$ and $\hat{d}_t$ are not sufficient to distinguish the static standing and sitting pose.
Although the scene context image can be helpful \cite{Jiang}, it is sensitive to the large variation of people's height, 
e.g. a kid's standing view point can be similar to an adult's sitting view point.  
To solve this problem, we propose to use the camera's height relative to the person's standing pose (denoted by $g_t$) in the motion representation. 
We aggregate the movement features $R$, $d$ and $g$ through time to construct the motion history image. 
Specifically, we concatenate the flattened $R_{t}-\mathcal{I}_{3\times3}$, the scaled transition vector $a\hat{d}_{t}$ and 
the scaled relative height $c(g_{t}-m)$,
where $a=15, m=0.5, c=0.3$.
Fig.~\ref{fig:motion-history} gives examples of the motion history images with the corresponding human poses. 
As is evident, the proposed motion representation captures the dynamics of the pose changes in both periodic or non-periodic movements.
We then construct a deep network, the motion feature net, to extract the features from the motion history image, shown in Fig.~\ref{fig:motionnet}. 

\vspace{-8pt}
\subsubsection{Space aware shape estimation} \label{sec:shapeest}
\vspace{-4pt}

\begin{figure}[tb]
	\centering
	\scalebox{0.7}{	
\includegraphics[width=0.125\textwidth]{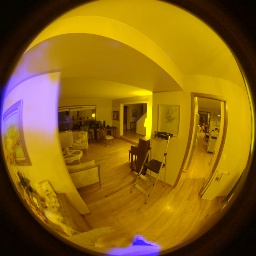}%
\includegraphics[width=0.125\textwidth]{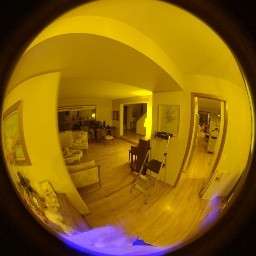}%
\includegraphics[width=0.125\textwidth]{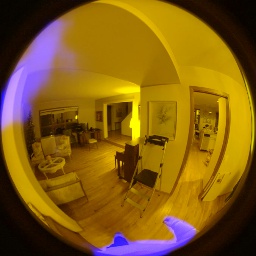}%
\includegraphics[width=0.125\textwidth]{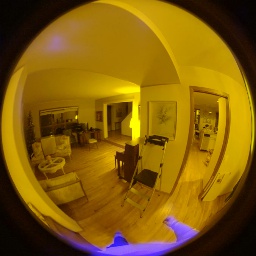}%
	}
\caption{Foreground shape estimation not using spatial map (odd number images) and using spatial map (even 
	number images).}
	\label{fig:shapenet-comparison}	
	\vspace{-10pt}
\end{figure}

Apart from the head motion, the foreground shape of the wearer is also closely coupled with the ego-head and ego-body poses. 
And it is particularly useful to disambiguate the upper body poses.   
To that end, we propose an efficient method to extract body shape.  
Unlike existing keypoint extraction scheme, from \cite{fb1, mo2cap2}, we argue that foreground body shape is a more suitable representation for our problem. 
As shown in Fig.~\ref{fig:shapenet-comparison}, in the human vision span, the wearer is often barely 
visible in the camera's FOV and there are often very few visible body keypoints. 
Keypoint estimation is thus a much harder task than the overall shape extraction.
In such setting, the foreground body shape often contains more information about the possible body poses than the isolated keypoints. 
For instance, if only two hands and part of the arms are visible, the keypoints would give only the hand locations
while the foreground body shape also indicates how the arm is positioned in the space.
The foreground shape can also be computed more efficiently and thus more suitable for real-time applications.

\begin{figure}[tb]
\centering
\includegraphics[width=0.5\textwidth]{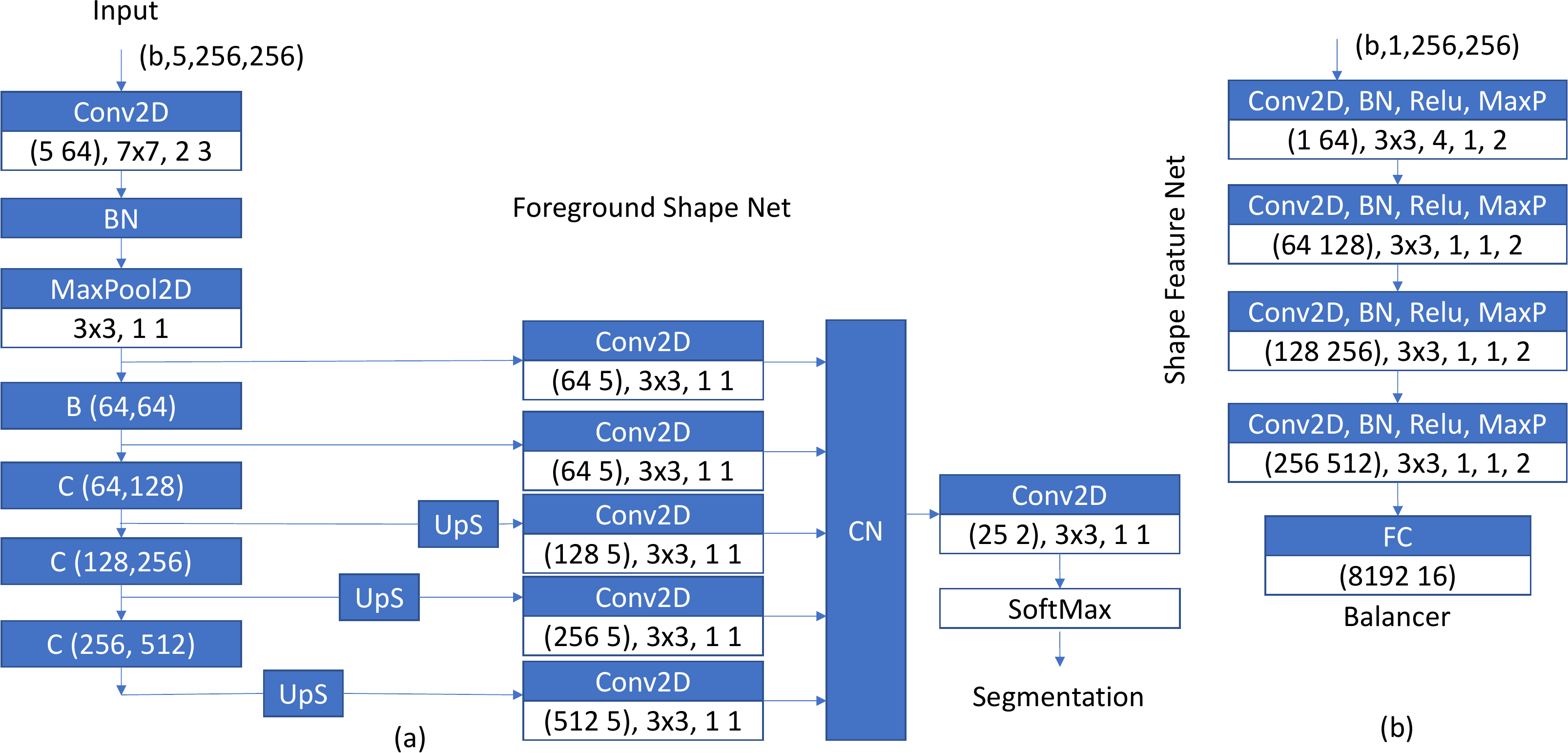}
	\caption{(a) The shape net. UpS block: bilinear upsampling layer. 
		Target resolution is $256\times256$. CN layer concatenates features from different scales along the channel dimension.
		Blocks B(.) and C(.) are defined in Fig.~\ref{fig:motionnet}.
		 (b) Shape feature extraction net.}
	\label{fig:shapenet}
	\vspace{-6pt}
\end{figure}

The proposed body shape network is shown in Fig~\ref{fig:shapenet}(a).
The shape net is fully convolutional and thus if we directly use the fisheye image as the input, 
we would obtain a spatial invariant estimation, which is undesirable. 
Since the wearer foreground is mostly concentrated at the lower part of the image and the arms would often appear in specific regions, 
the segmentation network should preferably be spatially variant.
To this end, we construct two more spatial grids: the normalized $x$ and $y$ coordinate maps, 
and concatenate them with the input image along depth dimension to generate a $256\times256\times5$ tensor.
These extra spatial maps help incorporate the spatial prior into the network during the training and inference. 
Fig.~\ref{fig:shapenet-comparison} shows the effects of these spatial constraints. 
The spatial map helps not only reduce the false alarms, but also correct missing detections in the foreground.
In this paper, we threshold the foreground probability map with 0.5 to obtain the final foreground shape representation.  
The foreground shape then passes through a small CNN, shown in Fig.~\ref{fig:shapenet}(b), for feature extraction.

\subsubsection{Feature balancing, fusion and initial egopose estimation} \label{sec:fusion}

\begin{figure}[tb]
	\includegraphics[width=0.5\linewidth]{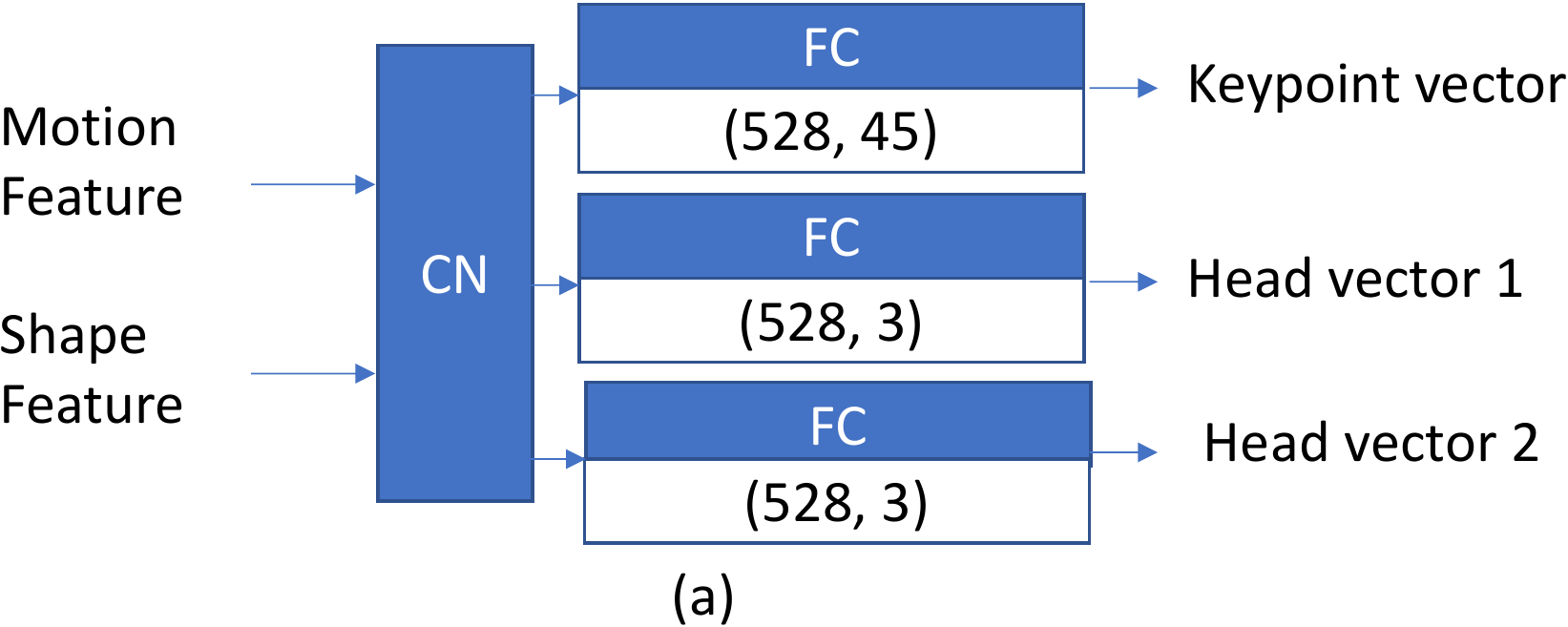} \hspace{10pt}
\rotatebox{90}{\includegraphics[width=0.25\linewidth]{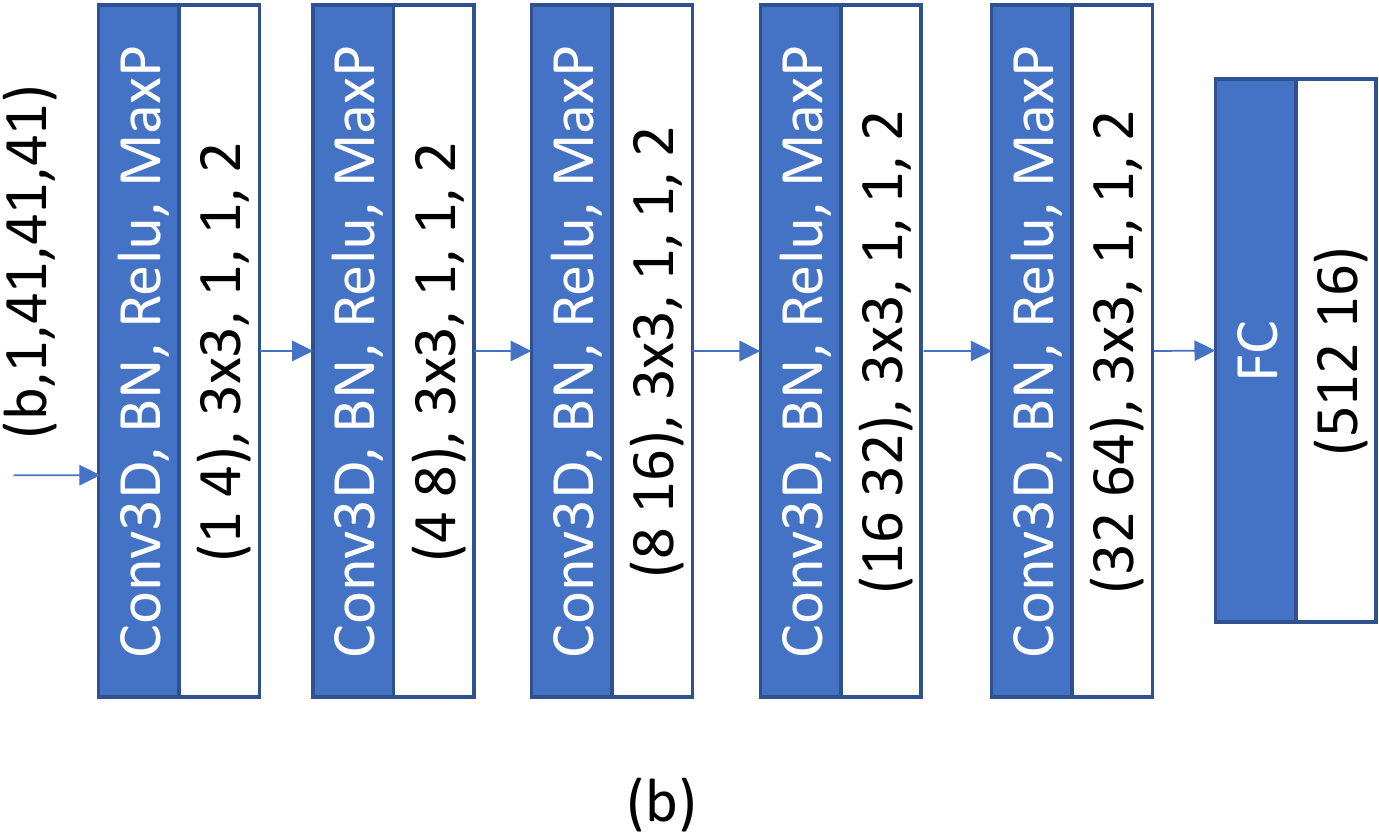}}
	\caption{(a) Fusion network, (b) 3D shape feature network.}
	\label{fig:fusion}
	\vspace{-8pt}
\end{figure}

We fuse the dynamic features from Section \ref{sec:motion} with shape features from Section \ref{sec:shapeest} for robust egopose estimation.
A simple strategy is to directly concatenate them and process the concatenation through a regression network.
Unfortunately, this leads to poor results, and it is important to balance the two sets of features.
To this end, we use a fully connected network, the balancer as shown in Fig.~\ref{fig:shapenet}(b), to reduce the dimensions of shape features and then do the concatenation, 
thereby implicitly balancing the weight between two features.
It turns out that the shape features can be quite low dimensional (e.g. $16$d), while the movement features are long (e.g., $512$d). 
With shorter input, there would be fewer neurons in the fully connected layer that are connected to it, 
and thus it has less voting power for the output. 
This scheme also has the effect of smoothing out the noisy shape observations.
Once these adjustments are done, the concatenated motion feature with the balanced shape feature are 
fed to three fully connected networks to infer the pose vector and the two head orientation vectors as shown in Fig.~\ref{fig:fusion}(a).

\subsection{Stage 2: 3D egopose refinement} \label{sec:stage2}

\begin{figure}[tb]
	\centering
	{
	 \fboxsep=0pt	
	 \fbox{
		\scalebox{-1}[1]{\includegraphics[width=0.06\textwidth]{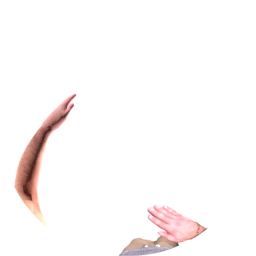}}}%
	\includegraphics[width=0.075\textwidth]{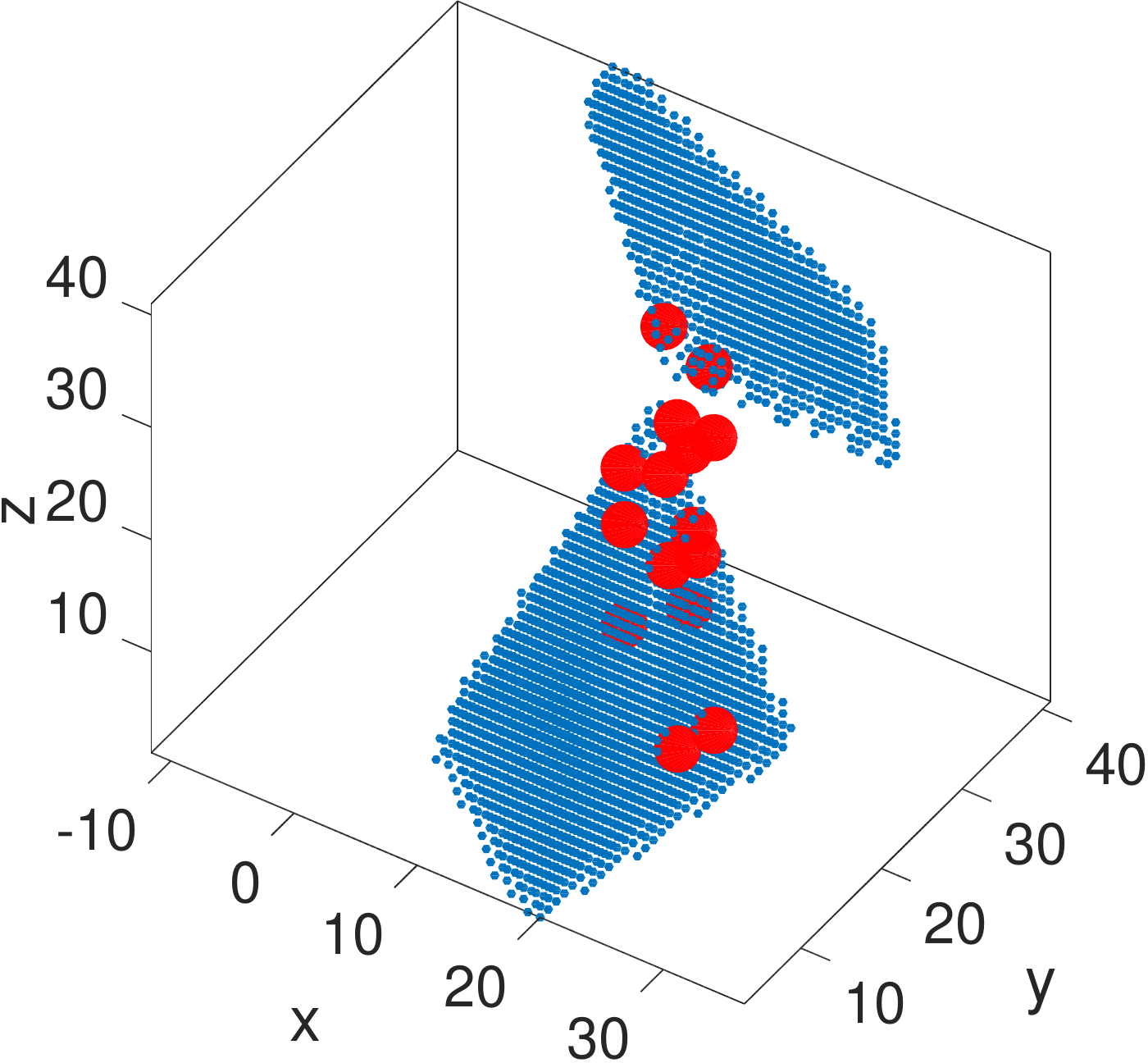}\hspace{5pt}%
	\fbox{\scalebox{-1}[1]{\includegraphics[width=0.06\textwidth]{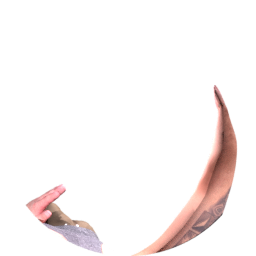}}}%
	\includegraphics[width=0.06\textwidth]{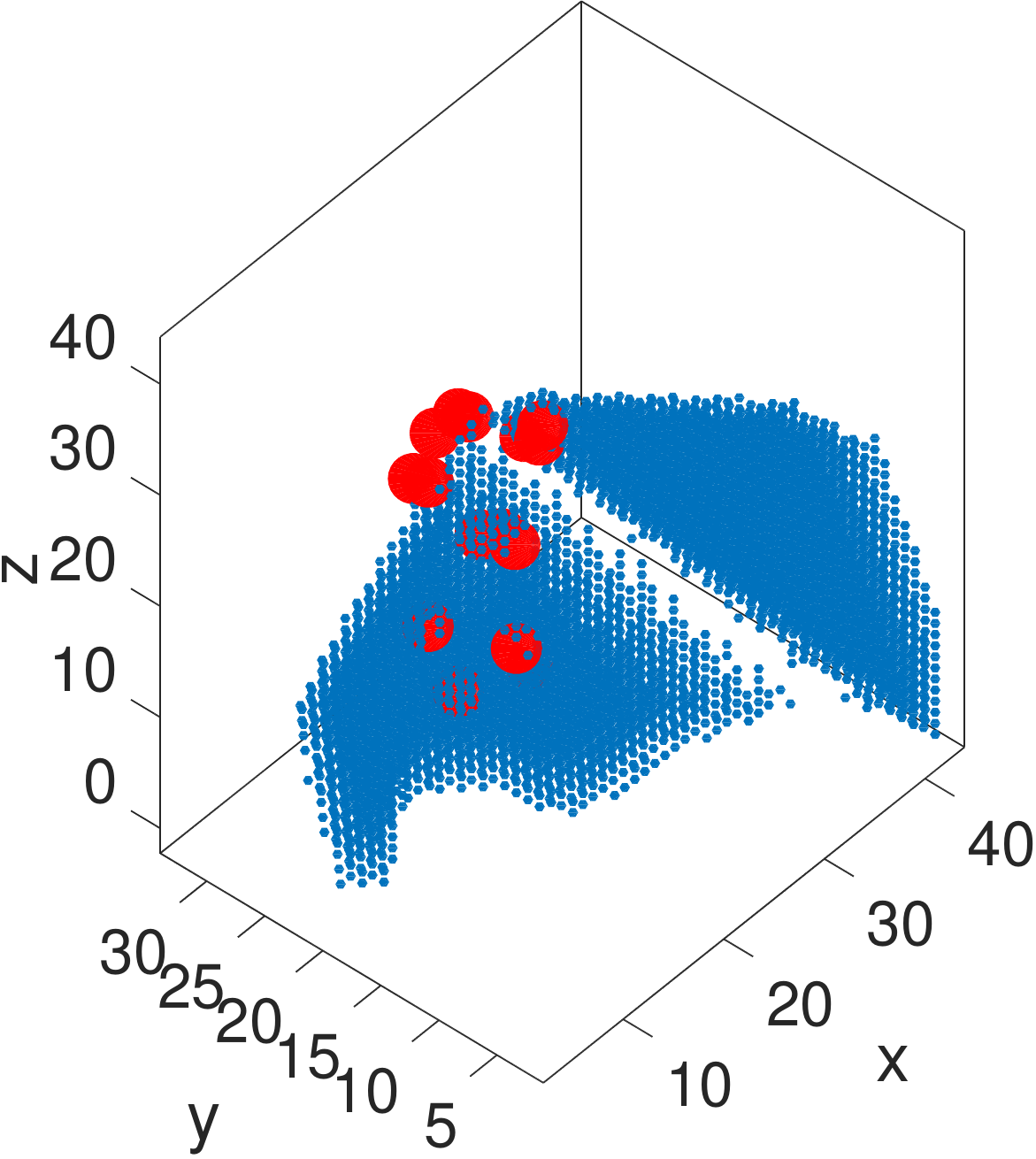}\hspace{5pt}%
	\fbox{\scalebox{-1}[1]{\includegraphics[width=0.06\textwidth]{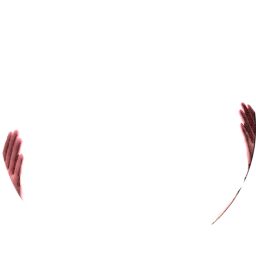}}}%
	\includegraphics[width=0.1\textwidth]{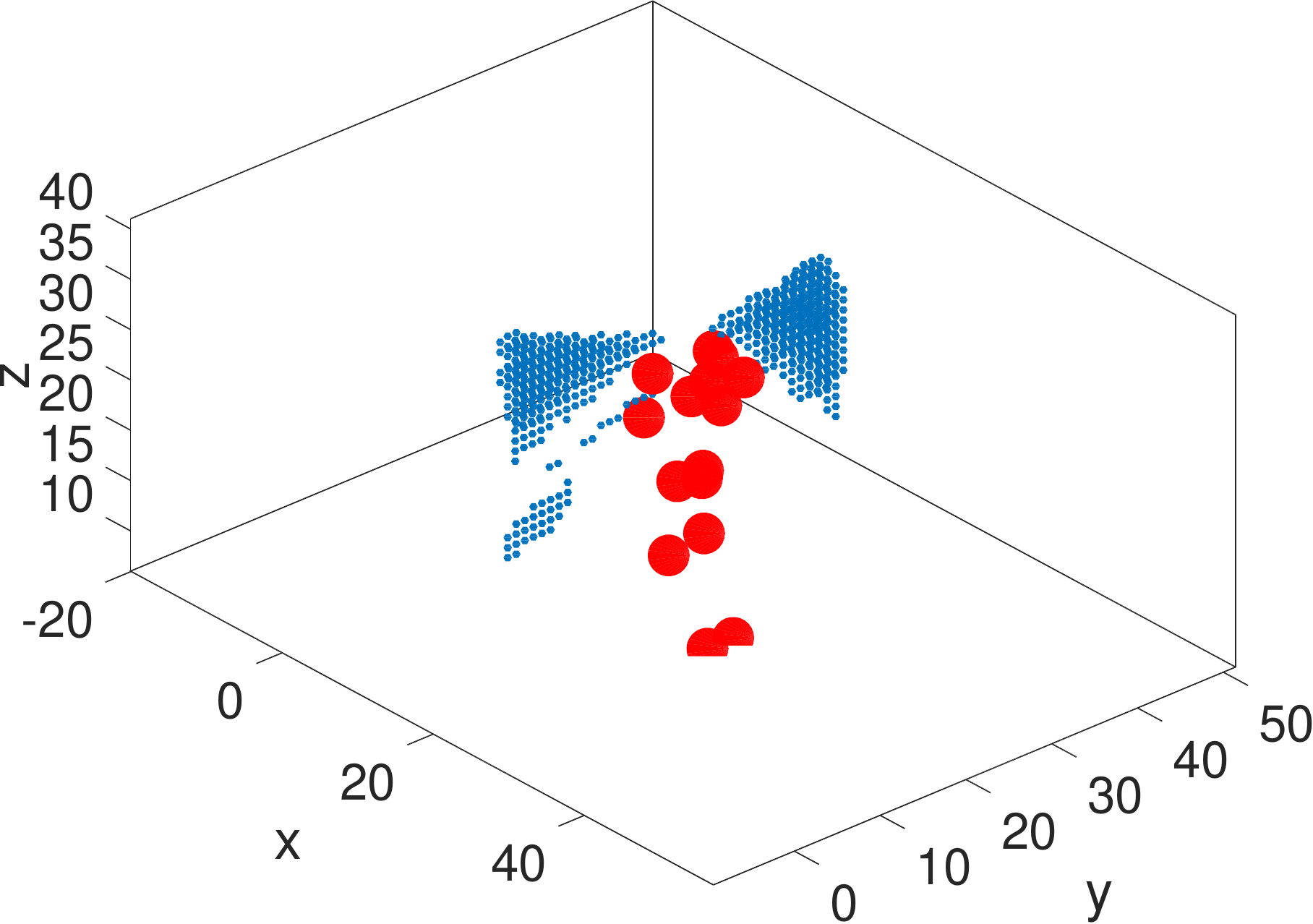}%
        }  
\caption{The pose volume representation. Odd columns: foreground image with foreground mask in the alpha channel. 
	Even columns: reconstructed pose volume (in blue) from Stage $1$ head pose and the foreground map estimation. 
	The read dots are the refined pose estimates from Stage $2$.}
	\label{fig:posevol}
	\vspace{-8pt}
\end{figure}

Given an estimate of the egopose, we further refine it by fixing the head pose estimation from Stage $1$, and re-estimating full body 3D pose.
Using the head/camera pose and foreground shape estimations from Stage $1$, 
we construct a 3D volume by back-projecting the foreground pixels in a 2-cubic-meter volume, as shown in Fig.~\ref{fig:posevol}.
The volume is descritized into a $41\times41\times41$ 3D matrix.
We assign value $1$ if a voxel projects to the wearer foreground and $0$ otherwise.
This volume thus explicitly represents a 3D
body shape envelop given the current head pose and body shape estimations.
We then pass the 3D shape representation to a 3D CNN, shown in Fig.~\ref{fig:fusion}(b), for feature extraction. 
The resulting features are flattened and concatenated with the motion feature, the initial 3D pose estimation, 
and then fed to a fully connected network for 3D body pose estimation. 
This refinement regression network has similar structure to the fusion network in Fig.~\ref{fig:fusion}(a) 
where the input now also includes the initial 3D keypoint estimation and the output is body pose estimation {\it alone}.
In Fig.~\ref{fig:posevol}, we overlay the refined 3D poses in the volume.
With this explicit 3D representation that directly captures the 3D geometry, we are able to achieve better body pose estimation.

\subsection{Model training and loss function} \label{sec:lossfunc}

We first train Stage $1$ (refer to Fig.~\ref{fig:sysarch}), and depending on the estimation on training data results, we subsequently train Stage $2$. 
We use the L1 norm to quantify the errors in body keypoints and head orientation estimations.
\begin{equation} \label{eq:ld}
L_d = |\mathbf{b}-\mathbf{b}_g| + |\mathbf{h}-\mathbf{h}_g|
\end{equation}	
where $\mathbf{b}$ and $\mathbf{b}_g$ are the flattened body keypoint 3D coordinates and their ground truth, 
$\mathbf{h}$ is the head orientation vector (concatenation of the vectors $\mathbf{f}$ and $\mathbf{u}$), 
and $\mathbf{h}_g$ is its corresponding ground truth.
To improve the generalization, we further include several regularization terms that constrain the structure of the regression results.
The two head orientation vectors are orthonormal, and so, we minimize
\begin{equation} \label{eq:lo}
	L_o = |\mathbf{f} \cdot \mathbf{u}| + | \hspace{1pt}||\mathbf{f}||^2-1| + |\hspace{1pt}||\mathbf{u}||^2-1|
\end{equation}
where $\cdot$ is the inner product of two vectors and $||.||$ is the L2 norm. We also enforce the body length symmetry constraints. 
Let $l^{(i)}$ and $l^{(j)}$ be a pair of symmetrical bone lengths and the set of the symmetrical bones is $\mathcal{P}$. We minimize
\begin{equation} \label{eq:ls}
	L_s = \sum_{(i,j)\in \mathcal{P}}|l^{(i)} -  l^{(j)}|
\end{equation}
%
We also enforce the consistency of the head pose, body pose and body shape maps.
From the head pose, we compute the camera local coordinate system. 
With the equidistant fisheye camera model, let $(x_k,y_k), k=1..K$ be the 2D projections of the 3D body keypoints. We minimize 
\begin{equation} \label{eq:lc}
	L_c = \sum_{k=1}^K [\min(D(y_k, x_k) - q, 0) + q]
\end{equation} 
where $D$ is the distance transform of the binary body shape map and $q$ is a truncation threshold e.g. 20 pixels.
With $\alpha$, $\beta$ set to $0.01$ and $\gamma$ to $0.001$, the final loss function is 
\begin{equation}
L = L_d + \alpha L_o + \beta L_s + \gamma L_c
\end{equation}
Note that for Stage $2$, the head vector related terms are removed from the loss.
   
\subsection{Leveraging synthetic data} \label{sec:syn}

\begin{figure}[tbh]
	\centering
 \includegraphics[width=0.09\textwidth, height=0.12\textwidth]{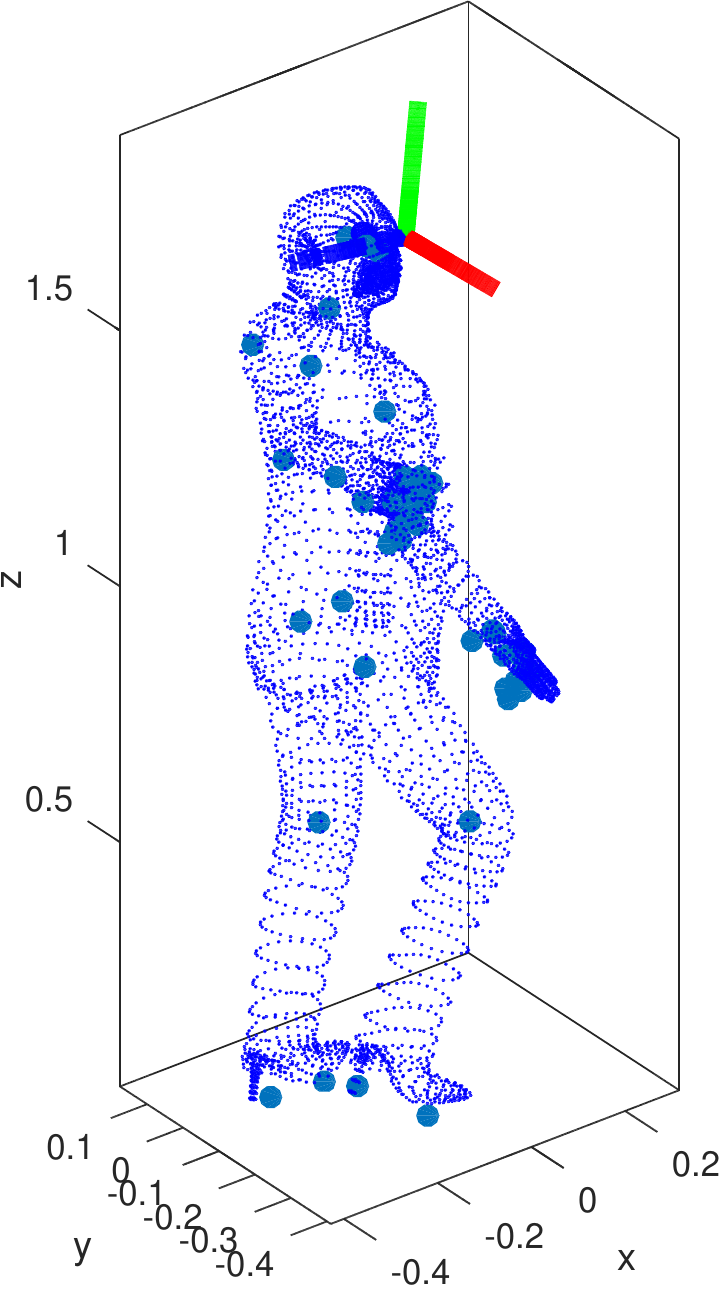}%
 \includegraphics[width=0.09\textwidth, height=0.12\textwidth]{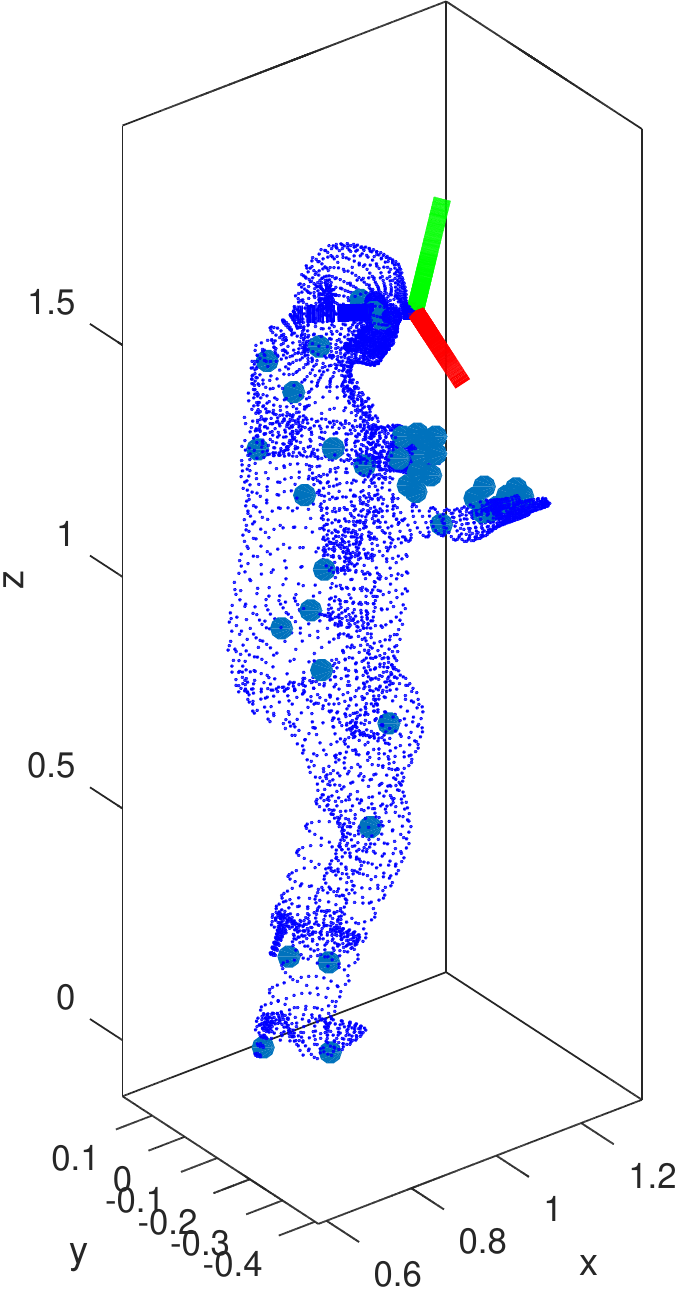}%
 \includegraphics[width=0.09\textwidth, height=0.12\textwidth]{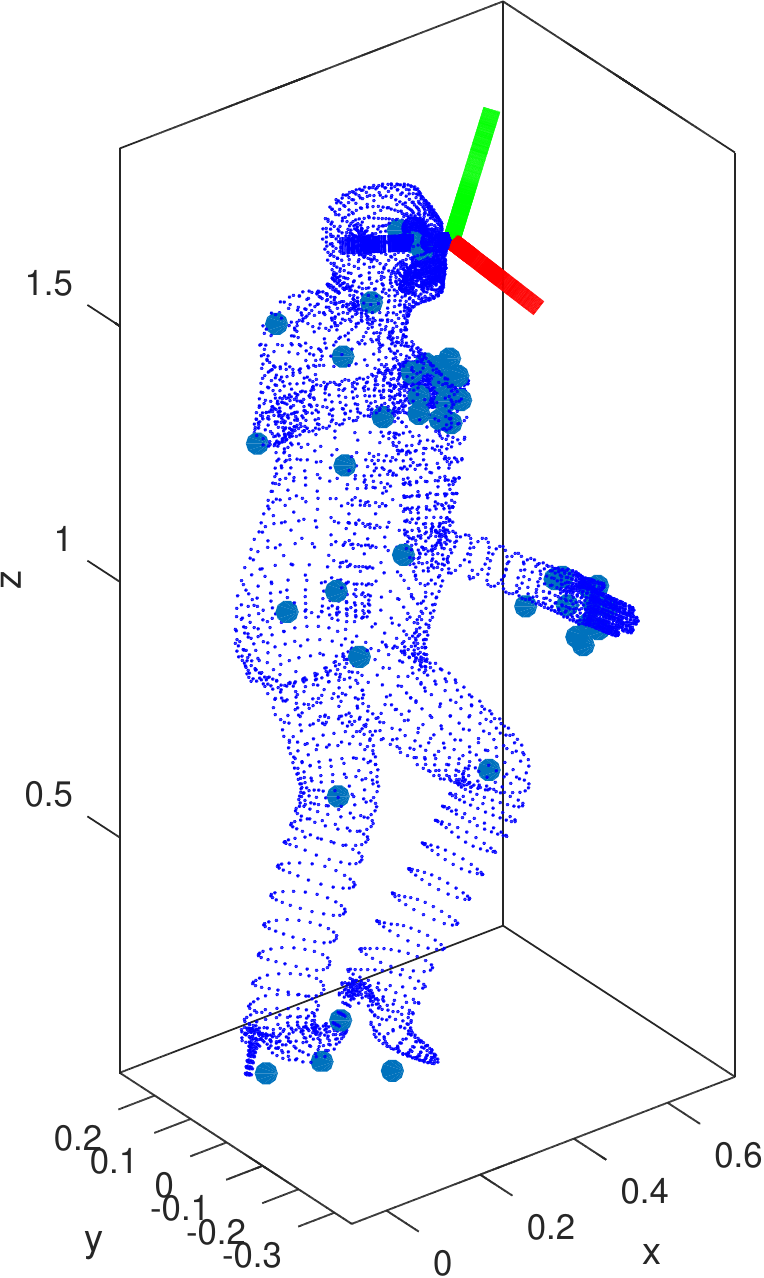}%
 \includegraphics[width=0.09\textwidth, height=0.12\textwidth]{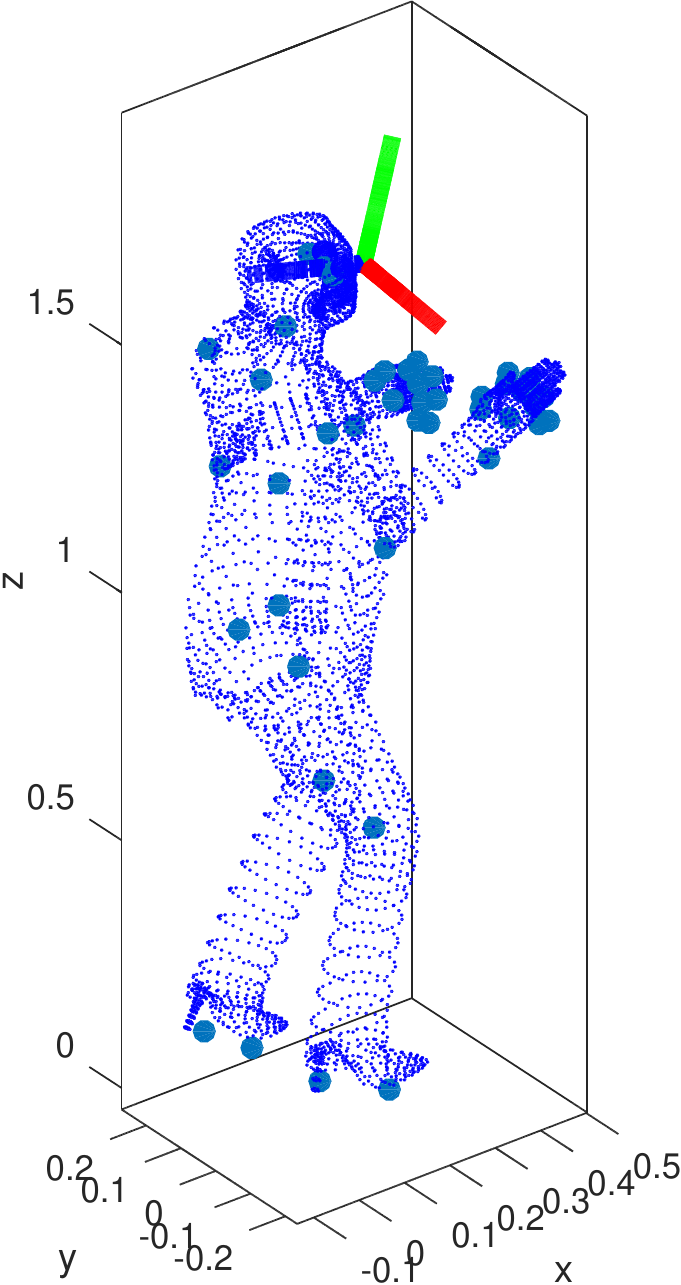}%
 \includegraphics[width=0.09\textwidth, height=0.12\textwidth]{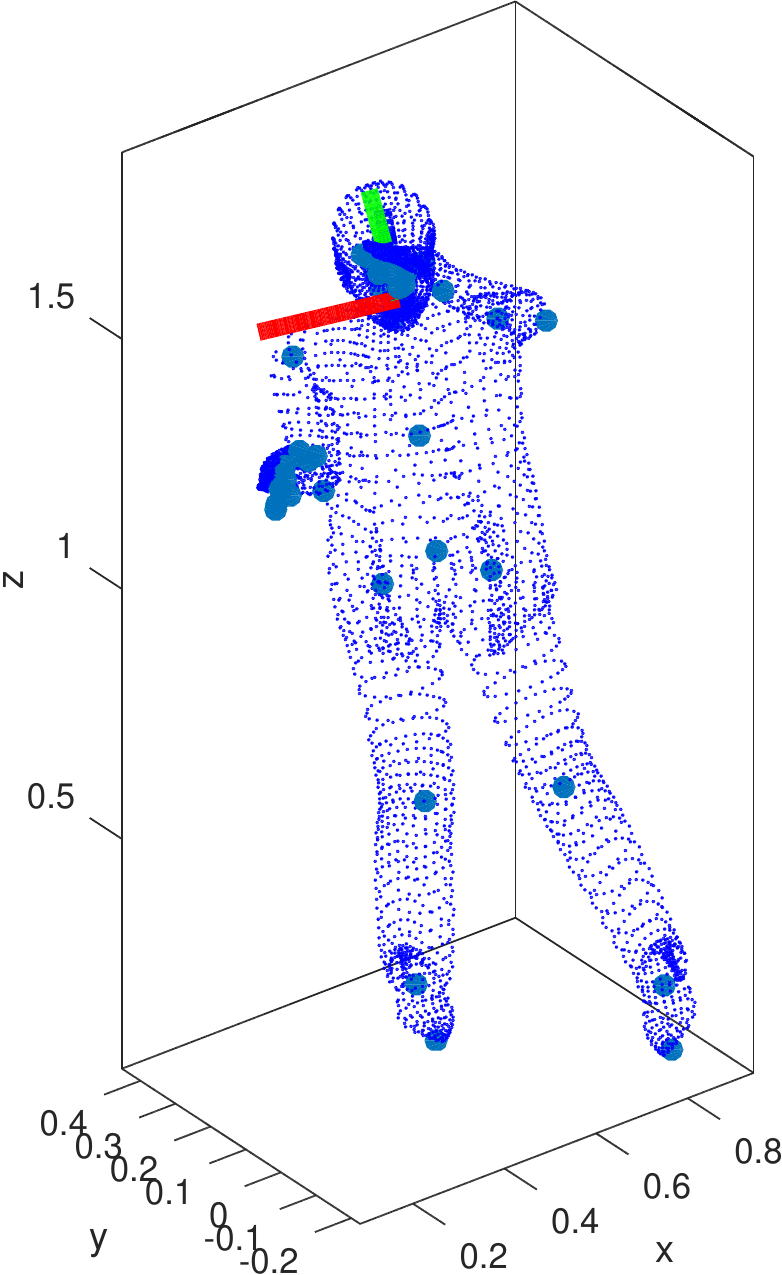}%
\linebreak
	\framebox{\includegraphics[width=0.0783\textwidth]{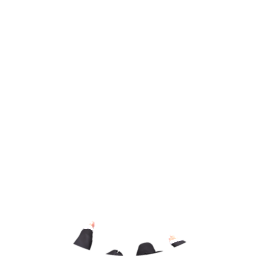}}%
	\framebox{\includegraphics[width=0.0783\textwidth]{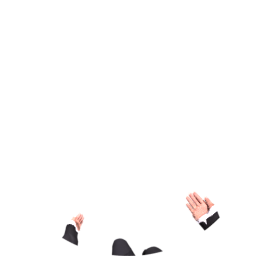}}%
	\framebox{\includegraphics[width=0.0783\textwidth]{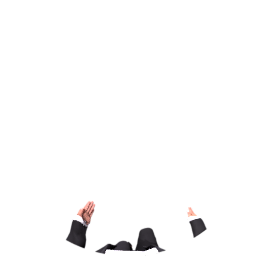}}%
	\framebox{\includegraphics[width=0.0783\textwidth]{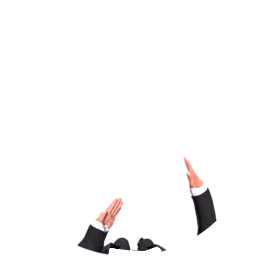}}%
	\framebox{\includegraphics[width=0.0783\textwidth]{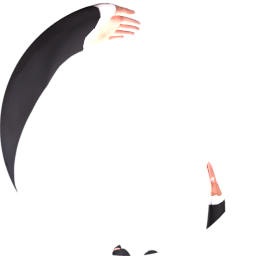}}%
\caption{Synthetic data. Row 1: synthetic person meshes, keypoints, head orientation. 
Camera's local coordinate system is shown as $3$ lines red ($x$), green ($y$) and blue ($z$).
	Row 2: rendered person image in the head-mounted camera's virtual FOV. 
	The alpha channel of the image gives the person foreground mask. 
	}
	\label{fig:syn}
	\vspace{-8pt}
\end{figure}

It is challenging to capture a large set of synchronized head-mounted camera video and the corresponding `matched' body mocap data. 
Hence, we leverage a total of $2548$ CMU mocap sequences \cite{cmu-mocap} and Blender \cite{blender}, to generate synthetic training data. 
These sequences involve a few hundred different subjects, and the total length is approximately $10$hrs.
For each mocap sequence, we randomly choose a person mesh from $190$ different mesh models to generate the synthetic data.
An example synthetic person is shown in Fig.~\ref{fig:syn}, which illustrates
the body keypoints, body mesh, three axes of the camera local coordinate 
system, and the rendered person with alpha channel in the camera's view.

The data synthesis process follows these steps. We first re-target skeletons in mocap data to person mesh models to generate animations. 
We then rigidly attach a virtual front facing fisheye camera between two eyes of each person model. 
A motion history map is then computed using the virtual camera pose and position history in the animations. 
Using this camera setup, we render the camera view with an equidistant fisheye model.
The rendered image's alpha channel then gives the person's foreground mask. 
Note that, in our setting, the camera's $-z$ and $y$ axes are aligned with the two head orientation vectors. 
Overall, this provides high quality data for boosting training as well as validating the proposed egopose deep models.
Lastly, since this synthesized data are invariant to the scene and wearer's appearances, they can be easily generalized to real videos.


\section{Experiments} \label{sec:exp}

We evaluate the proposed method on both synthetic and real video data and compare it with some existing approaches.
Note that, as motivated in Section \ref{sec:intro}, our task is novel. Hence, there are no previous methods that serve as appropriate baselines.
We therefore use the state-of-the-art egopose methods with our inputs, 
and we also evaluate and compare different variations of the proposed method to justify our design choices.
The baselines include:
\begin{itemize}
		\vspace{-6pt}	
	\item \texttt{xr-egopose} \cite{fb1}: Designed to estimate wearer's pose using head-mounted downward fisheye cameras. 
	It extracts the 2D and 3D body keypoints at the same time. 
	This method needs to see the camera wearer to estimate the egopose. 
		\vspace{-8pt}
	\item \texttt{pd-egopose} \cite{cmu2}: Uses deep learning and an explicit control mechanism for egopose estimation.
	It uses an optical flow as the input and does not need to see the camera wearer in the FOV. 
		\vspace{-8pt}
	\item \texttt{MotionOnly, ShapeOnly, Stage1Only, NoHeight, Stage1RNN, HandMap}: 
	Multiple variations of the proposed system -- 
	using only Stage $1$ network and motion history image as input, using only the body shape, bypassing Stage $2$ pose refinement, 
	bypassing the height information in the motion history image, 
	using Stage $1$ network with an RNN structure instead, and
	using hand keypoint map instead of the body shape as the input.
	These baselines validate the necessity of each of the proposed components in the overall system architecture from Fig. \ref{fig:sysarch}.
		\vspace{-8pt}
	\item \texttt{AllStand} and \texttt{AllSit}: Two special cases that always give a standard standing pose or sitting pose.	
\end{itemize}
\vspace{-5pt}
We use the body and head pose estimation errors to quantify the egopose estimation accuracy.
The body pose estimation error is the average Euclidean distance between the estimated 3D keypoints 
and the ground truth keypoints in the normalized coordinate system.
During training and testing, the ground truth 3D body poses are normalized to have a body height around 170 centimeters.
The head pose estimation error is quantified by the angles between the two estimated head orientations and the ground truth directions.

\subsection{Tests on synthetic data} \label{sec:syneval}

Recall the synthetic dataset setup from section \ref{sec:syn}. 
Among the full $2548$ synthetic sequences, we randomly pick $180$ of the sequences for training and another $60$ sequences for testing.
Such a setting is to reduce the chance that two mocap sequences share the same subject.
The body shape model is trained by pasting rendered foreground images on random background images from the ADE20K \cite{ade20k} dataset.
Note that the motion feature is obtained from the motion of the virtual camera on the virtual person's head.
For \texttt{xr-egopose}, we use the ground truth keypoint 2D heat map to replace the first stage fully convolutional network's output. 
We thus reasonably assume the initial keypoint heat map estimation network in \texttt{xr-egopose} can give a perfect result.
Our proposed method on the other hand uses the inferred foreground shape as the input.
As can be seen from the estimation result in Table.~\ref{tab:syn}, 
even though we give \texttt{xr-egopose} an advantage in the form of generally superior inputs, our method still improves the accuracy by $13\%$. 
This is not surprising because of the setting we are operating with: 
\texttt{xr-egopose} depends on the visible body parts which are now in general absent in a human vision span. 
The trends are similar with regard to \texttt{pd-egopose}. 
Note that the original \texttt{pd-egopose} requires the optical flow image as the input. 
For this synthetic experiment, we let the method take the enlarged motion history image as the surrogate for the optical flow. 
For evaluations on real data, as we will discuss in Section \ref{sec:realeval}, the optical flow images are used instead of these surrogates.
The outputs from \texttt{pd-egopose} are normalized for valid comparison. 
As shown in Table~\ref{tab:syn}, the proposed method gives much better result when estimating the body keypoints. 

\begin{figure}[tb]
	\centering
	\scalebox{0.95}{	

\setlength\tabcolsep{1pt}	
\begin{tabularx}{\linewidth}{c X }	
	\rotatebox{90}{\hspace{4pt}{\scriptsize Ground Truth}}	& \includegraphics[width=0.125\linewidth, height=0.2\linewidth]{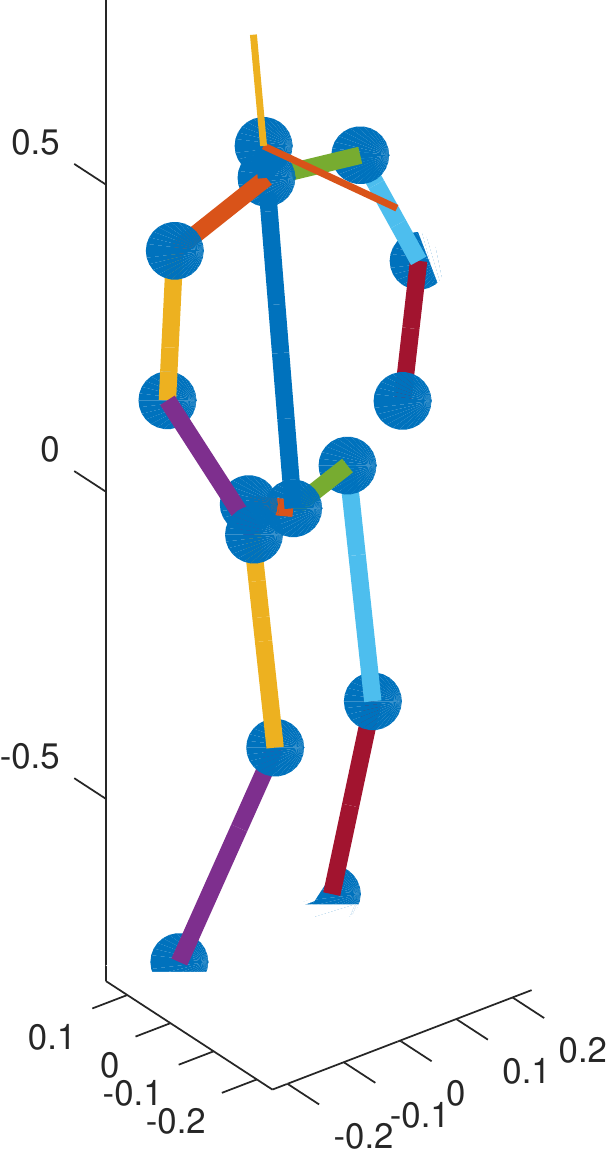}%
\includegraphics[width=0.125\linewidth, height=0.2\linewidth]{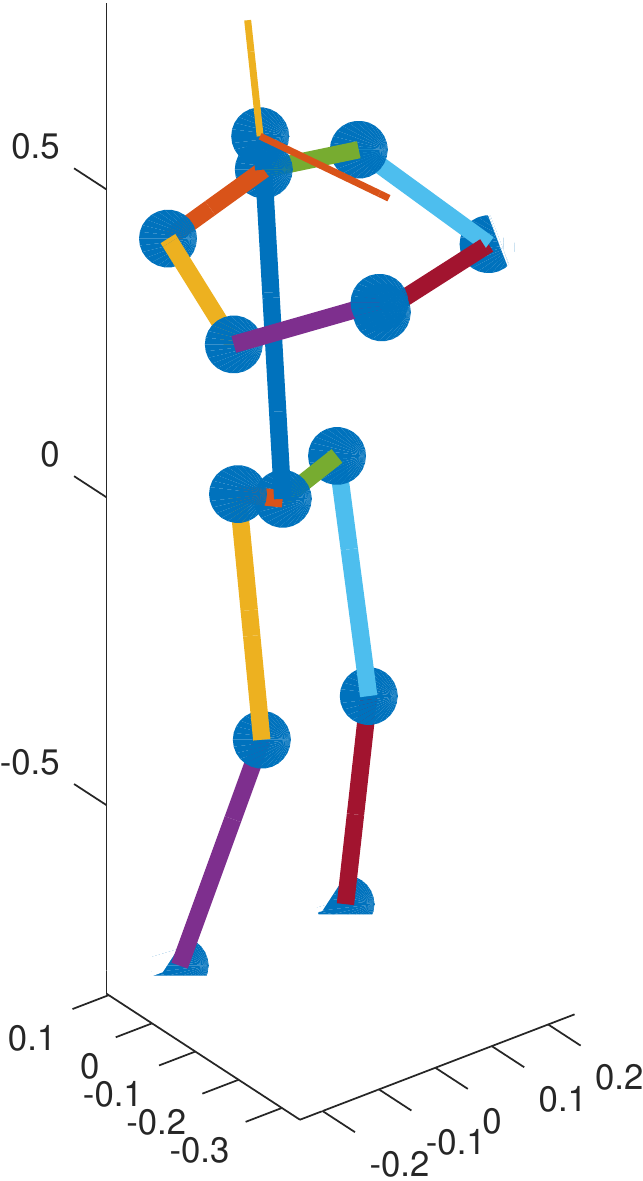}%
\includegraphics[width=0.125\linewidth, height=0.2\linewidth]{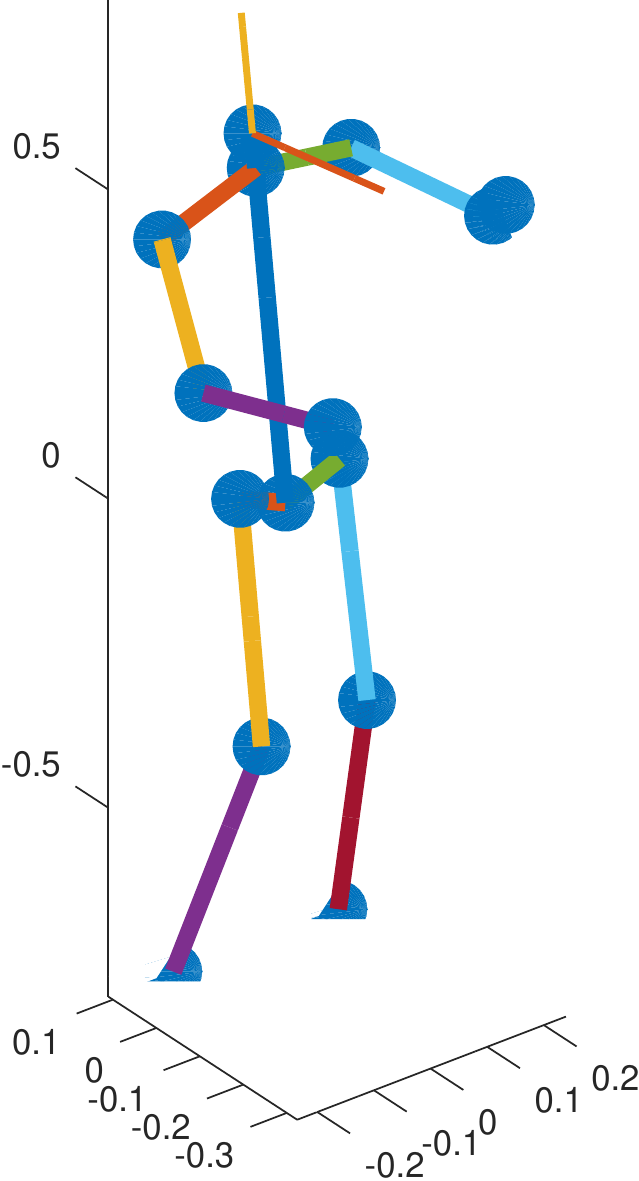}%
\includegraphics[width=0.125\linewidth, height=0.2\linewidth]{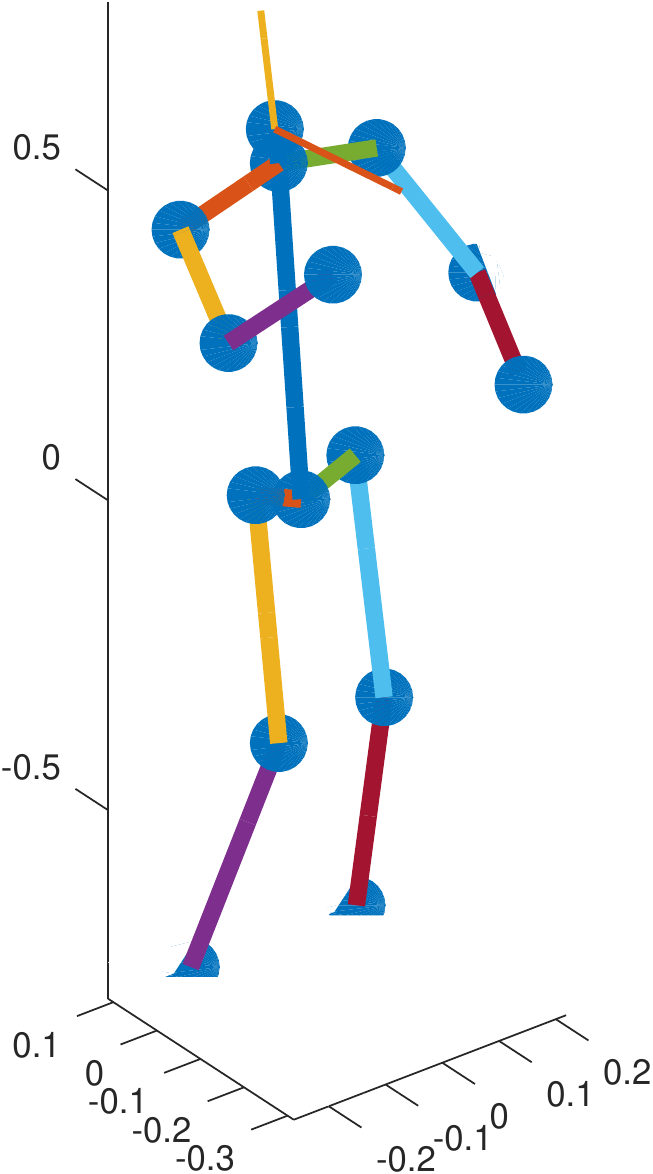}%
\includegraphics[width=0.125\linewidth, height=0.2\linewidth]{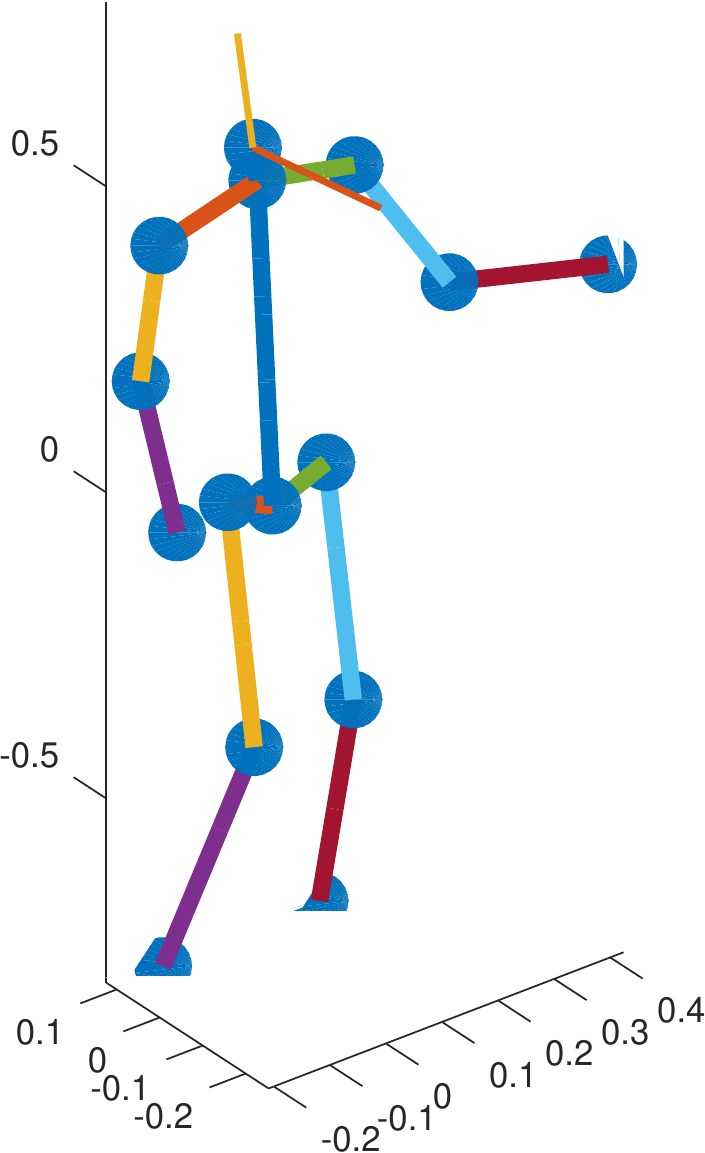}%
\includegraphics[width=0.125\linewidth, height=0.2\linewidth]{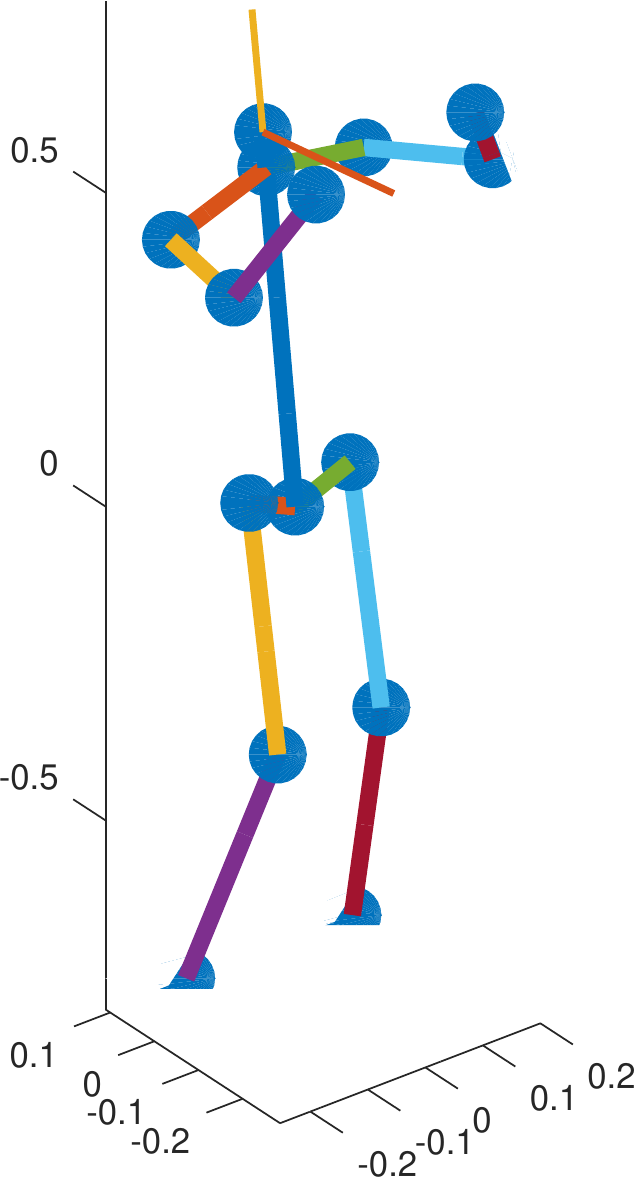}%
\includegraphics[width=0.125\linewidth, height=0.2\linewidth]{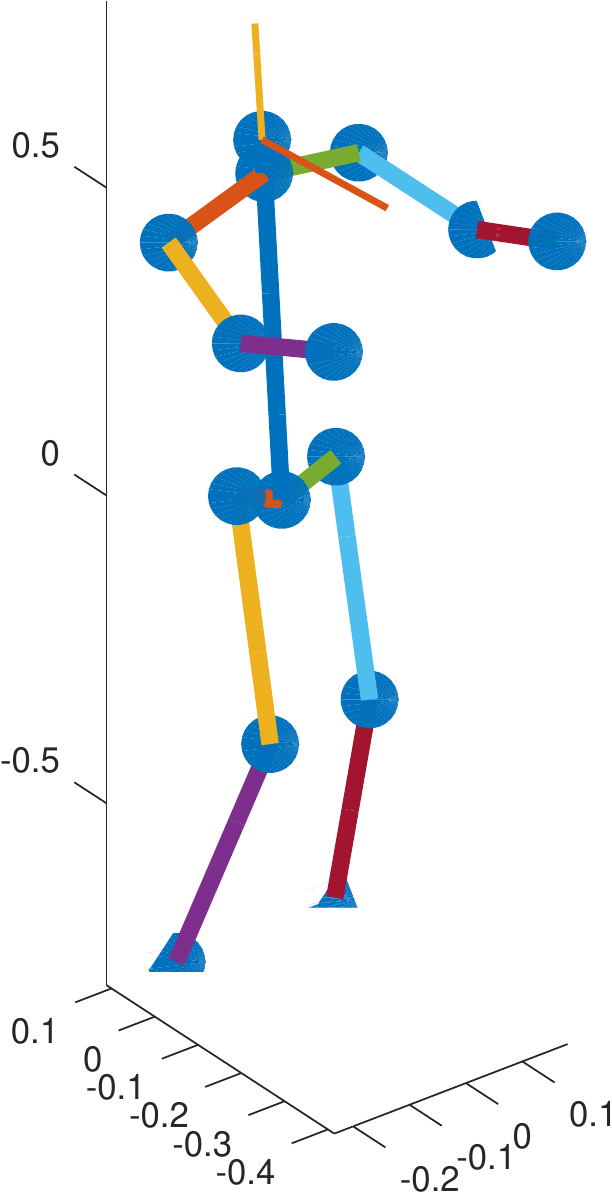}%
\includegraphics[width=0.125\linewidth, height=0.2\linewidth]{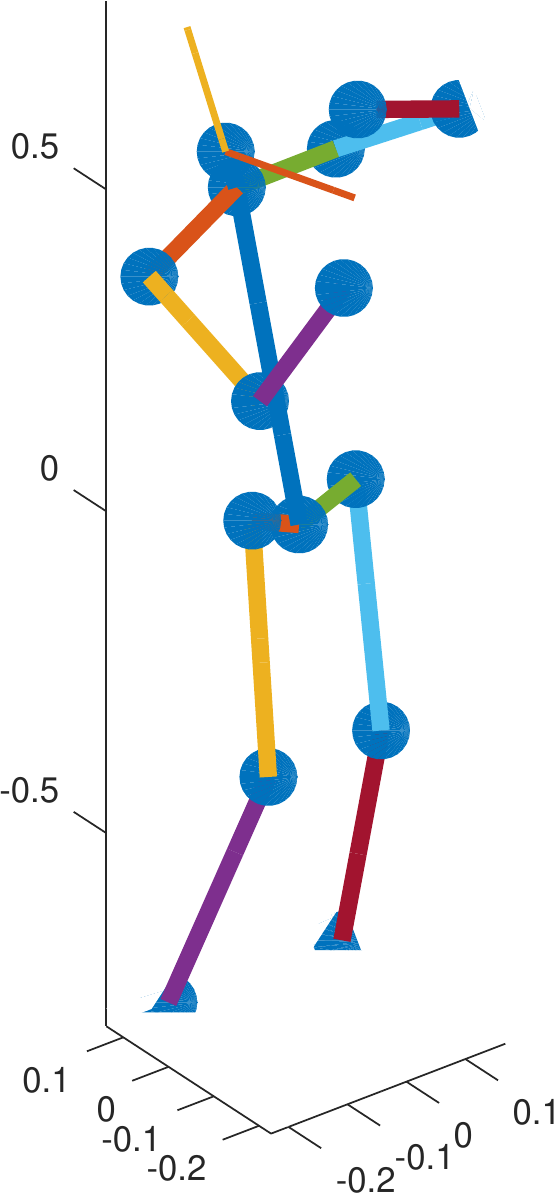}\\
 \rotatebox{90}{\hspace{10pt}{\scriptsize Full Model}} & \includegraphics[width=0.125\linewidth, height=0.2\linewidth]{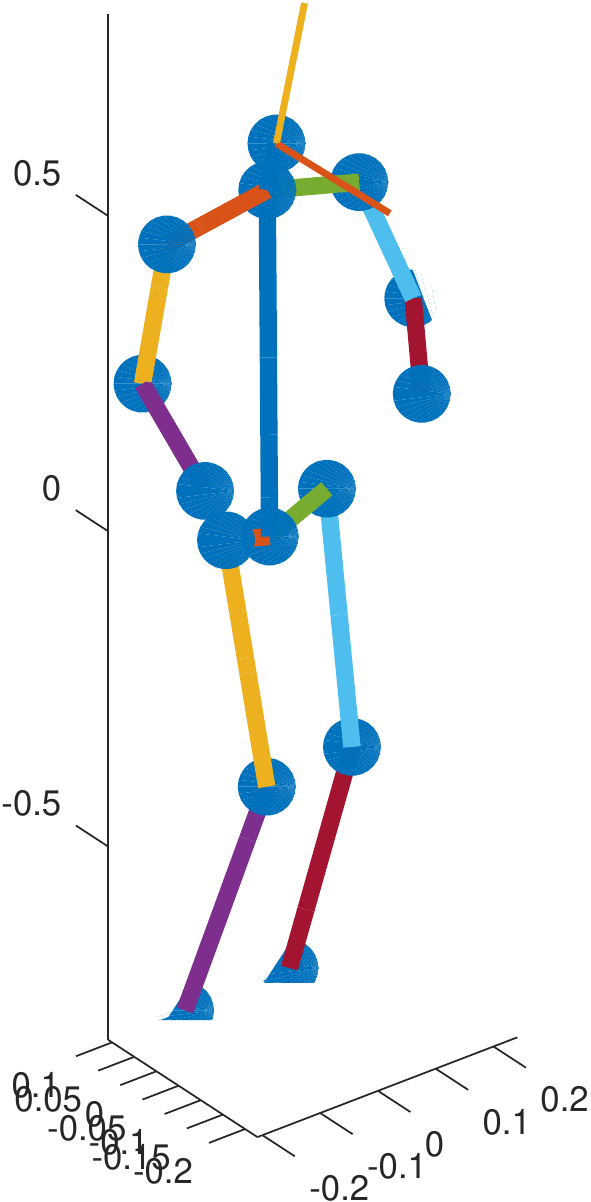}%
\includegraphics[width=0.125\linewidth, height=0.2\linewidth]{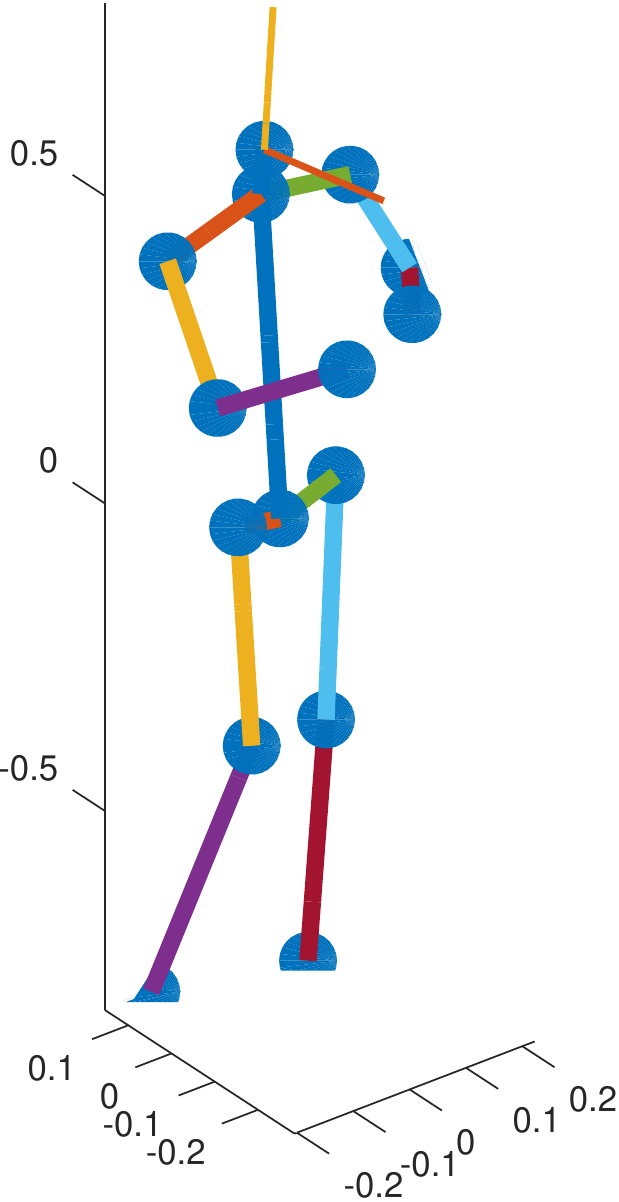}%
\includegraphics[width=0.125\linewidth, height=0.2\linewidth]{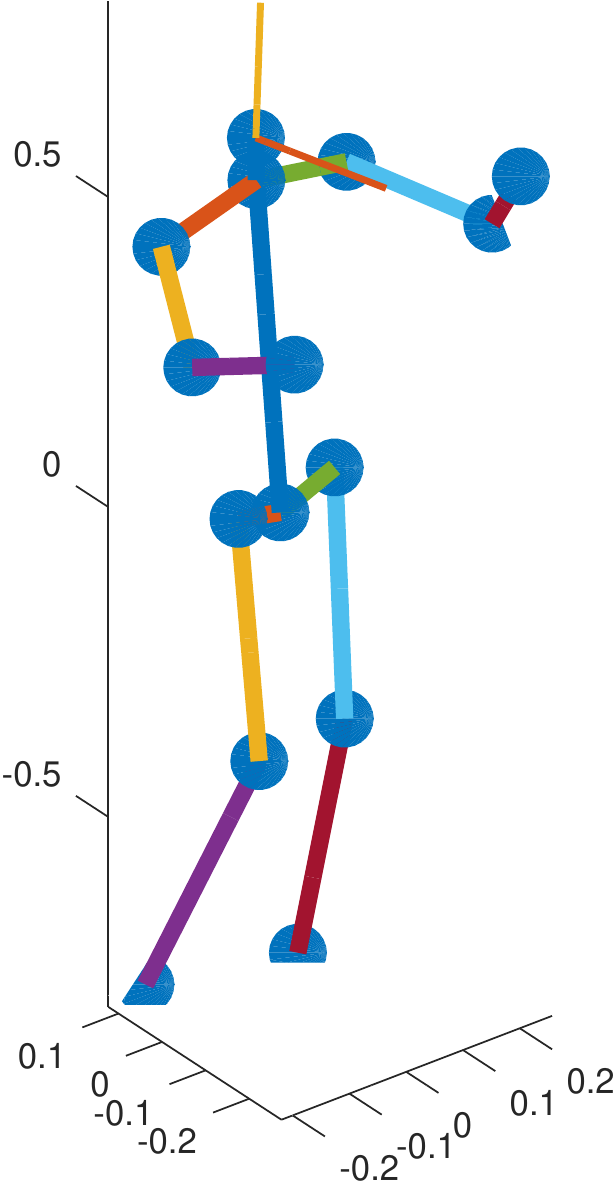}%
\includegraphics[width=0.125\linewidth, height=0.2\linewidth]{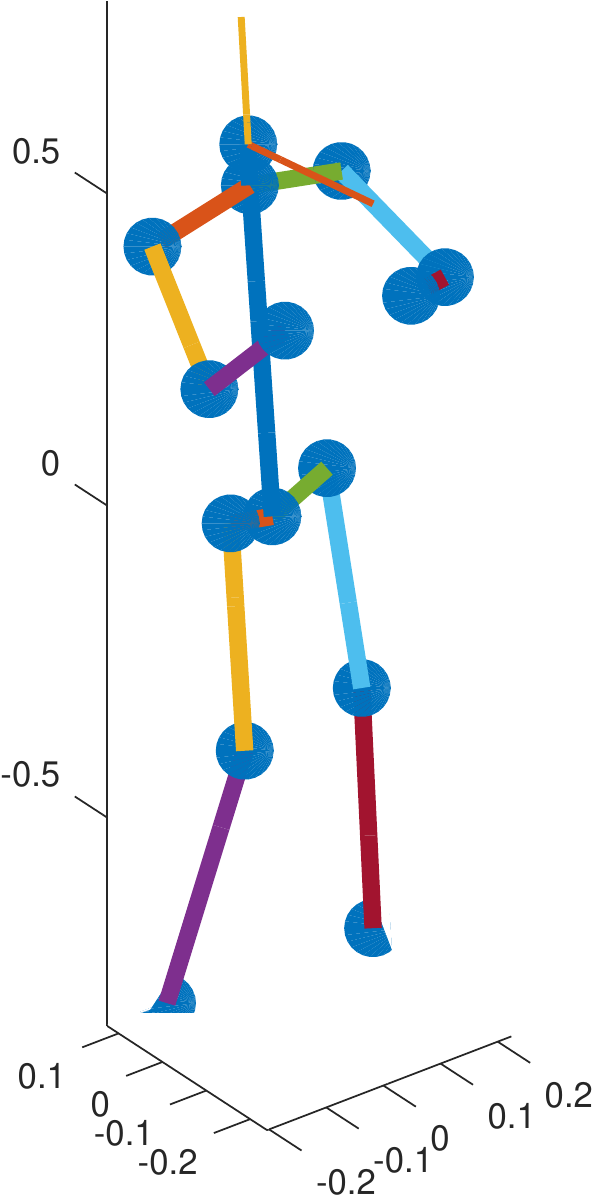}%
\includegraphics[width=0.125\linewidth, height=0.2\linewidth]{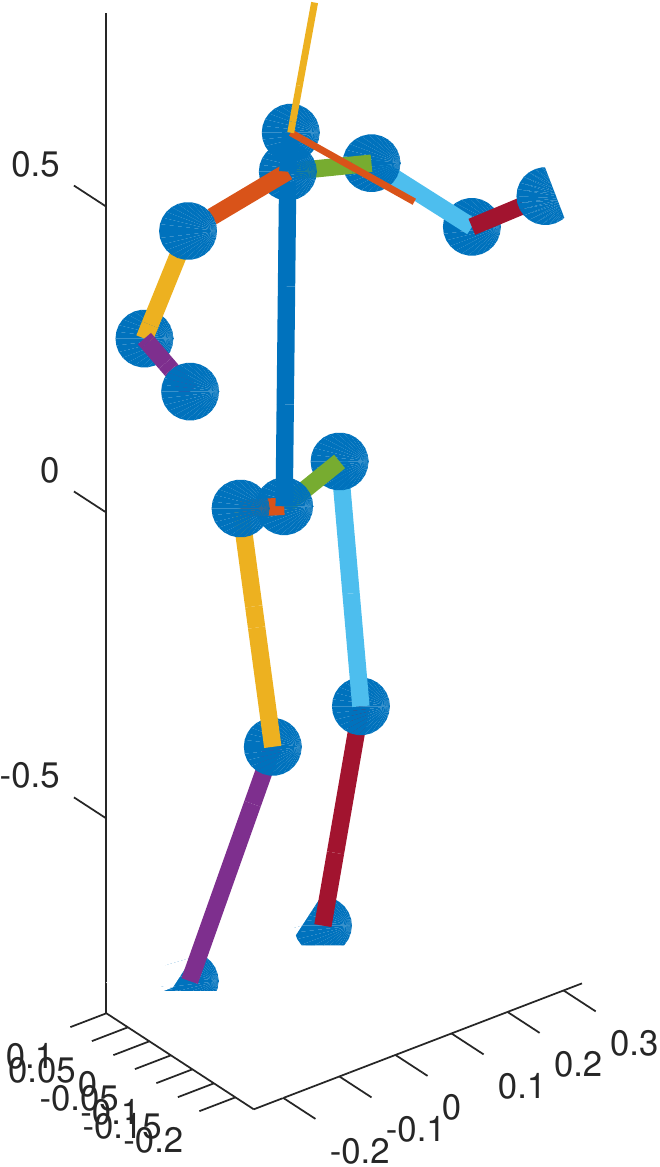}%
\includegraphics[width=0.125\linewidth, height=0.2\linewidth]{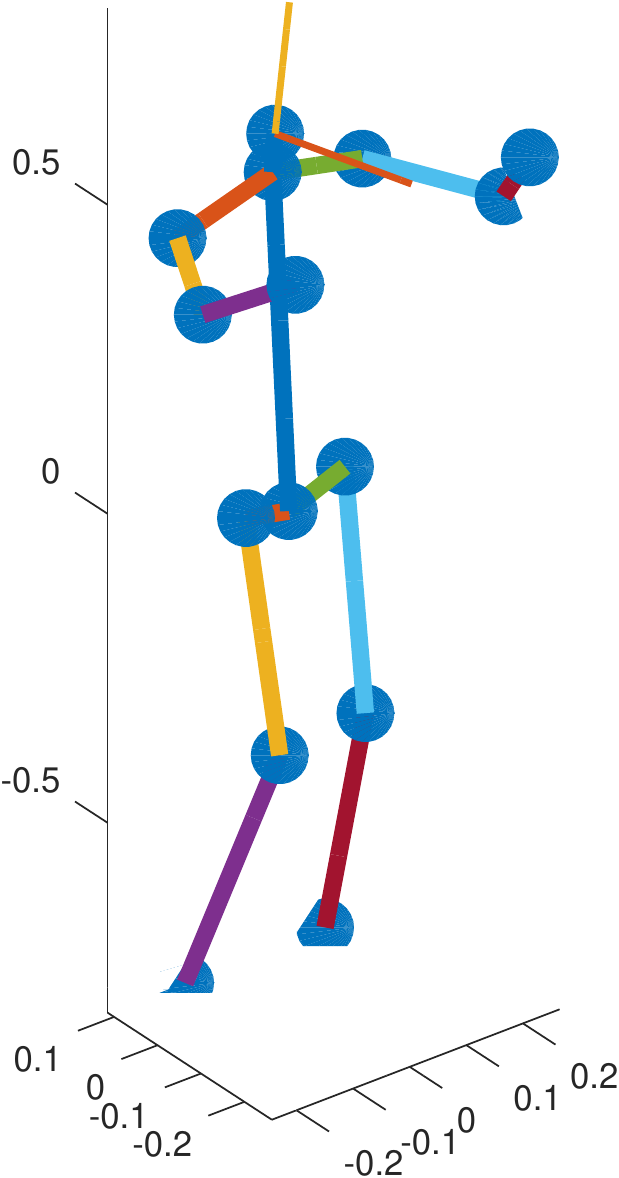}%
\includegraphics[width=0.125\linewidth, height=0.2\linewidth]{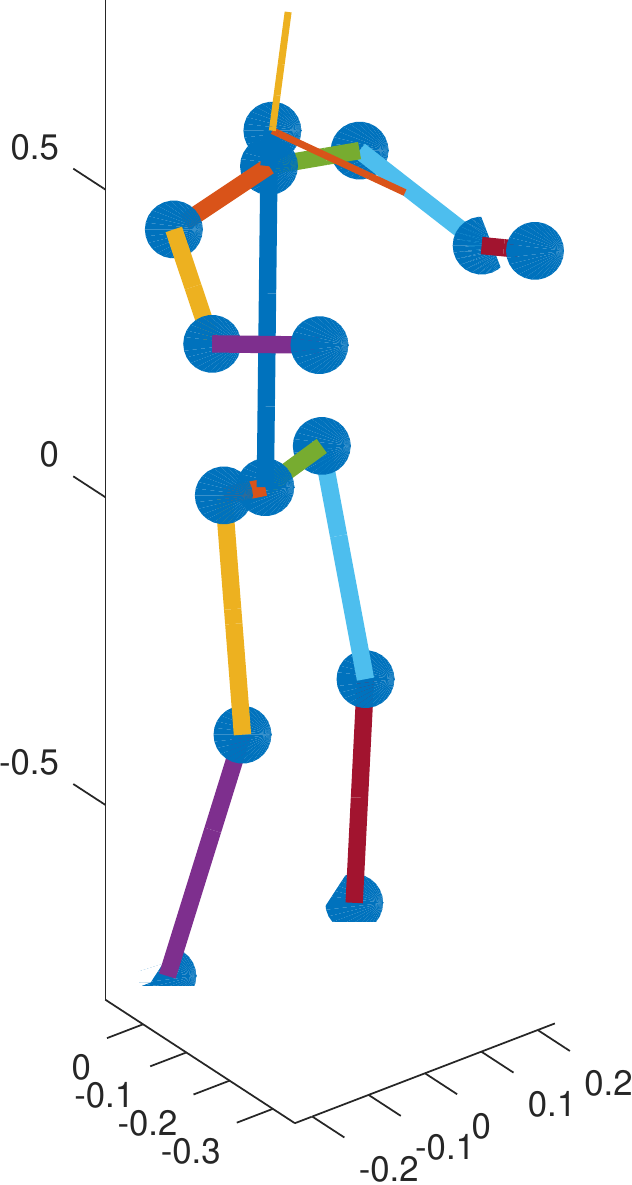}%
\includegraphics[width=0.125\linewidth, height=0.2\linewidth]{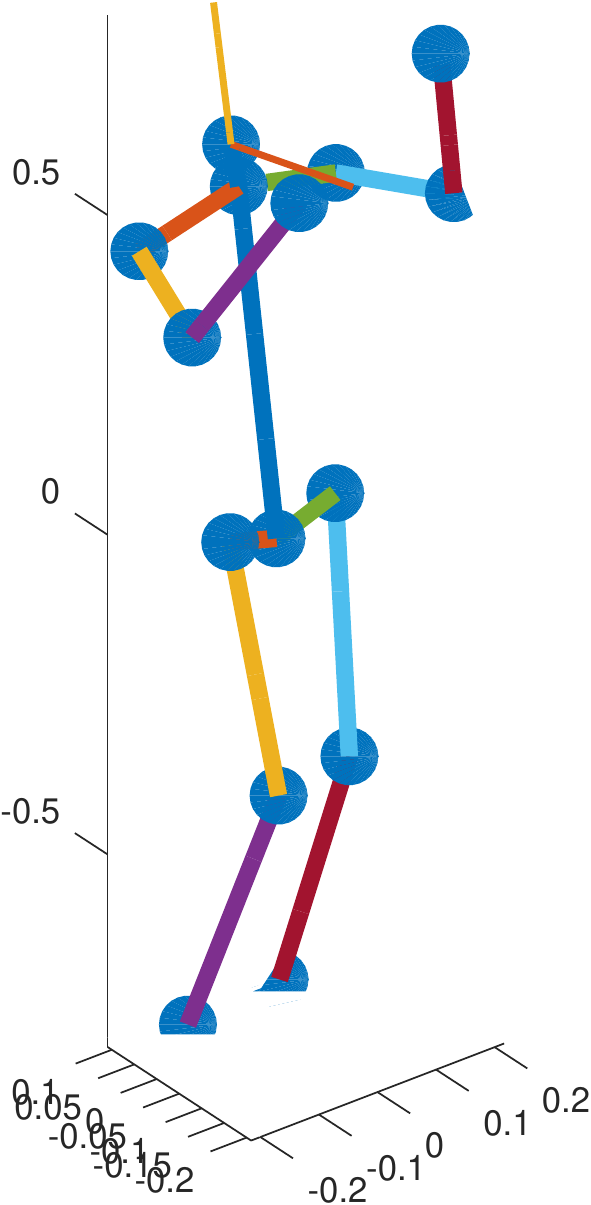}\\
 \rotatebox{90}{\hspace{6pt}{\scriptsize MotionOnly}}	 & \includegraphics[width=0.125\linewidth, height=0.2\linewidth]{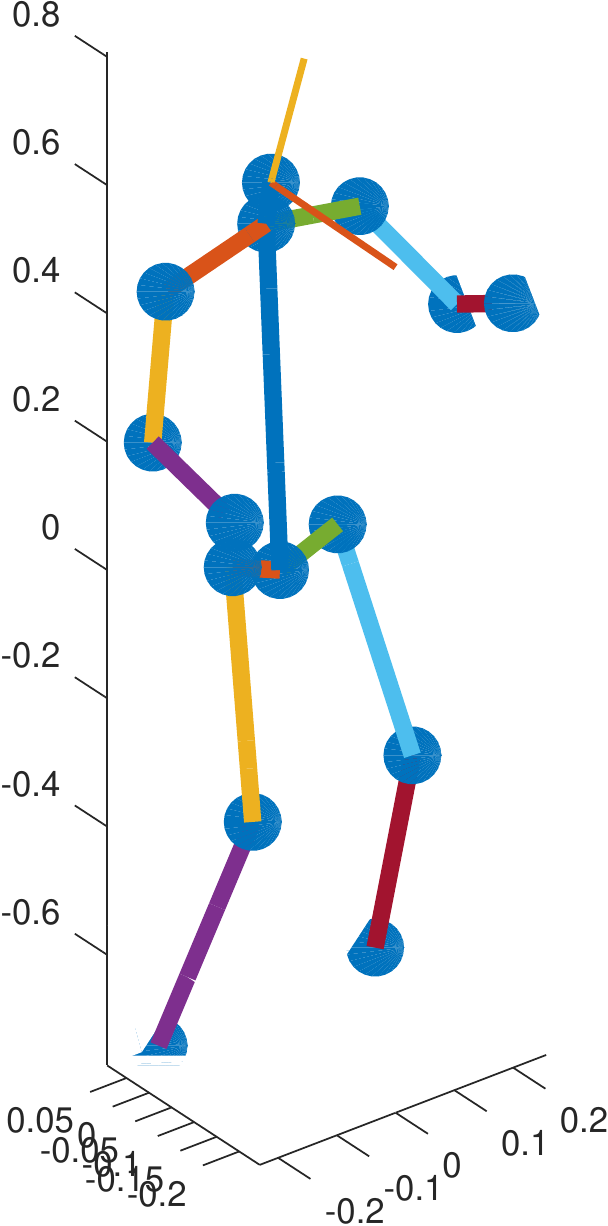}%
\includegraphics[width=0.125\linewidth, height=0.2\linewidth]{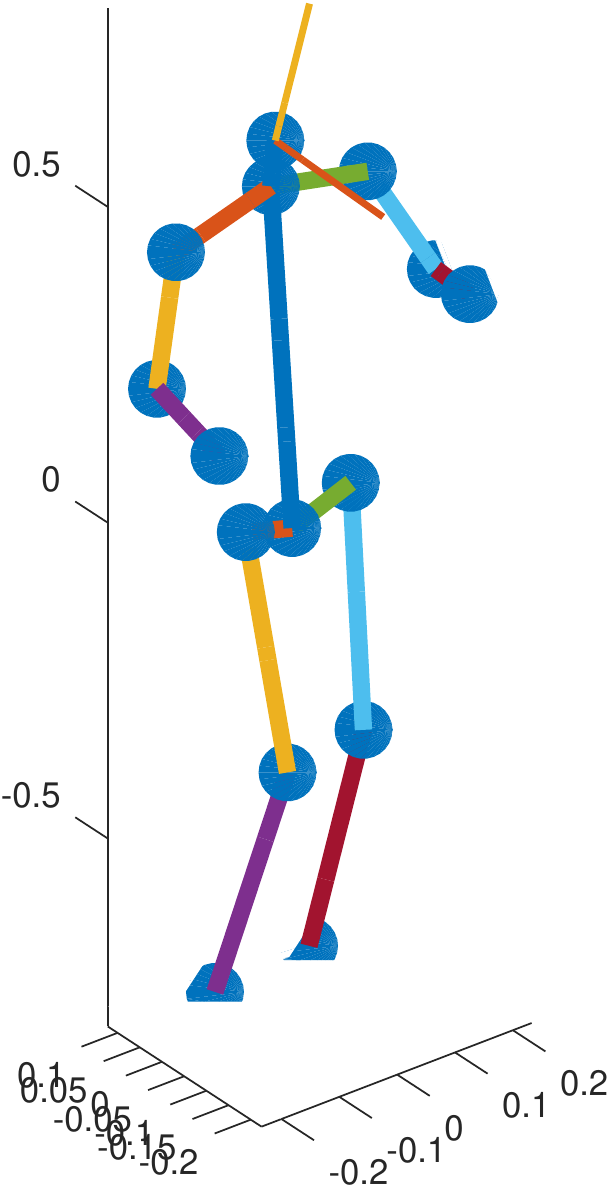}%
\includegraphics[width=0.125\linewidth, height=0.2\linewidth]{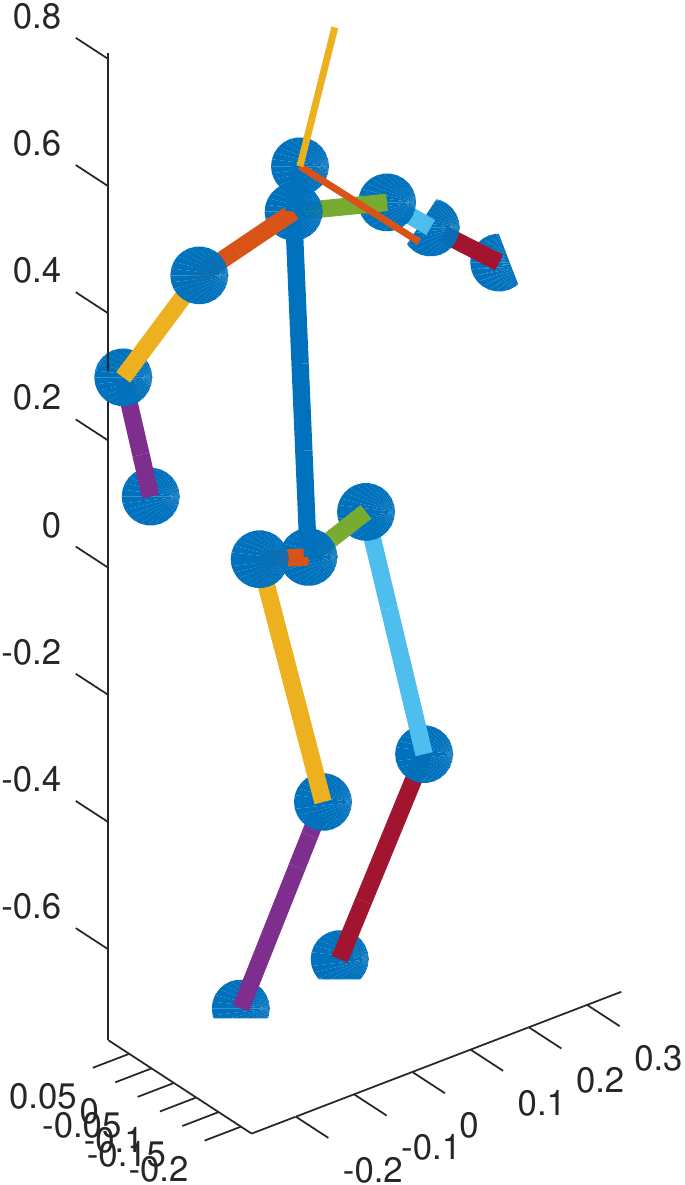}%
\includegraphics[width=0.125\linewidth, height=0.2\linewidth]{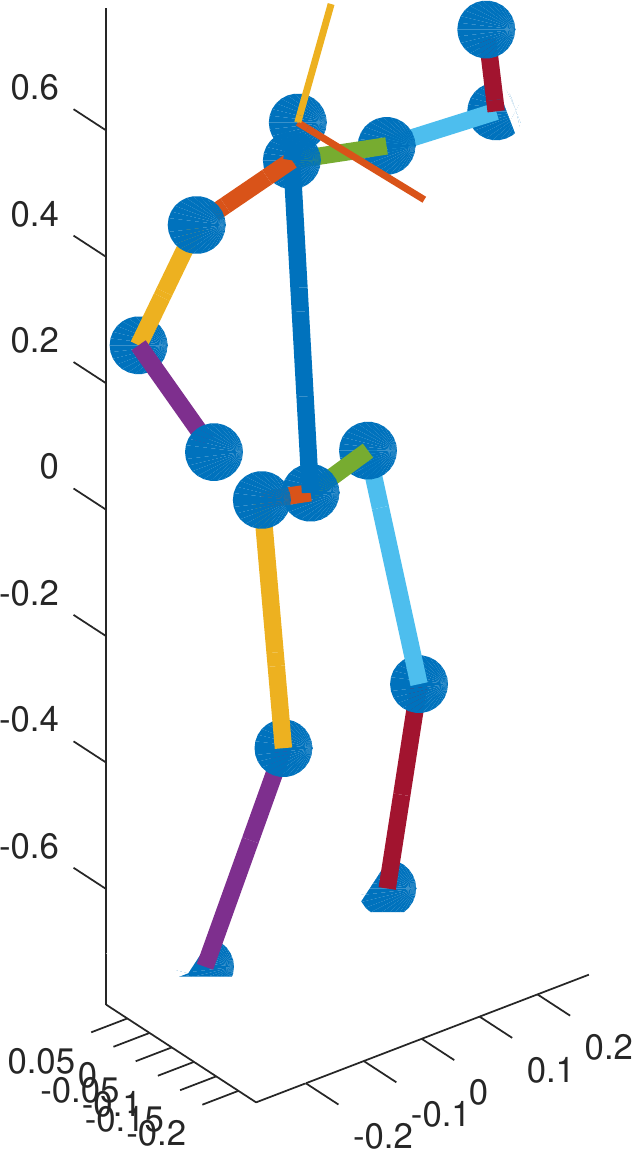}%
\includegraphics[width=0.125\linewidth, height=0.2\linewidth]{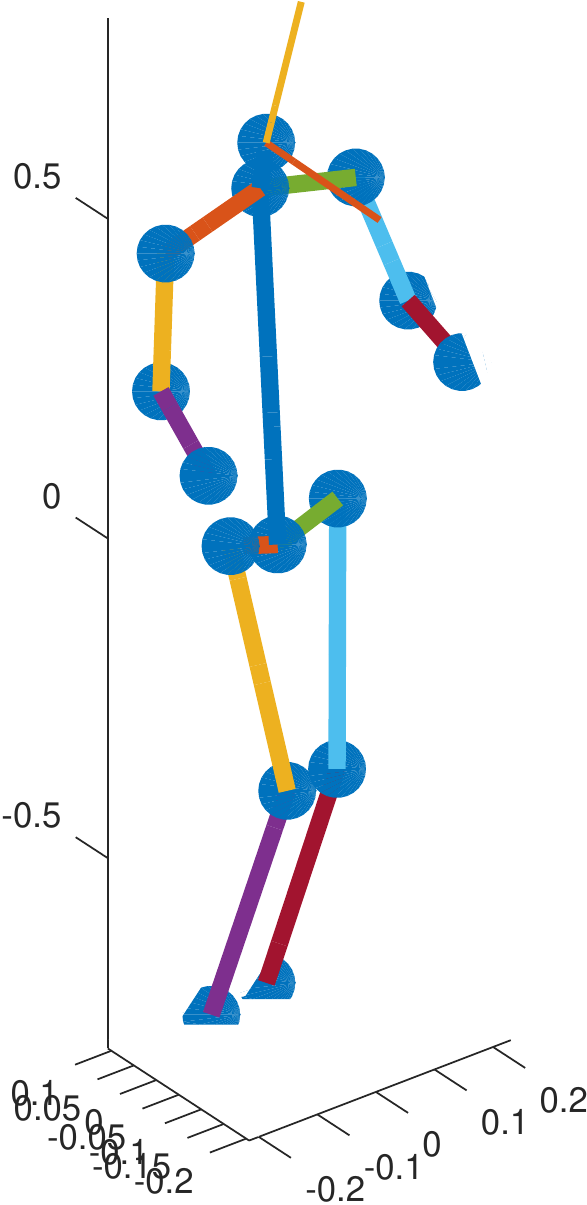}%
\includegraphics[width=0.125\linewidth, height=0.2\linewidth]{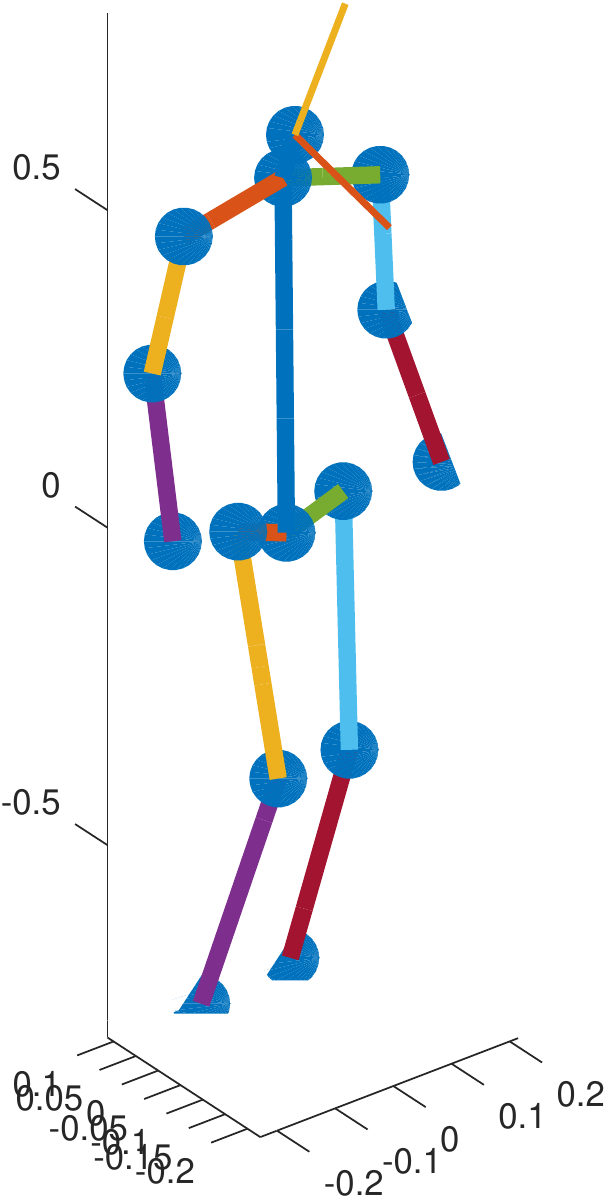}%
\includegraphics[width=0.125\linewidth, height=0.2\linewidth]{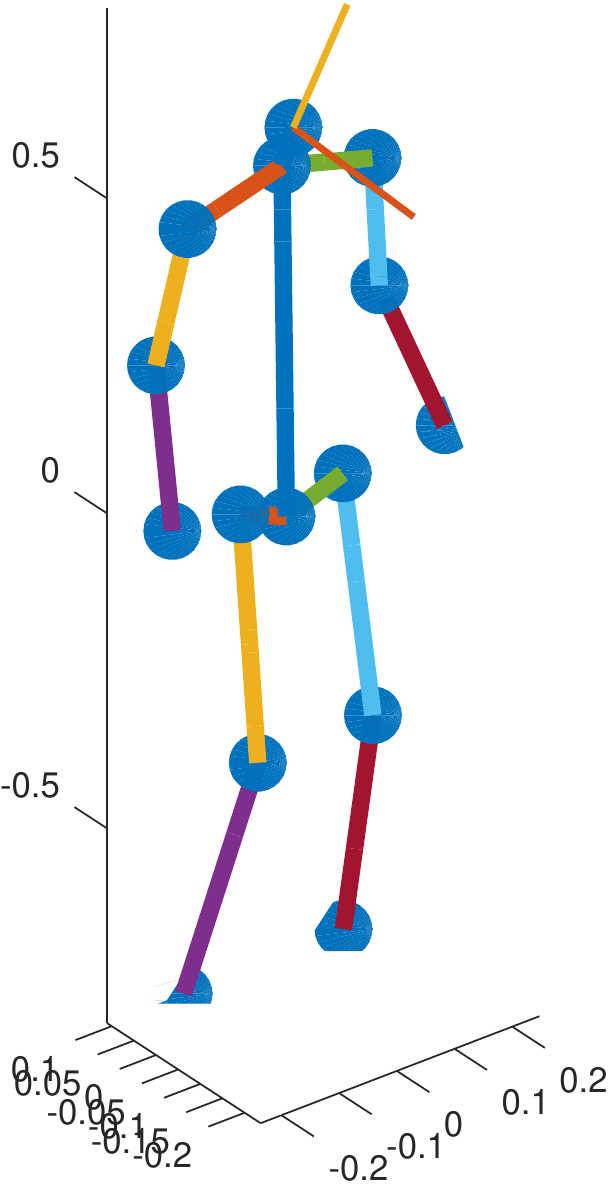}%
\includegraphics[width=0.125\linewidth, height=0.2\linewidth]{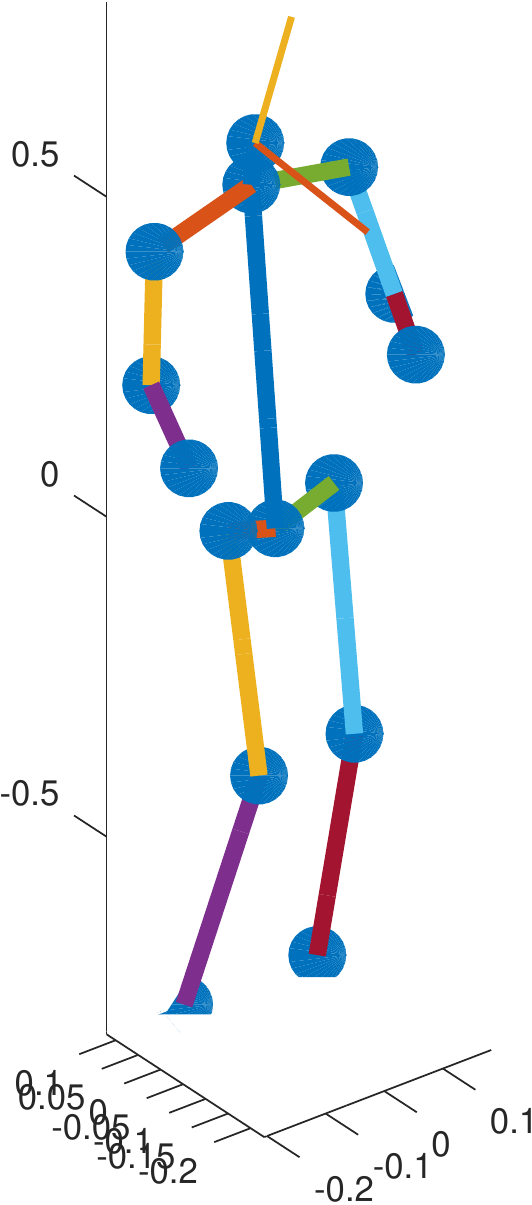}%
\end{tabularx}}
\caption{ Comparison with variations of the proposed method. Row one: ground truth body and head poses.
Row two: the result of the proposed method. 
	Row three: the result of \texttt{MotionOnly} method.
}
	\label{fig:comp1}
	\vspace{-10pt}
\end{figure}

\begin{figure}[tb]
	\centering
	\scalebox{0.95}{	
	\setlength\tabcolsep{1pt}
\begin{tabularx}{\linewidth}{c X }
\rotatebox{90}{\hspace{4pt}{\scriptsize Ground Truth}}  & \includegraphics[width=0.125\linewidth, height=0.2\linewidth]{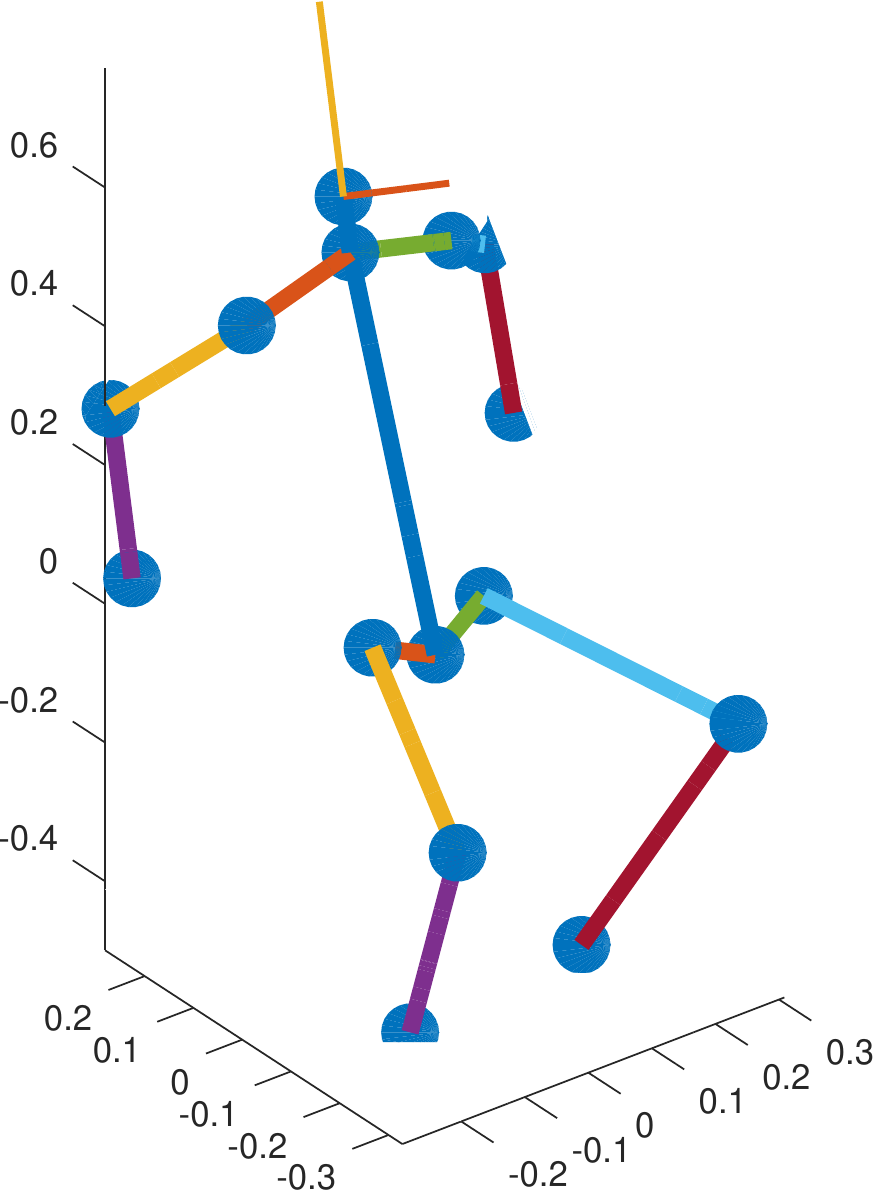}%
\includegraphics[width=0.125\linewidth, height=0.2\linewidth]{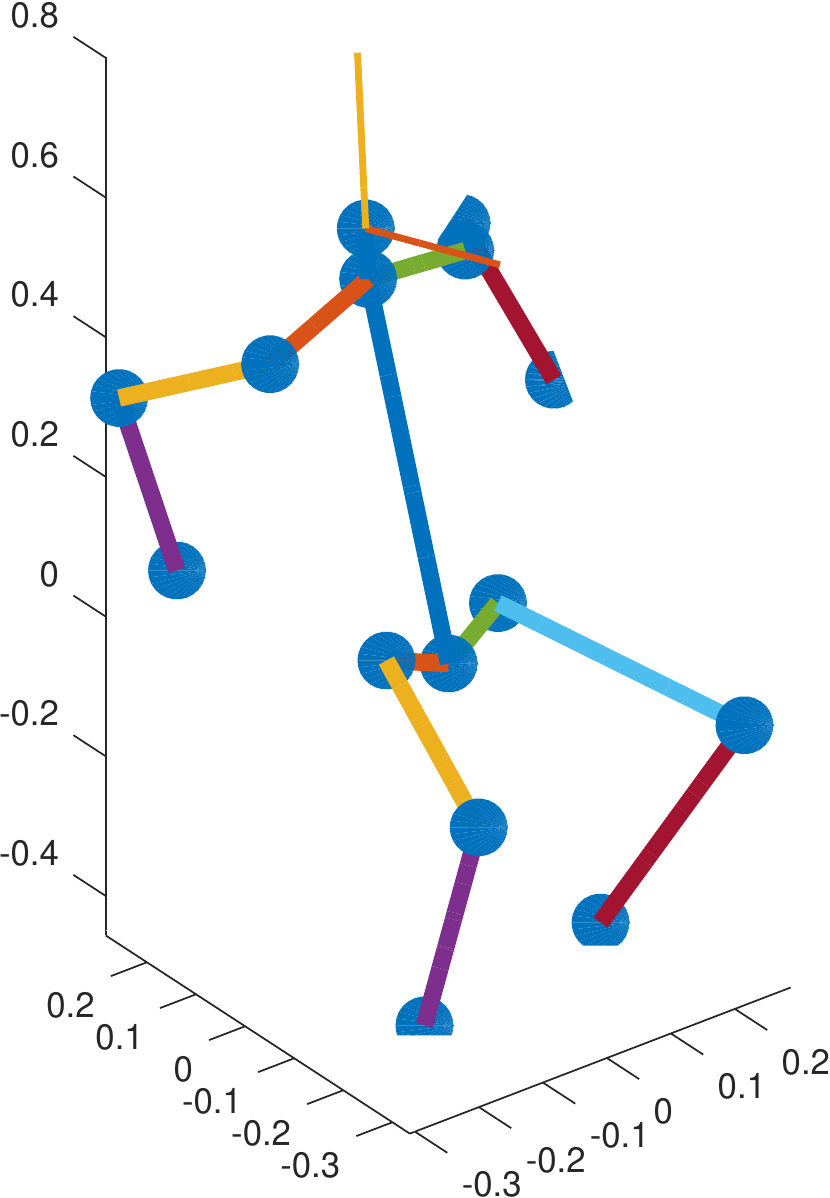}%
\includegraphics[width=0.125\linewidth, height=0.2\linewidth]{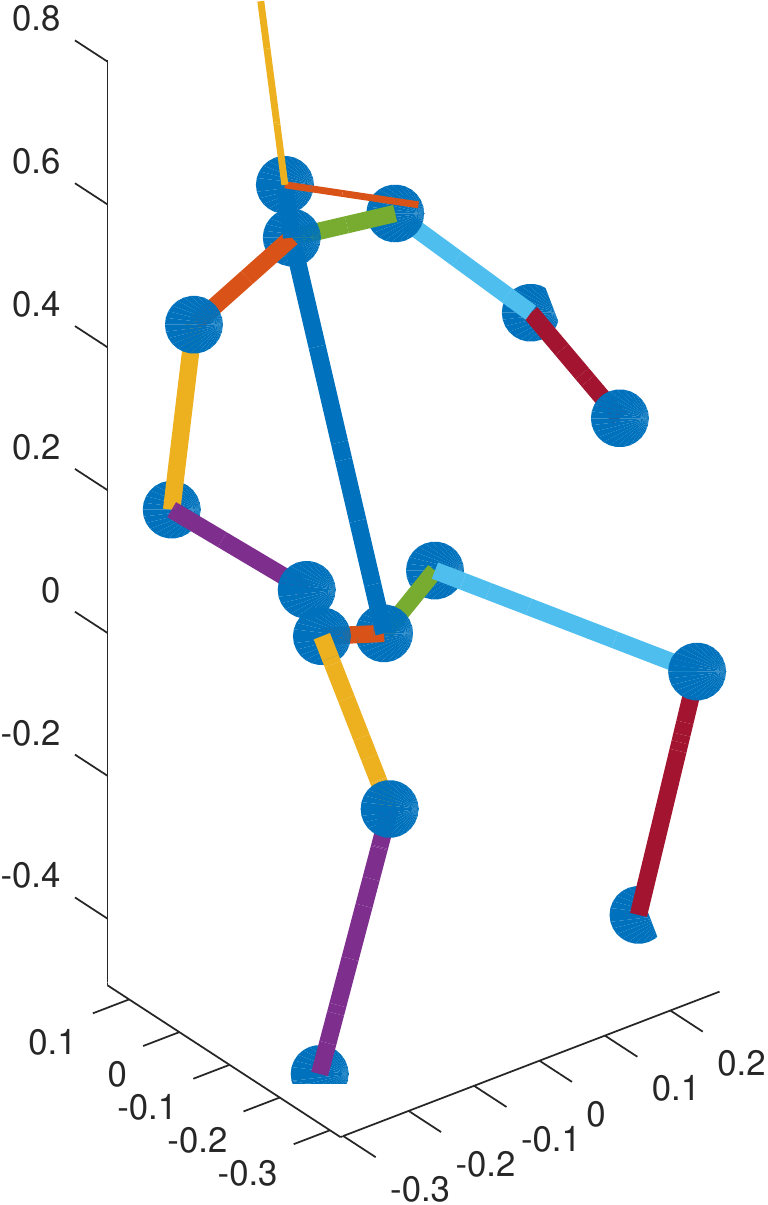}%
\includegraphics[width=0.125\linewidth, height=0.2\linewidth]{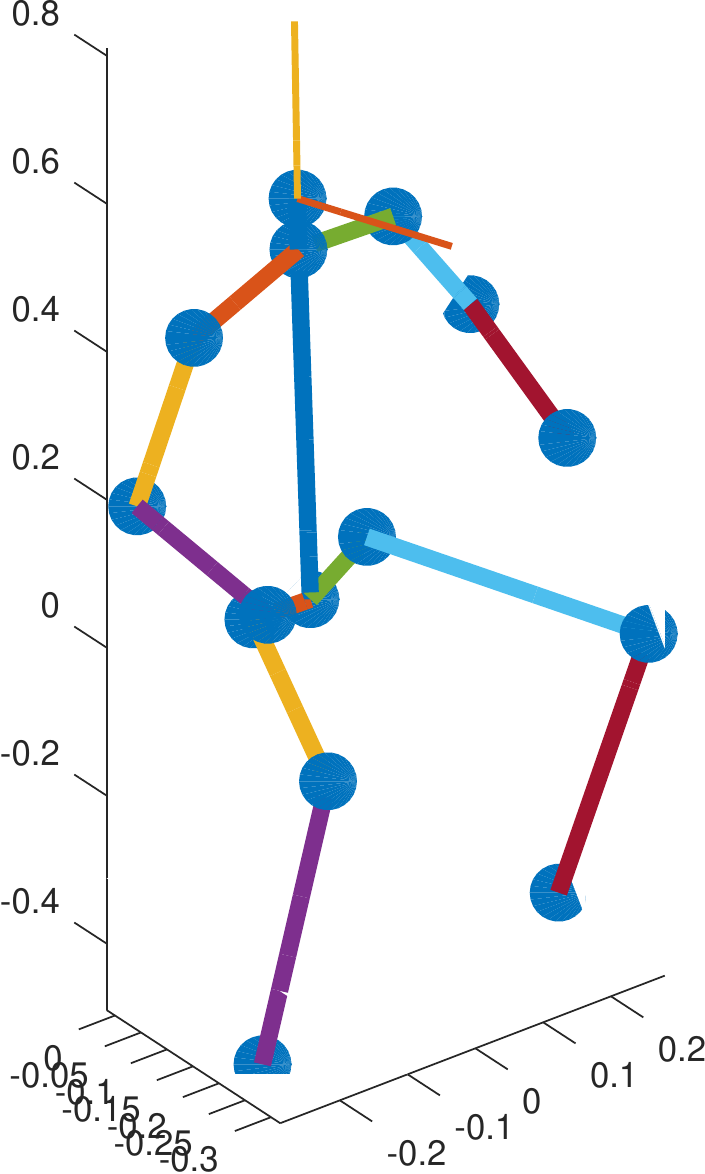}%
\includegraphics[width=0.125\linewidth, height=0.2\linewidth]{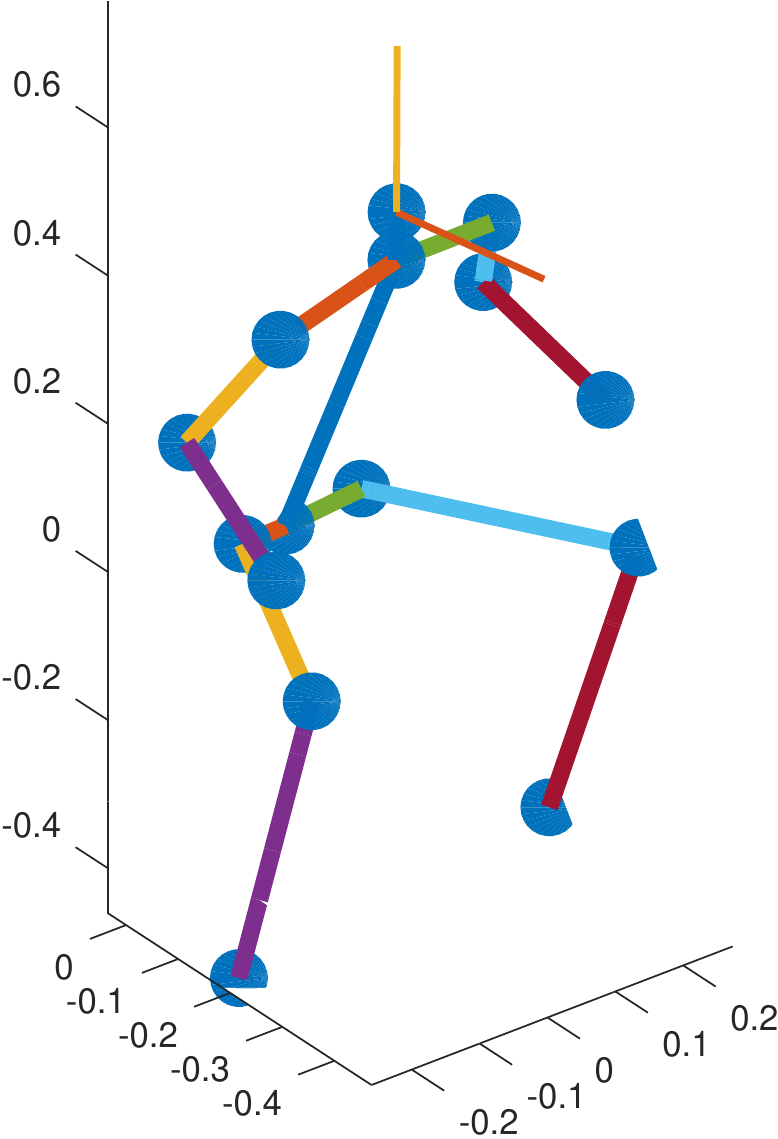}%
\includegraphics[width=0.125\linewidth, height=0.2\linewidth]{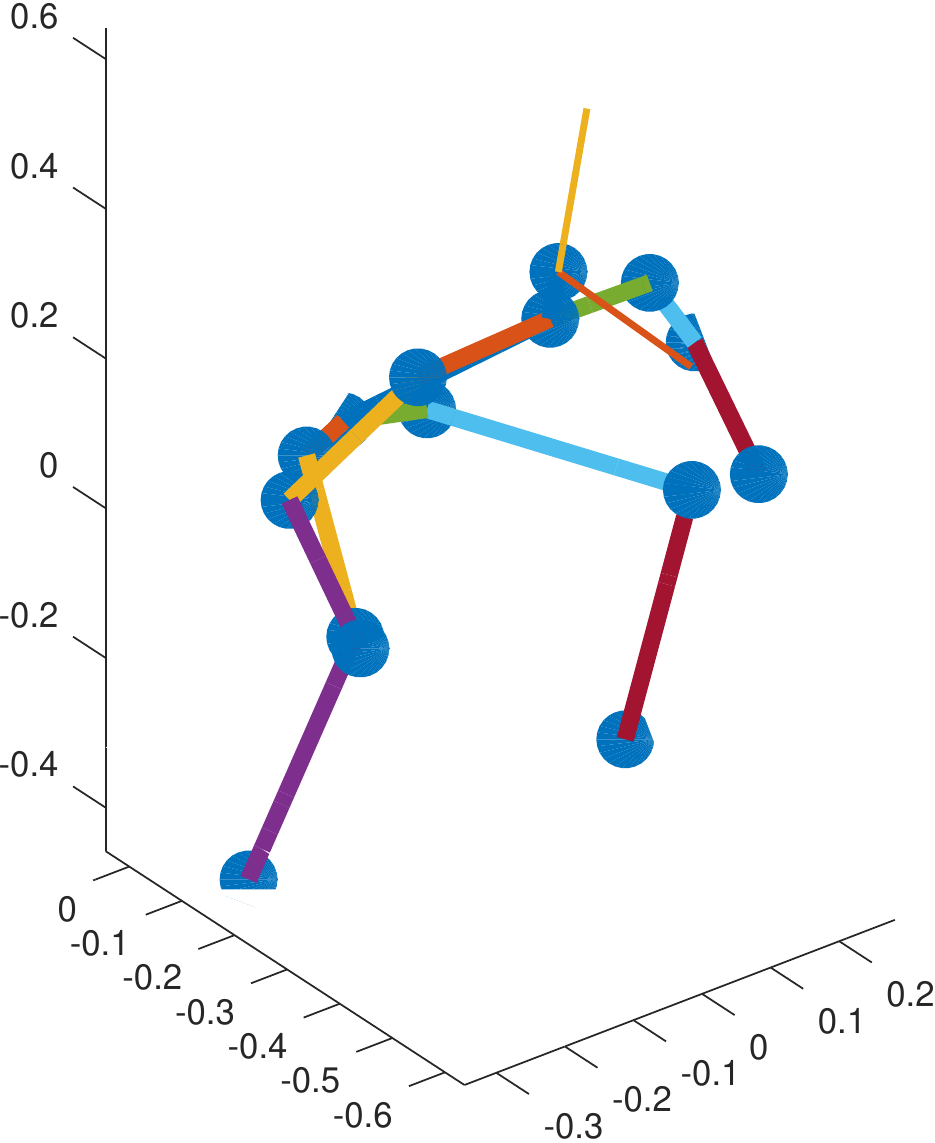}%
\includegraphics[width=0.125\linewidth, height=0.2\linewidth]{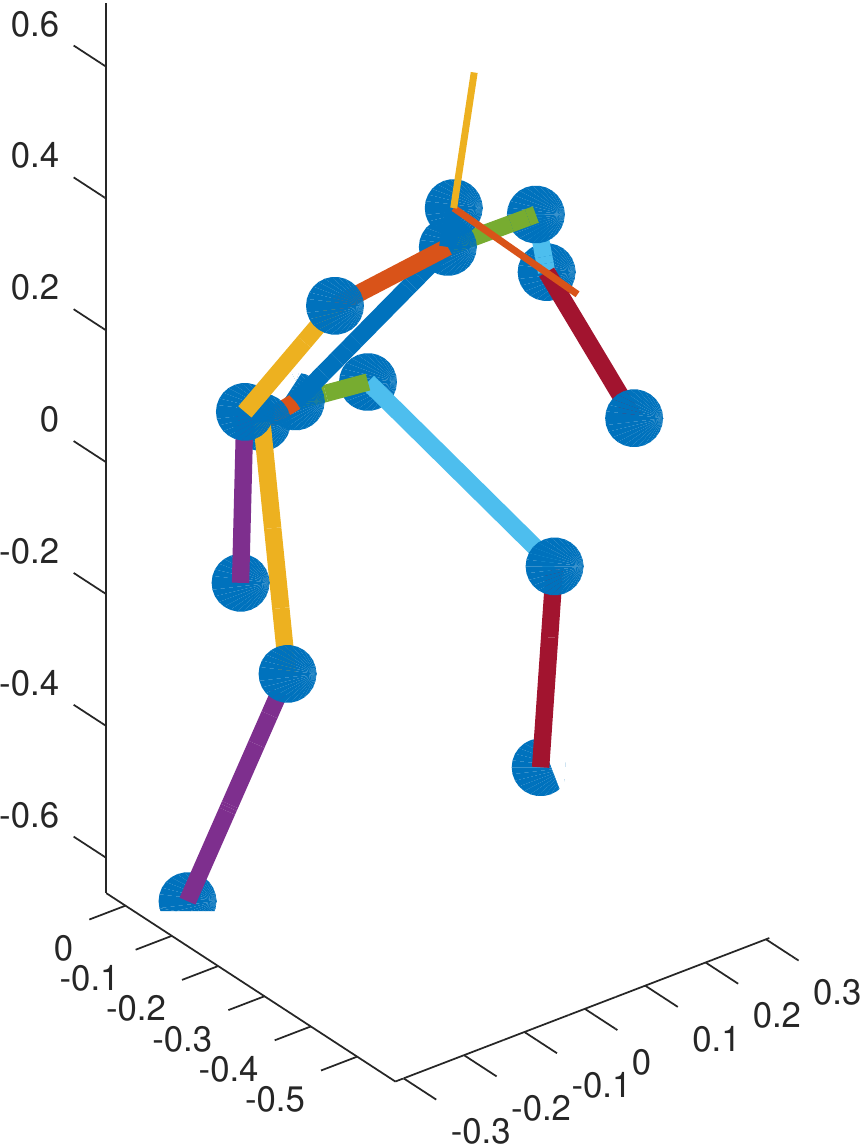}%
\includegraphics[width=0.125\linewidth, height=0.2\linewidth]{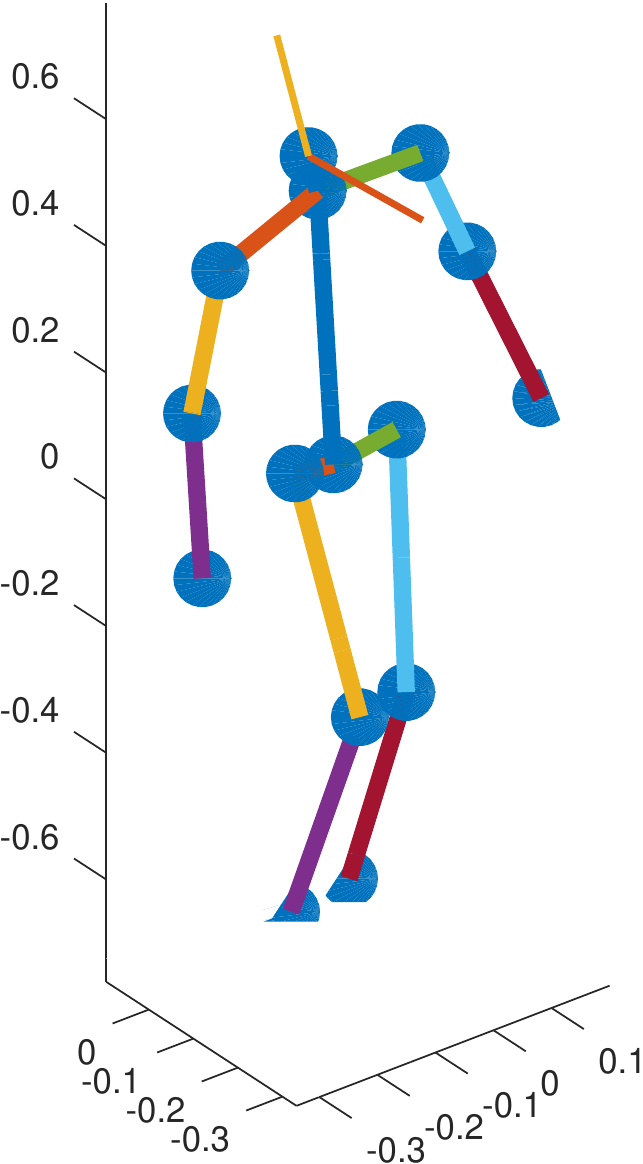}\\
 \rotatebox{90}{\hspace{10pt}{\scriptsize Full Model}} &
\includegraphics[width=0.125\linewidth, height=0.2\linewidth]{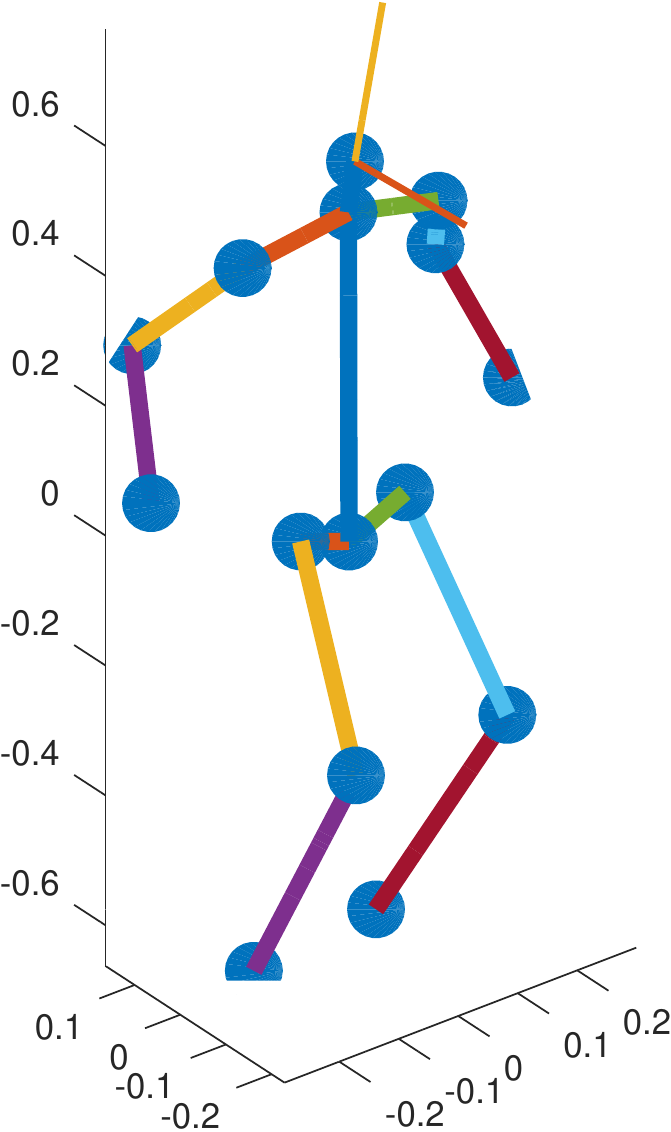}%
\includegraphics[width=0.125\linewidth, height=0.2\linewidth]{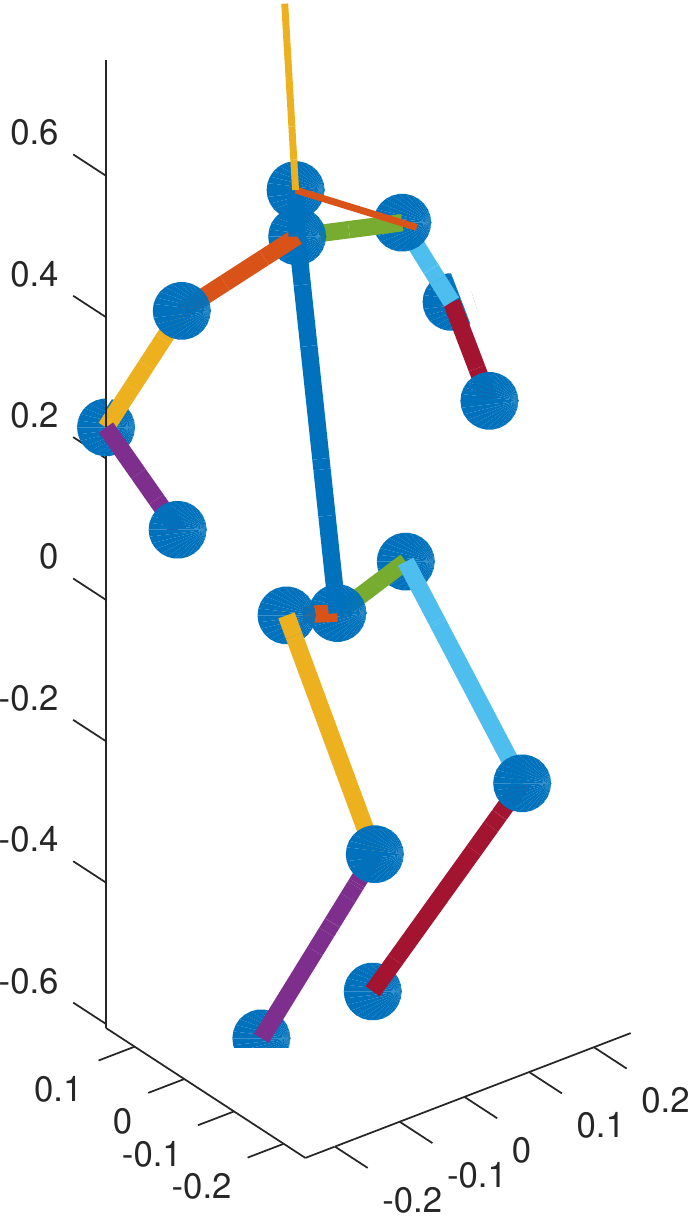}%
\includegraphics[width=0.125\linewidth, height=0.2\linewidth]{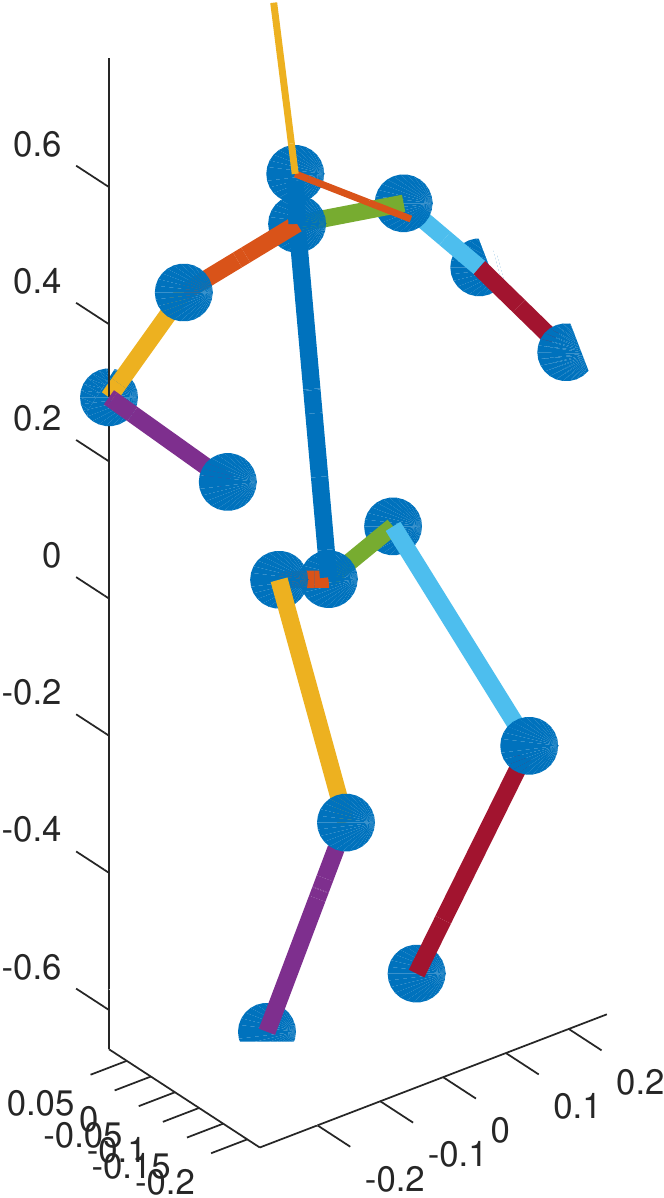}%
\includegraphics[width=0.125\linewidth, height=0.2\linewidth]{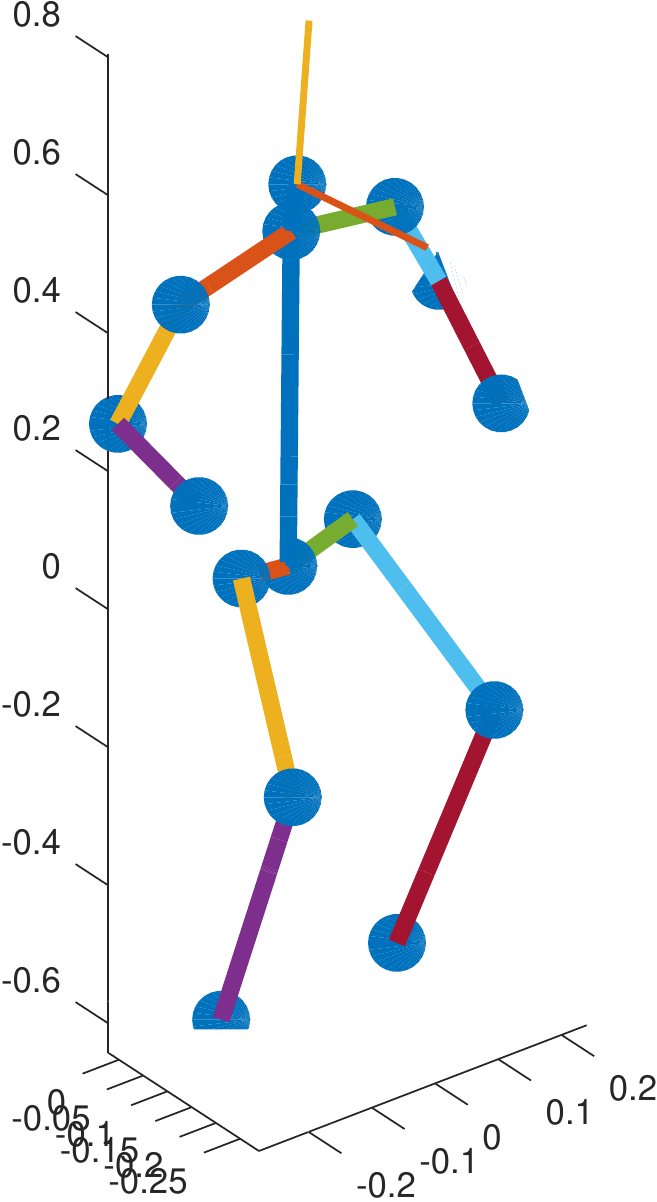}%
\includegraphics[width=0.125\linewidth, height=0.2\linewidth]{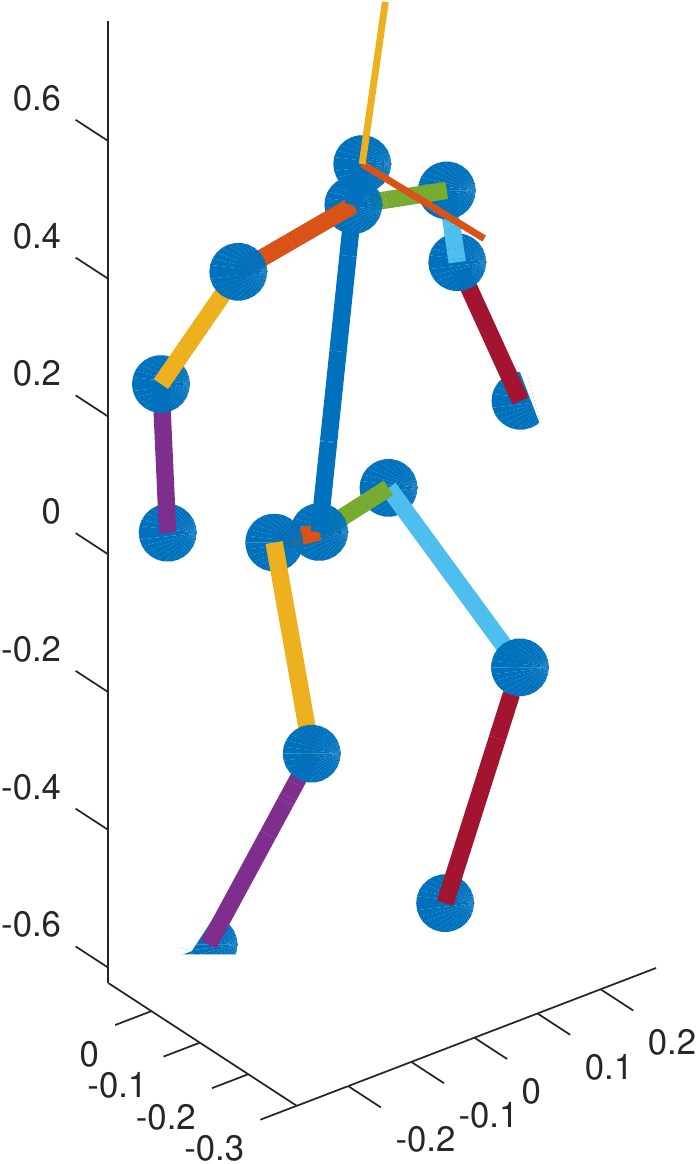}%
\includegraphics[width=0.125\linewidth, height=0.2\linewidth]{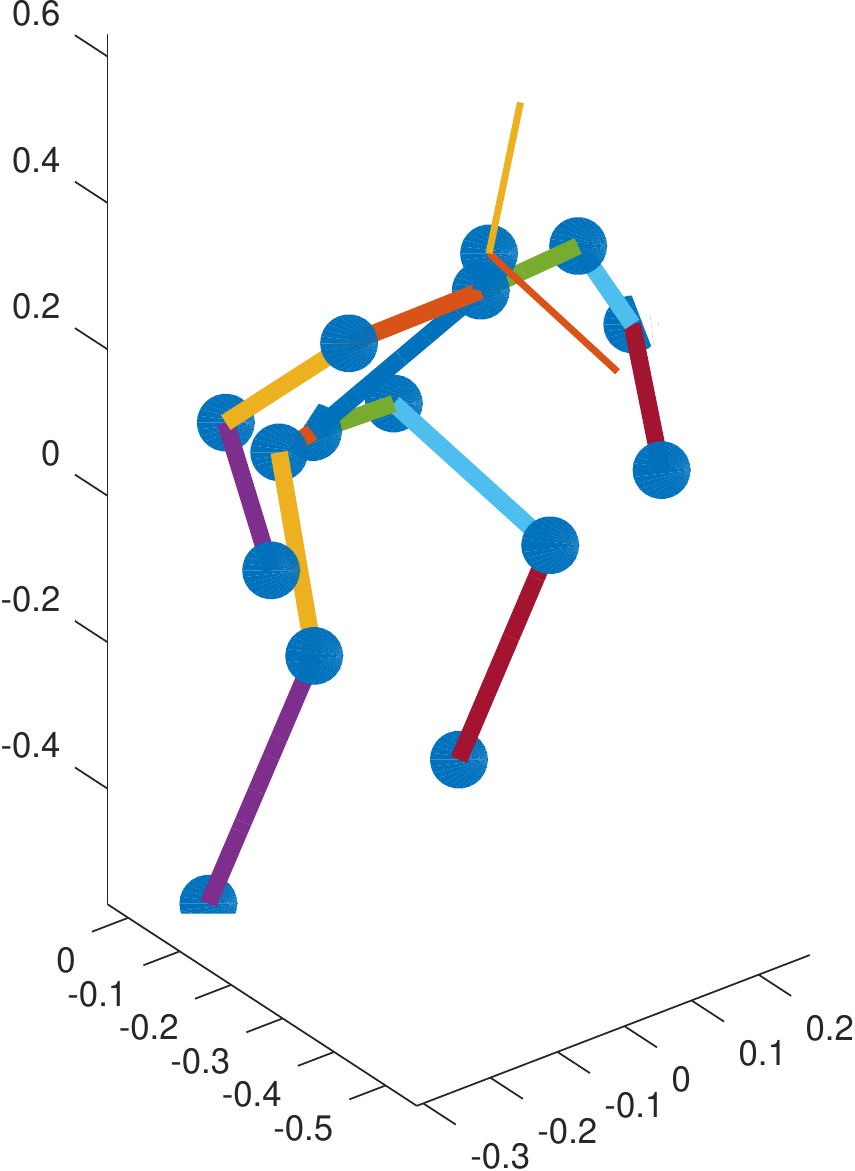}%
\includegraphics[width=0.125\linewidth, height=0.2\linewidth]{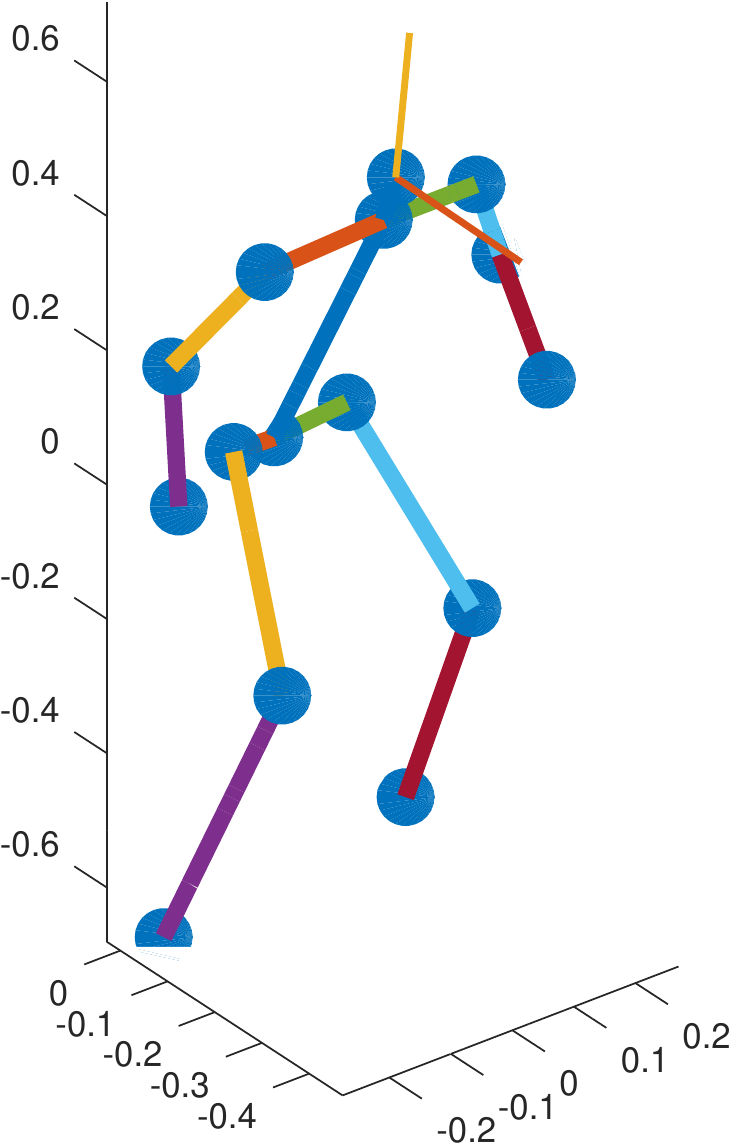}%
\includegraphics[width=0.125\linewidth, height=0.2\linewidth]{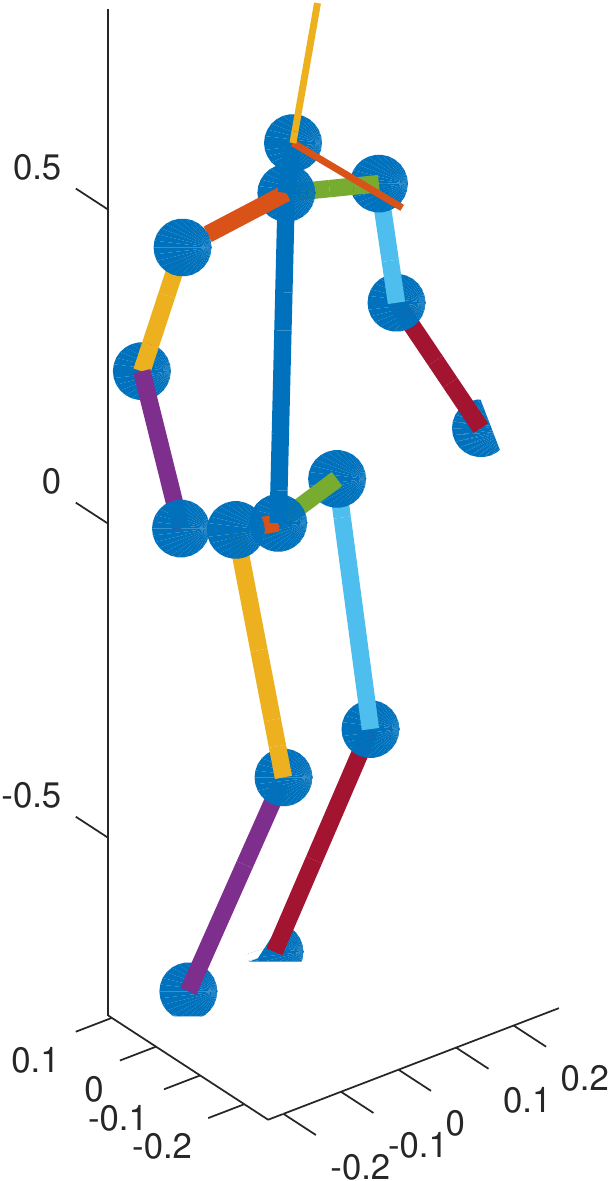}\\
	\rotatebox{90}{\hspace{10pt}{\scriptsize ShapeOnly}} & \includegraphics[width=0.125\linewidth, height=0.2\linewidth]{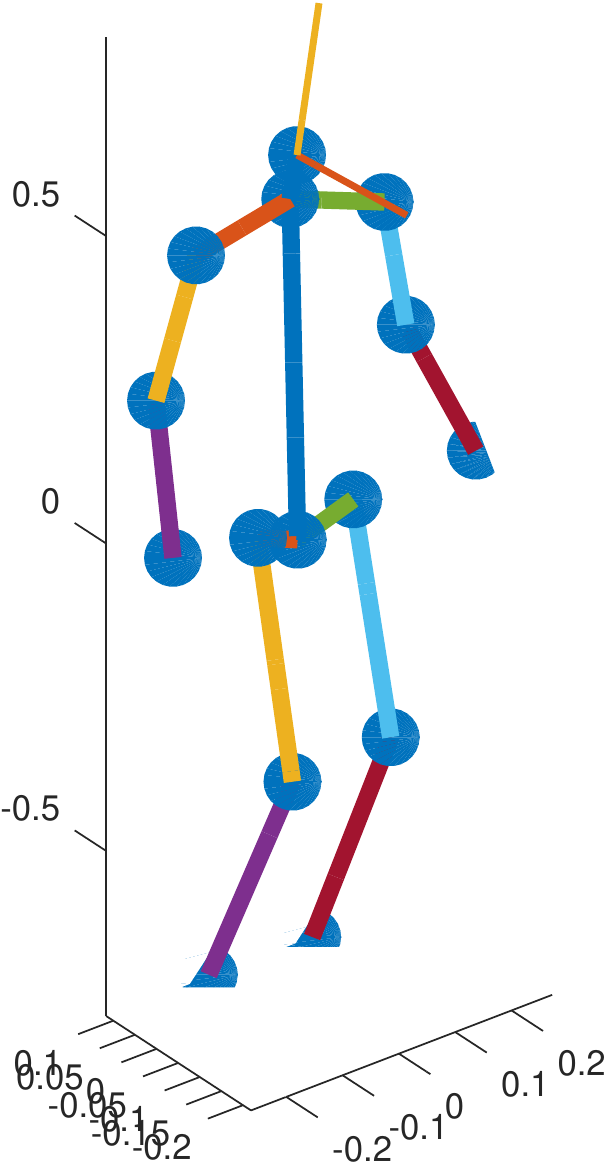}%
\includegraphics[width=0.125\linewidth, height=0.2\linewidth]{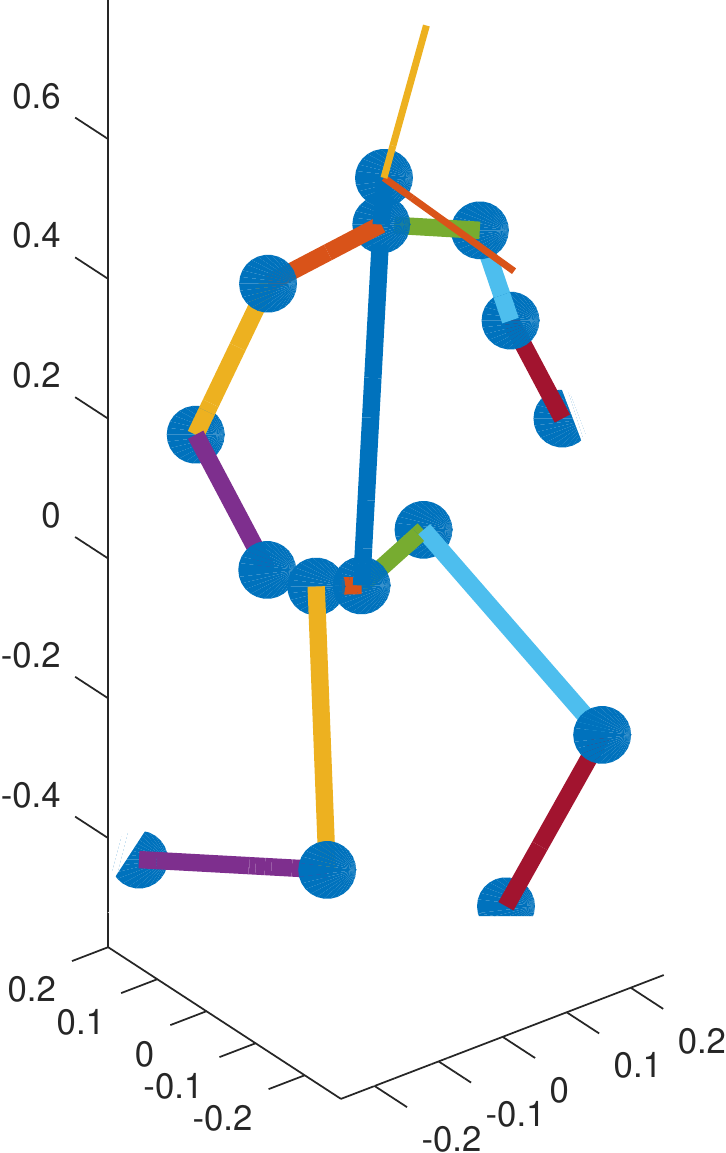}%
\includegraphics[width=0.125\linewidth, height=0.2\linewidth]{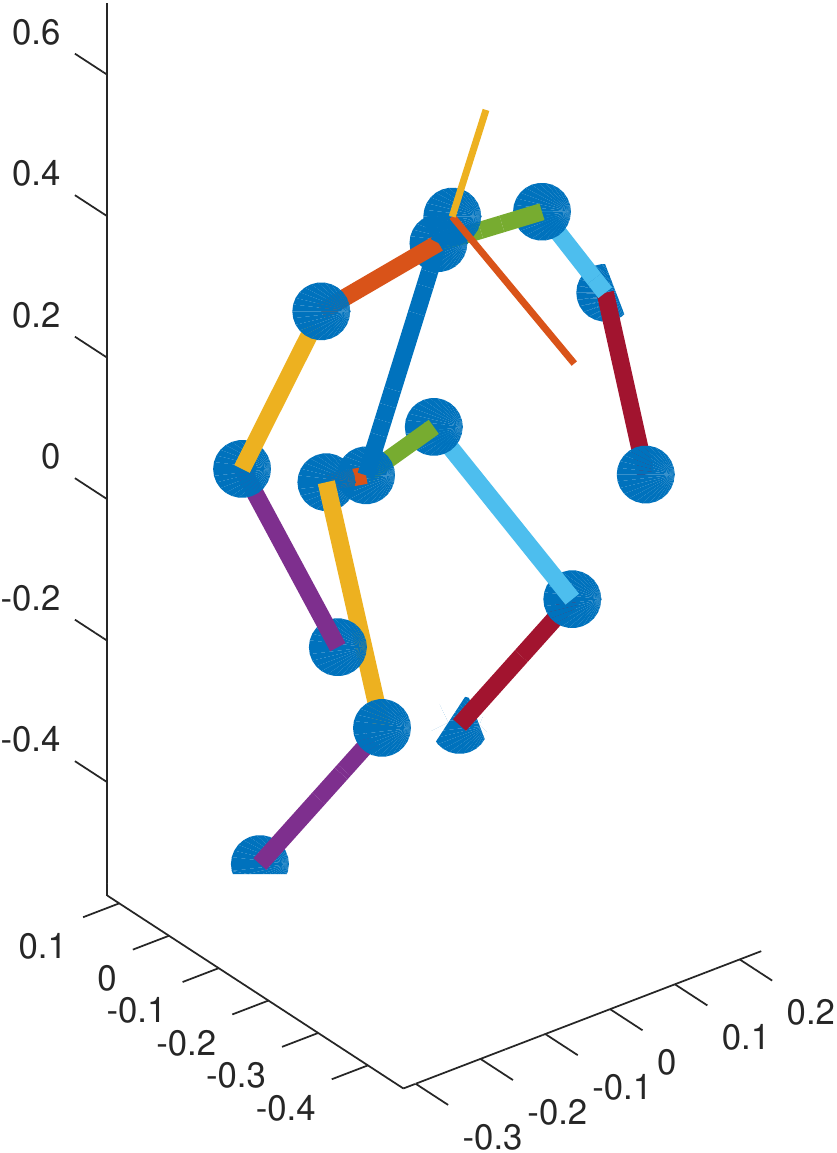}%
\includegraphics[width=0.125\linewidth, height=0.2\linewidth]{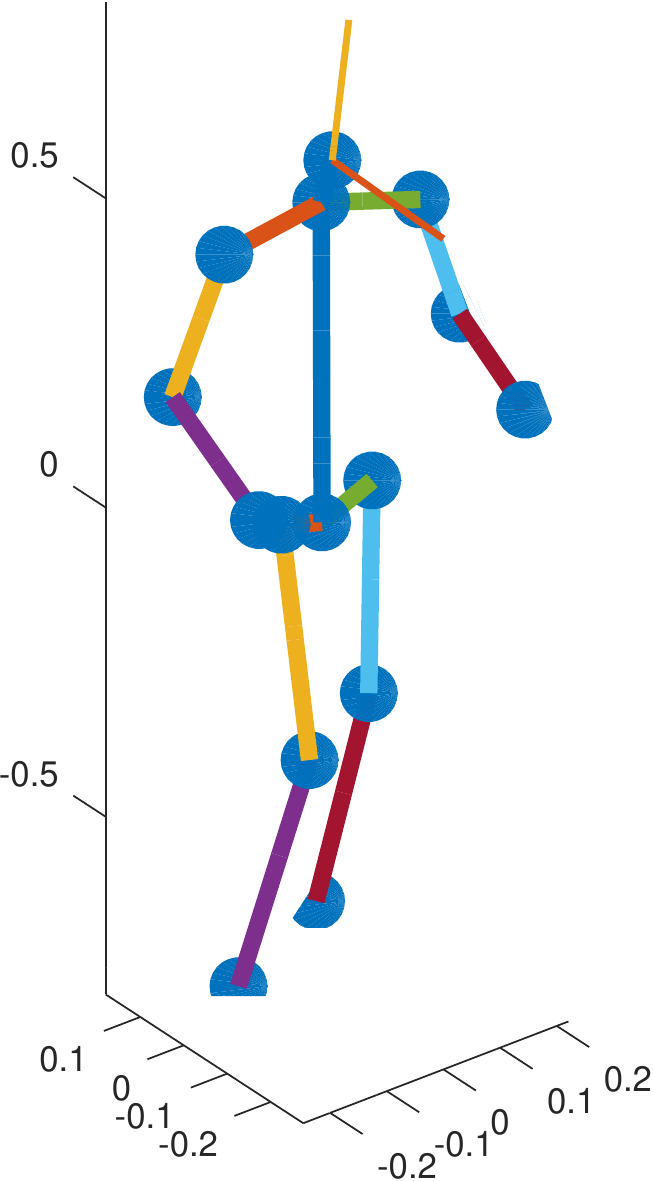}%
\includegraphics[width=0.125\linewidth, height=0.2\linewidth]{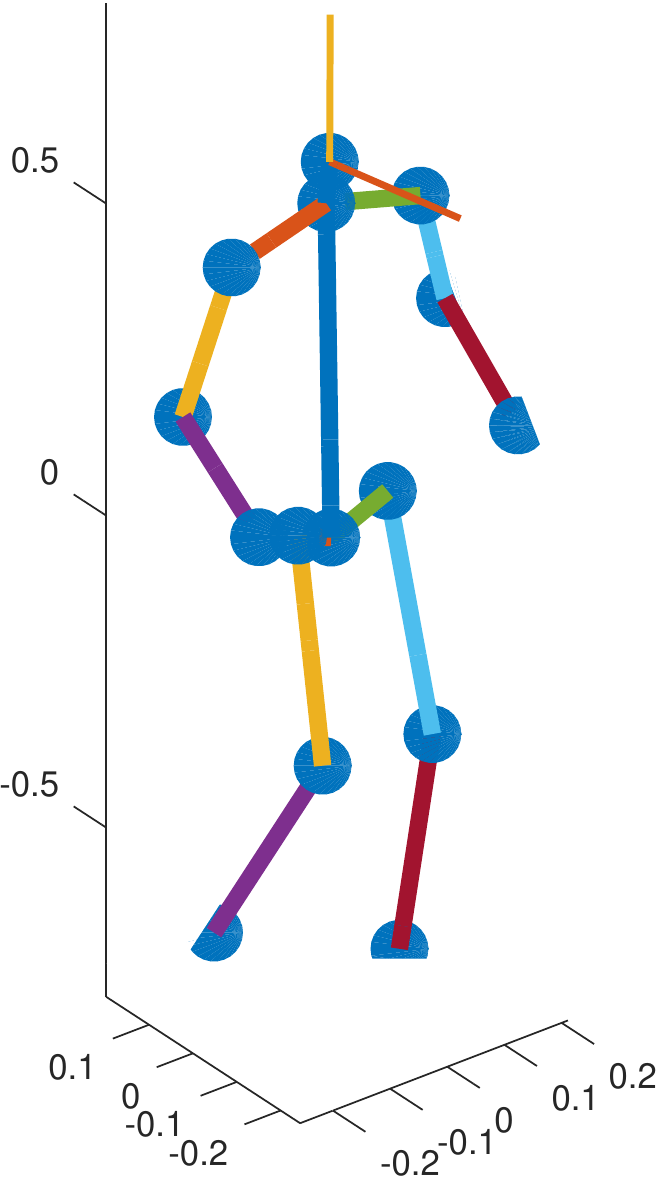}%
\includegraphics[width=0.125\linewidth, height=0.2\linewidth]{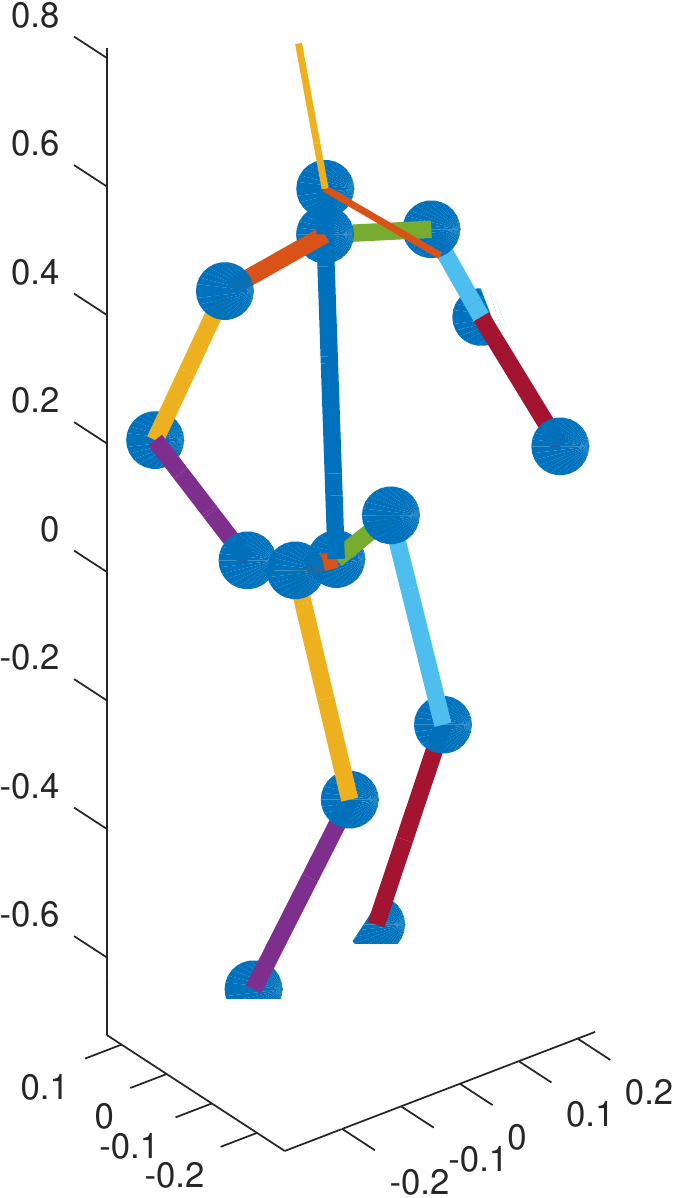}%
\includegraphics[width=0.125\linewidth, height=0.2\linewidth]{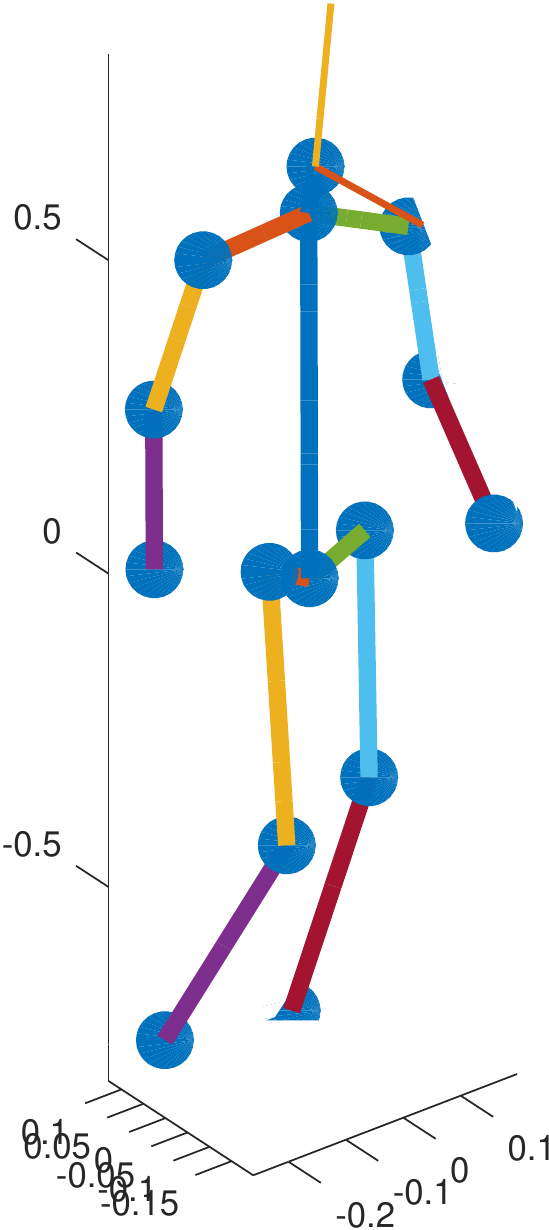}%
\includegraphics[width=0.125\linewidth, height=0.2\linewidth]{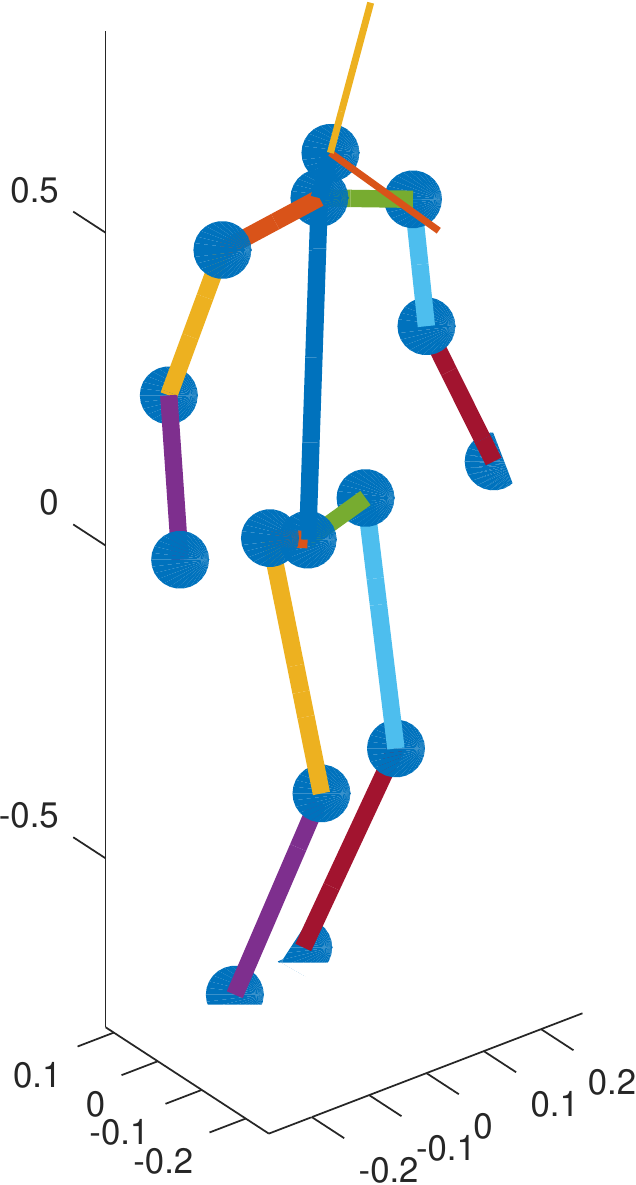}\\
	\rotatebox{90}{\hspace{10pt}{\scriptsize NoHeight}} &
\includegraphics[width=0.125\linewidth, height=0.2\linewidth]{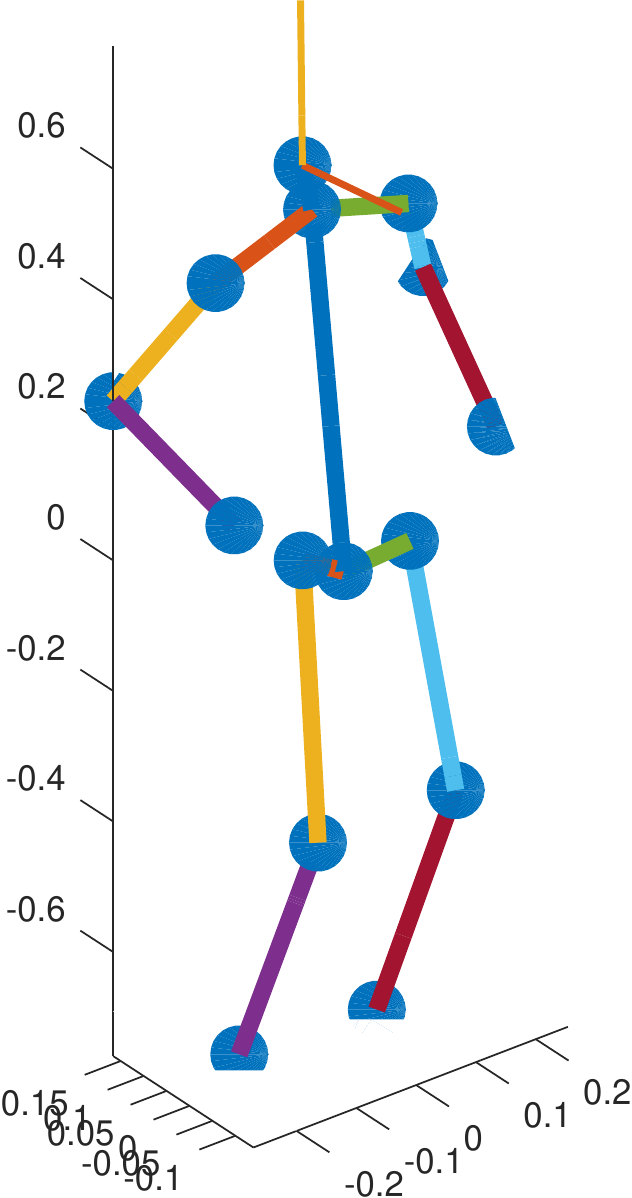}%
\includegraphics[width=0.125\linewidth, height=0.2\linewidth]{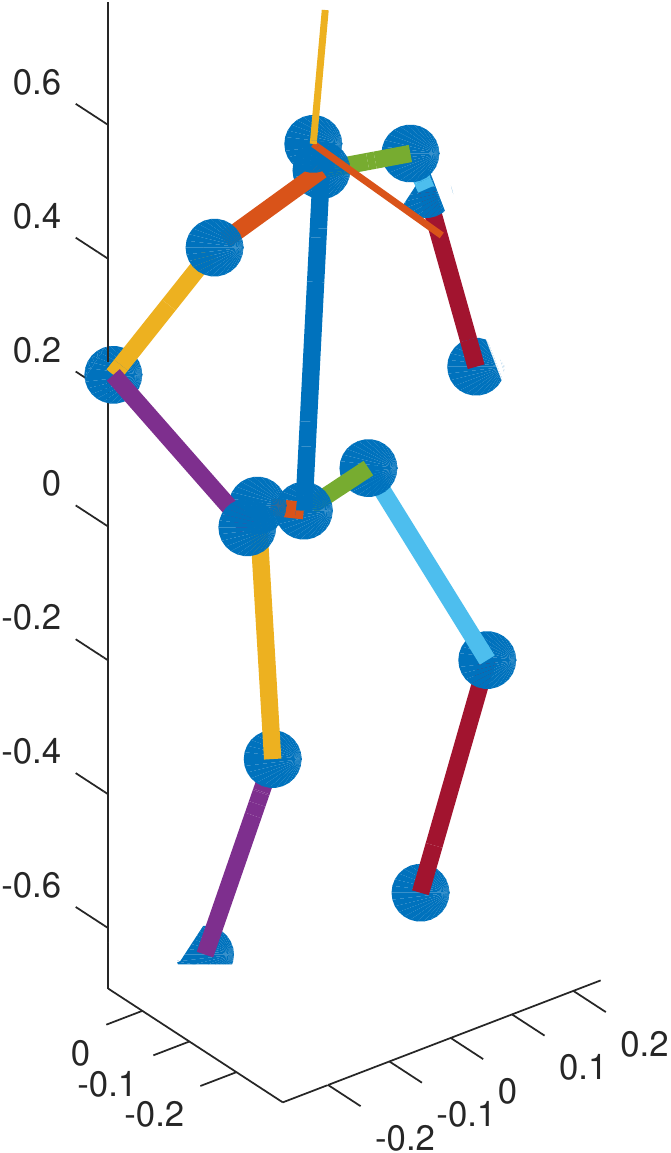}%
\includegraphics[width=0.125\linewidth, height=0.2\linewidth]{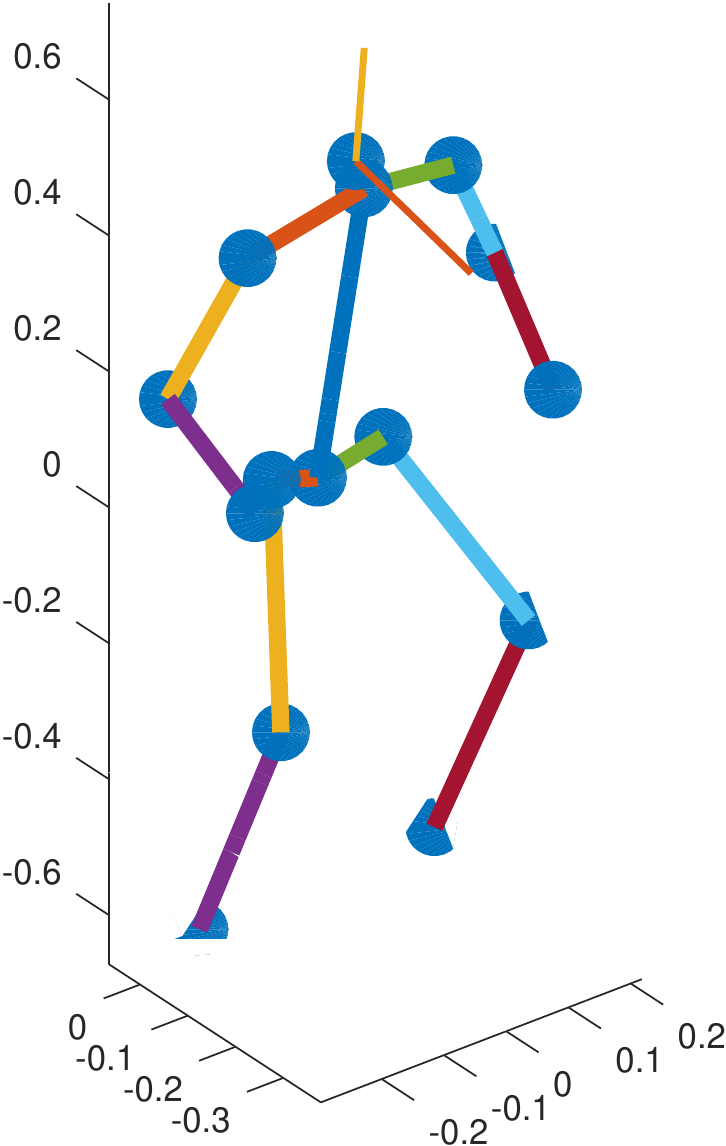}%
\includegraphics[width=0.125\linewidth, height=0.2\linewidth]{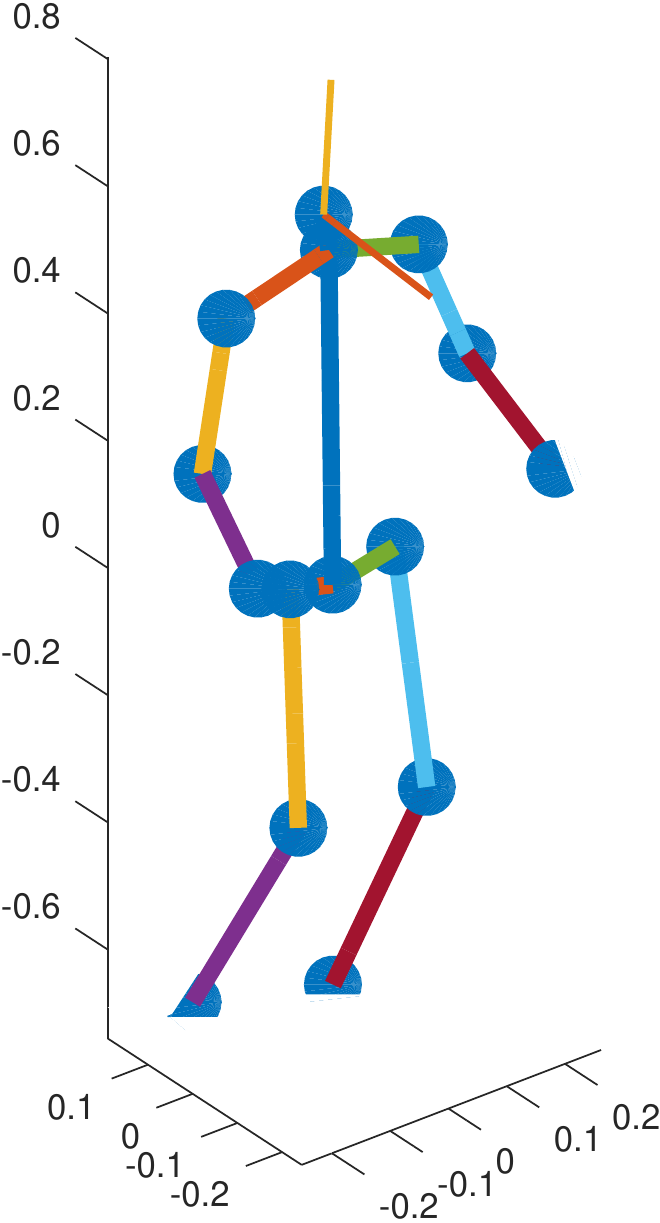}%
\includegraphics[width=0.125\linewidth, height=0.2\linewidth]{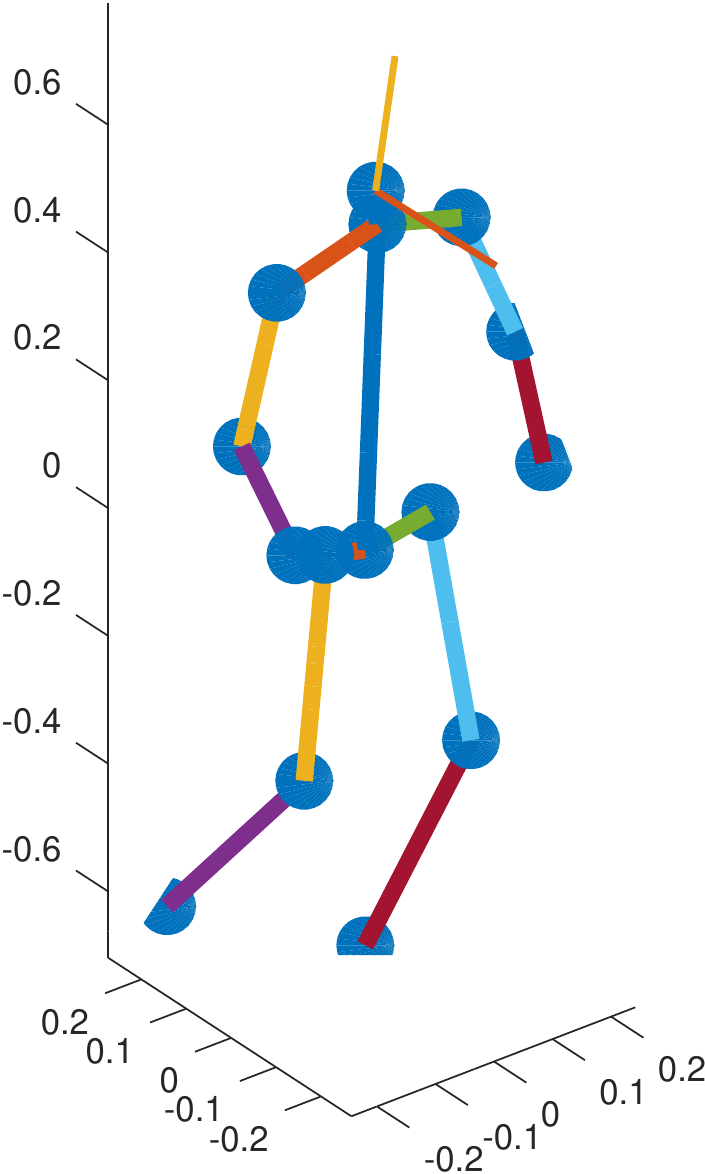}%
\includegraphics[width=0.125\linewidth, height=0.2\linewidth]{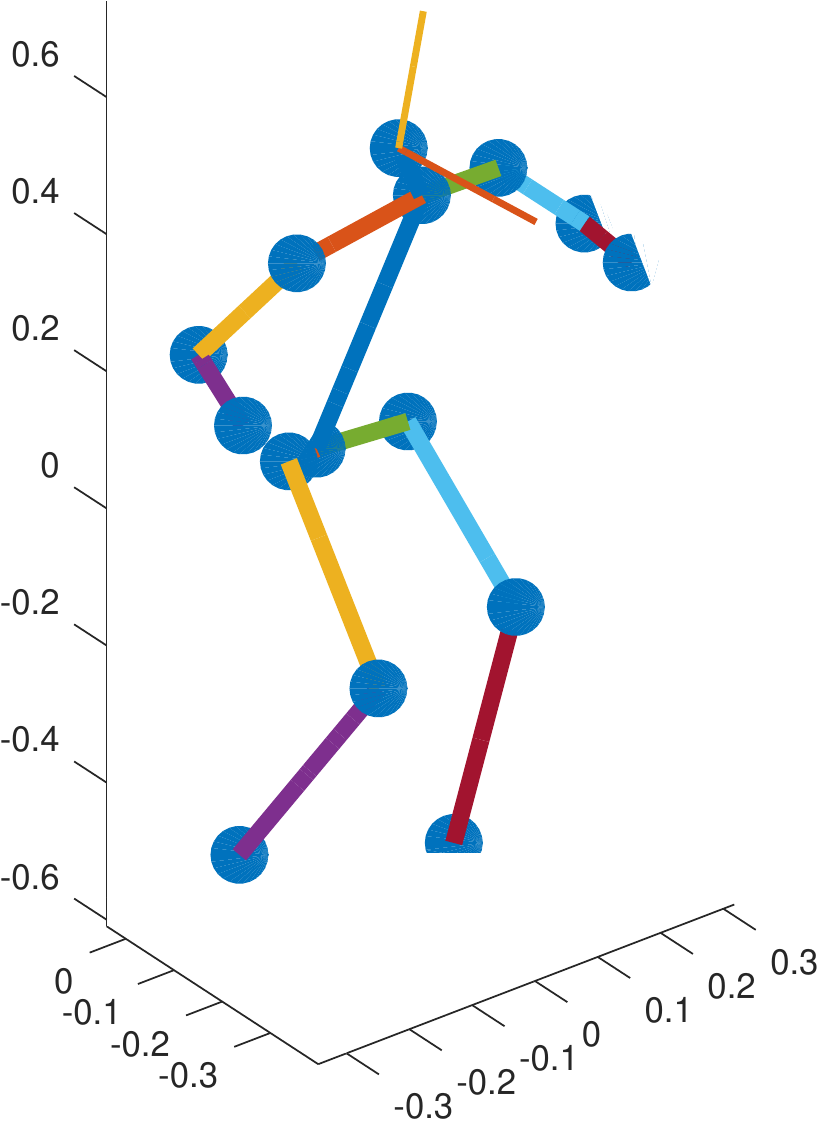}%
\includegraphics[width=0.125\linewidth, height=0.2\linewidth]{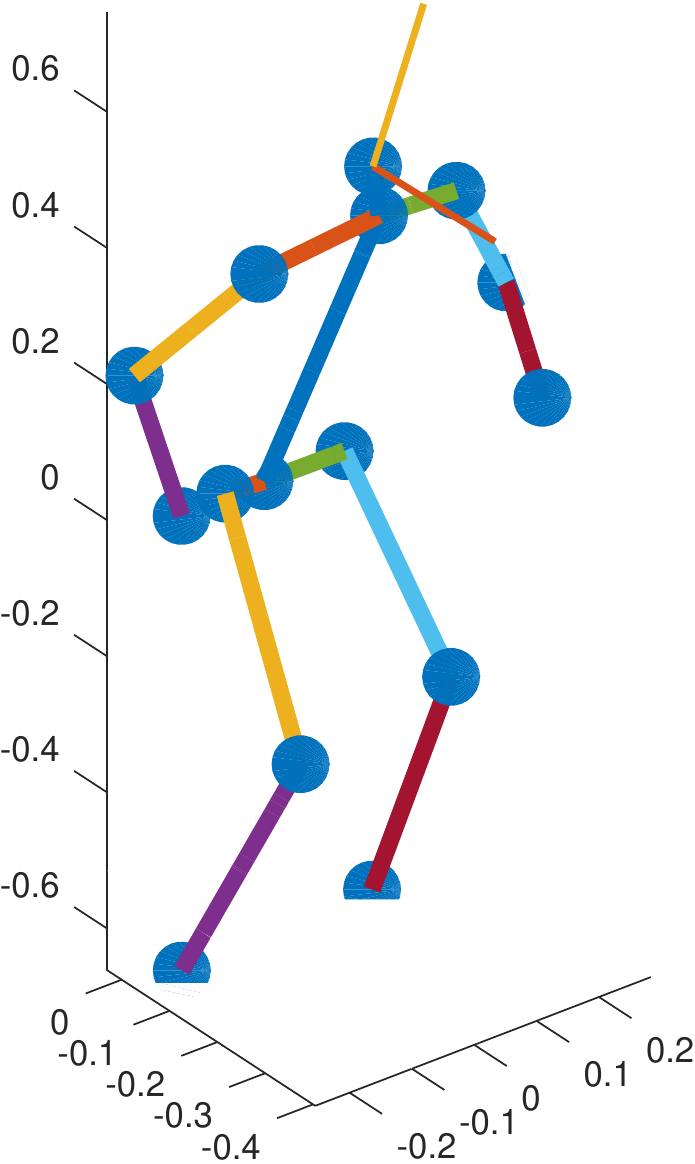}%
\includegraphics[width=0.125\linewidth, height=0.2\linewidth]{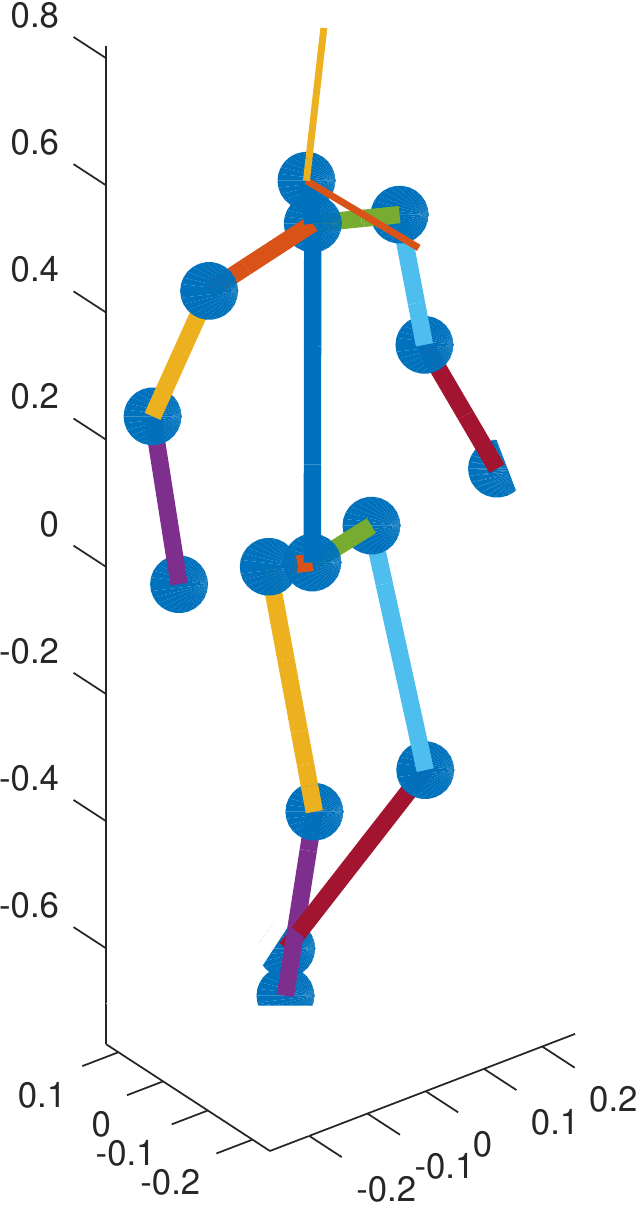}%
\end{tabularx}}
	\caption{ Comparison with variations of the proposed method. Row one: the ground truth. Row two: the proposed method's result. Row three: \texttt{ShapeOnly}'s result. 
	Row four: \texttt{NoHeight}'s result.
}
	\label{fig:comp2}
	\vspace{-15pt}
\end{figure}

Table~\ref{tab:syn} also shows that the full model leads to overall best results compared to the different variations 
(special cases resulting from changing different components of the system). 
Figs.~\ref{fig:comp1} and \ref{fig:comp2} further illustrate this behavior.  
We see that the motion only or the shape only methods give inferior results.
While motion information is important for estimating the lower body pose, the body shape can help improve the upper body pose estimation.
And independently either cannot achieve good performance. 
The results also confirm that the proposed two-stage approach indeed improves accuracy compared to a single stage method without pose refinement.
The RNN network structure gives slightly better results for the 3D keypoints estimation.
When using the refinement network, the non-recurrent version gives better results for both keypoints and head orientation. 
Our design thus chooses the simpler non-recurrent network. 
The inclusion of camera height in the motion history map also attributes to these more accurate results. 
Lastly, the performance of \texttt{AllStand} and \texttt{AllSit} (from Table~\ref{tab:syn}) shows that a naive method gives much larger errors for the head and body pose estimation.
It also retrospectively confirms that the metric used in evaluations is meaningful in quantifying egopose estimation quality.  

\begin{figure*}[tb]
	\centering
	\scalebox{0.975}{	
\setlength\tabcolsep{1pt}
\begin{tabularx}{\linewidth}{c X }
	\rotatebox{90}{\hspace{0pt}{\tiny }} &	
\includegraphics[width=0.05\linewidth, height=0.05\linewidth]{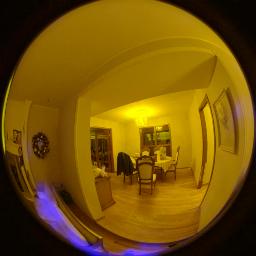}%
\includegraphics[width=0.05\linewidth, height=0.05\linewidth]{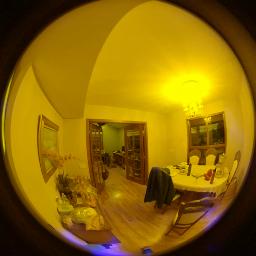}%
\includegraphics[width=0.05\linewidth, height=0.05\linewidth]{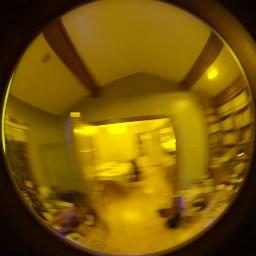}%
\includegraphics[width=0.05\linewidth, height=0.05\linewidth]{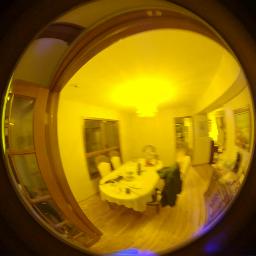}%
\includegraphics[width=0.05\linewidth, height=0.05\linewidth]{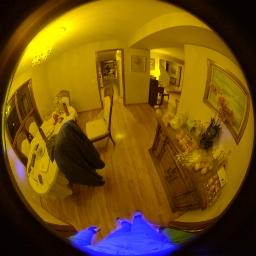}%
\includegraphics[width=0.05\linewidth, height=0.05\linewidth]{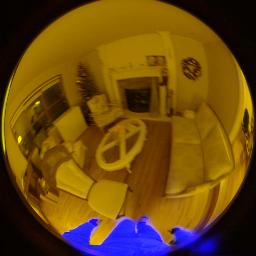}%
\includegraphics[width=0.05\linewidth, height=0.05\linewidth]{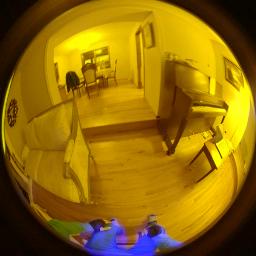}%
\includegraphics[width=0.05\linewidth, height=0.05\linewidth]{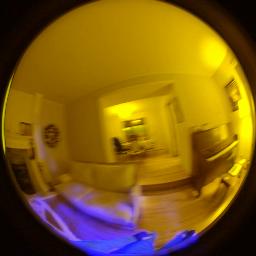}%
\includegraphics[width=0.05\linewidth, height=0.05\linewidth]{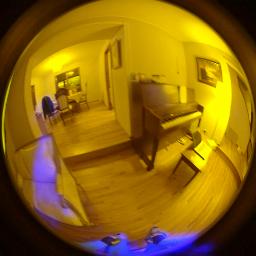}%
\includegraphics[width=0.05\linewidth, height=0.05\linewidth]{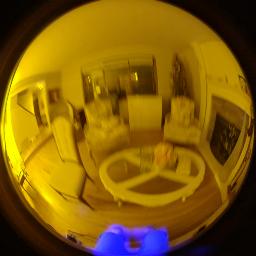}%
\includegraphics[width=0.05\linewidth, height=0.05\linewidth]{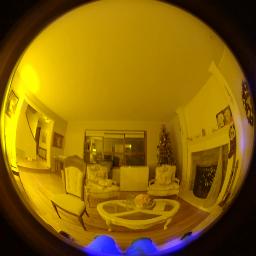}%
\includegraphics[width=0.05\linewidth, height=0.05\linewidth]{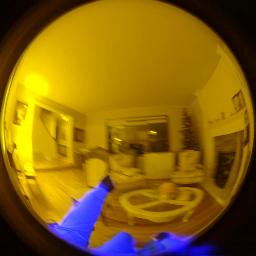}%
\includegraphics[width=0.05\linewidth, height=0.05\linewidth]{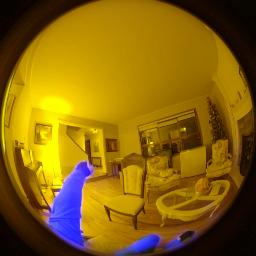}%
\includegraphics[width=0.05\linewidth, height=0.05\linewidth]{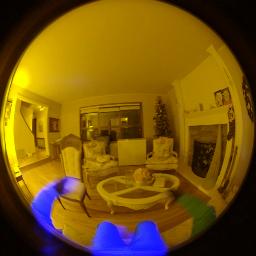}%
\includegraphics[width=0.05\linewidth, height=0.05\linewidth]{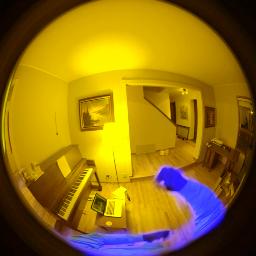}%
\includegraphics[width=0.05\linewidth, height=0.05\linewidth]{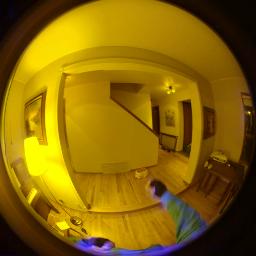}%
\includegraphics[width=0.05\linewidth, height=0.05\linewidth]{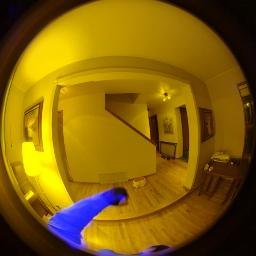}%
\includegraphics[width=0.05\linewidth, height=0.05\linewidth]{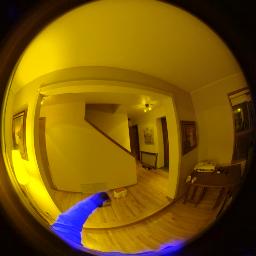}%
\includegraphics[width=0.05\linewidth, height=0.05\linewidth]{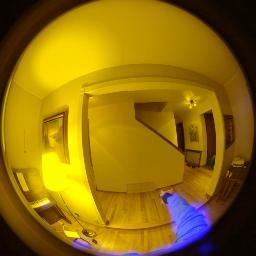}%
\includegraphics[width=0.05\linewidth, height=0.05\linewidth]{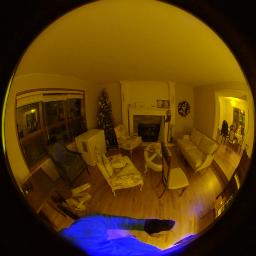}%
\\
 \rotatebox{90}{\hspace{4pt}{\tiny Ground Truth}} &
\includegraphics[width=0.05\linewidth, height=0.075\linewidth]{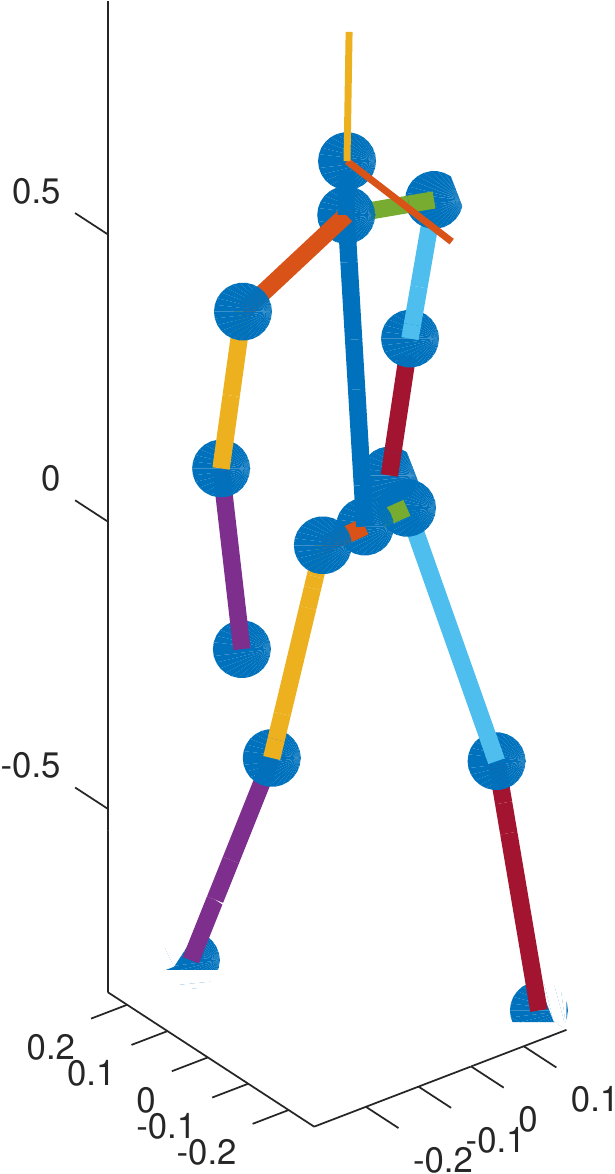}%
\includegraphics[width=0.05\linewidth, height=0.075\linewidth]{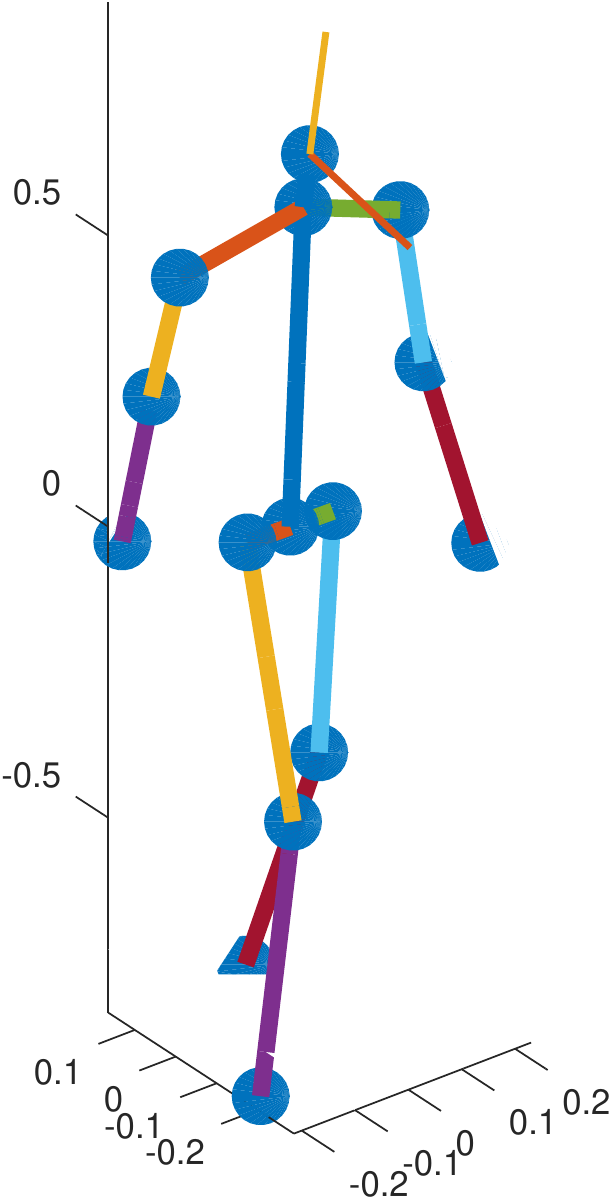}%
\includegraphics[width=0.05\linewidth, height=0.075\linewidth]{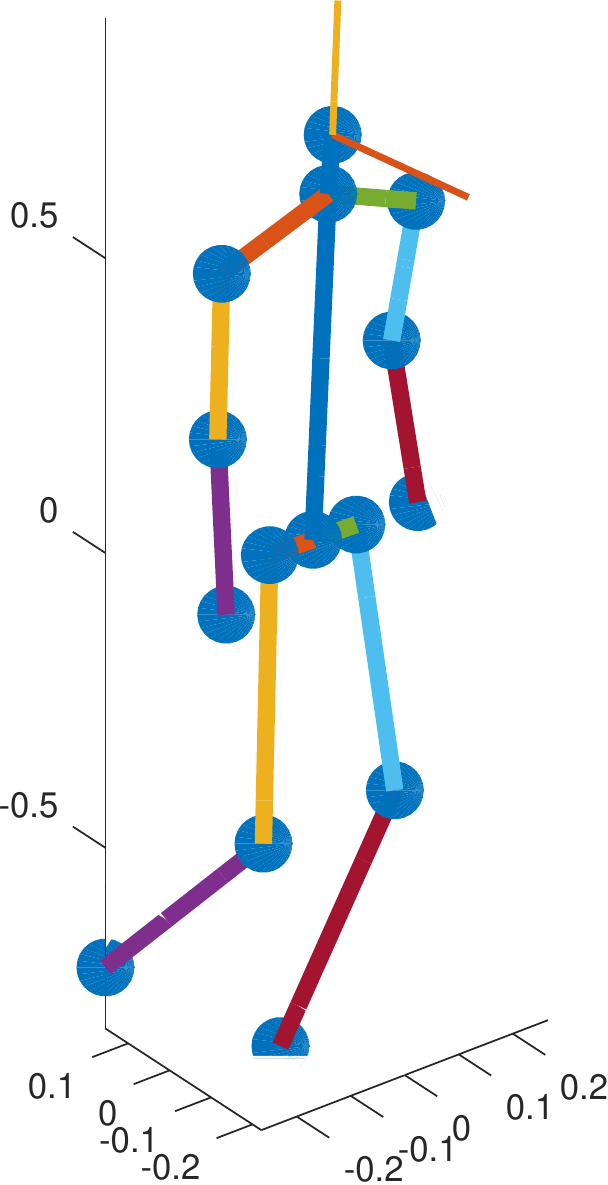}%
\includegraphics[width=0.05\linewidth, height=0.075\linewidth]{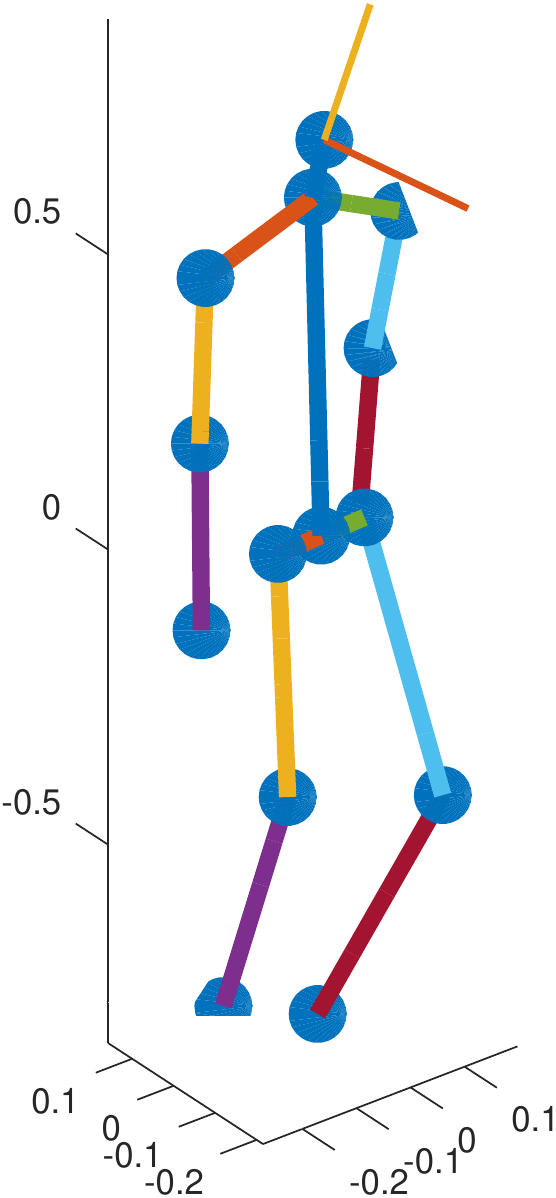}%
\includegraphics[width=0.05\linewidth, height=0.075\linewidth]{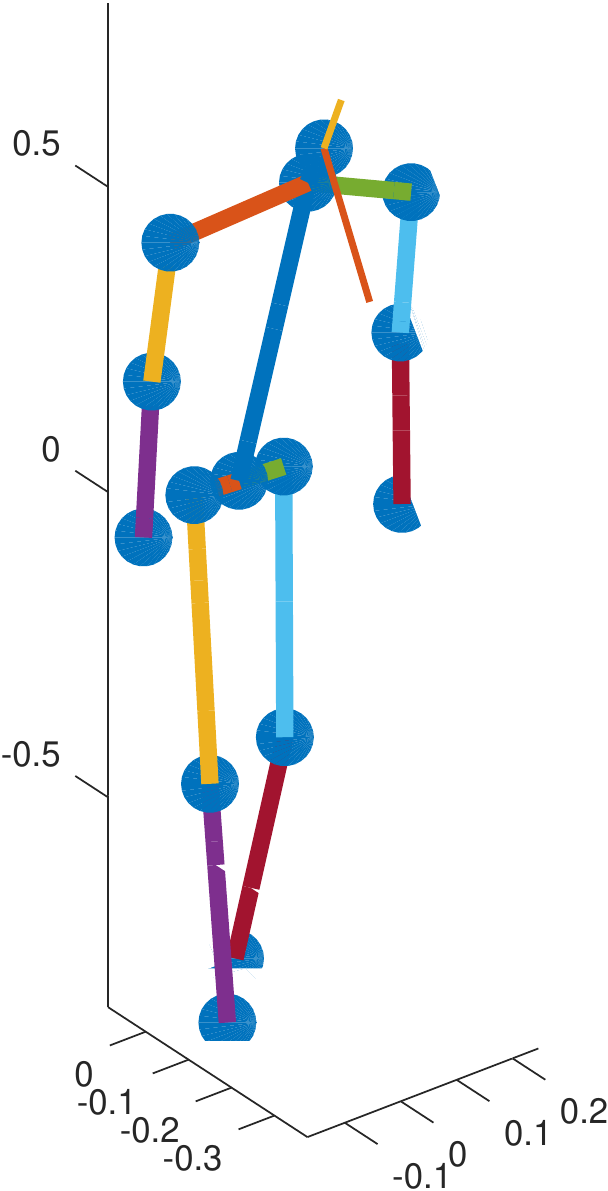}%
\includegraphics[width=0.05\linewidth, height=0.075\linewidth]{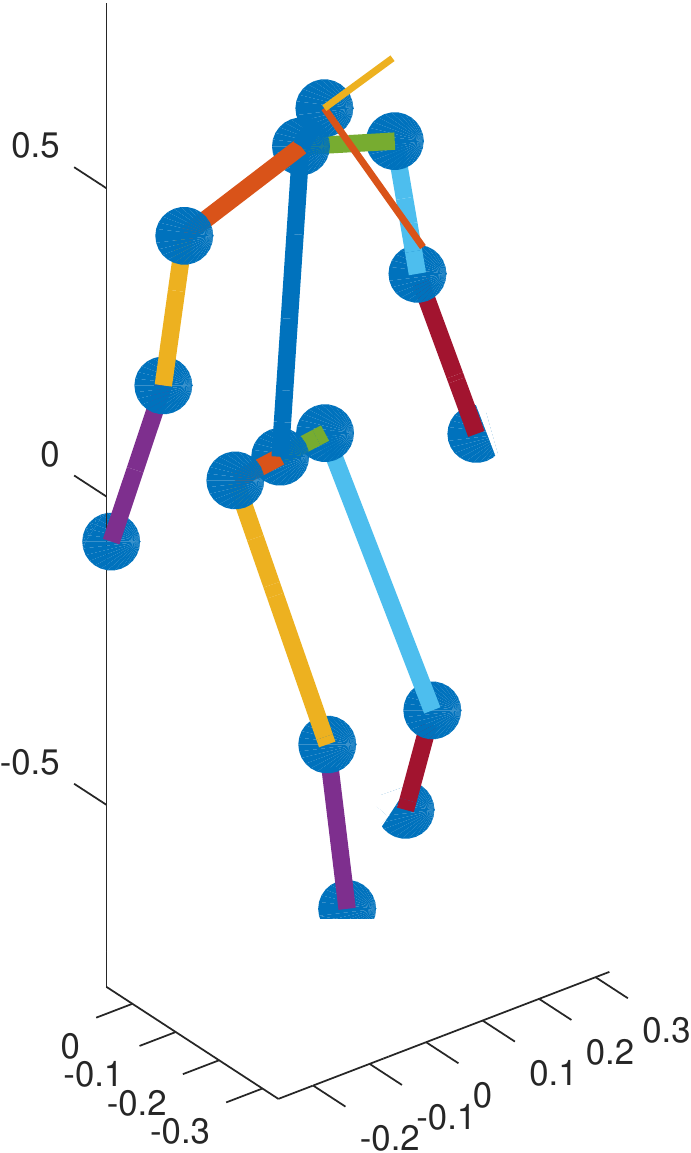}%
\includegraphics[width=0.05\linewidth, height=0.075\linewidth]{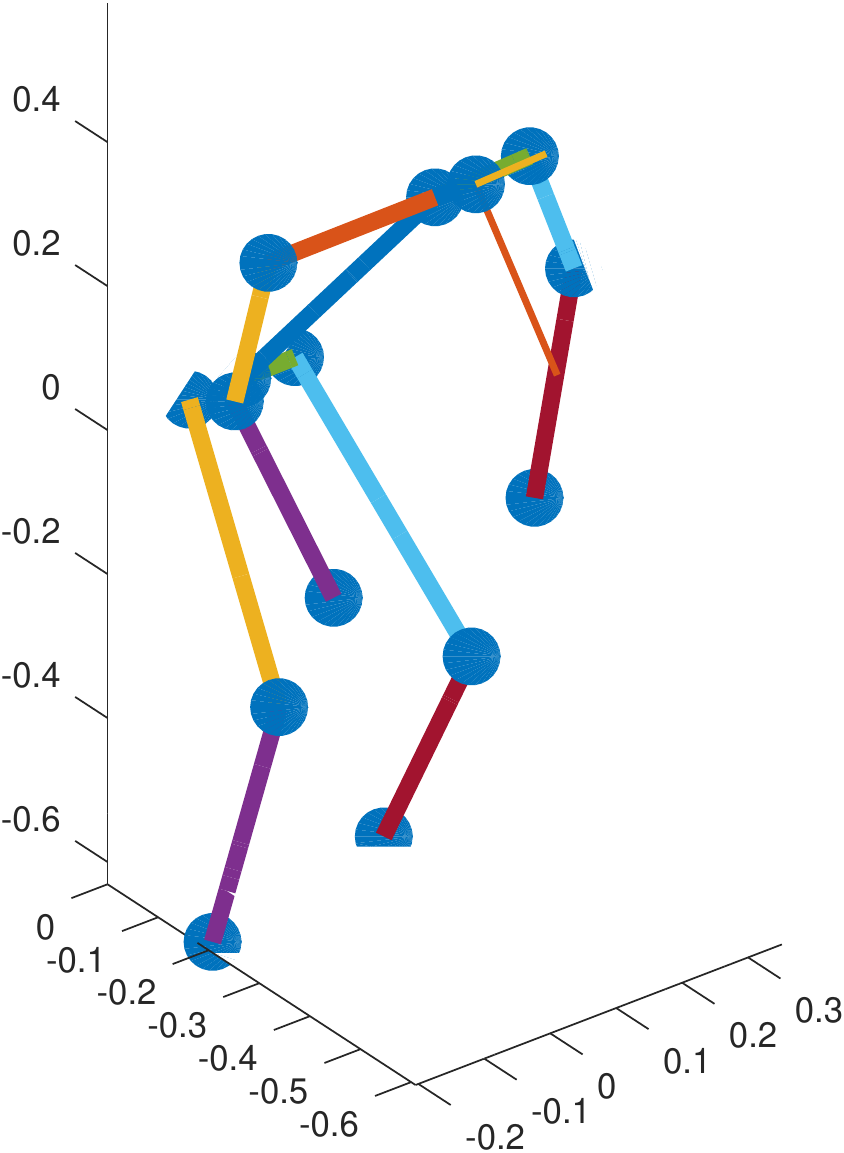}%
\includegraphics[width=0.05\linewidth, height=0.075\linewidth]{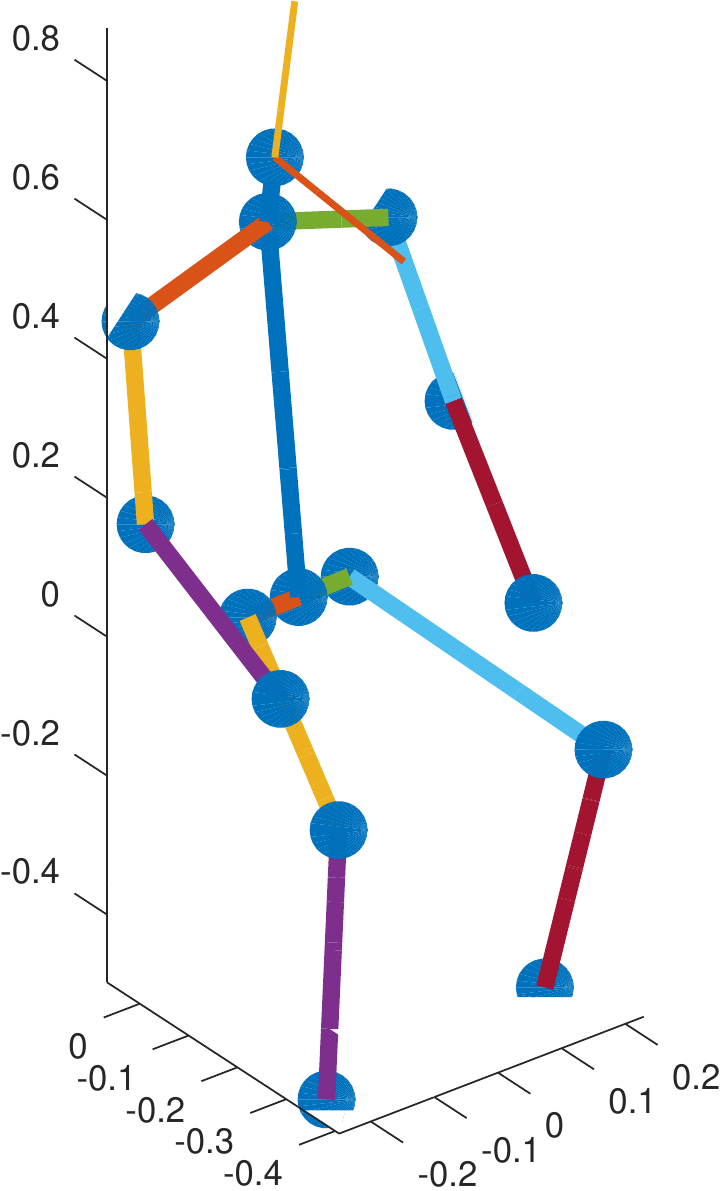}%
\includegraphics[width=0.05\linewidth, height=0.075\linewidth]{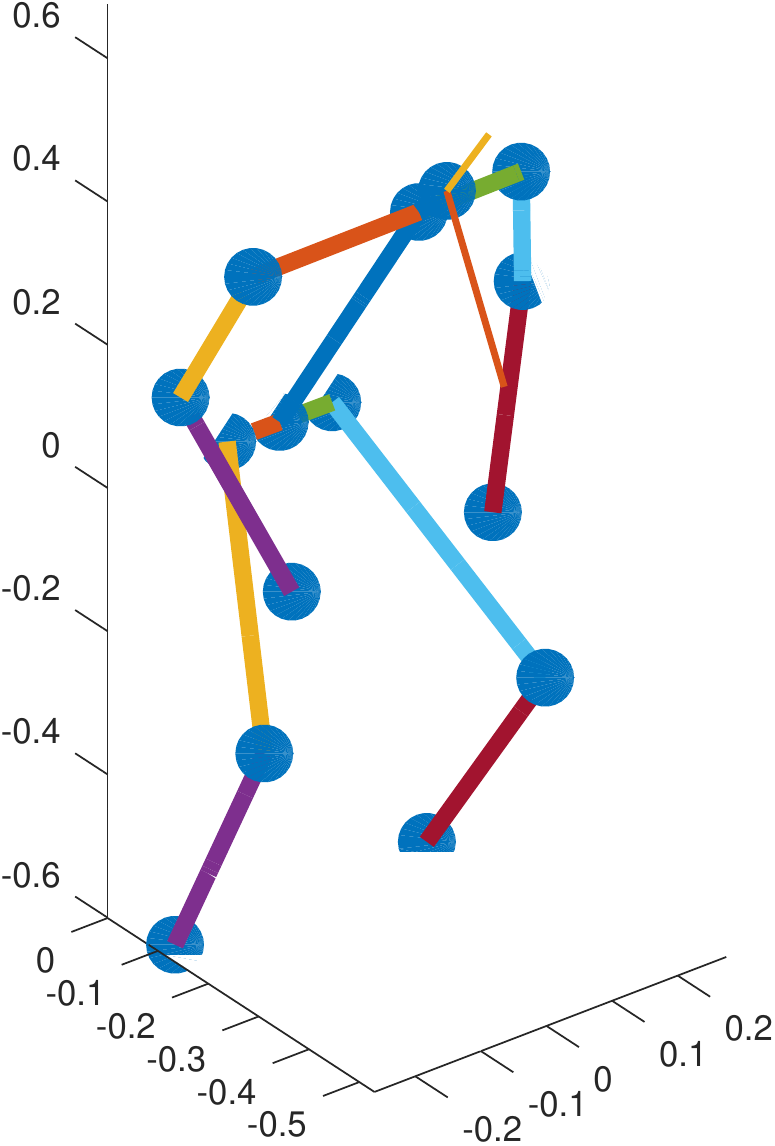}%
\includegraphics[width=0.05\linewidth, height=0.075\linewidth]{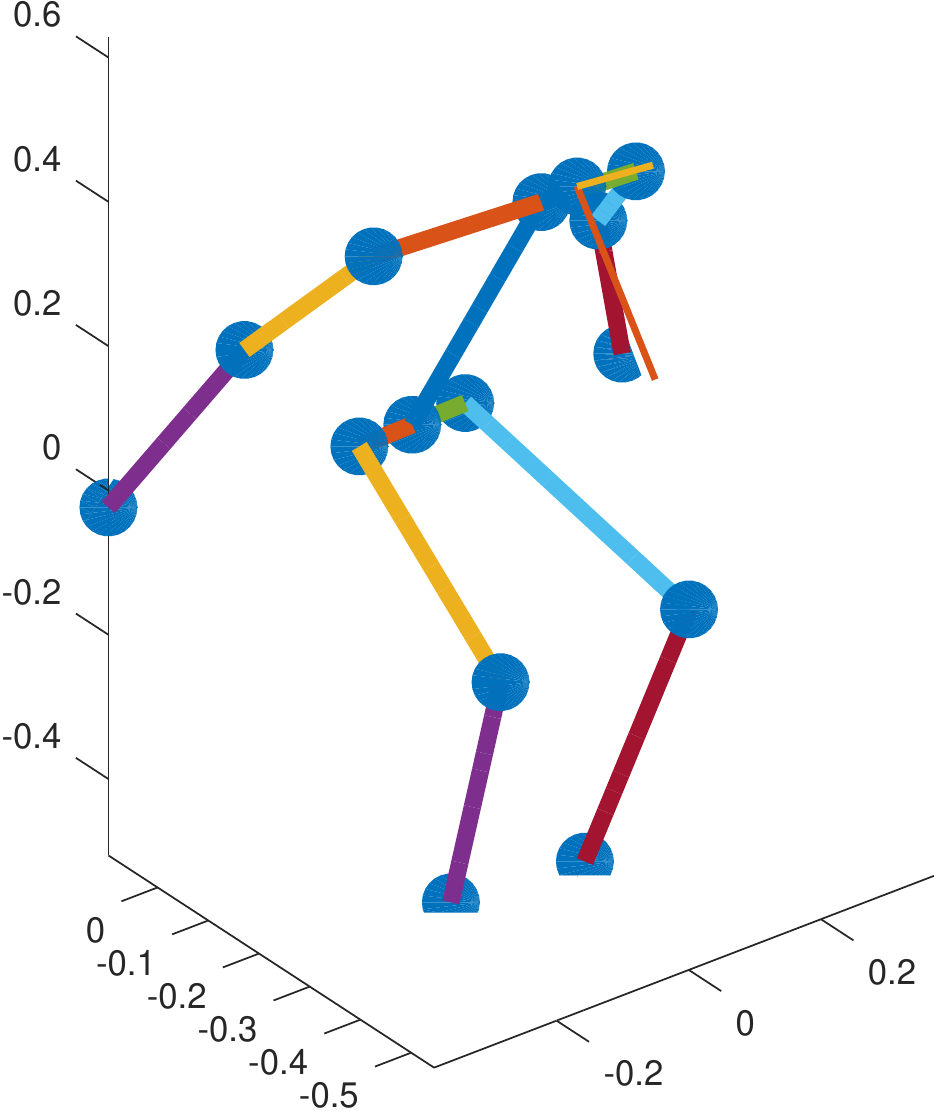}%
\includegraphics[width=0.05\linewidth, height=0.075\linewidth]{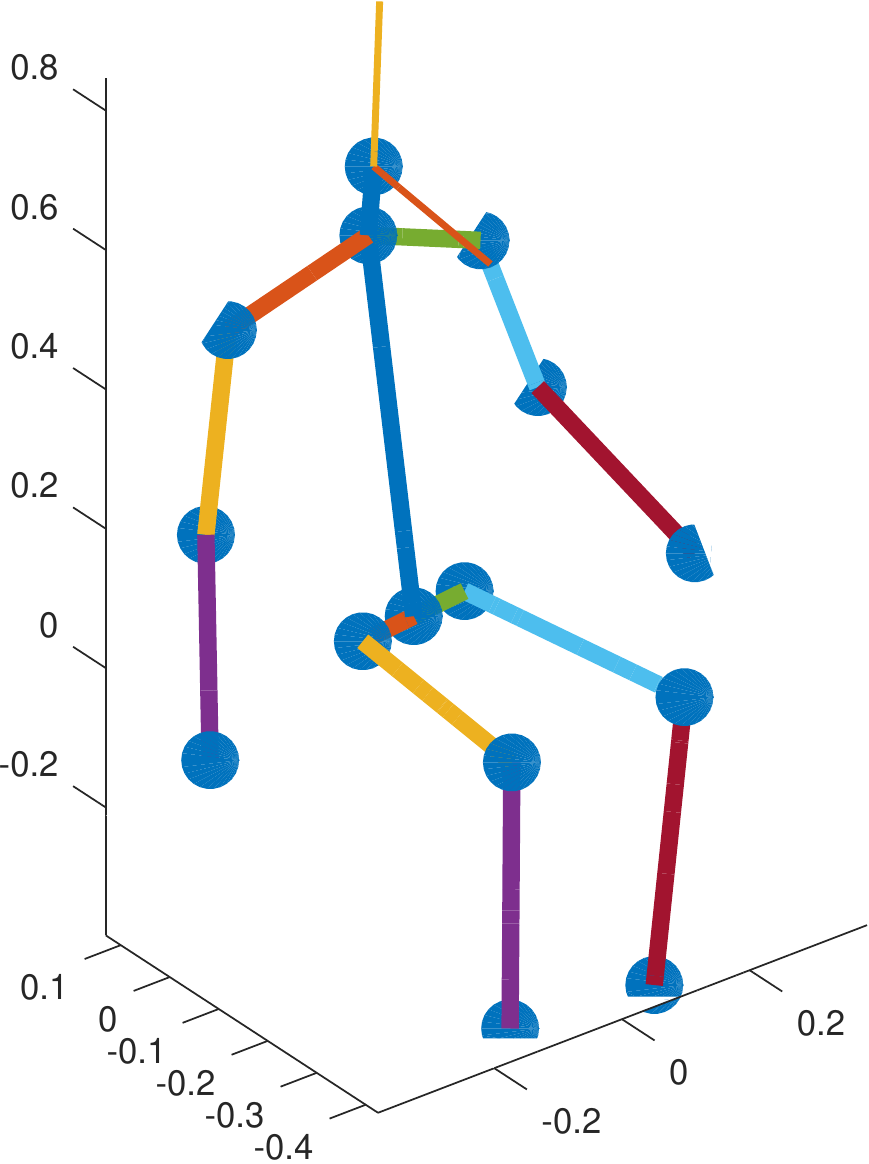}%
\includegraphics[width=0.05\linewidth, height=0.075\linewidth]{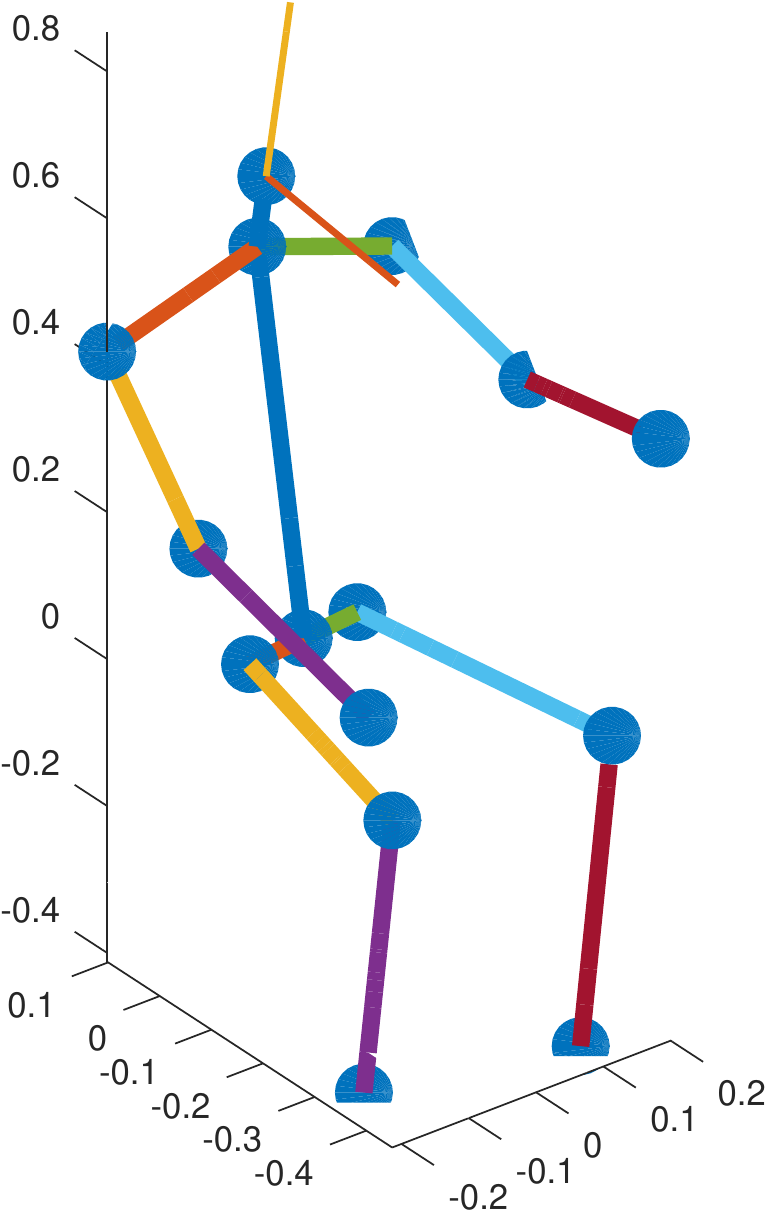}%
\includegraphics[width=0.05\linewidth, height=0.075\linewidth]{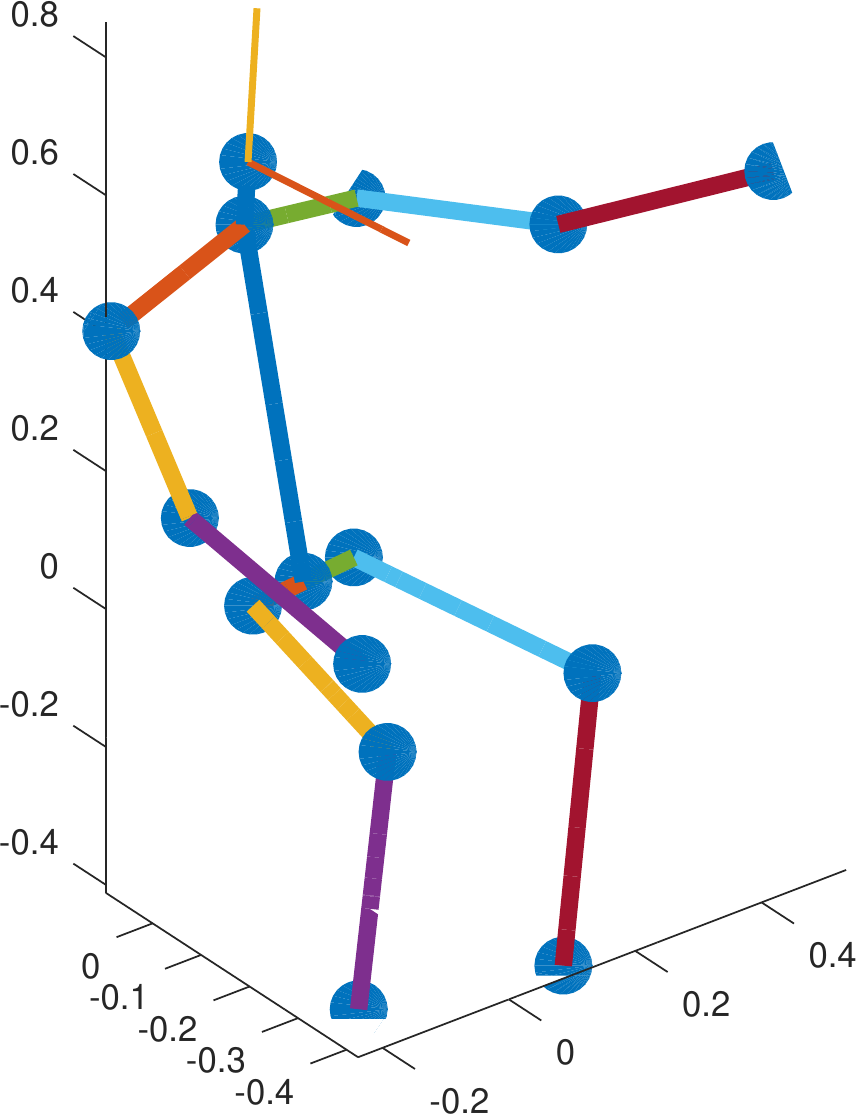}%
\includegraphics[width=0.05\linewidth, height=0.075\linewidth]{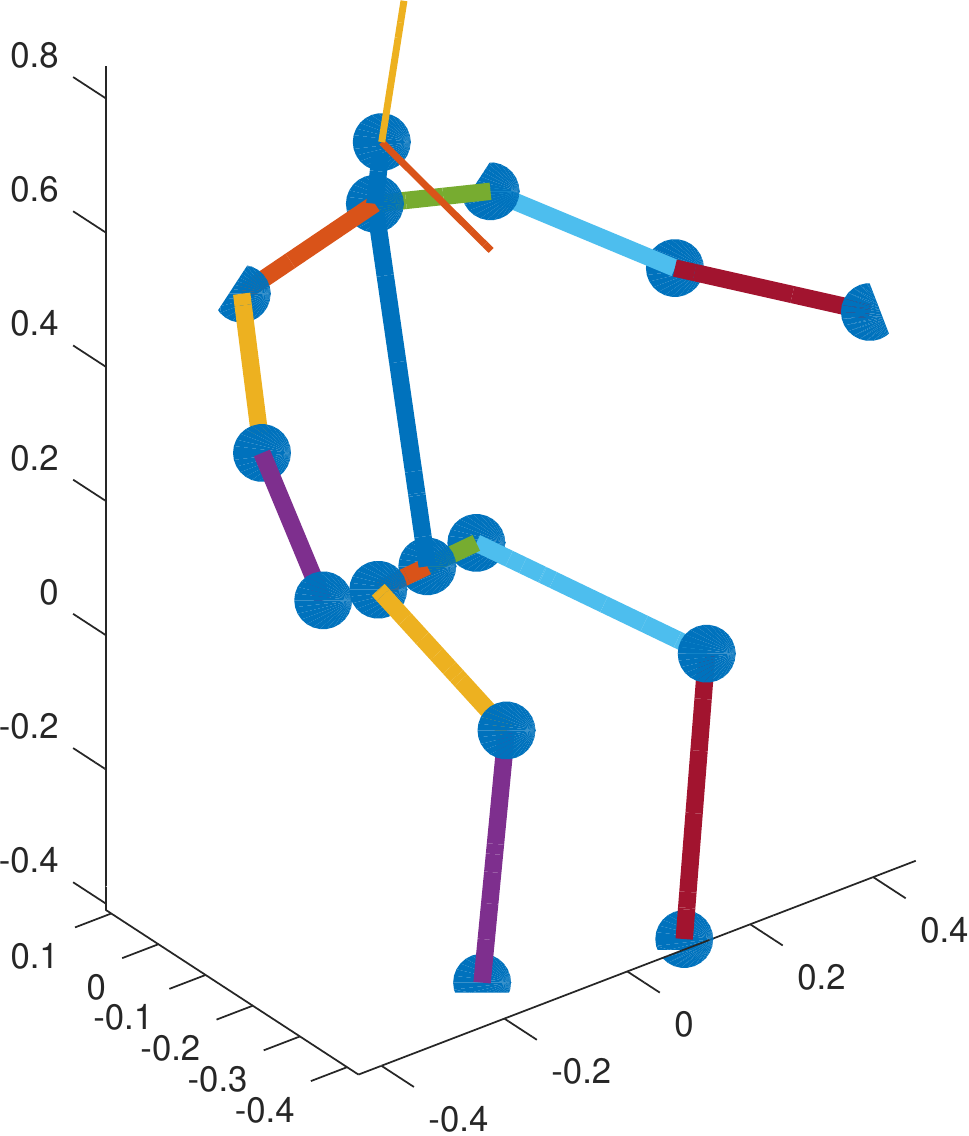}%
\includegraphics[width=0.05\linewidth, height=0.075\linewidth]{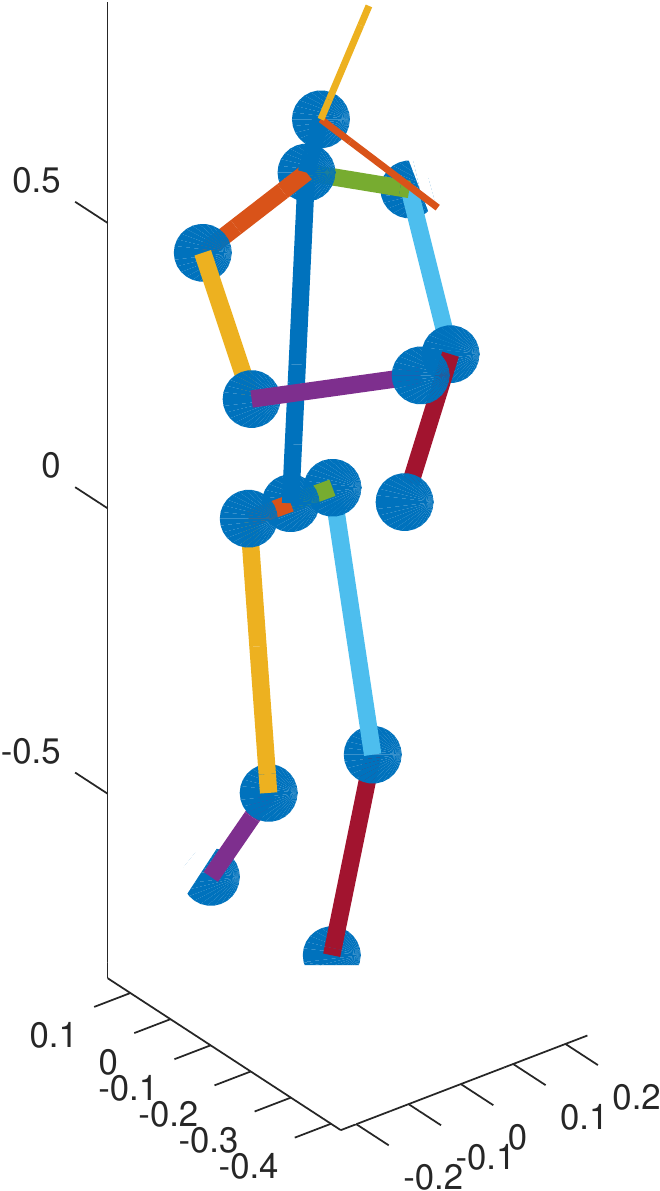}%
\includegraphics[width=0.05\linewidth, height=0.075\linewidth]{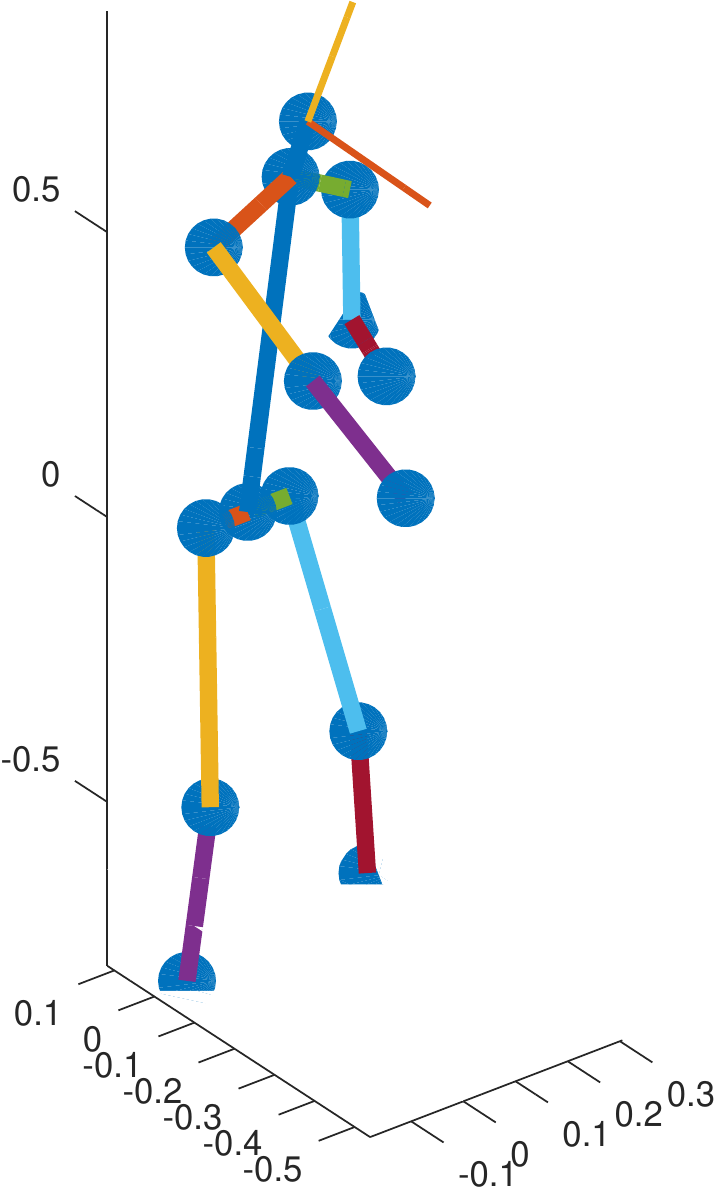}%
\includegraphics[width=0.05\linewidth, height=0.075\linewidth]{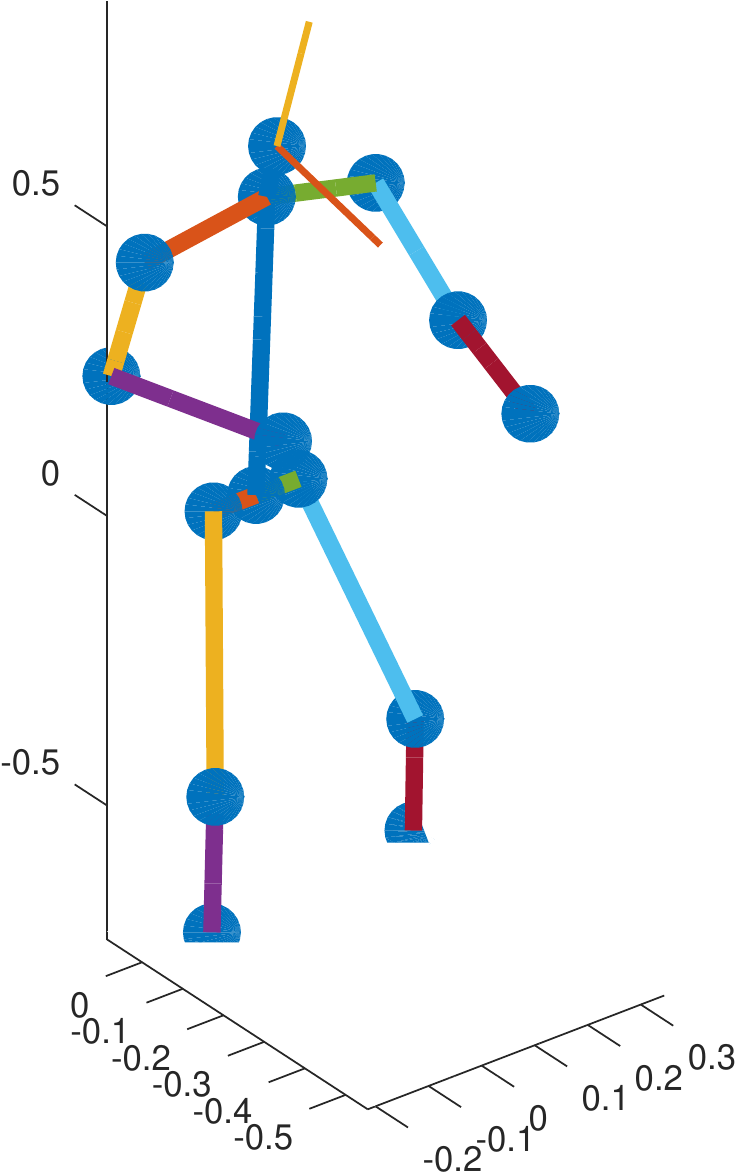}%
\includegraphics[width=0.05\linewidth, height=0.075\linewidth]{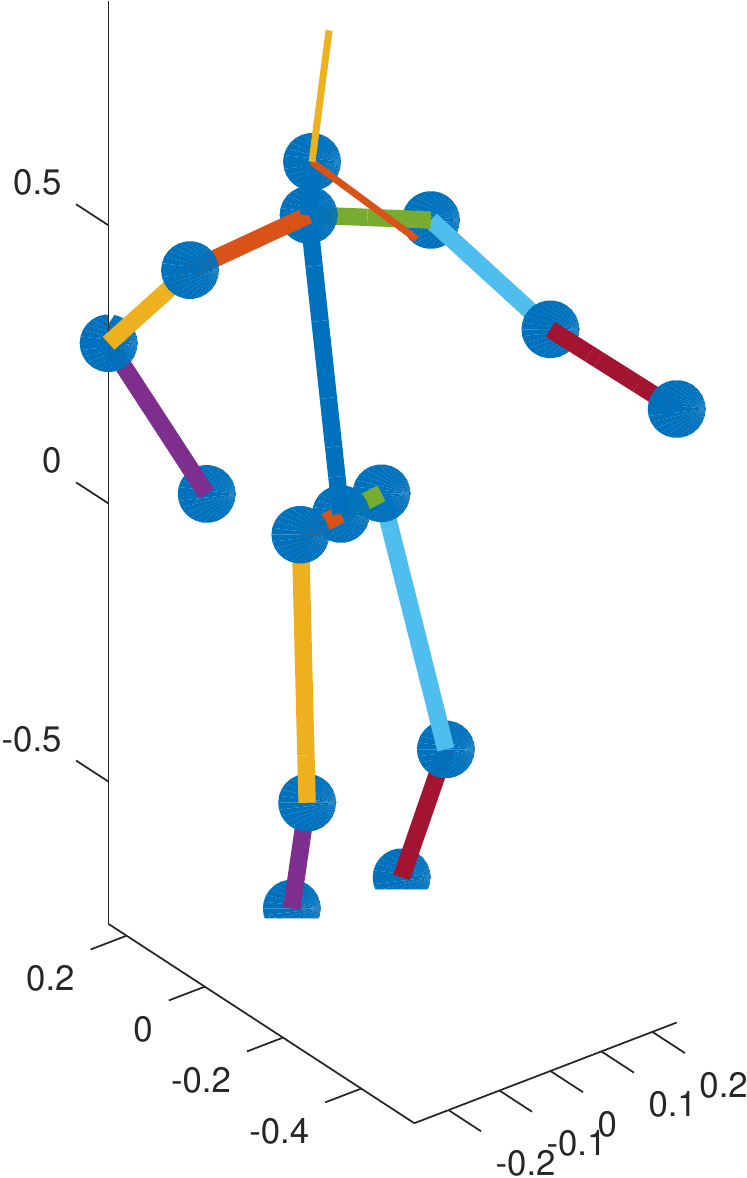}%
\includegraphics[width=0.05\linewidth, height=0.075\linewidth]{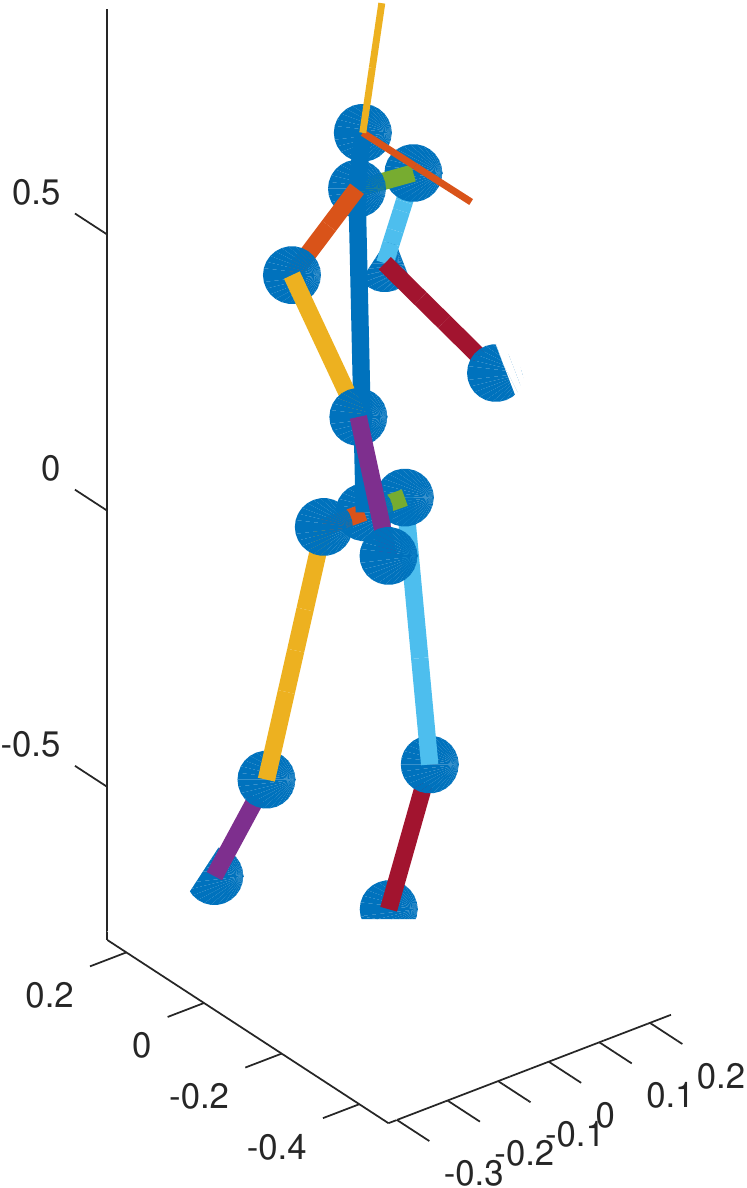}%
\includegraphics[width=0.05\linewidth, height=0.075\linewidth]{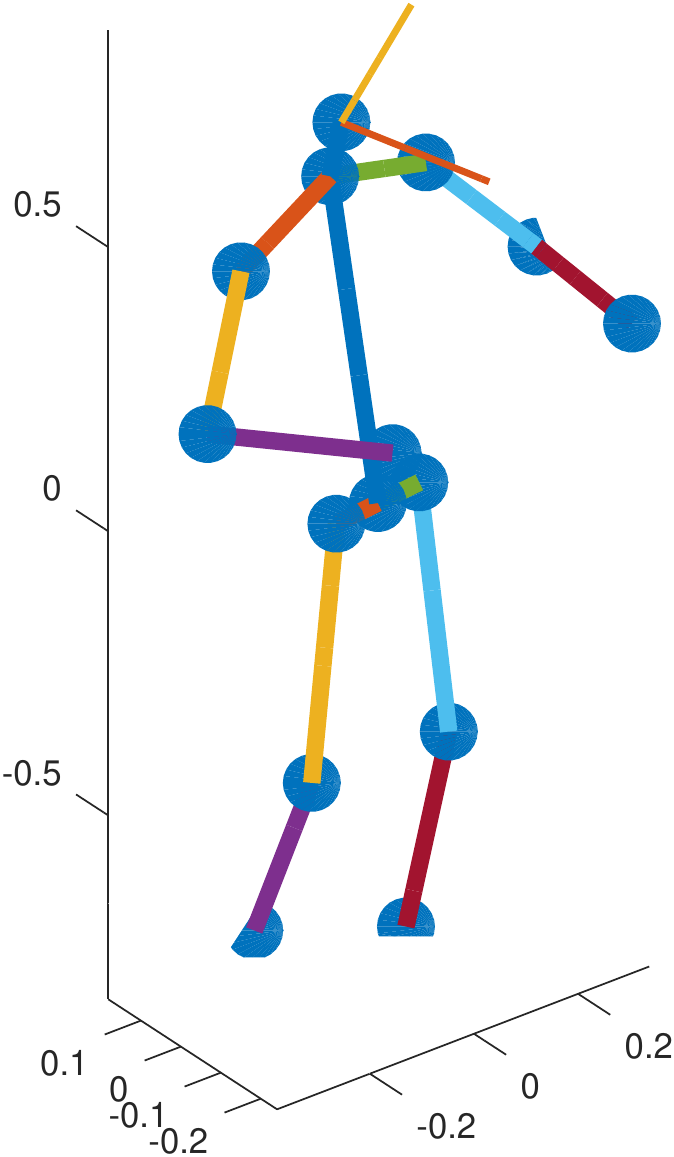}%
\\
 \rotatebox{90}{\hspace{15pt}{\tiny Ours}} &
\includegraphics[width=0.05\linewidth, height=0.075\linewidth]{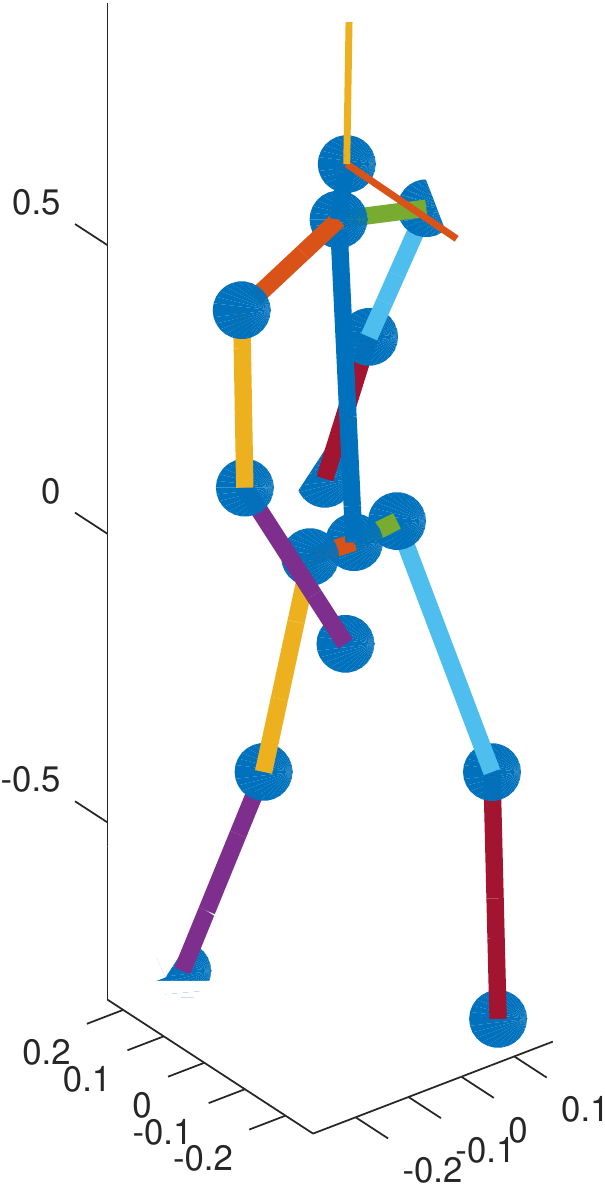}%
\includegraphics[width=0.05\linewidth, height=0.075\linewidth]{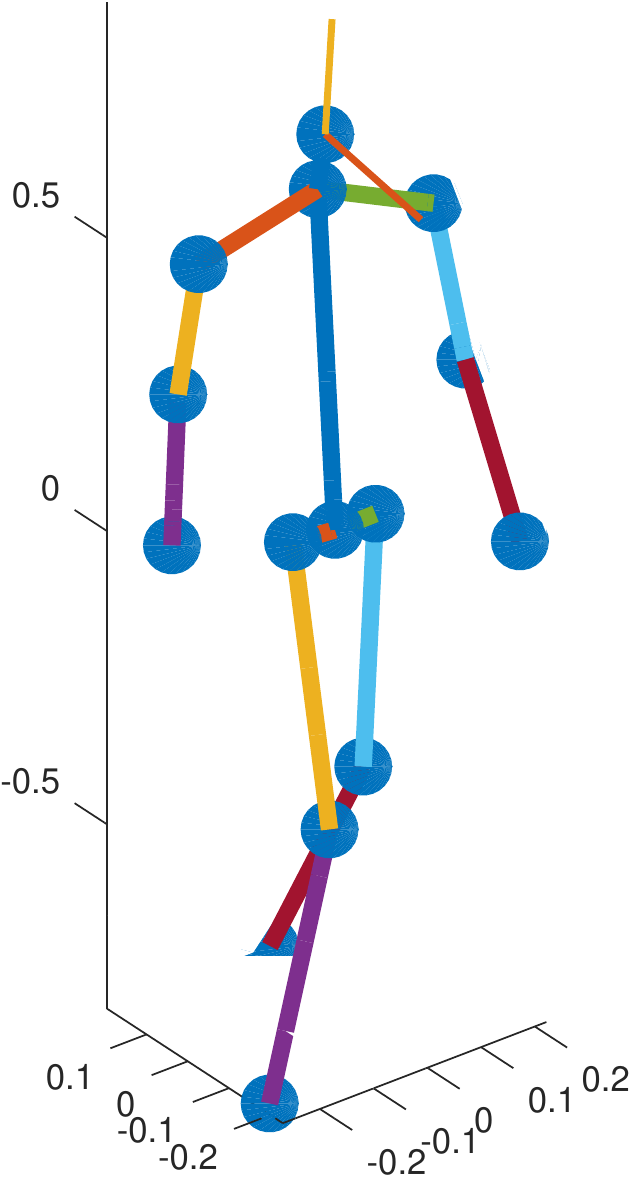}%
\includegraphics[width=0.05\linewidth, height=0.075\linewidth]{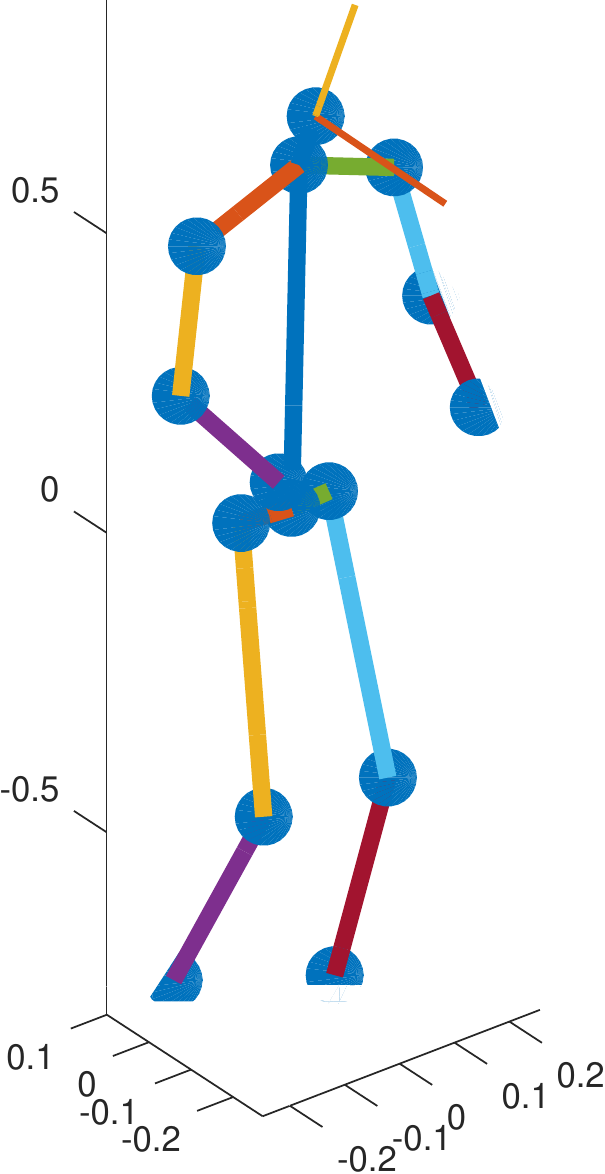}%
\includegraphics[width=0.05\linewidth, height=0.075\linewidth]{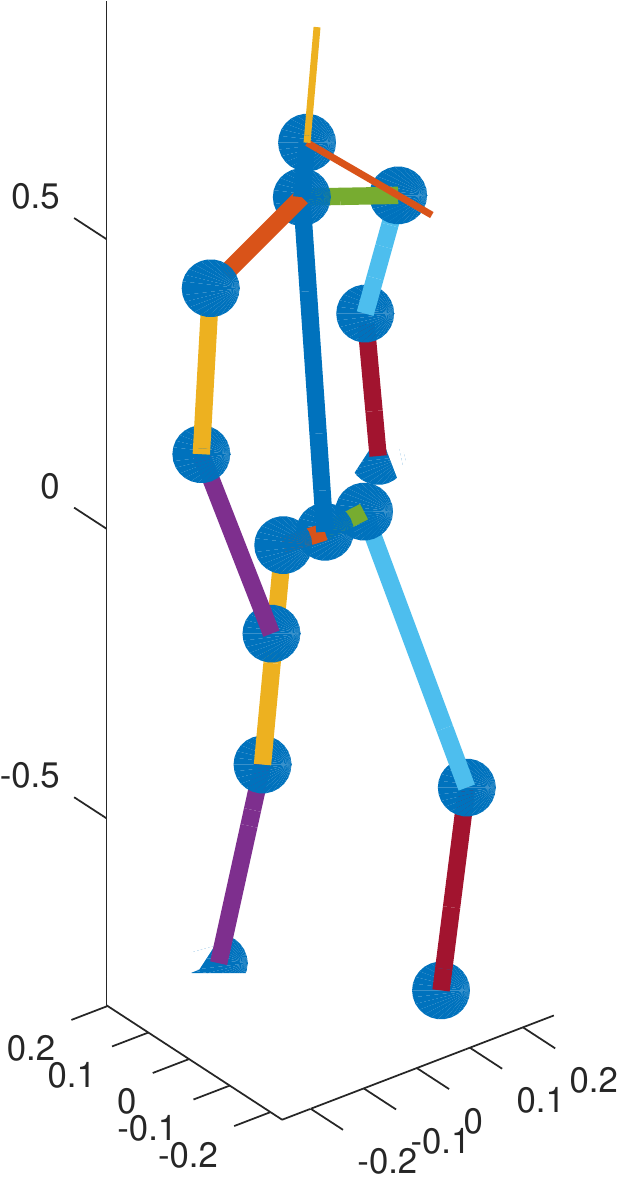}%
\includegraphics[width=0.05\linewidth, height=0.075\linewidth]{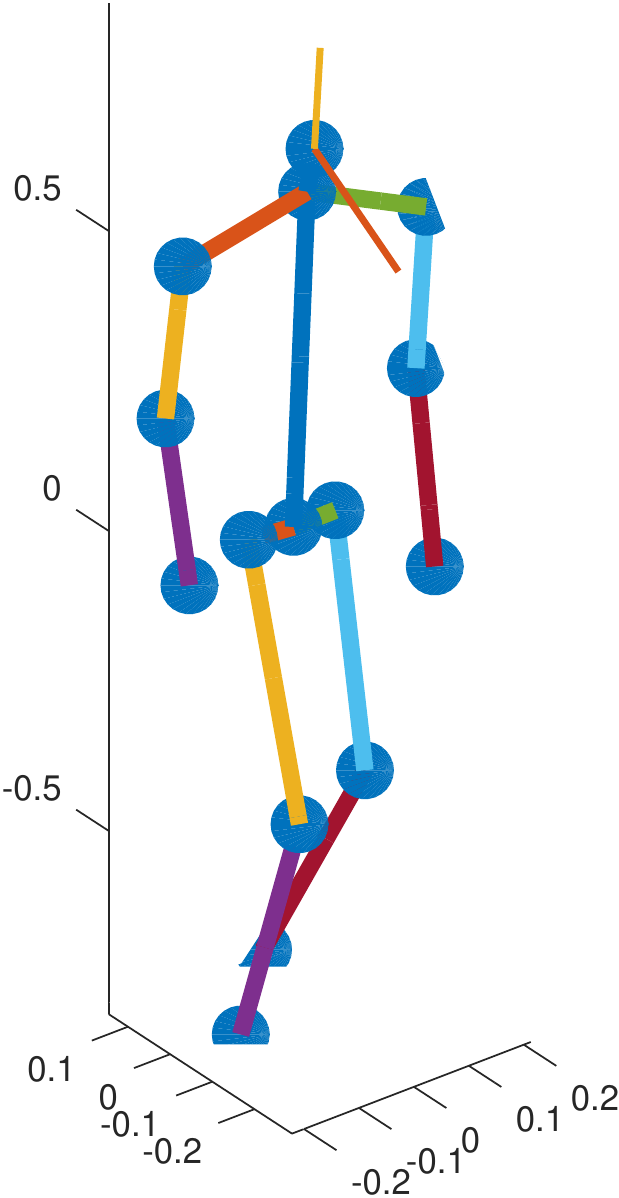}%
\includegraphics[width=0.05\linewidth, height=0.075\linewidth]{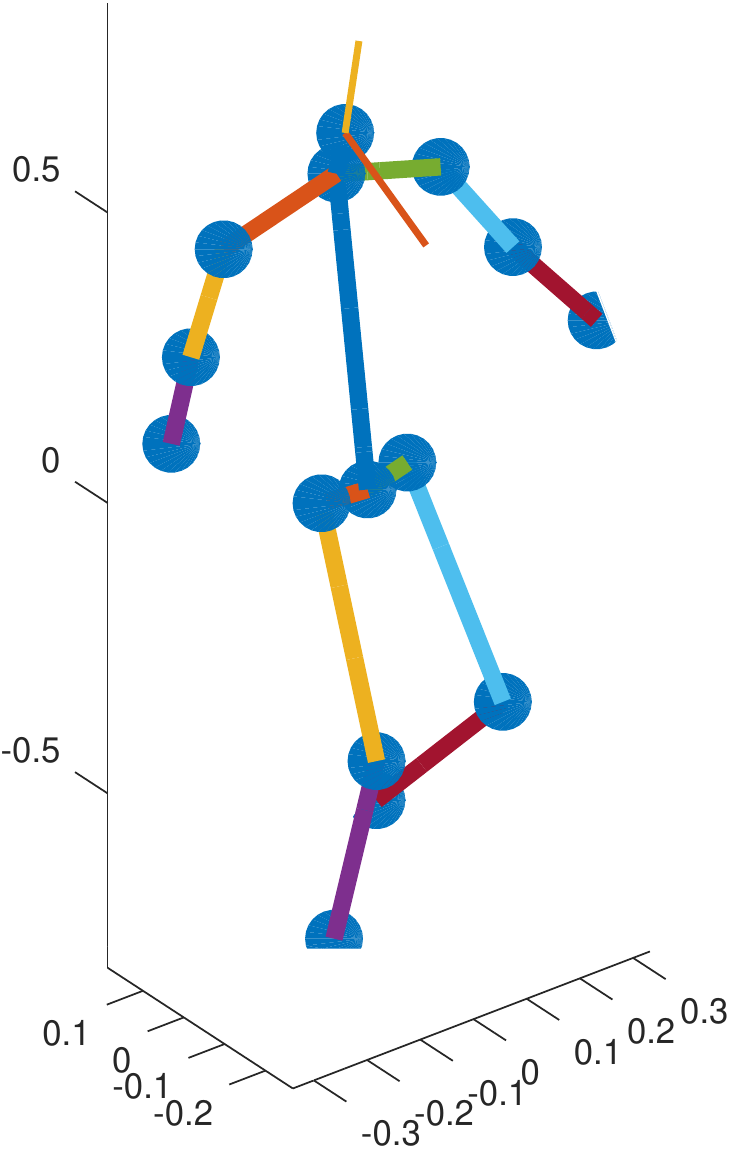}%
\includegraphics[width=0.05\linewidth, height=0.075\linewidth]{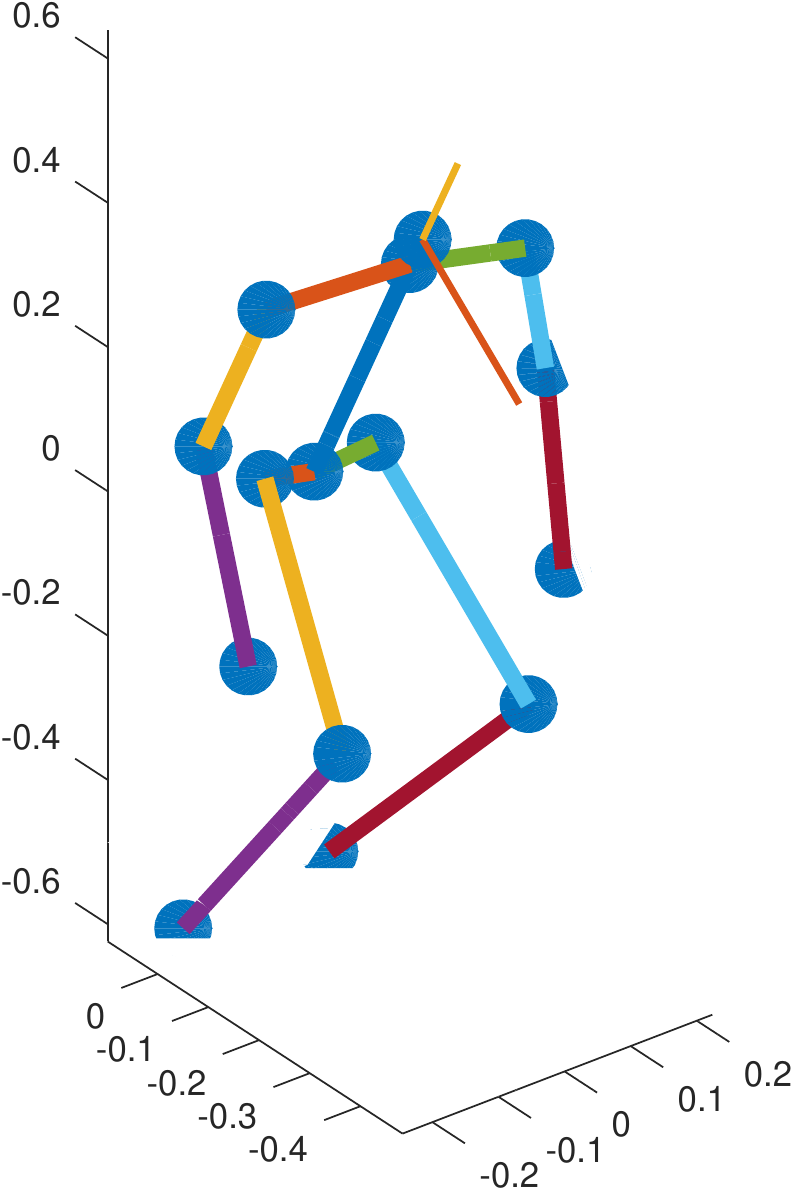}%
\includegraphics[width=0.05\linewidth, height=0.075\linewidth]{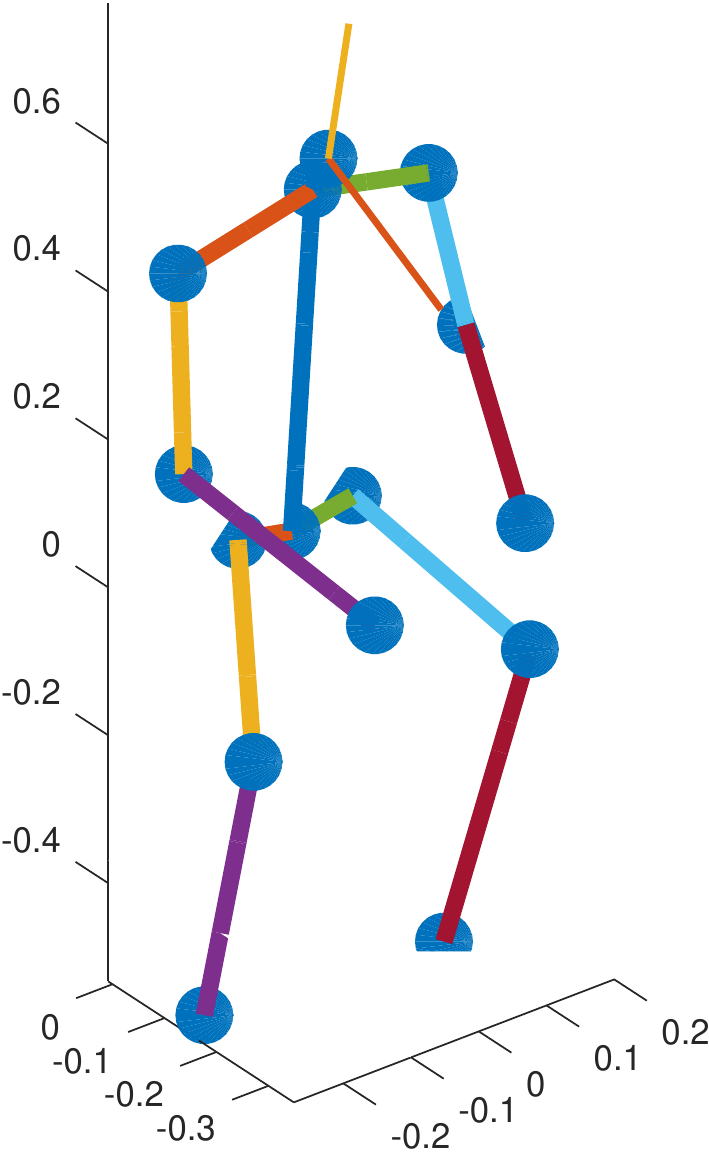}%
\includegraphics[width=0.05\linewidth, height=0.075\linewidth]{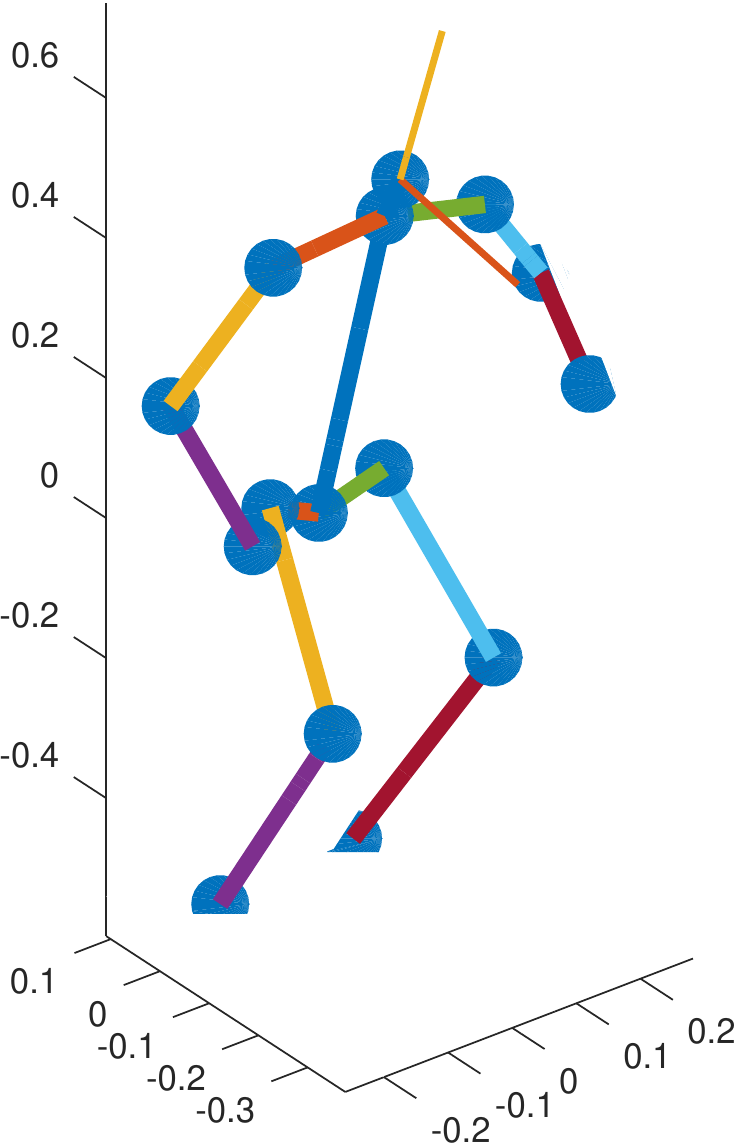}%
\includegraphics[width=0.05\linewidth, height=0.075\linewidth]{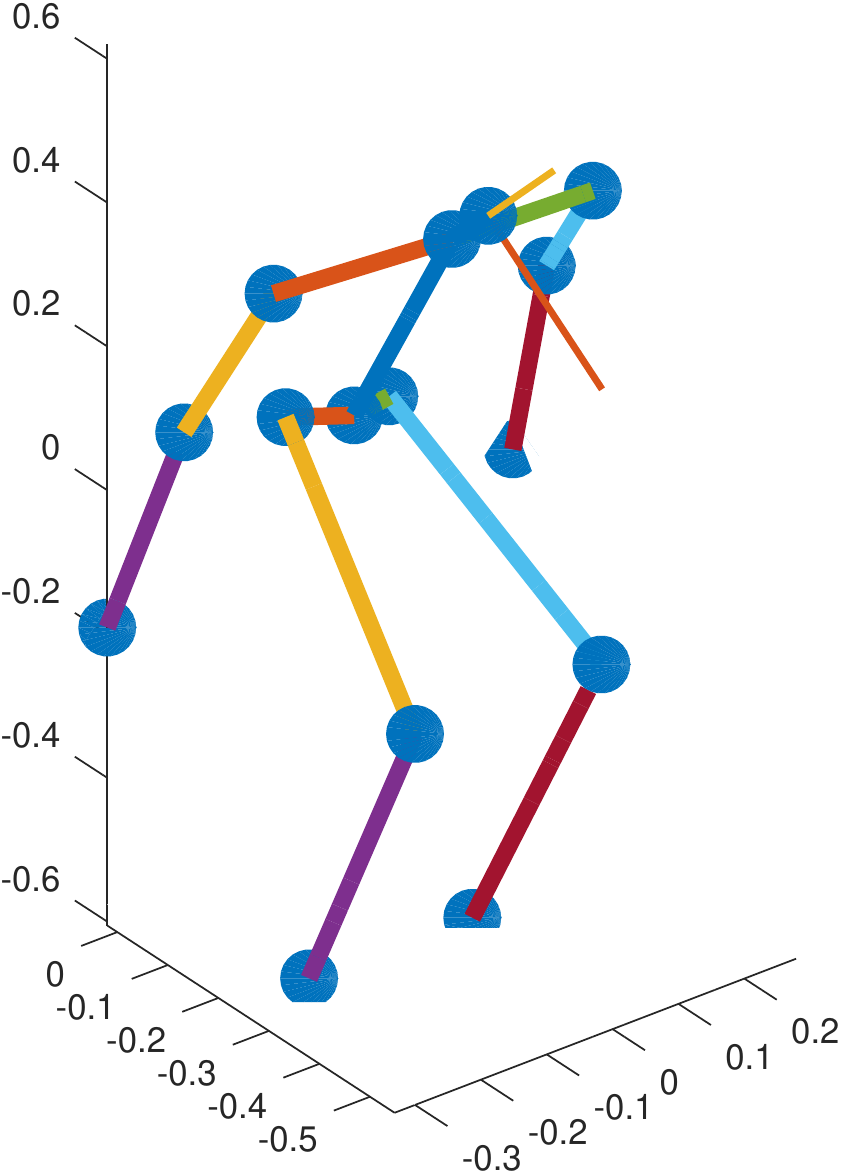}%
\includegraphics[width=0.05\linewidth, height=0.075\linewidth]{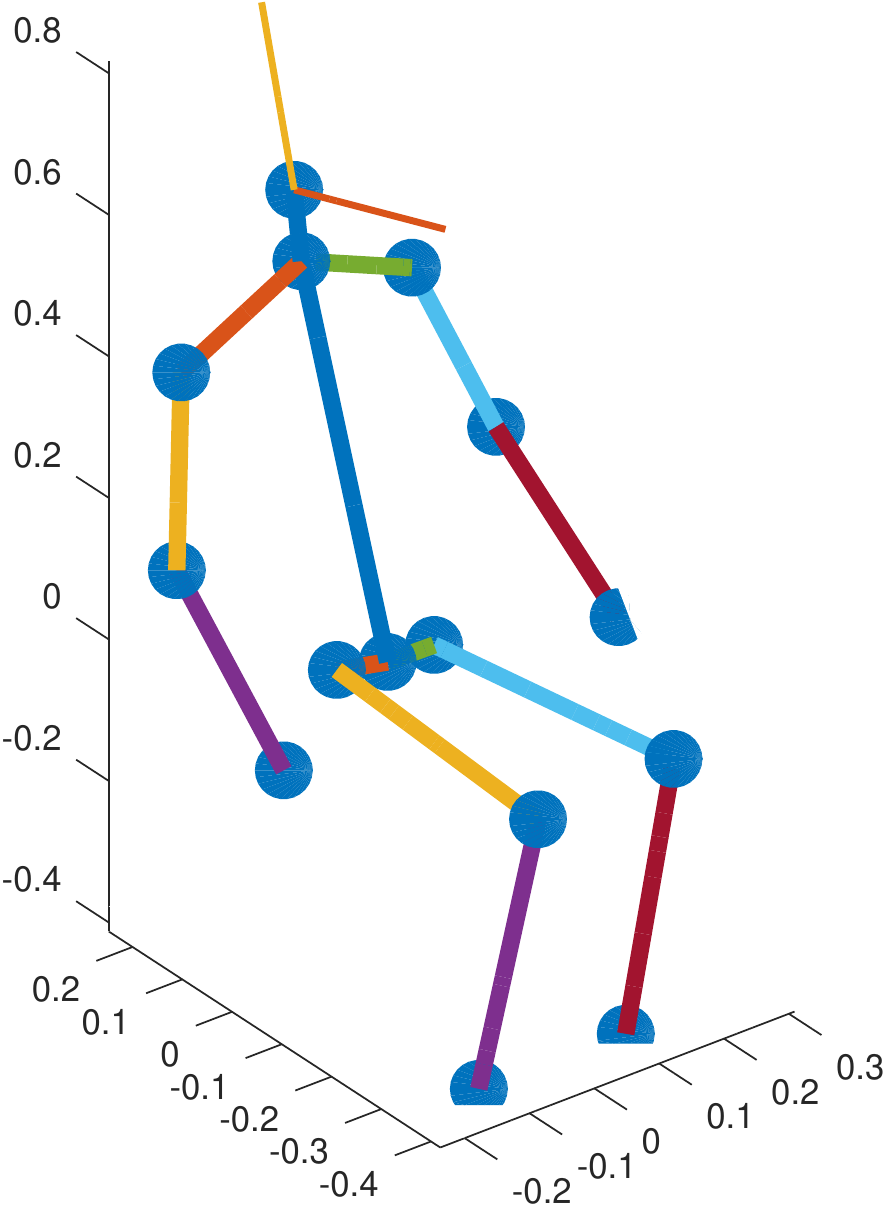}%
\includegraphics[width=0.05\linewidth, height=0.075\linewidth]{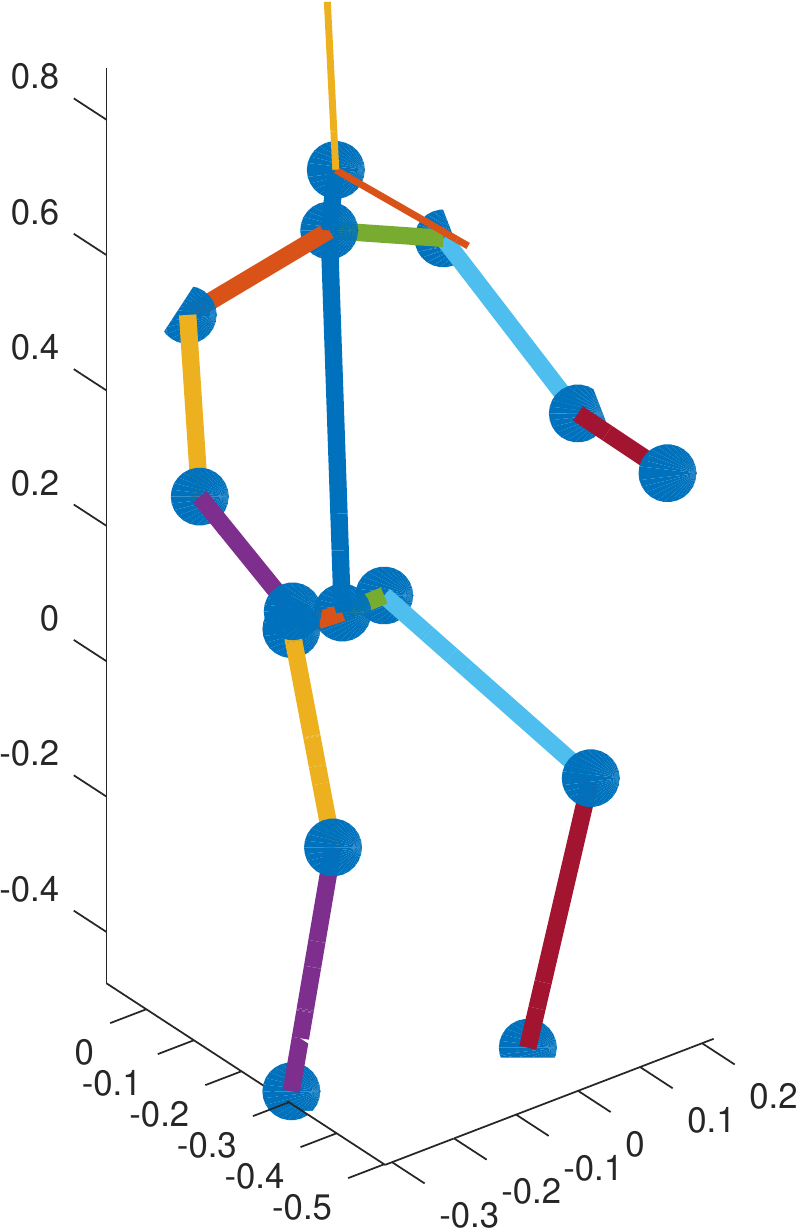}%
\includegraphics[width=0.05\linewidth, height=0.075\linewidth]{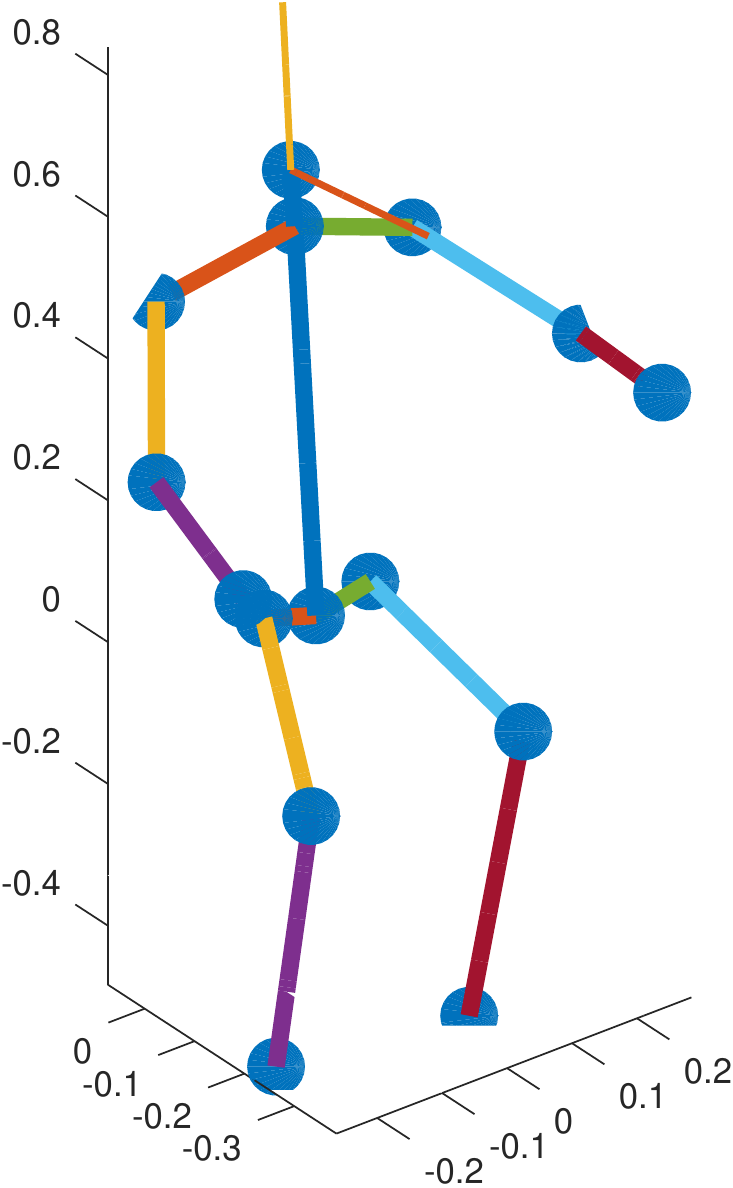}%
\includegraphics[width=0.05\linewidth, height=0.075\linewidth]{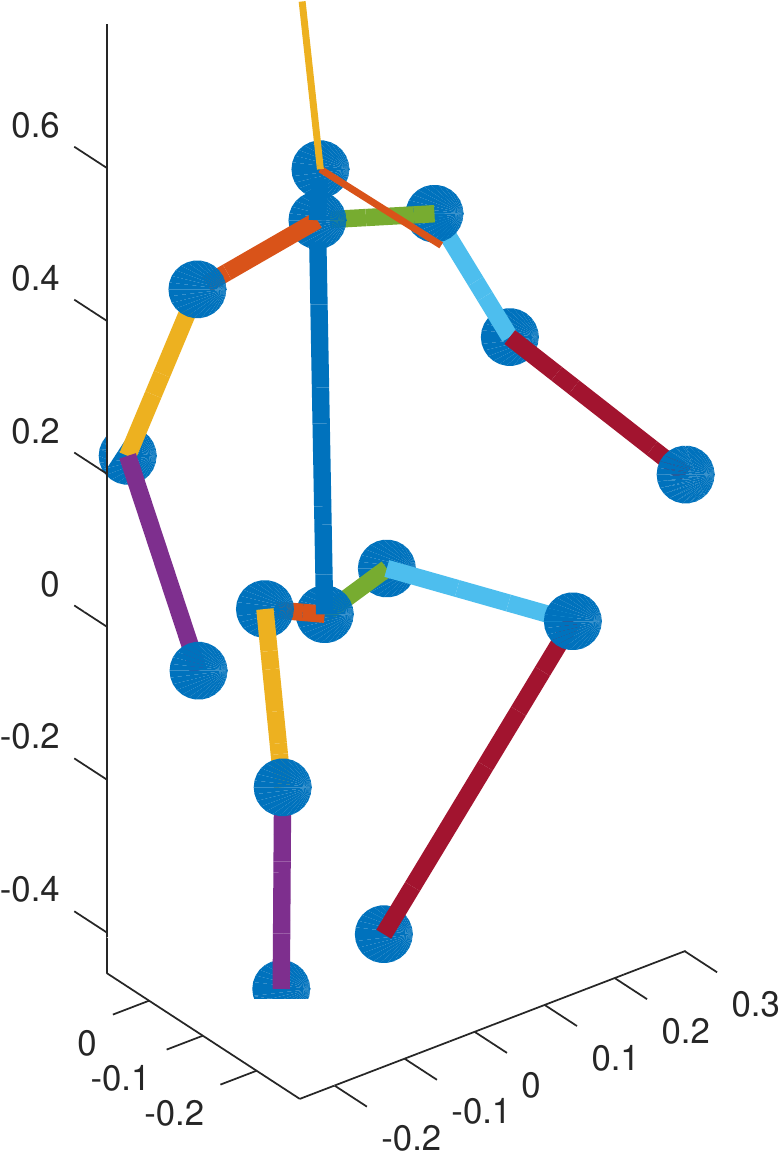}%
\includegraphics[width=0.05\linewidth, height=0.075\linewidth]{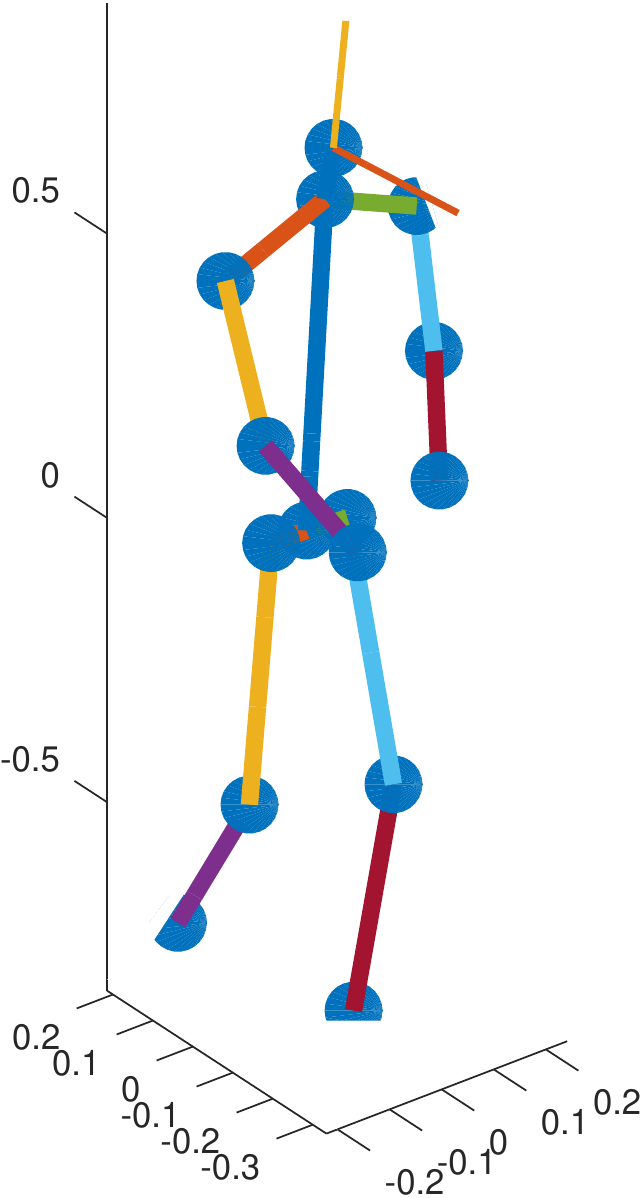}%
\includegraphics[width=0.05\linewidth, height=0.075\linewidth]{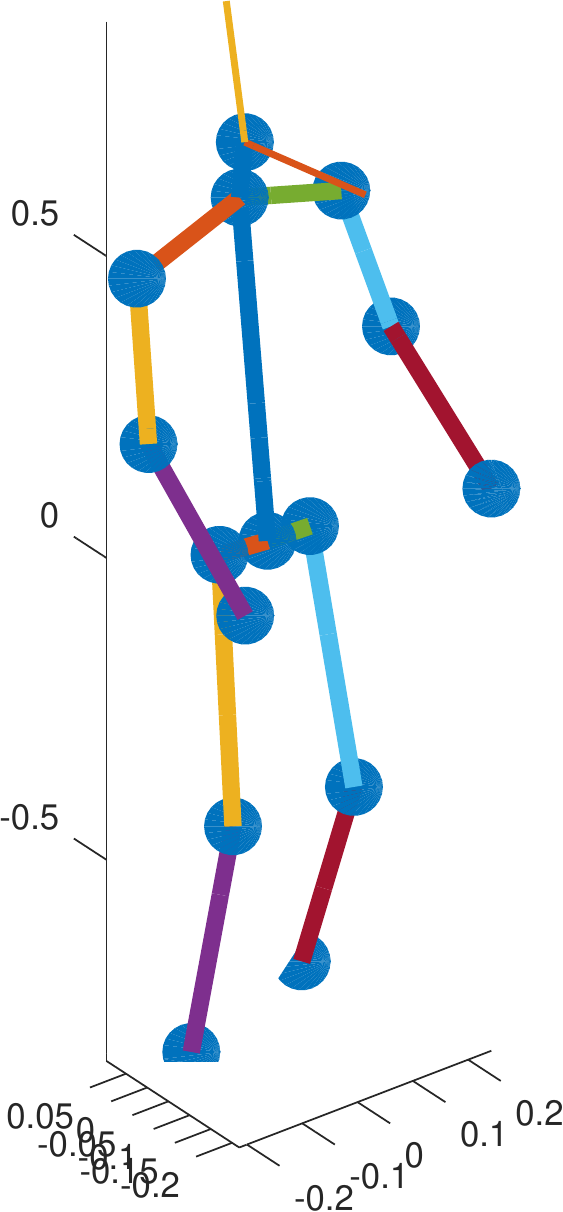}%
\includegraphics[width=0.05\linewidth, height=0.075\linewidth]{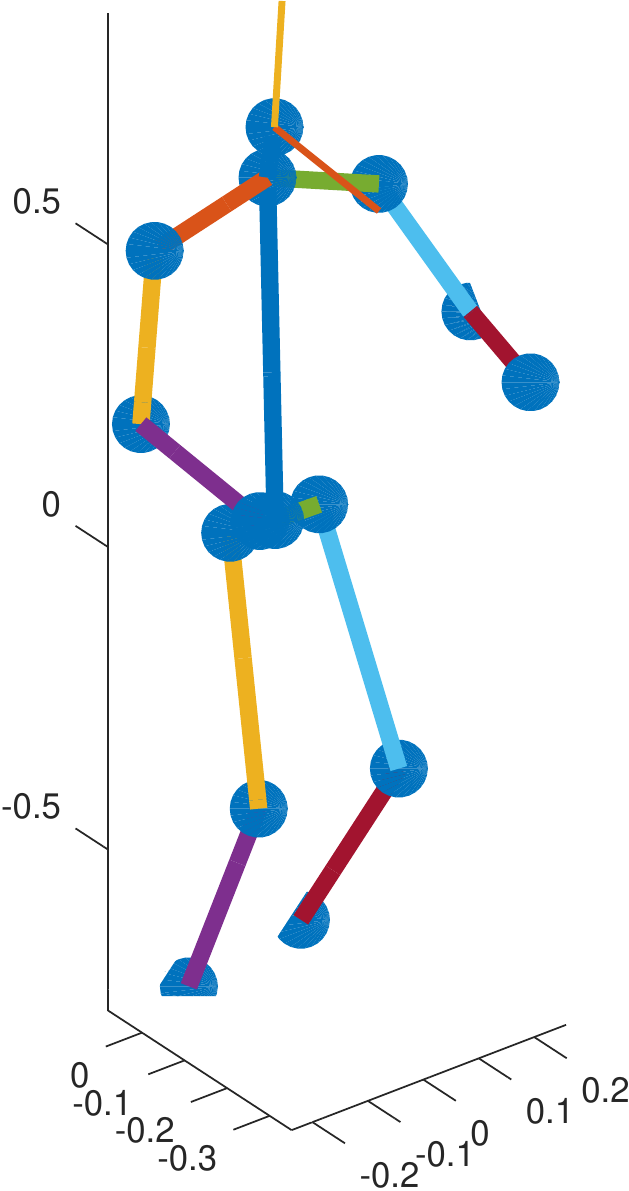}%
\includegraphics[width=0.05\linewidth, height=0.075\linewidth]{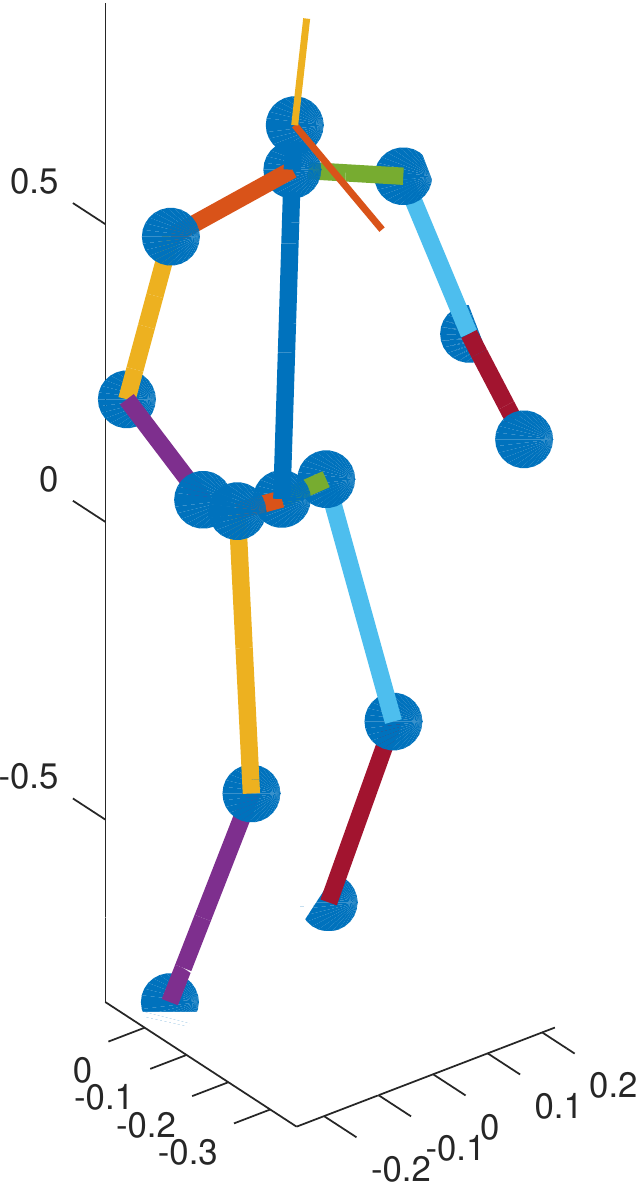}%
\includegraphics[width=0.05\linewidth, height=0.075\linewidth]{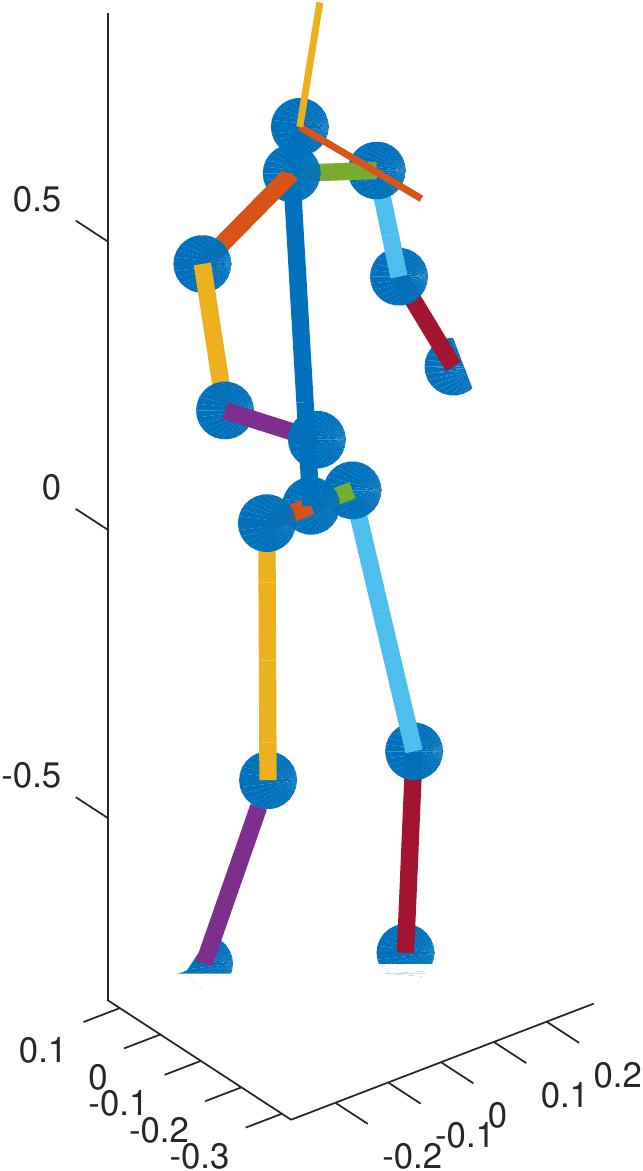}%
\includegraphics[width=0.05\linewidth, height=0.075\linewidth]{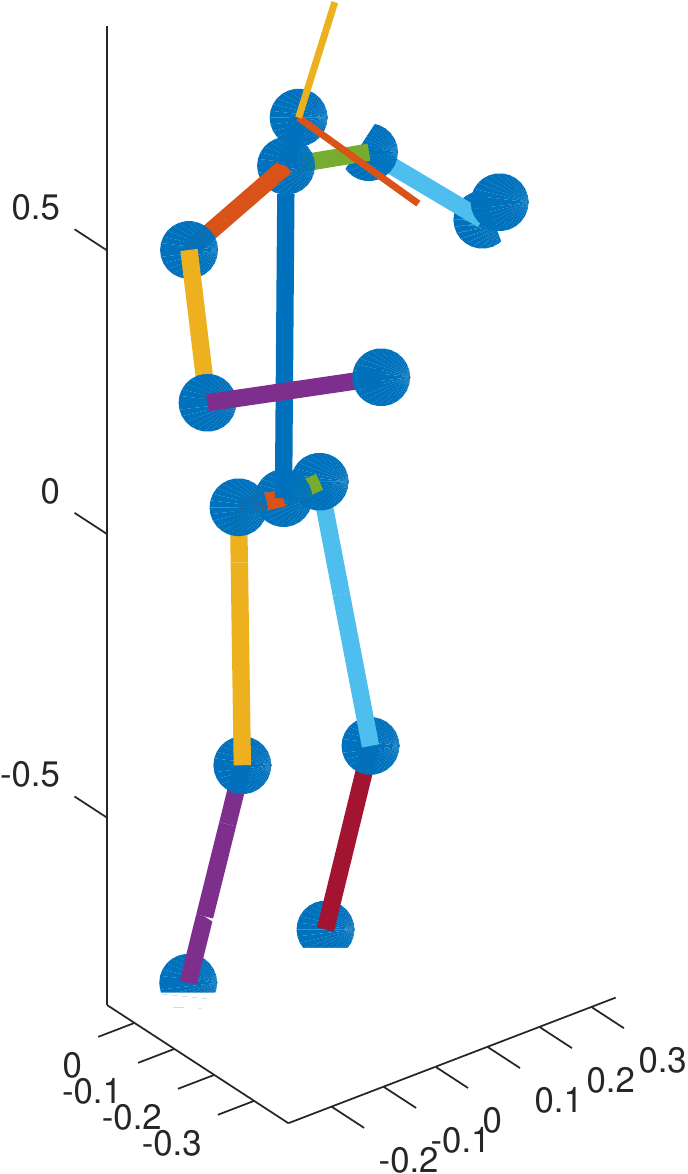}%
\\
 \rotatebox{90}{\hspace{8pt}{\tiny xr-egopose}} &
\includegraphics[width=0.05\linewidth, height=0.075\linewidth]{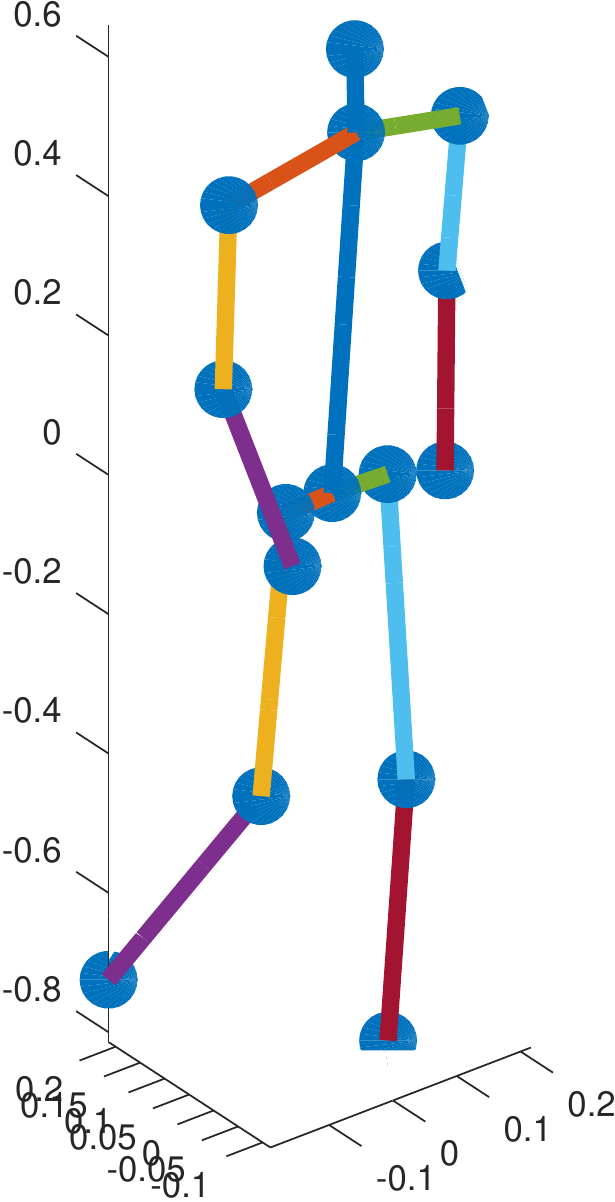}%
\includegraphics[width=0.05\linewidth, height=0.075\linewidth]{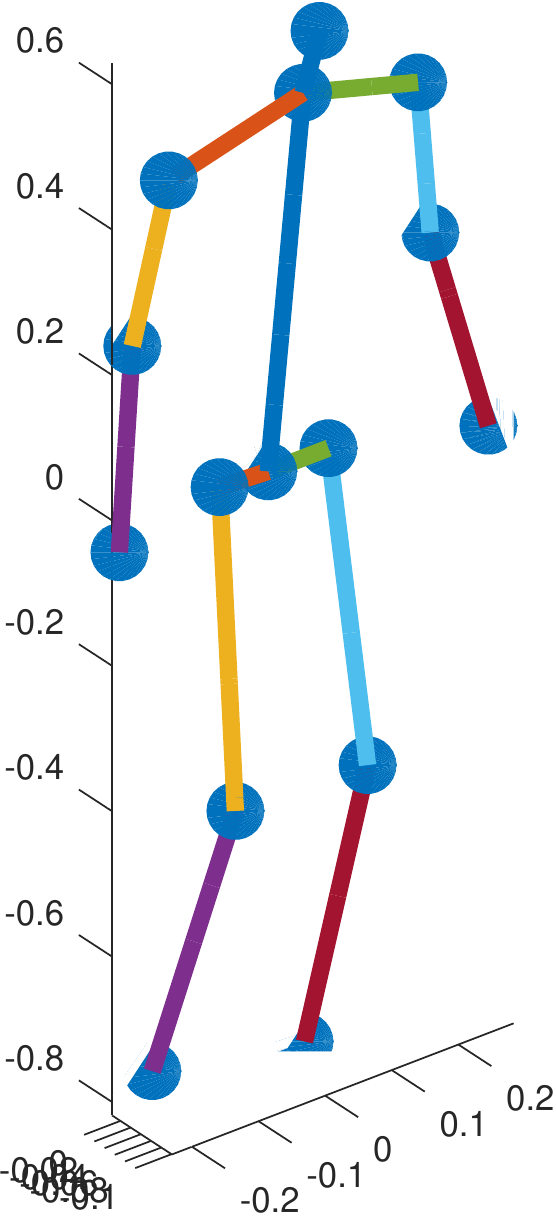}%
\includegraphics[width=0.05\linewidth, height=0.075\linewidth]{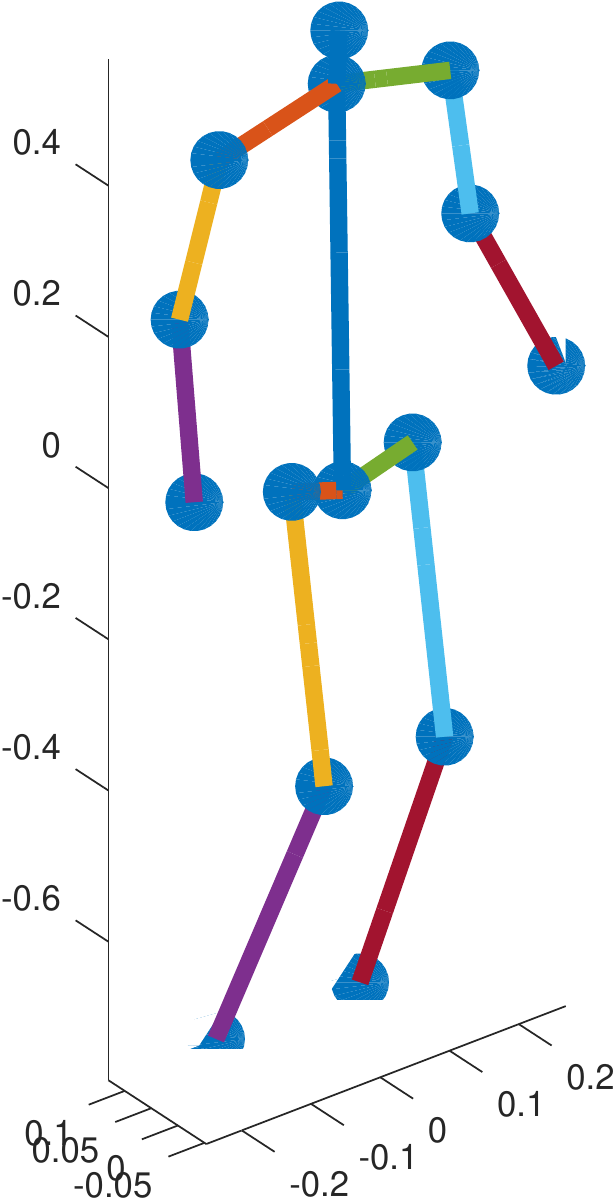}%
\includegraphics[width=0.05\linewidth, height=0.075\linewidth]{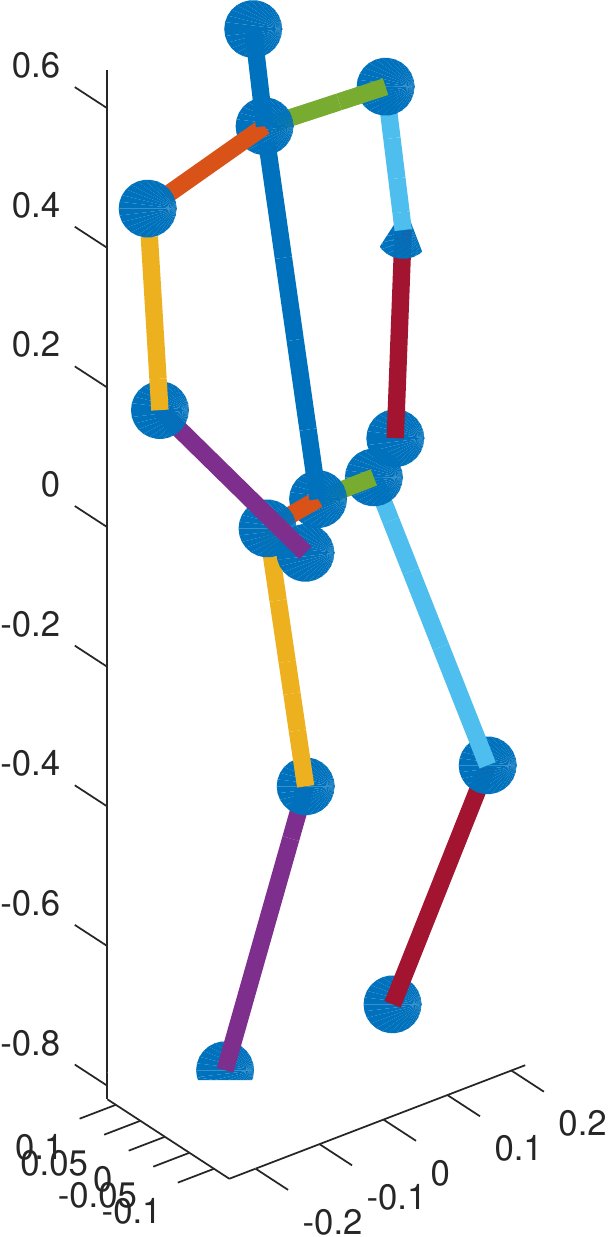}%
\includegraphics[width=0.05\linewidth, height=0.075\linewidth]{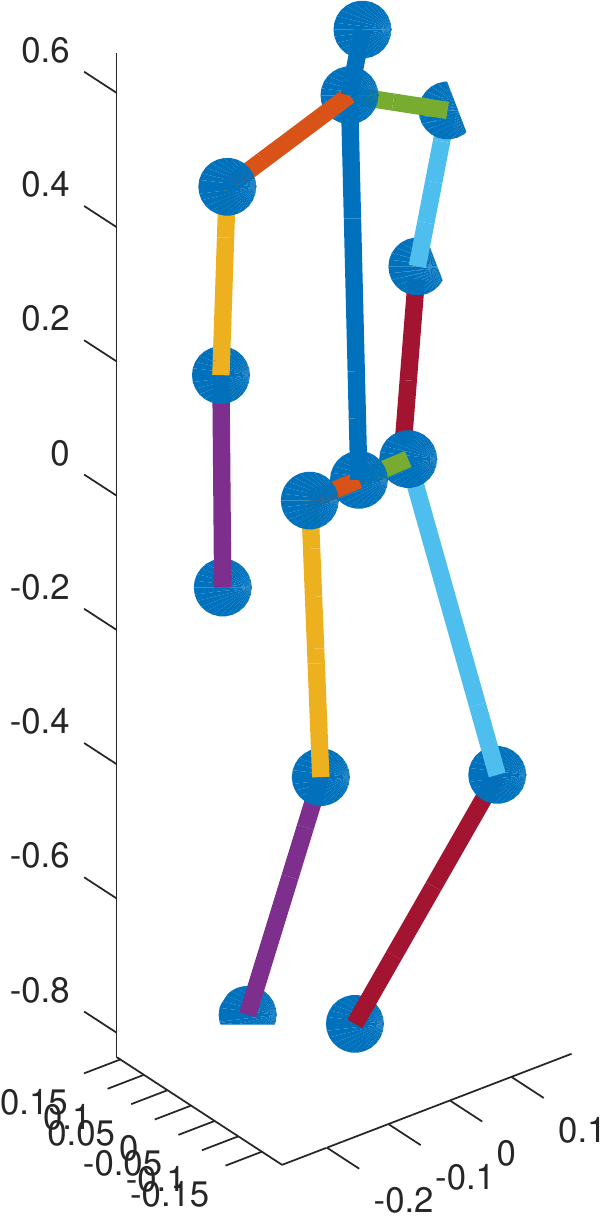}%
\includegraphics[width=0.05\linewidth, height=0.075\linewidth]{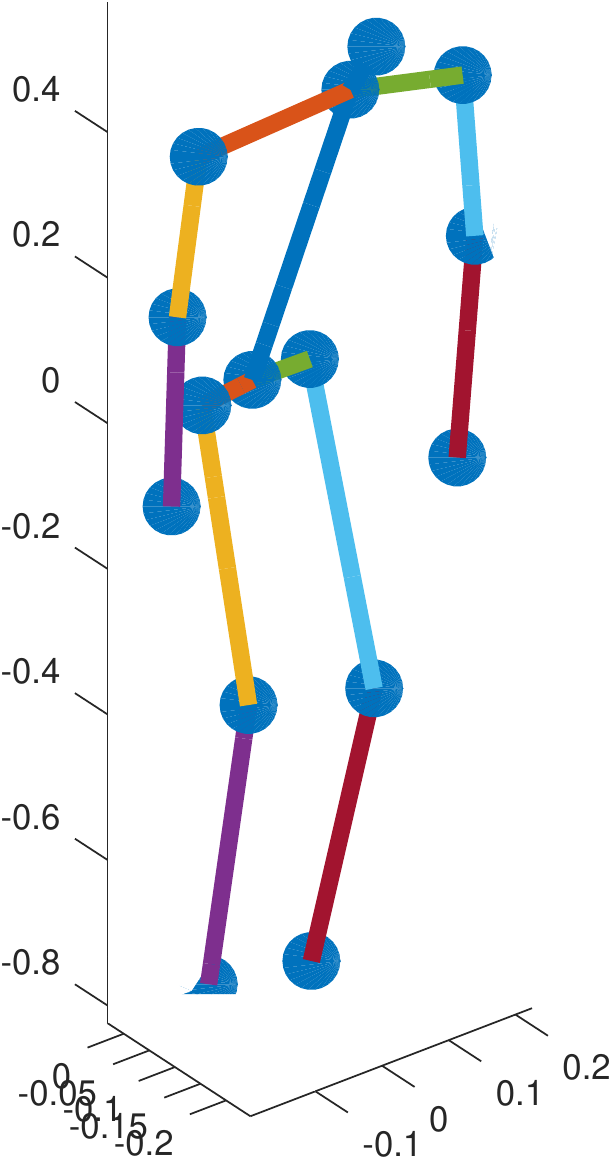}%
\includegraphics[width=0.05\linewidth, height=0.075\linewidth]{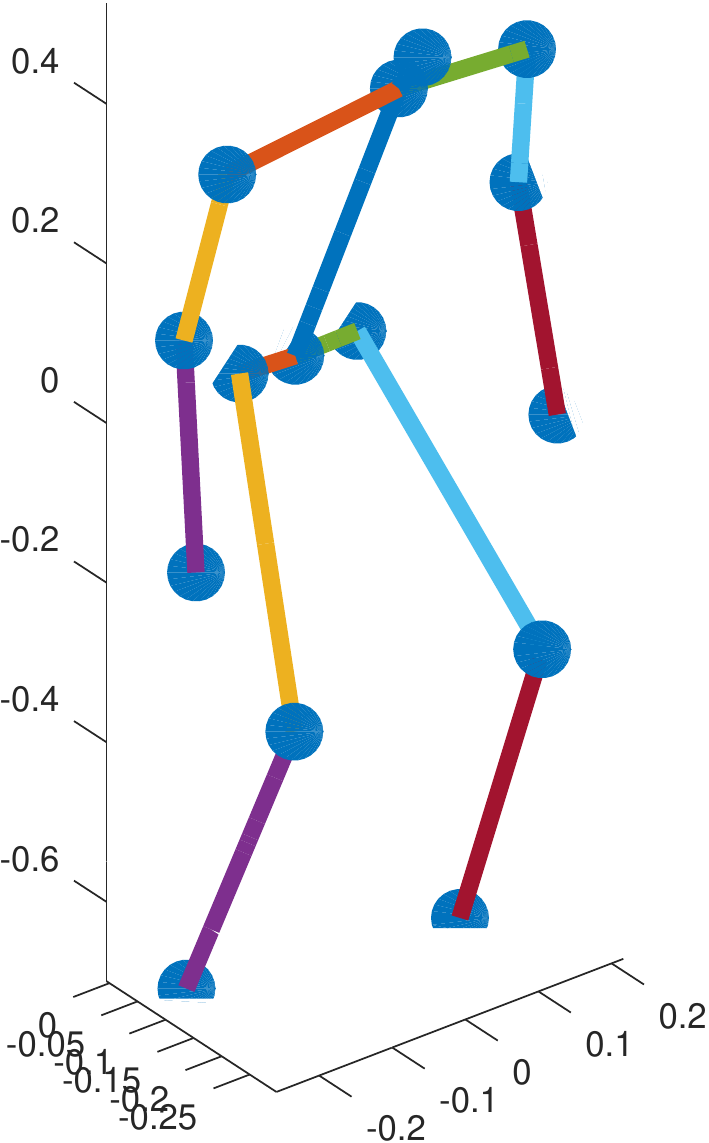}%
\includegraphics[width=0.05\linewidth, height=0.075\linewidth]{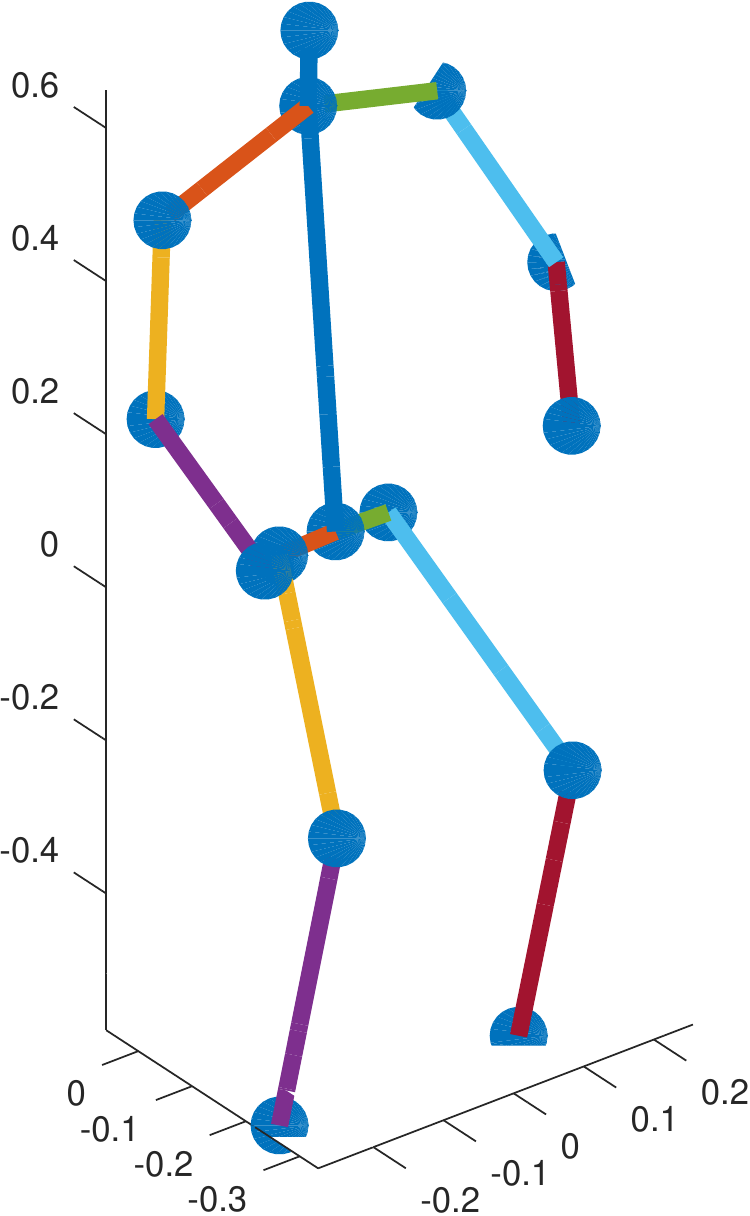}%
\includegraphics[width=0.05\linewidth, height=0.075\linewidth]{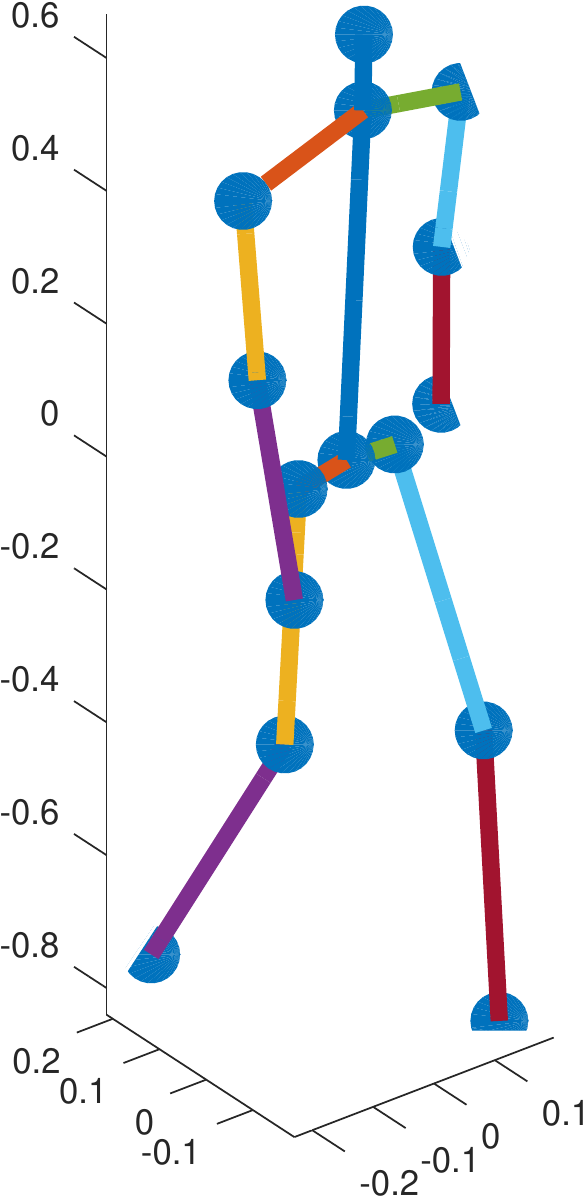}%
\includegraphics[width=0.05\linewidth, height=0.075\linewidth]{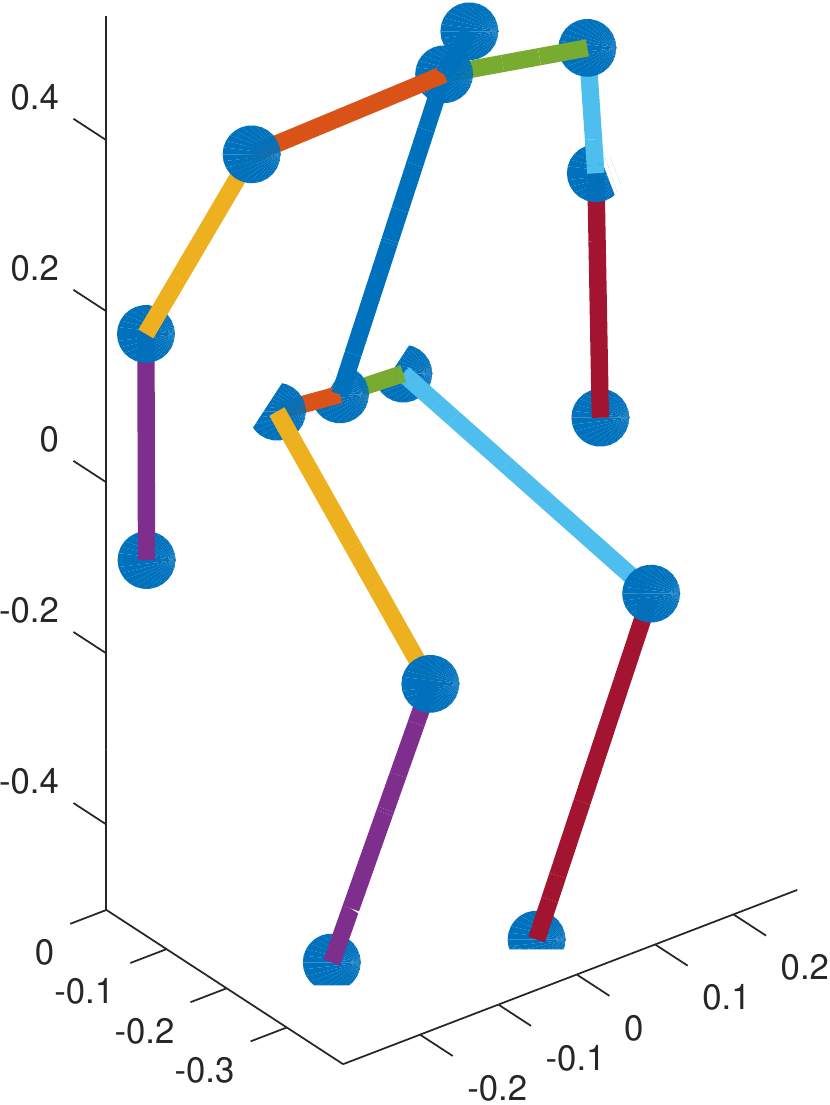}%
\includegraphics[width=0.05\linewidth, height=0.075\linewidth]{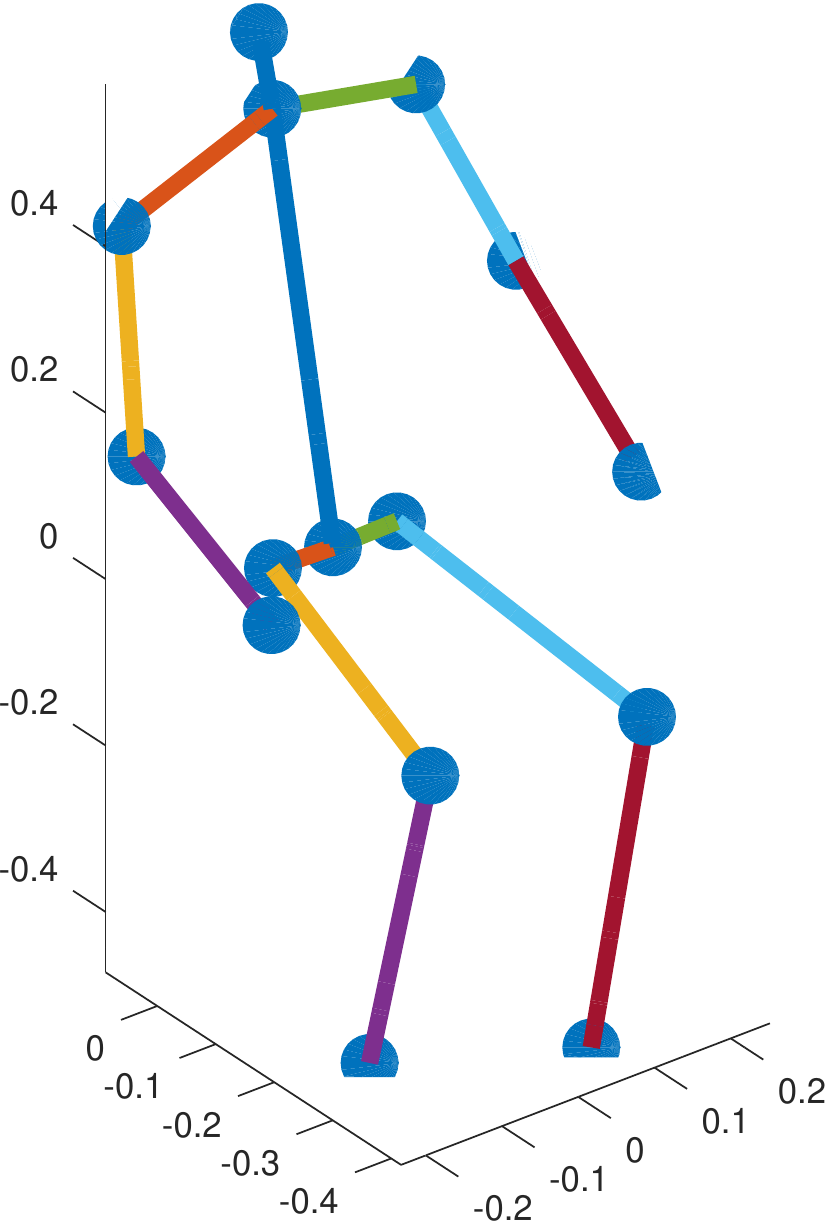}%
\includegraphics[width=0.05\linewidth, height=0.075\linewidth]{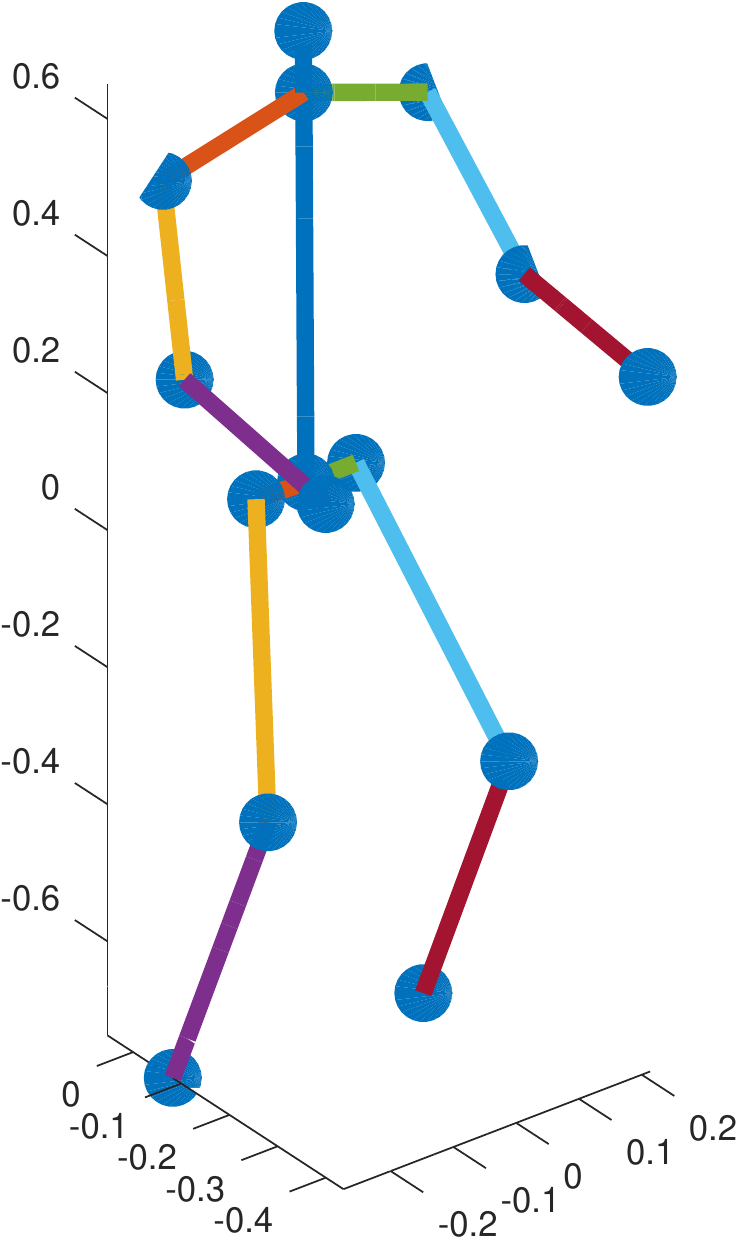}%
\includegraphics[width=0.05\linewidth, height=0.075\linewidth]{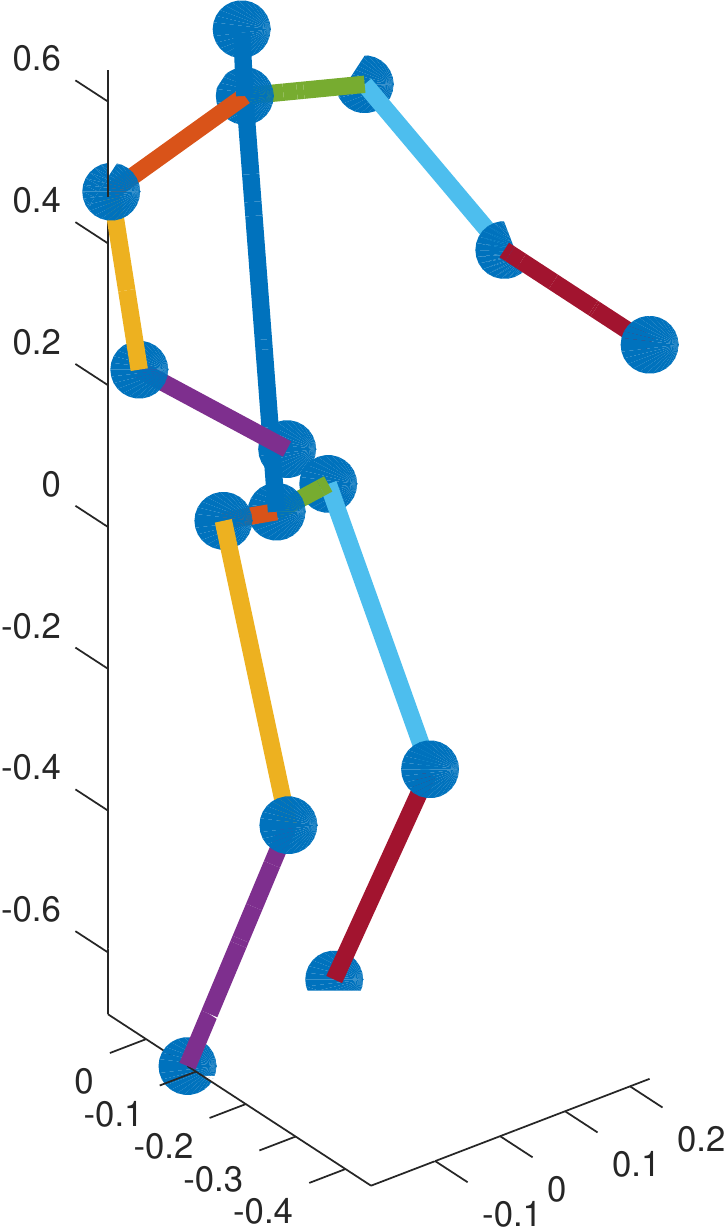}%
\includegraphics[width=0.05\linewidth, height=0.075\linewidth]{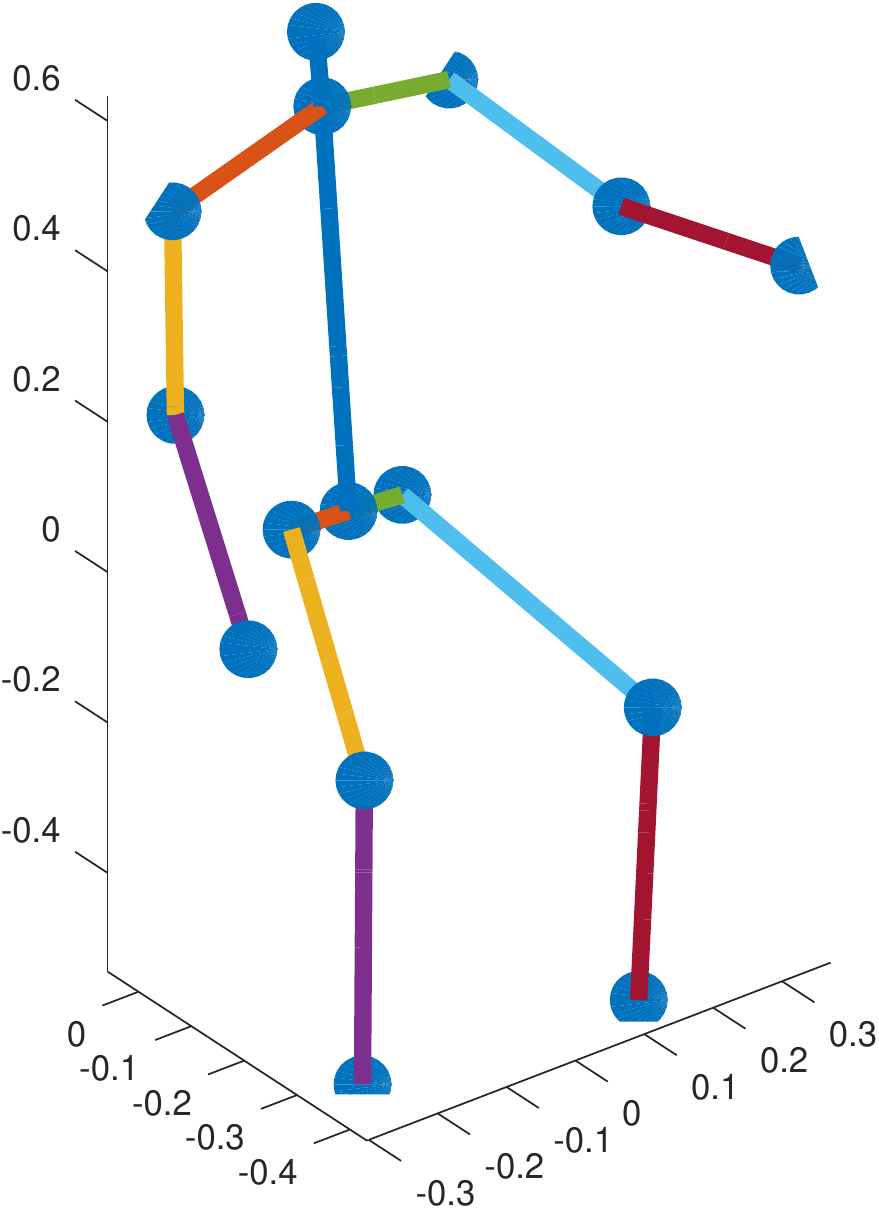}%
\includegraphics[width=0.05\linewidth, height=0.075\linewidth]{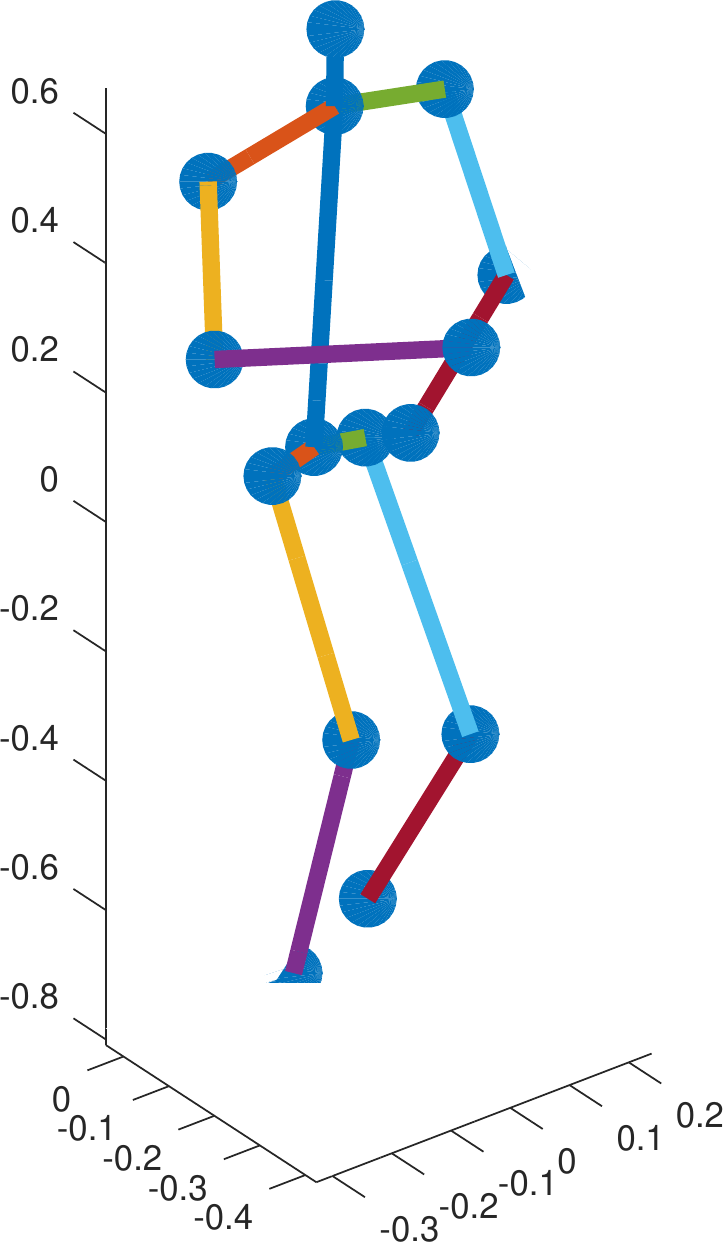}%
\includegraphics[width=0.05\linewidth, height=0.075\linewidth]{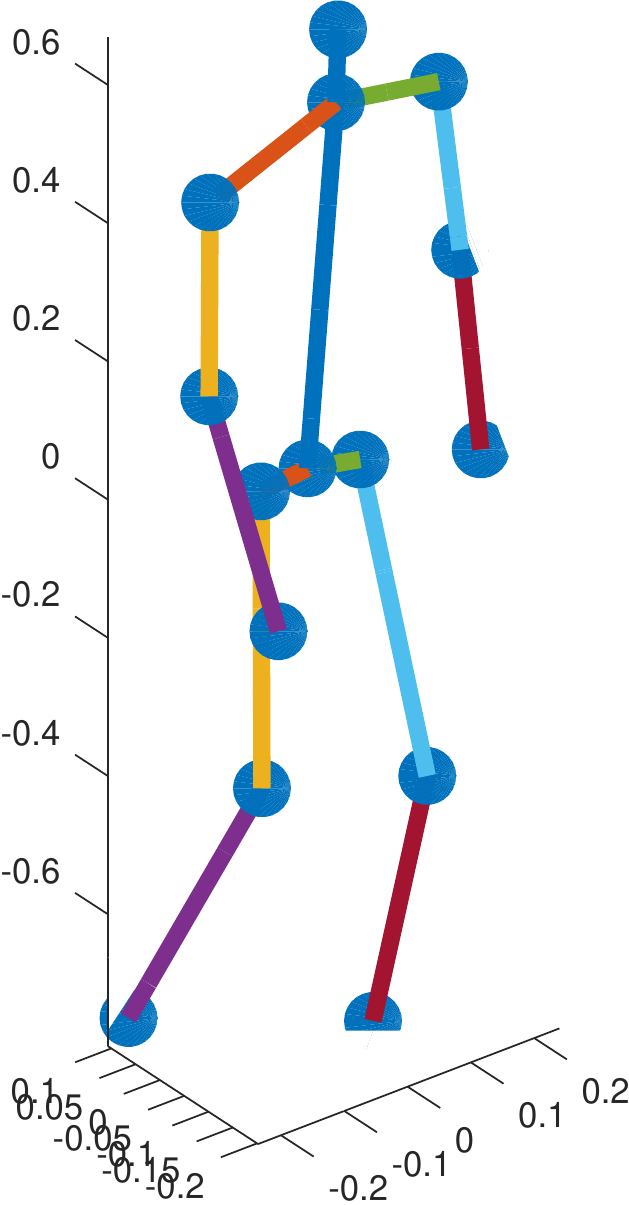}%
\includegraphics[width=0.05\linewidth, height=0.075\linewidth]{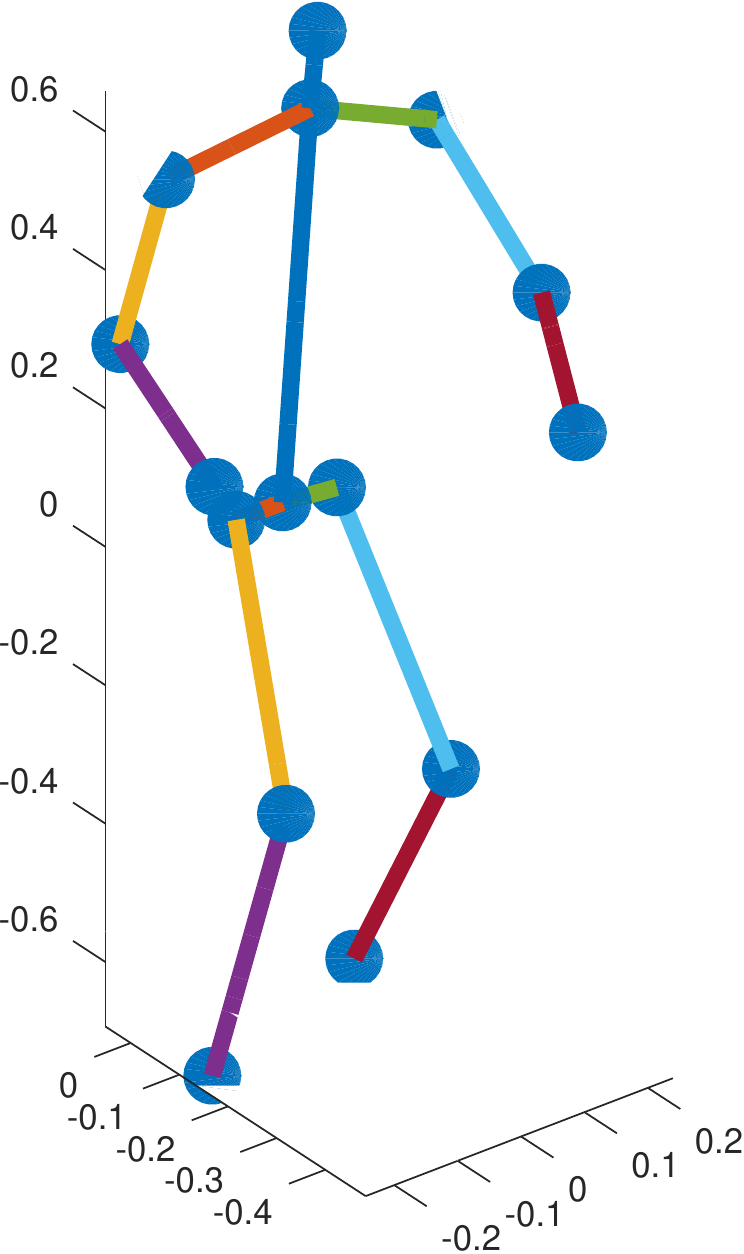}%
\includegraphics[width=0.05\linewidth, height=0.075\linewidth]{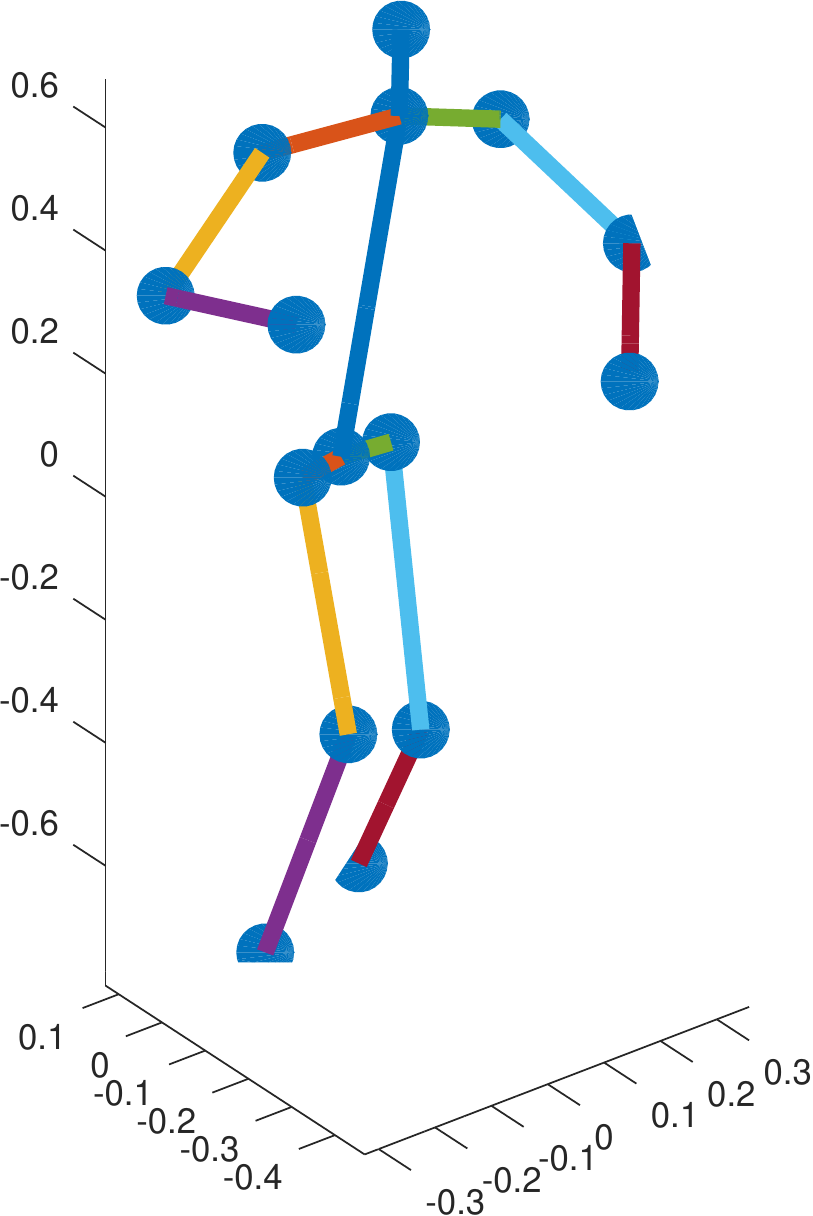}%
\includegraphics[width=0.05\linewidth, height=0.075\linewidth]{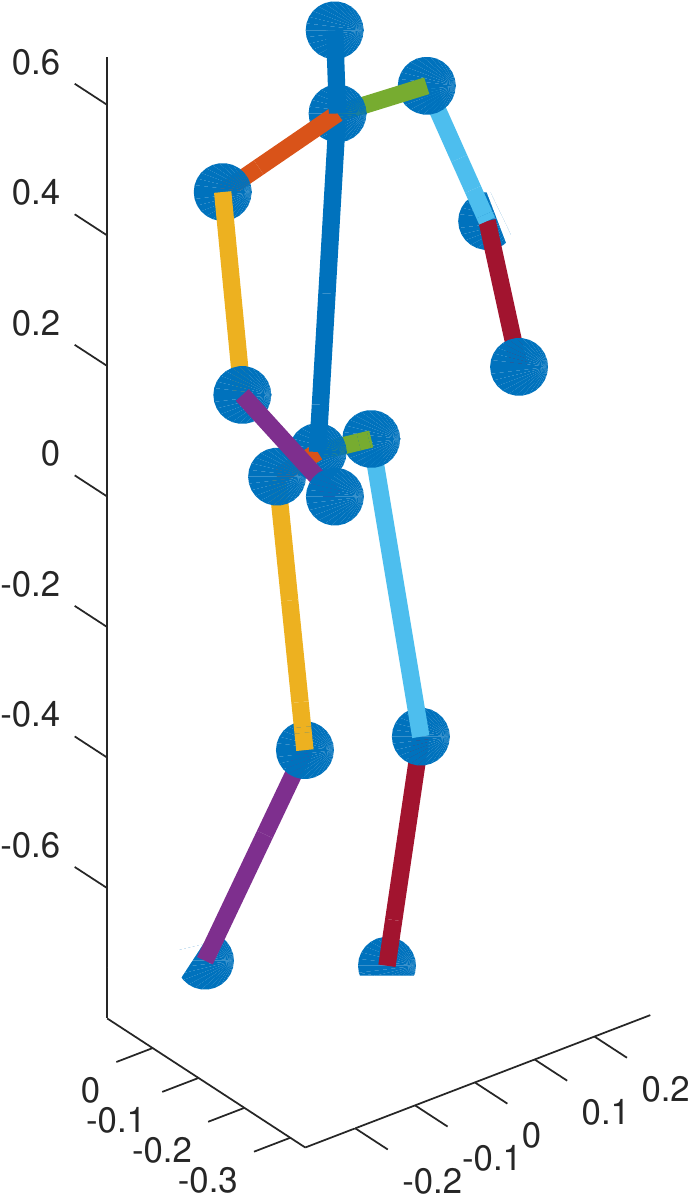}%
\includegraphics[width=0.05\linewidth, height=0.075\linewidth]{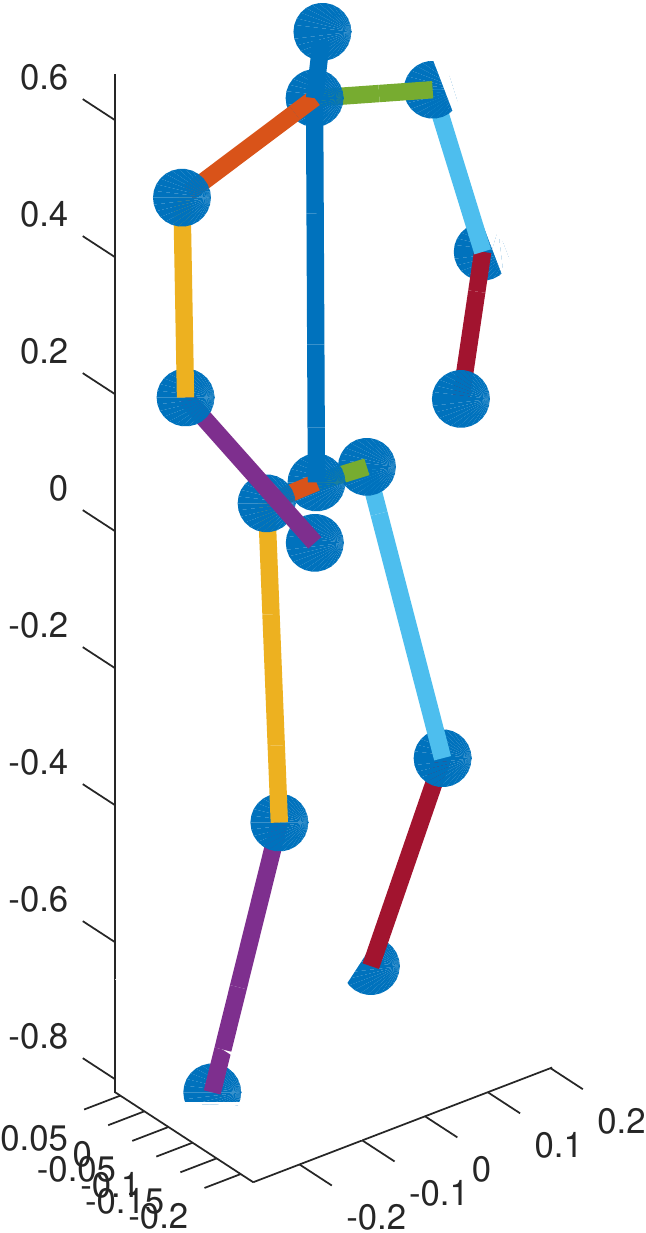}%
\\
 \rotatebox{90}{\hspace{8pt}{\tiny pd-egopose}} &
\includegraphics[width=0.05\linewidth, height=0.075\linewidth]{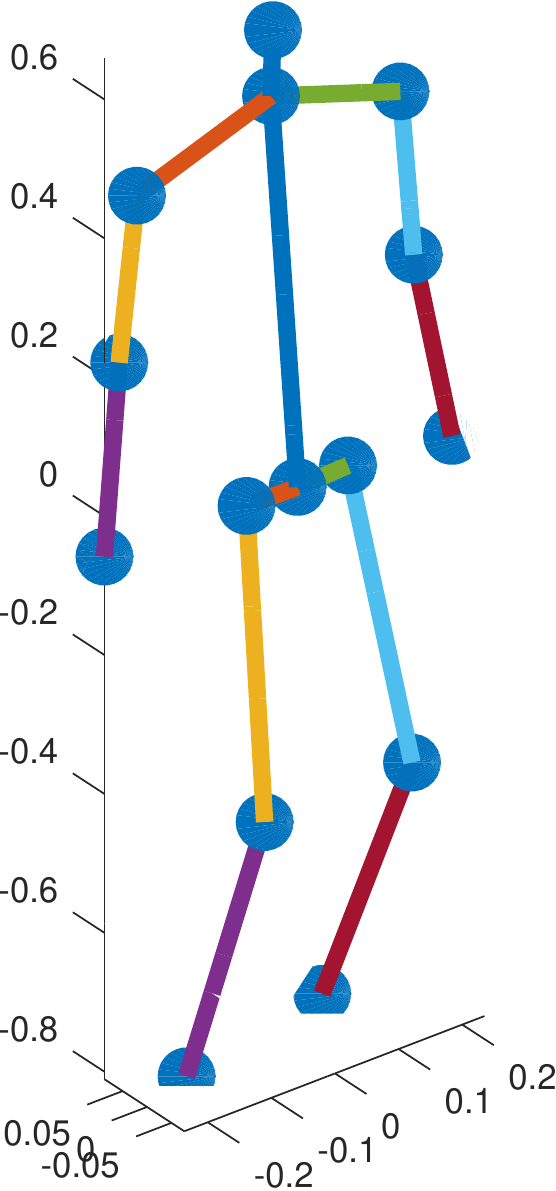}%
\includegraphics[width=0.05\linewidth, height=0.075\linewidth]{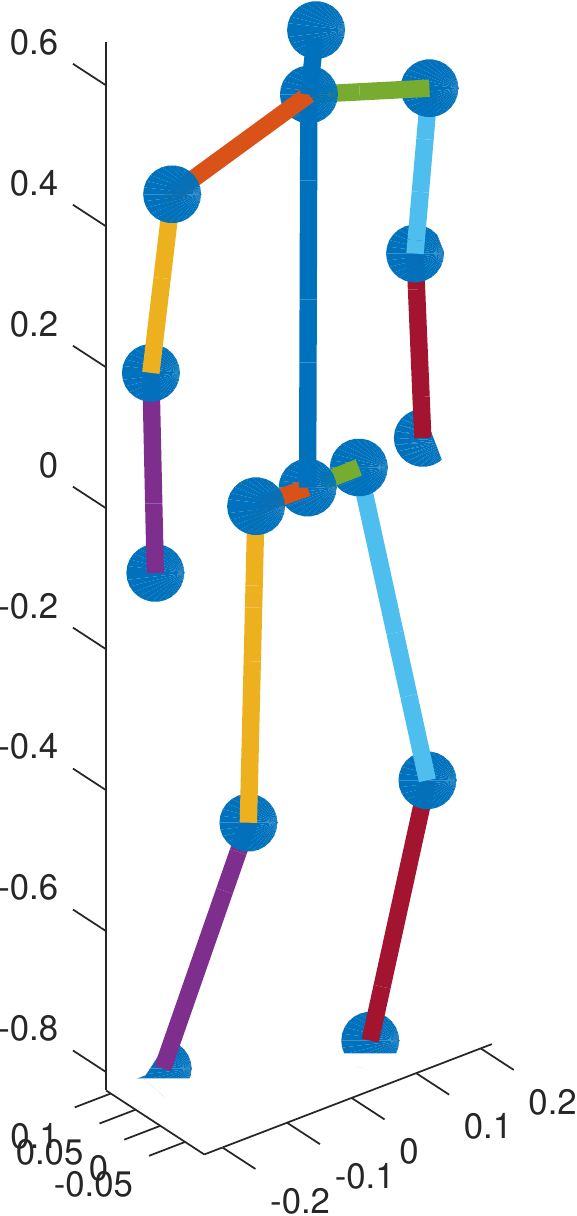}%
\includegraphics[width=0.05\linewidth, height=0.075\linewidth]{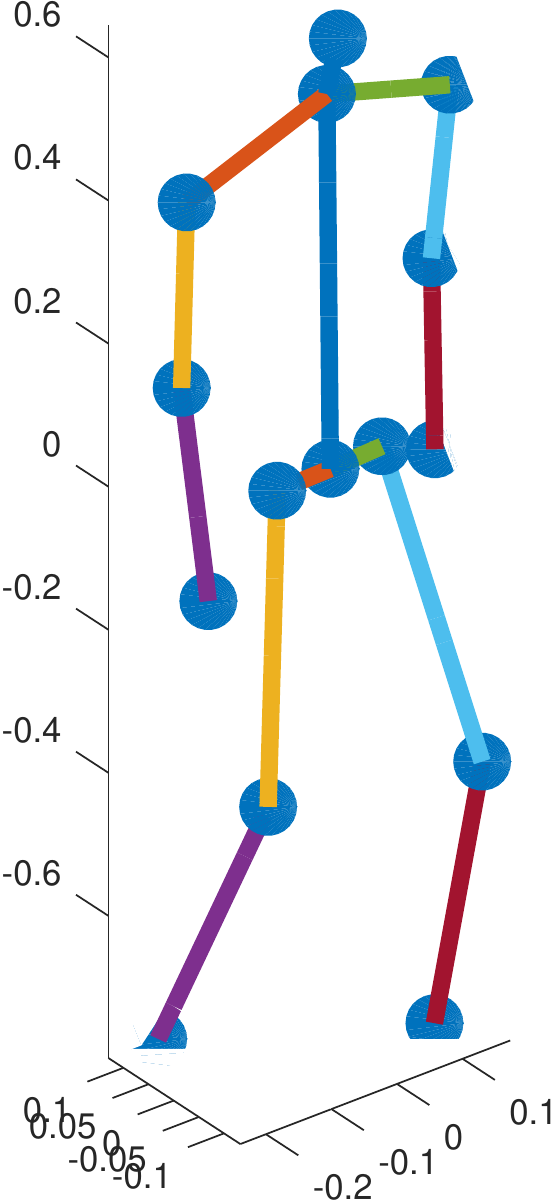}%
\includegraphics[width=0.05\linewidth, height=0.075\linewidth]{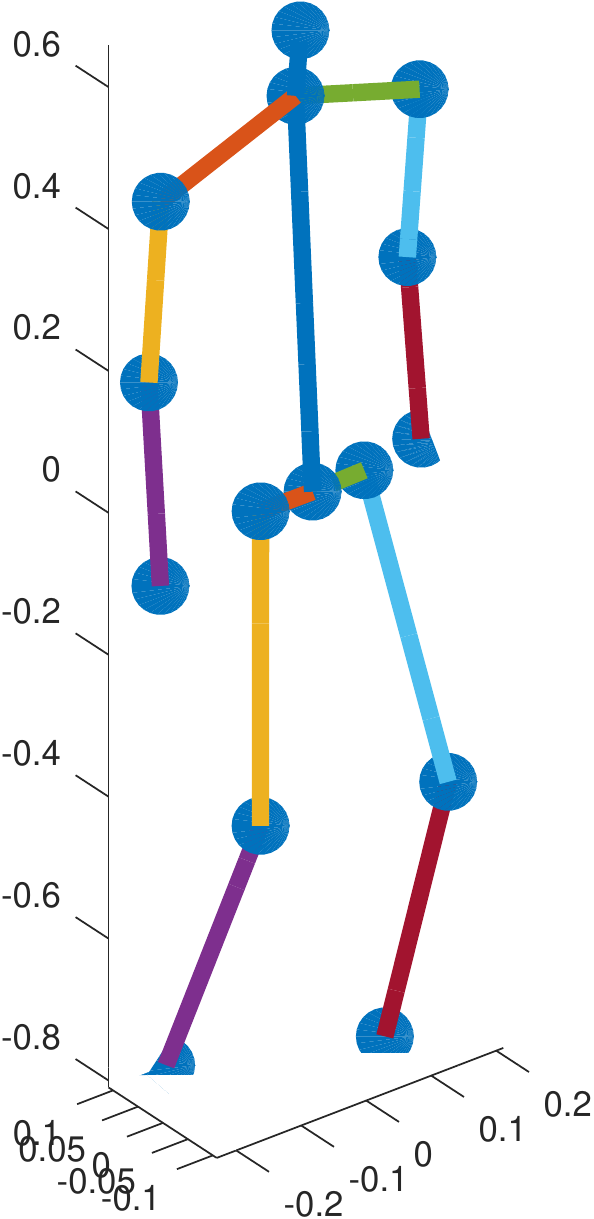}%
\includegraphics[width=0.05\linewidth, height=0.075\linewidth]{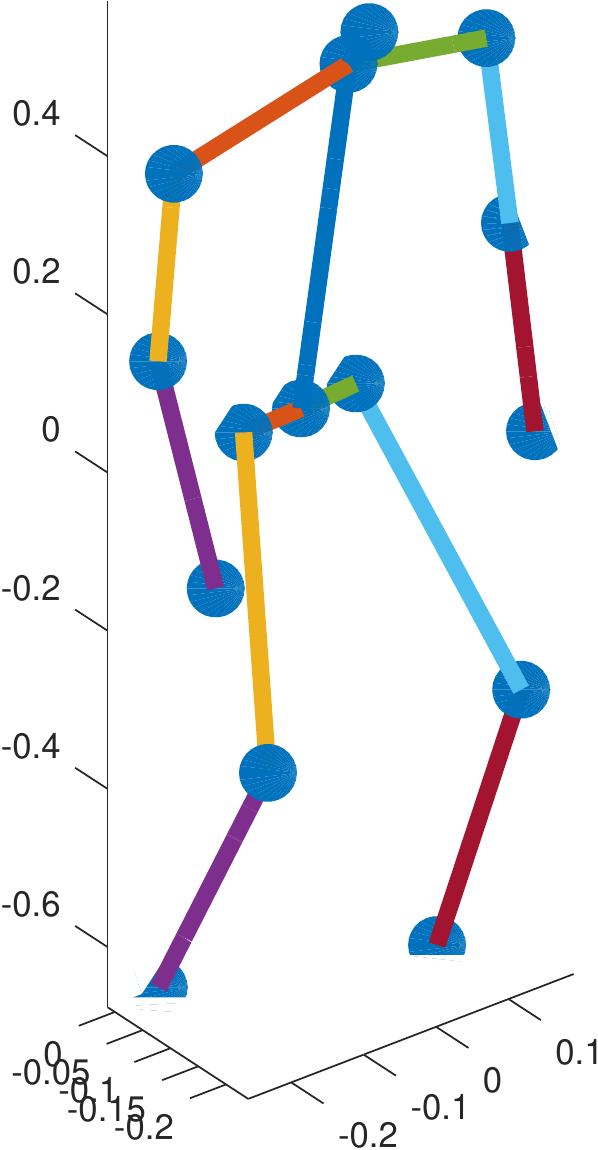}%
\includegraphics[width=0.05\linewidth, height=0.075\linewidth]{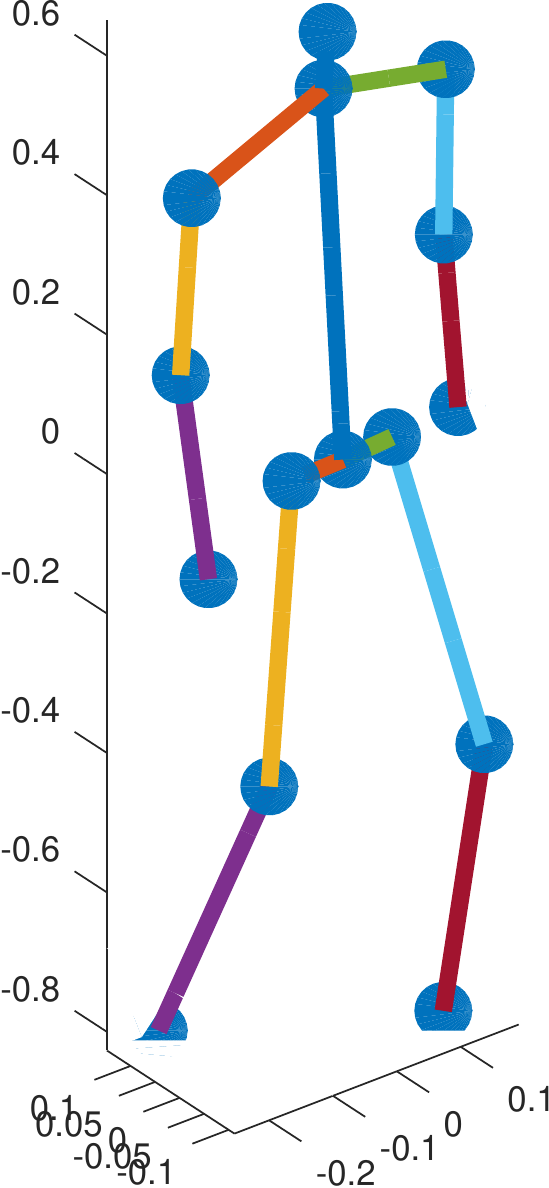}%
\includegraphics[width=0.05\linewidth, height=0.075\linewidth]{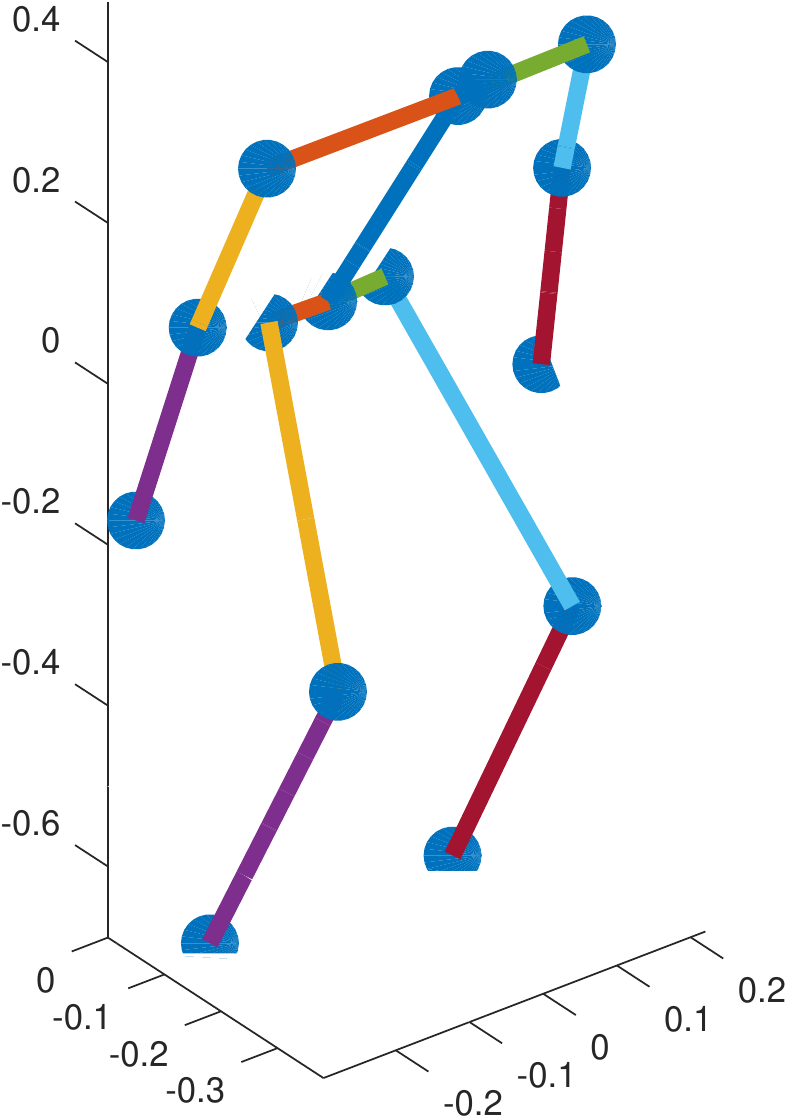}%
\includegraphics[width=0.05\linewidth, height=0.075\linewidth]{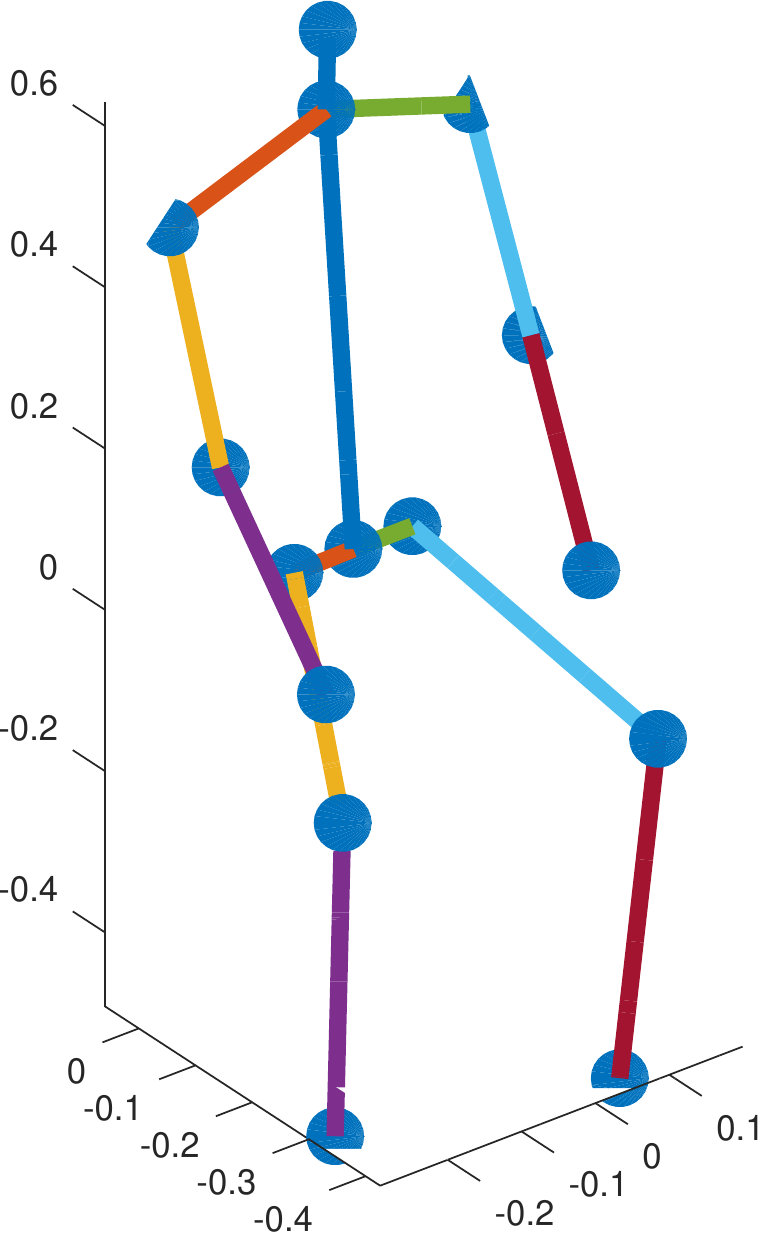}%
\includegraphics[width=0.05\linewidth, height=0.075\linewidth]{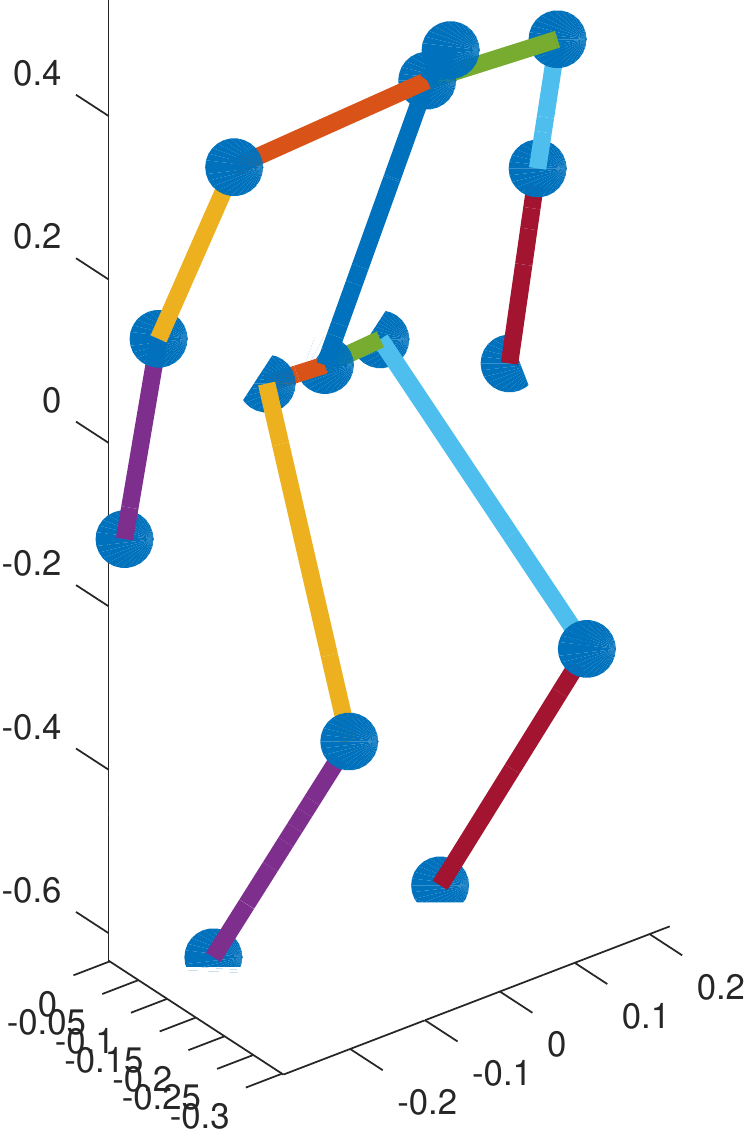}%
\includegraphics[width=0.05\linewidth, height=0.075\linewidth]{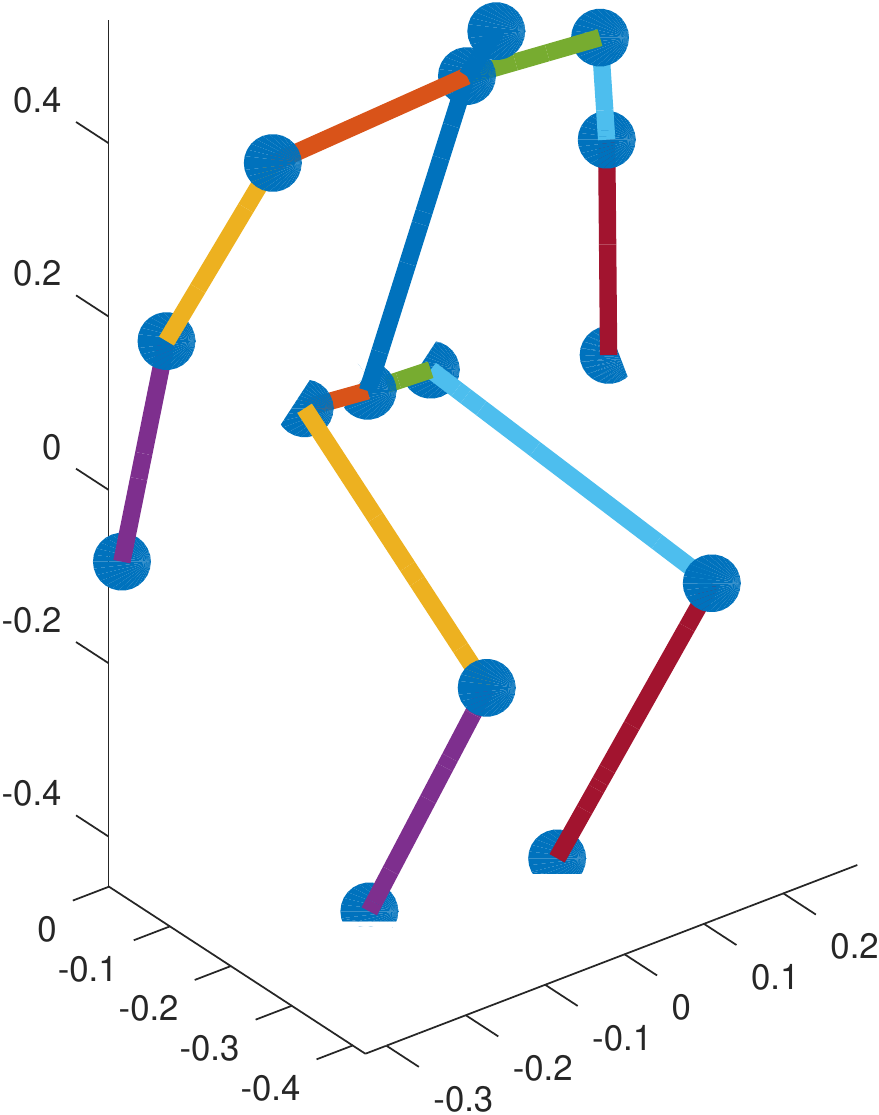}%
\includegraphics[width=0.05\linewidth, height=0.075\linewidth]{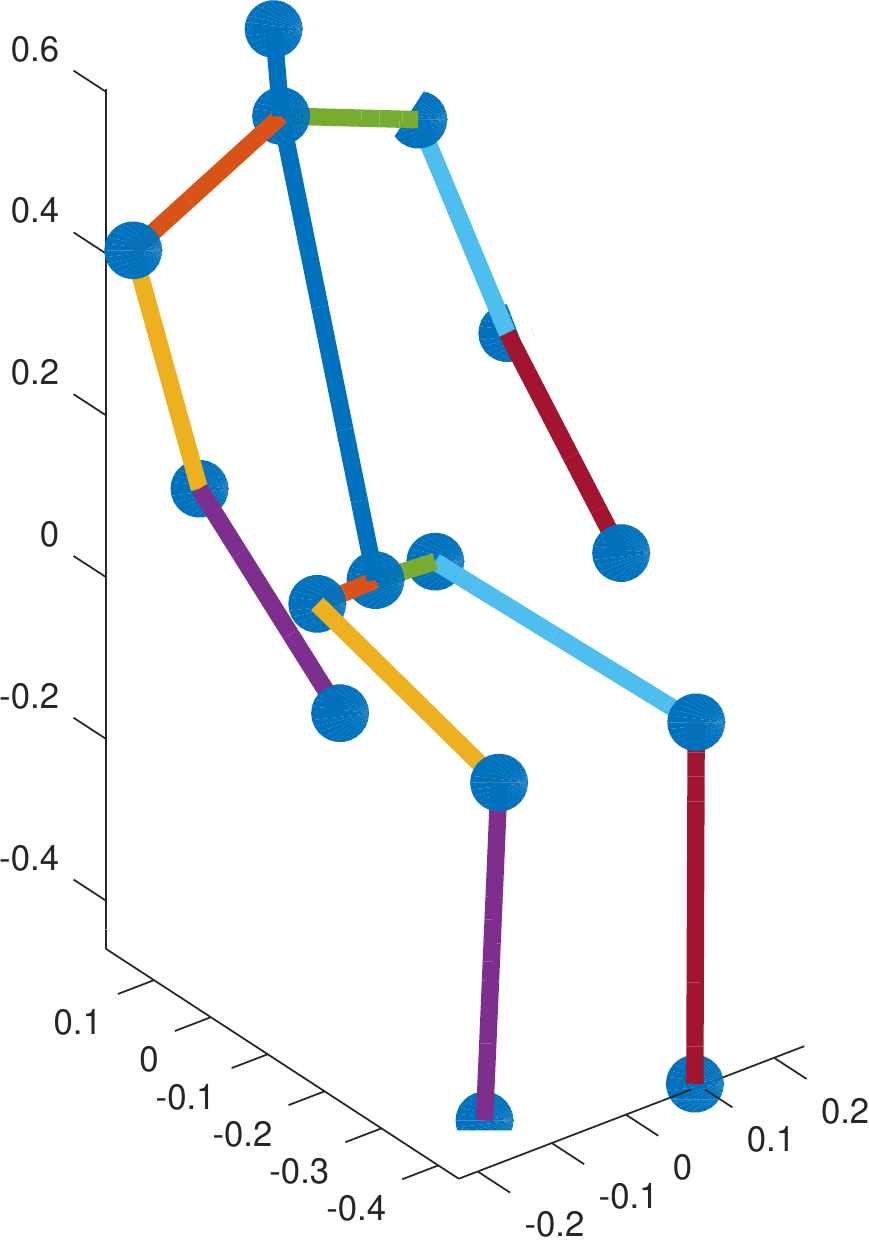}%
\includegraphics[width=0.05\linewidth, height=0.075\linewidth]{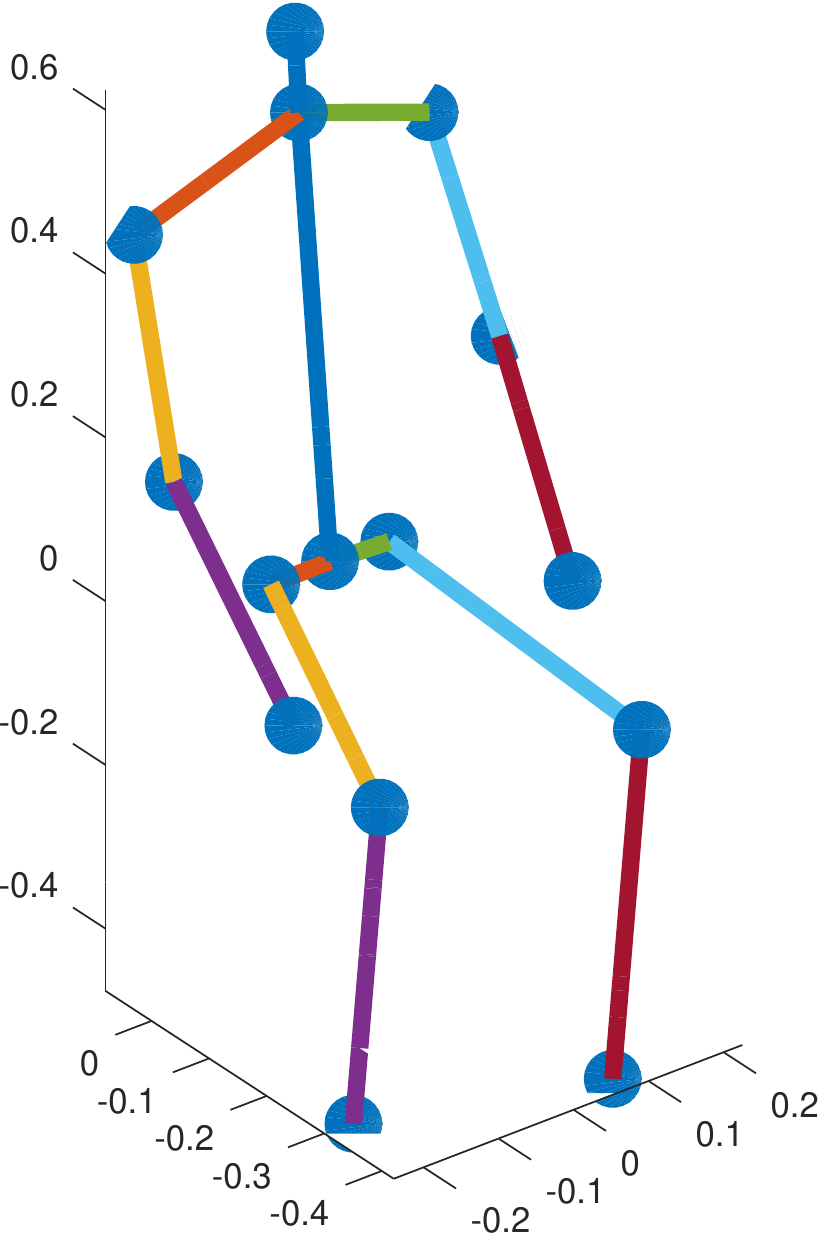}%
\includegraphics[width=0.05\linewidth, height=0.075\linewidth]{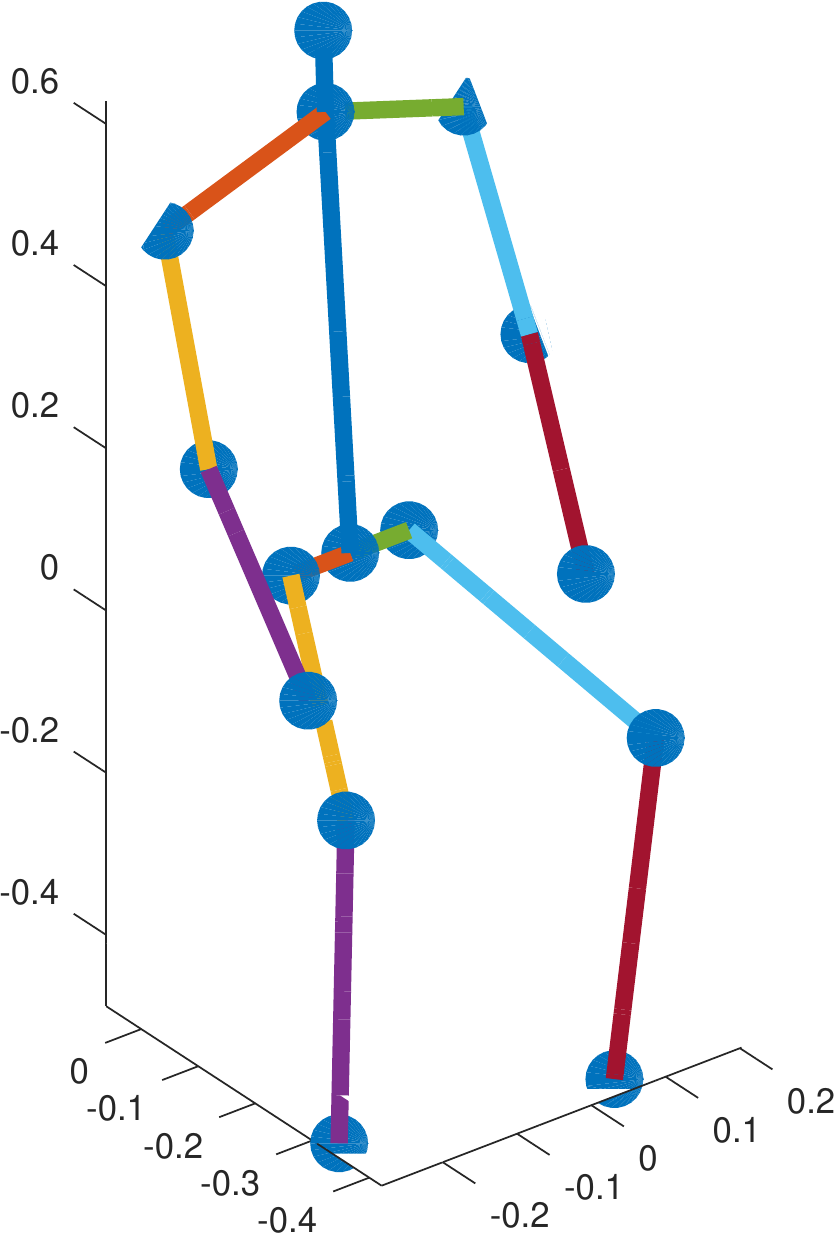}%
\includegraphics[width=0.05\linewidth, height=0.075\linewidth]{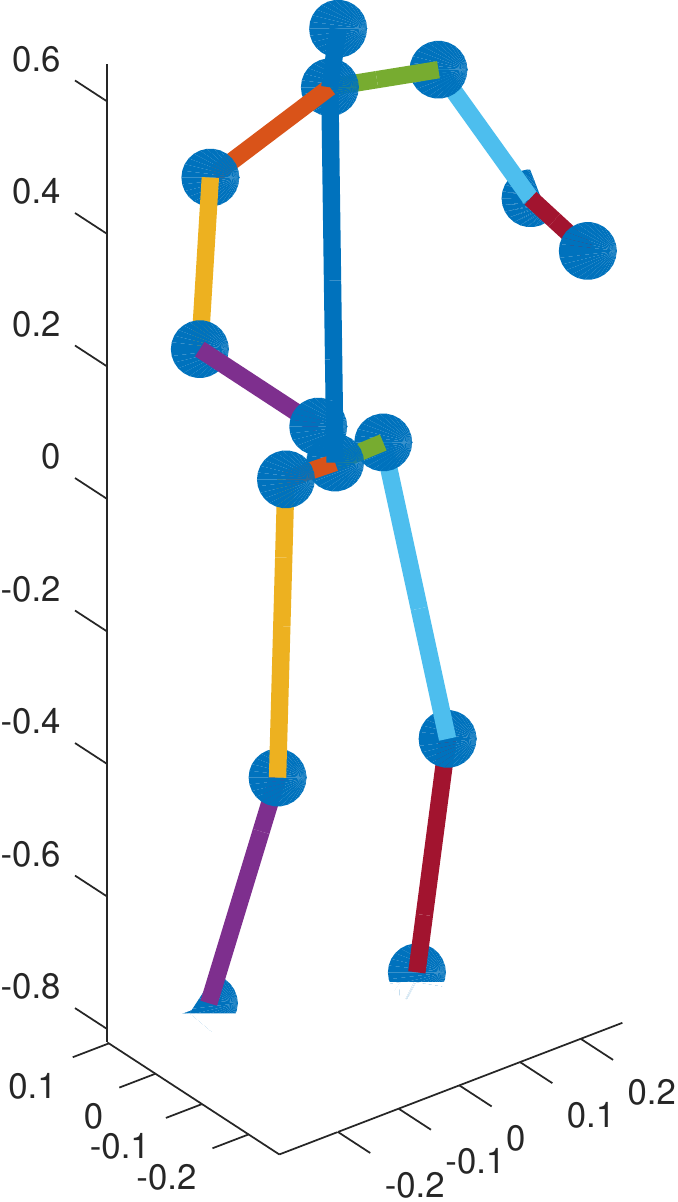}%
\includegraphics[width=0.05\linewidth, height=0.075\linewidth]{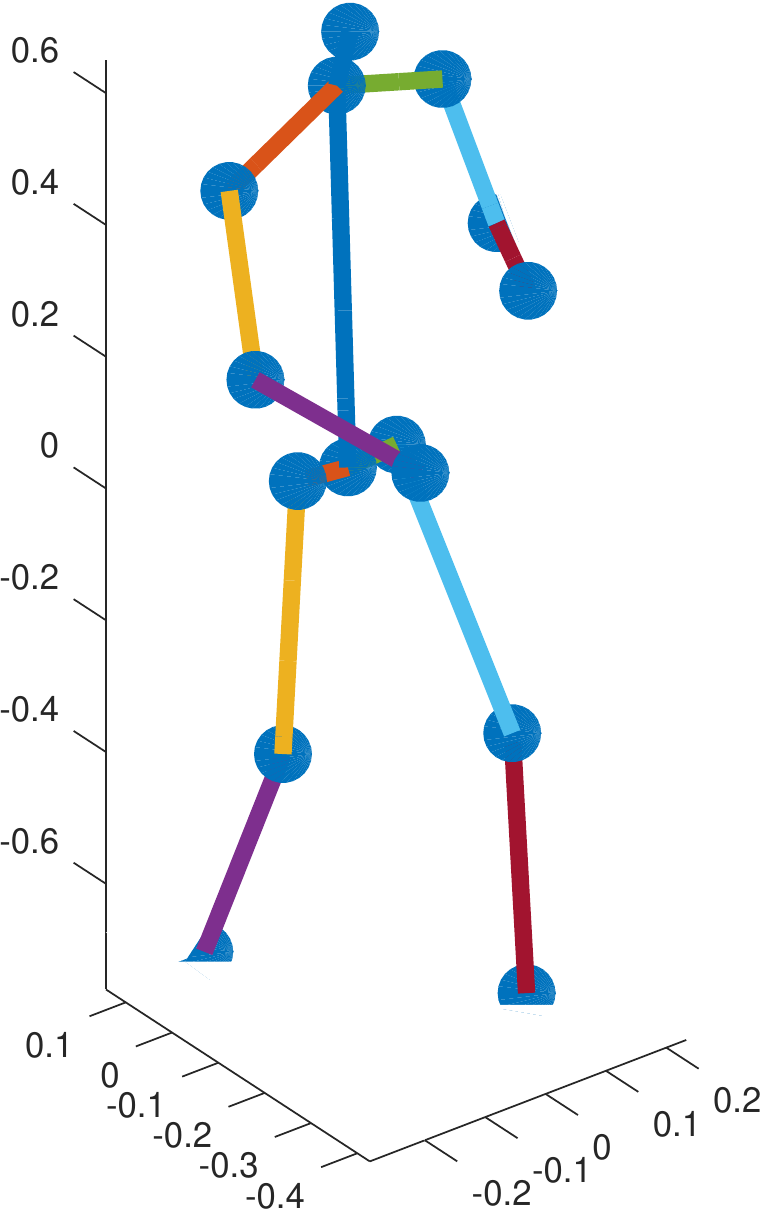}%
\includegraphics[width=0.05\linewidth, height=0.075\linewidth]{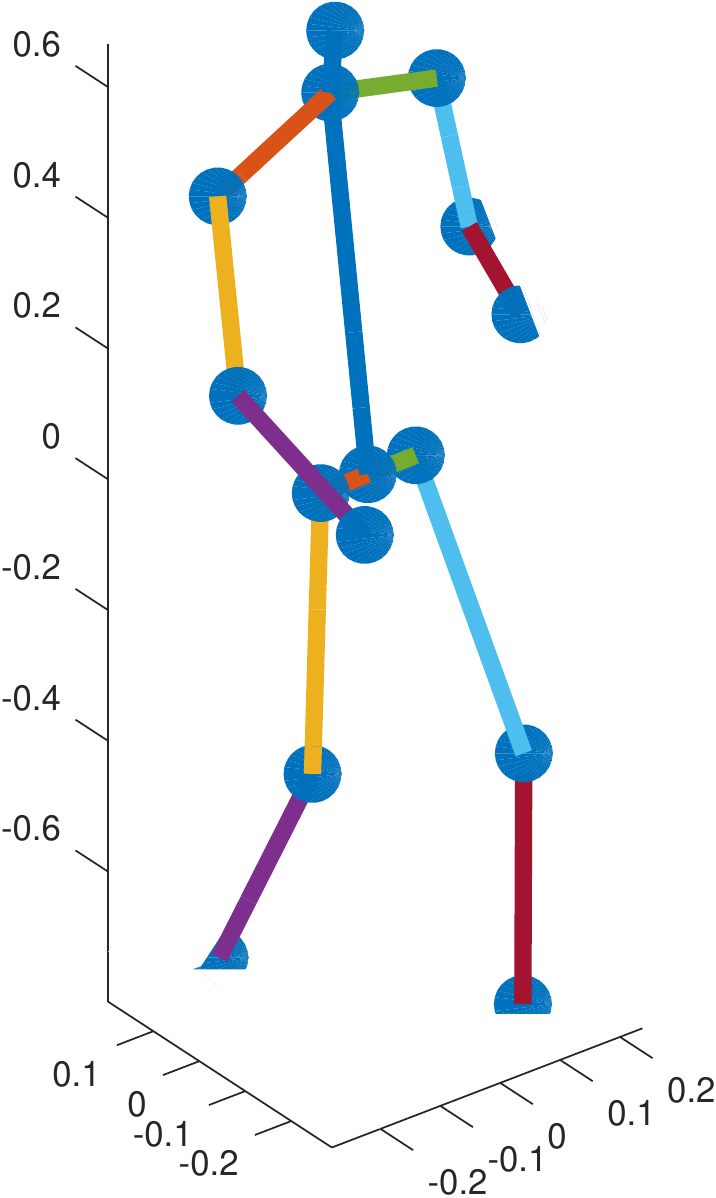}%
\includegraphics[width=0.05\linewidth, height=0.075\linewidth]{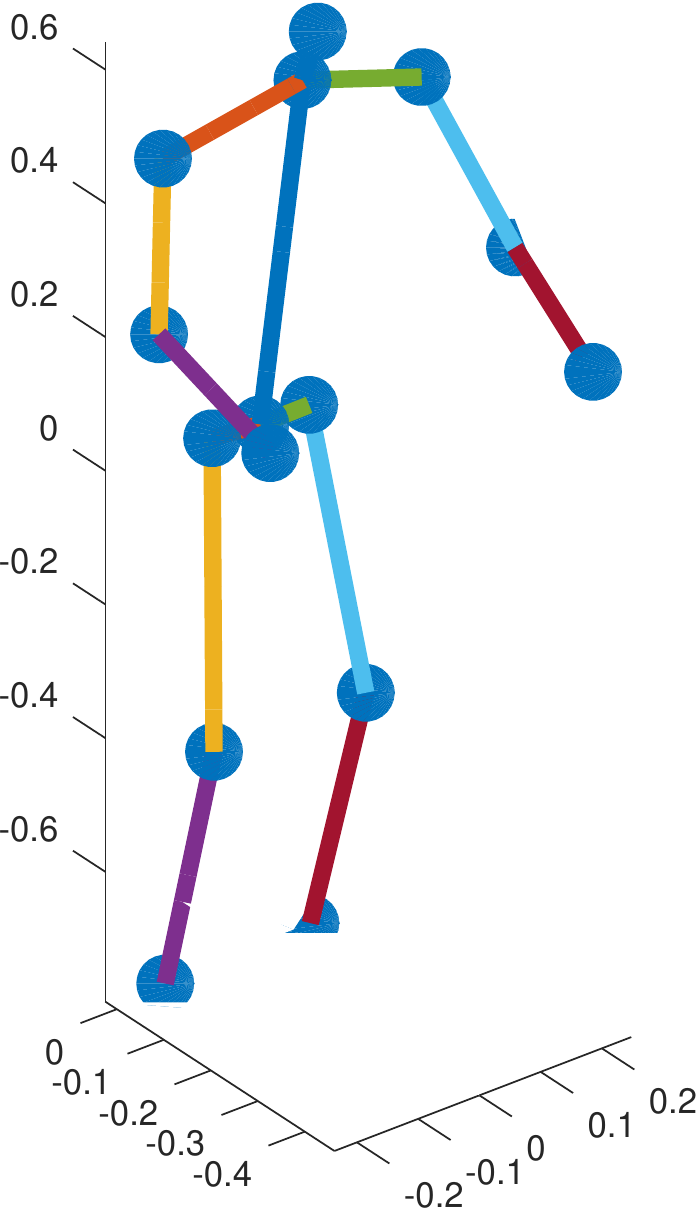}%
\includegraphics[width=0.05\linewidth, height=0.075\linewidth]{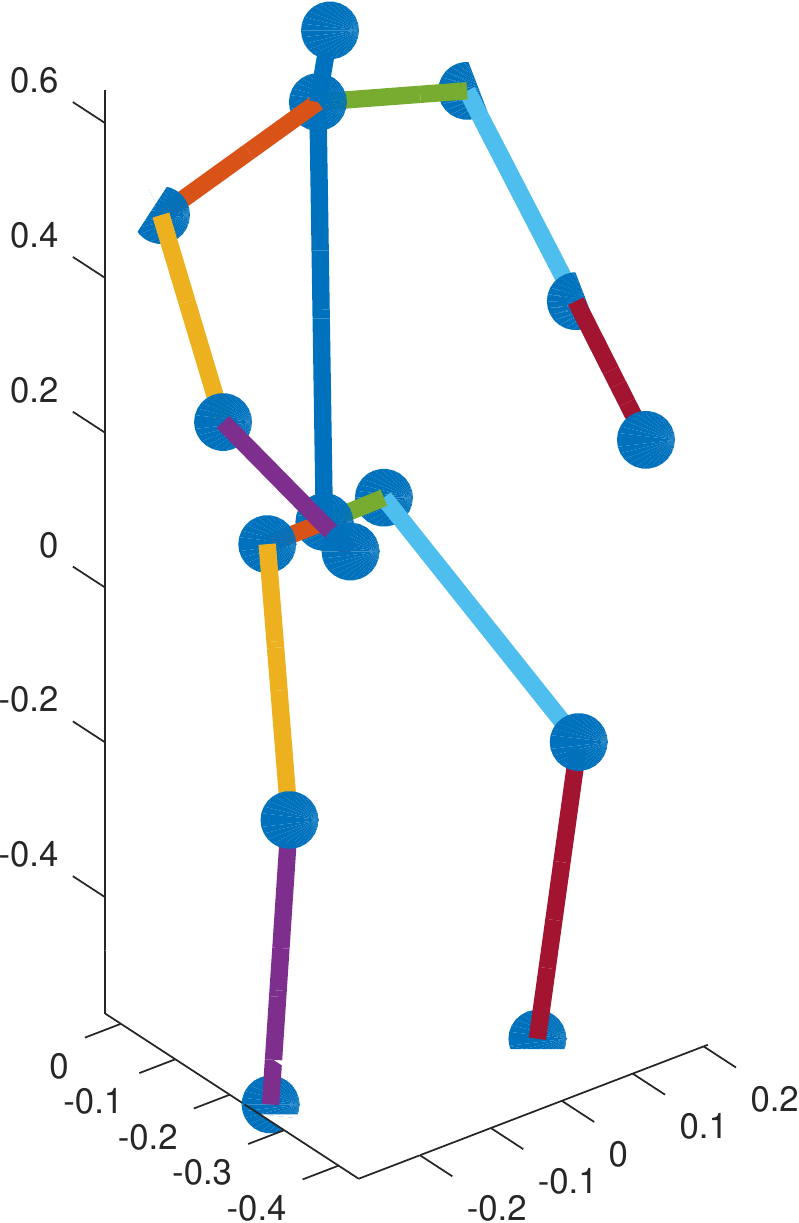}%
\includegraphics[width=0.05\linewidth, height=0.075\linewidth]{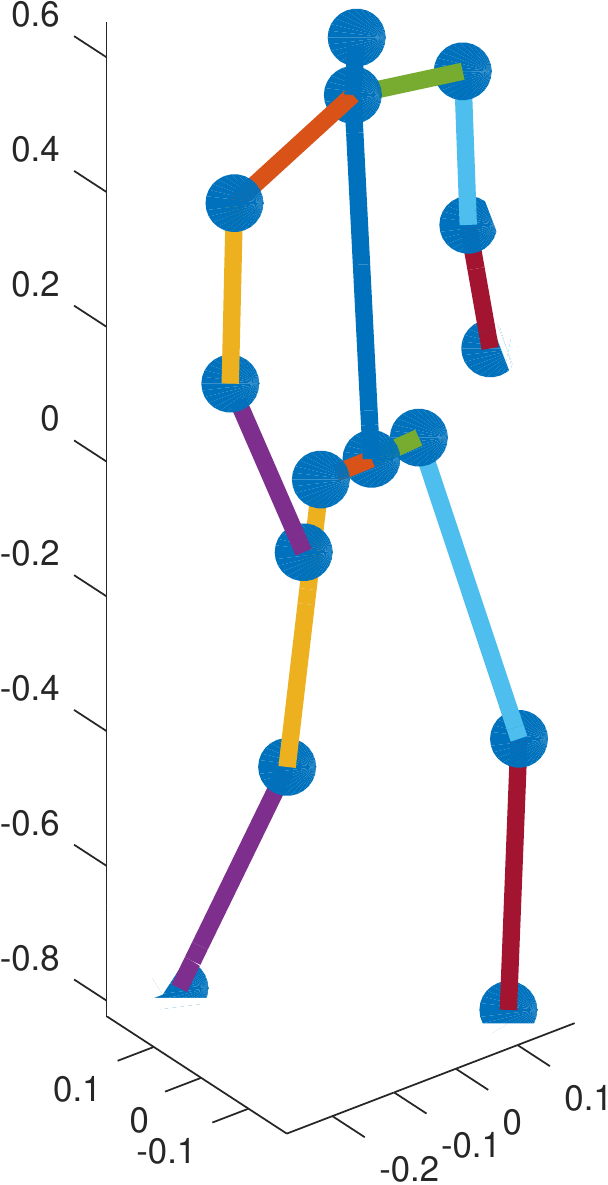}%
\includegraphics[width=0.05\linewidth, height=0.075\linewidth]{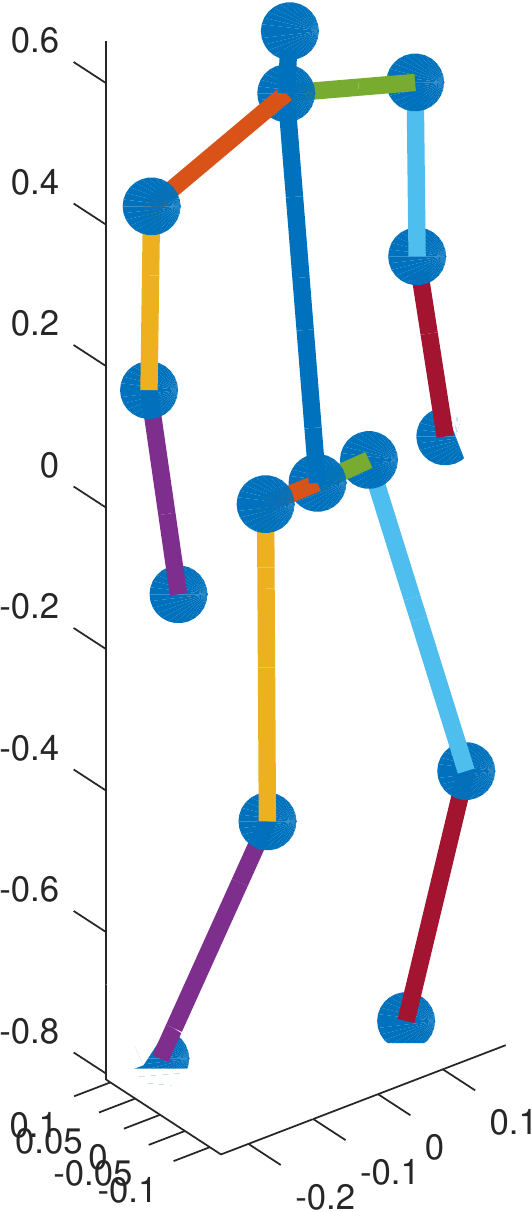}%
\\
 \rotatebox{90}{\hspace{0pt}{\tiny }} &
\includegraphics[width=0.05\linewidth, height=0.05\linewidth]{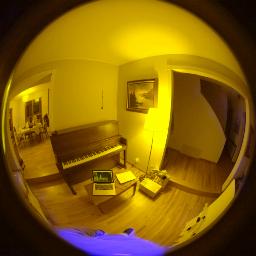}%
\includegraphics[width=0.05\linewidth, height=0.05\linewidth]{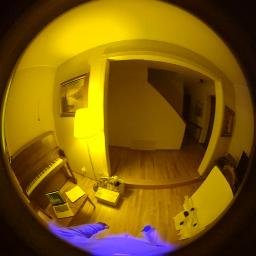}%
\includegraphics[width=0.05\linewidth, height=0.05\linewidth]{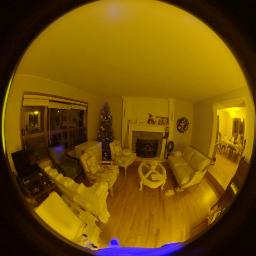}%
\includegraphics[width=0.05\linewidth, height=0.05\linewidth]{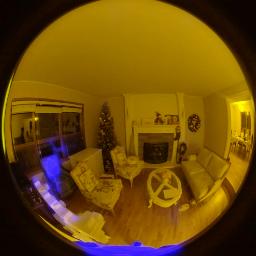}%
\includegraphics[width=0.05\linewidth, height=0.05\linewidth]{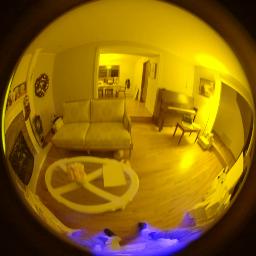}%
\includegraphics[width=0.05\linewidth, height=0.05\linewidth]{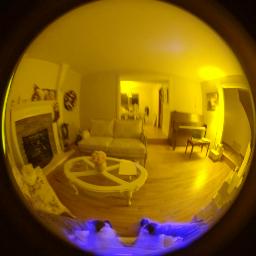}%
\includegraphics[width=0.05\linewidth, height=0.05\linewidth]{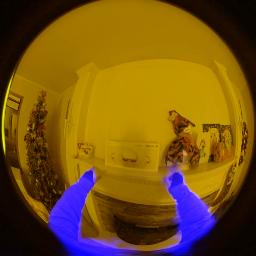}%
\includegraphics[width=0.05\linewidth, height=0.05\linewidth]{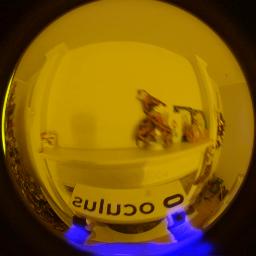}%
\includegraphics[width=0.05\linewidth, height=0.05\linewidth]{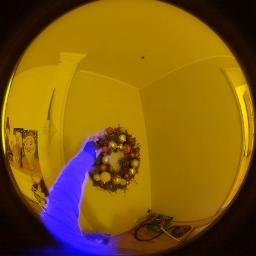}%
\includegraphics[width=0.05\linewidth, height=0.05\linewidth]{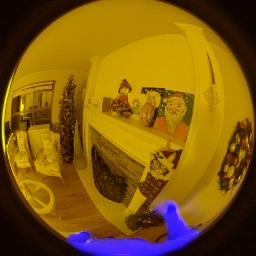}%
\includegraphics[width=0.05\linewidth, height=0.05\linewidth]{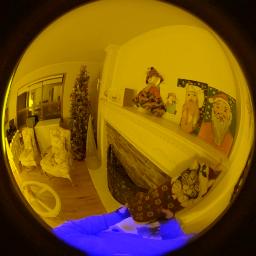}%
\includegraphics[width=0.05\linewidth, height=0.05\linewidth]{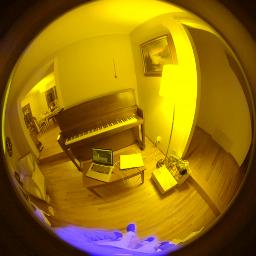}%
\includegraphics[width=0.05\linewidth, height=0.05\linewidth]{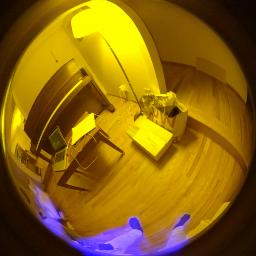}%
\includegraphics[width=0.05\linewidth, height=0.05\linewidth]{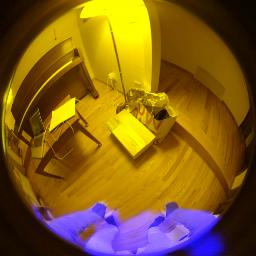}%
\includegraphics[width=0.05\linewidth, height=0.05\linewidth]{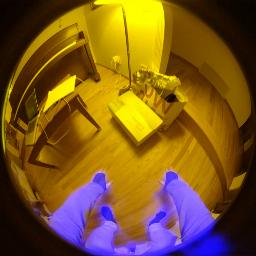}%
\includegraphics[width=0.05\linewidth, height=0.05\linewidth]{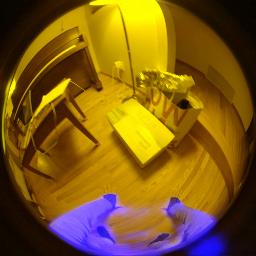}%
\includegraphics[width=0.05\linewidth, height=0.05\linewidth]{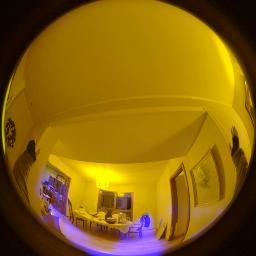}%
\includegraphics[width=0.05\linewidth, height=0.05\linewidth]{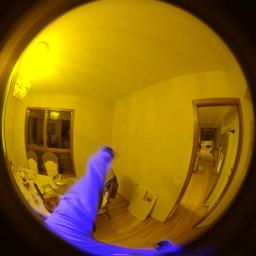}%
\includegraphics[width=0.05\linewidth, height=0.05\linewidth]{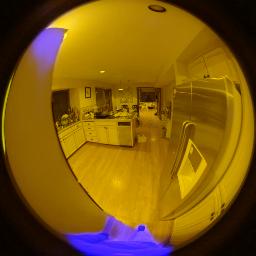}%
\includegraphics[width=0.05\linewidth, height=0.05\linewidth]{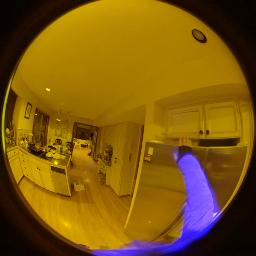}%
\\
 \rotatebox{90}{\hspace{4pt}{\tiny Ground Truth}} &
\includegraphics[width=0.05\linewidth, height=0.075\linewidth]{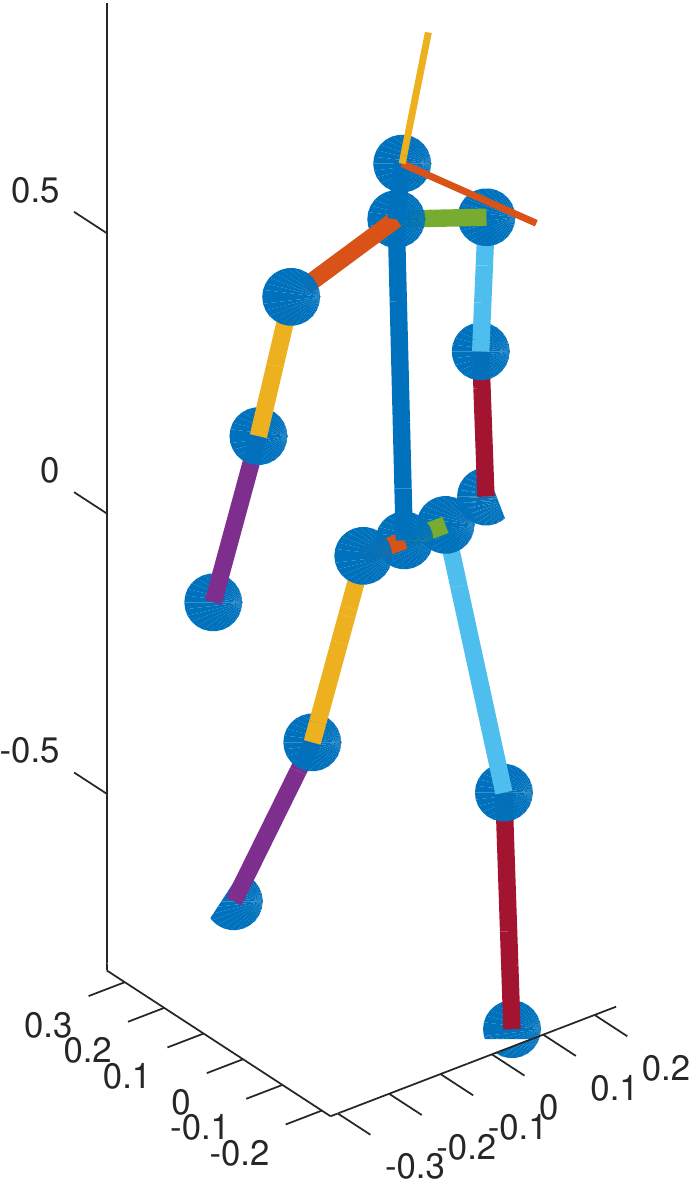}%
\includegraphics[width=0.05\linewidth, height=0.075\linewidth]{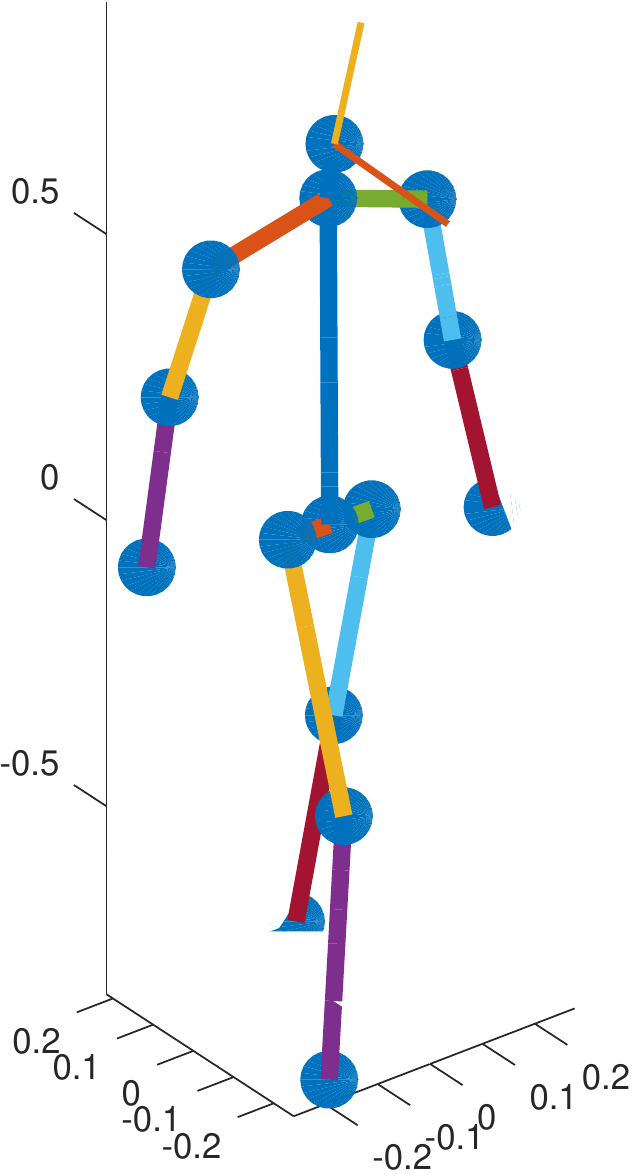}%
\includegraphics[width=0.05\linewidth, height=0.075\linewidth]{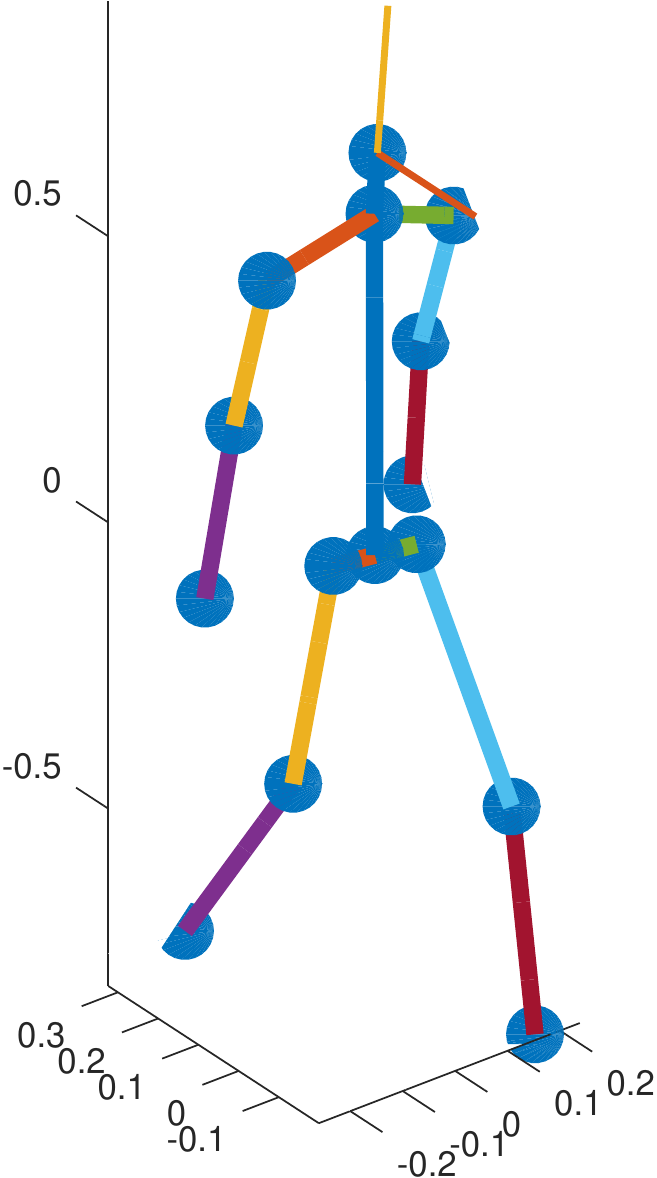}%
\includegraphics[width=0.05\linewidth, height=0.075\linewidth]{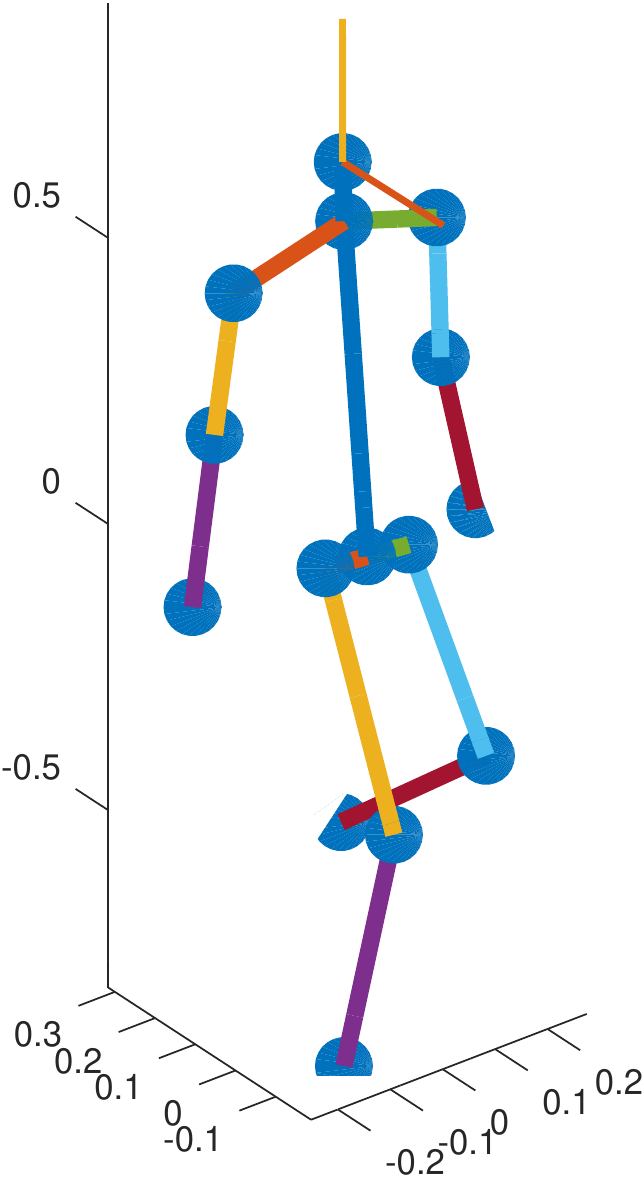}%
\includegraphics[width=0.05\linewidth, height=0.075\linewidth]{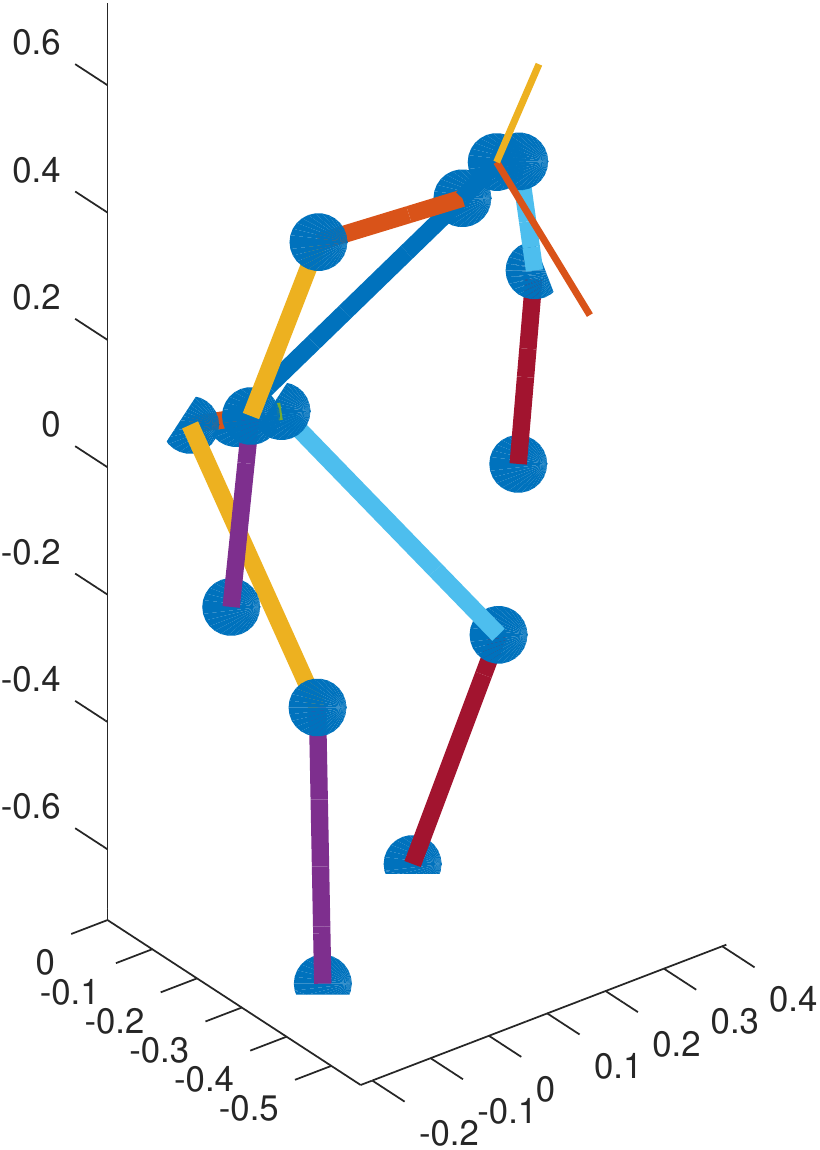}%
\includegraphics[width=0.05\linewidth, height=0.075\linewidth]{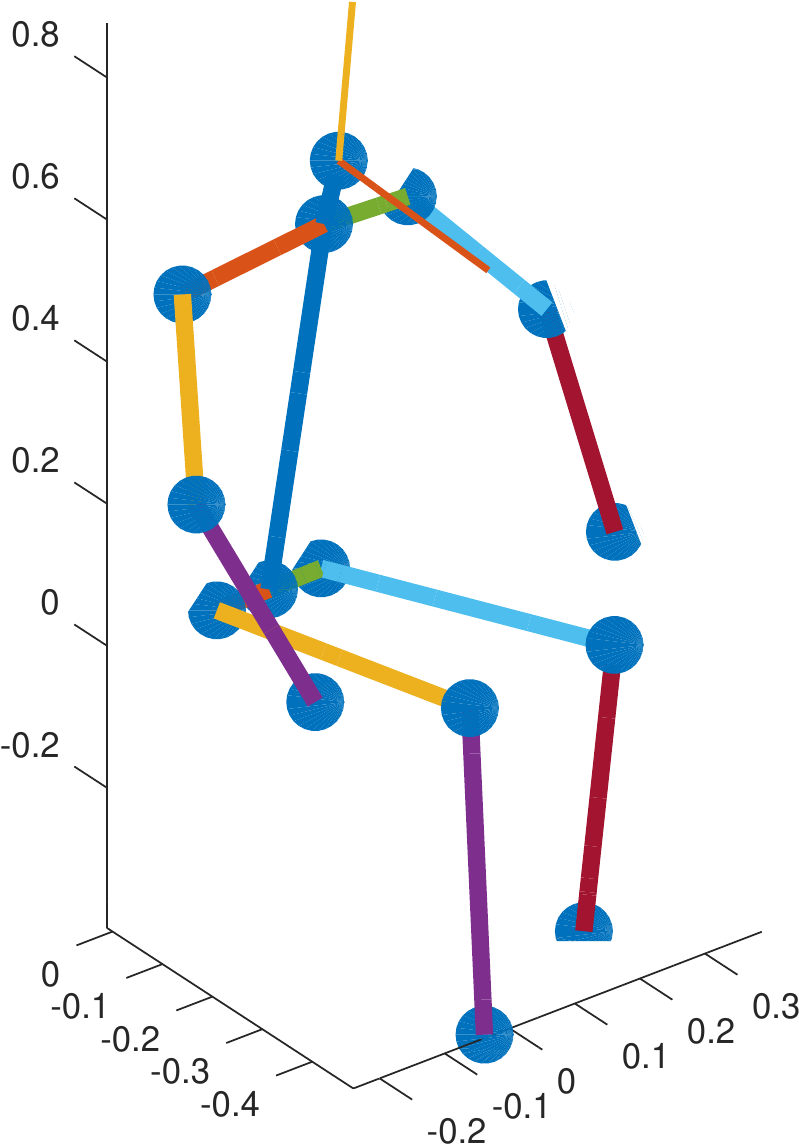}%
\includegraphics[width=0.05\linewidth, height=0.075\linewidth]{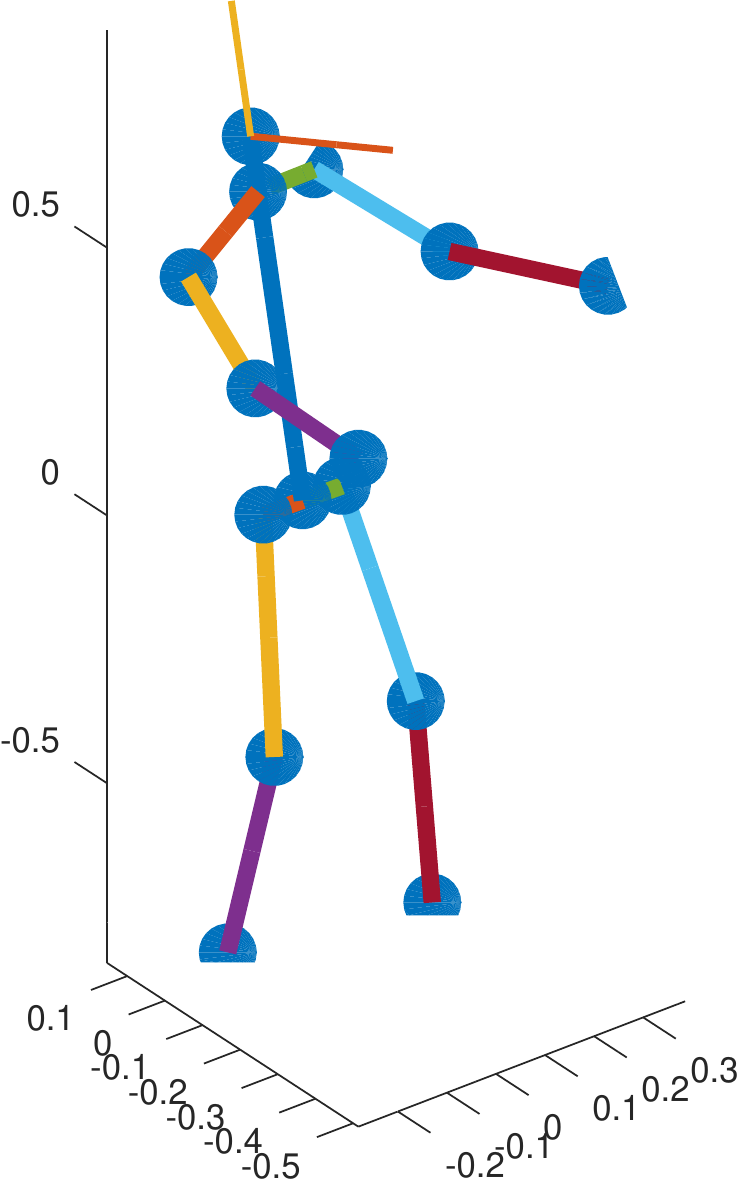}%
\includegraphics[width=0.05\linewidth, height=0.075\linewidth]{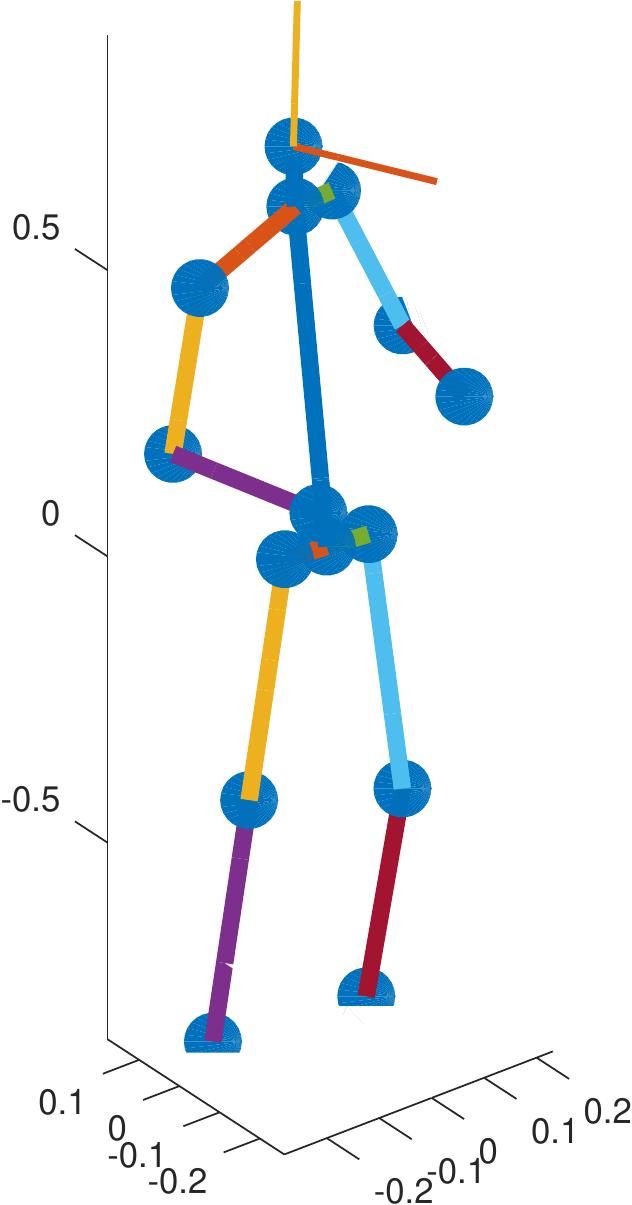}%
\includegraphics[width=0.05\linewidth, height=0.075\linewidth]{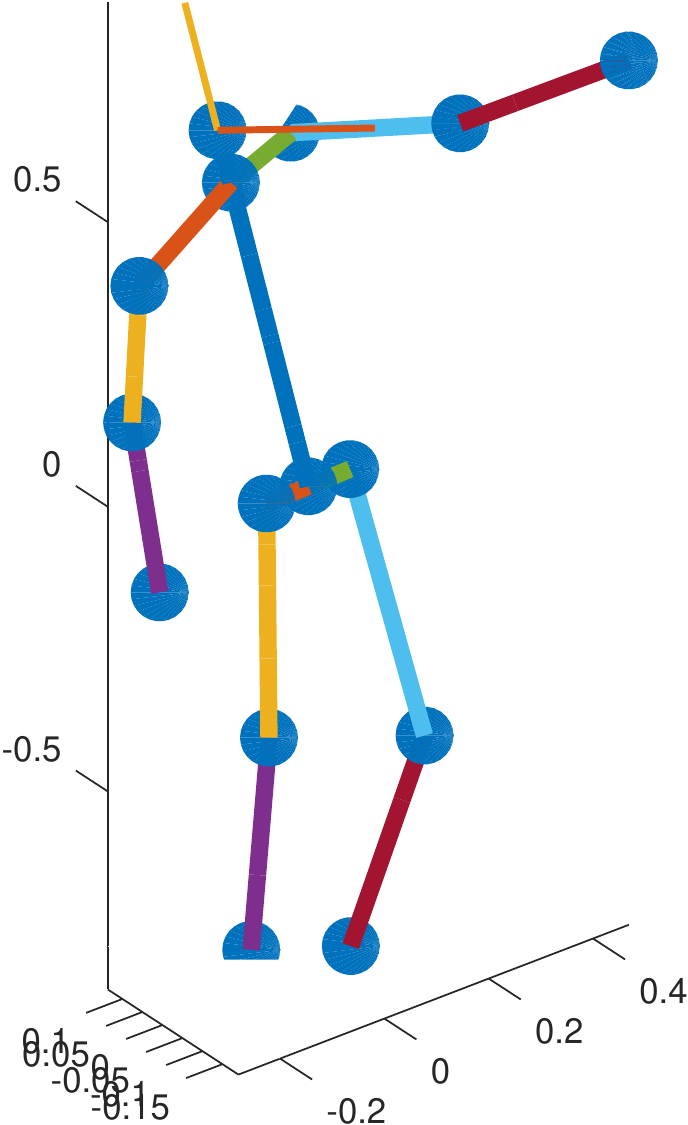}%
\includegraphics[width=0.05\linewidth, height=0.075\linewidth]{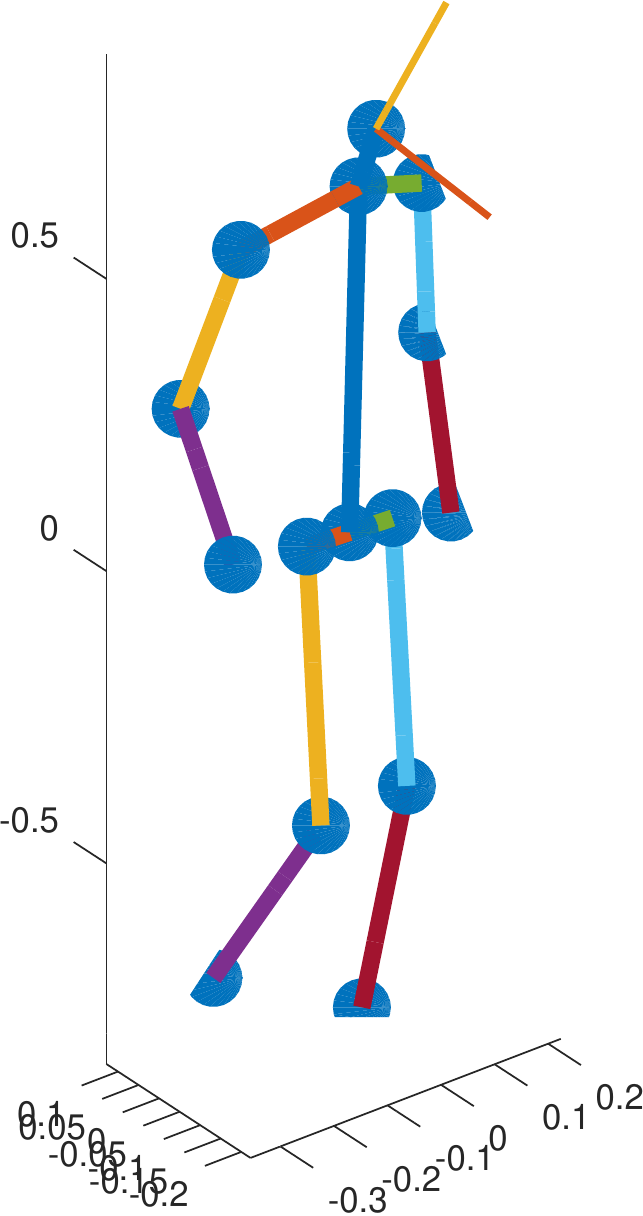}%
\includegraphics[width=0.05\linewidth, height=0.075\linewidth]{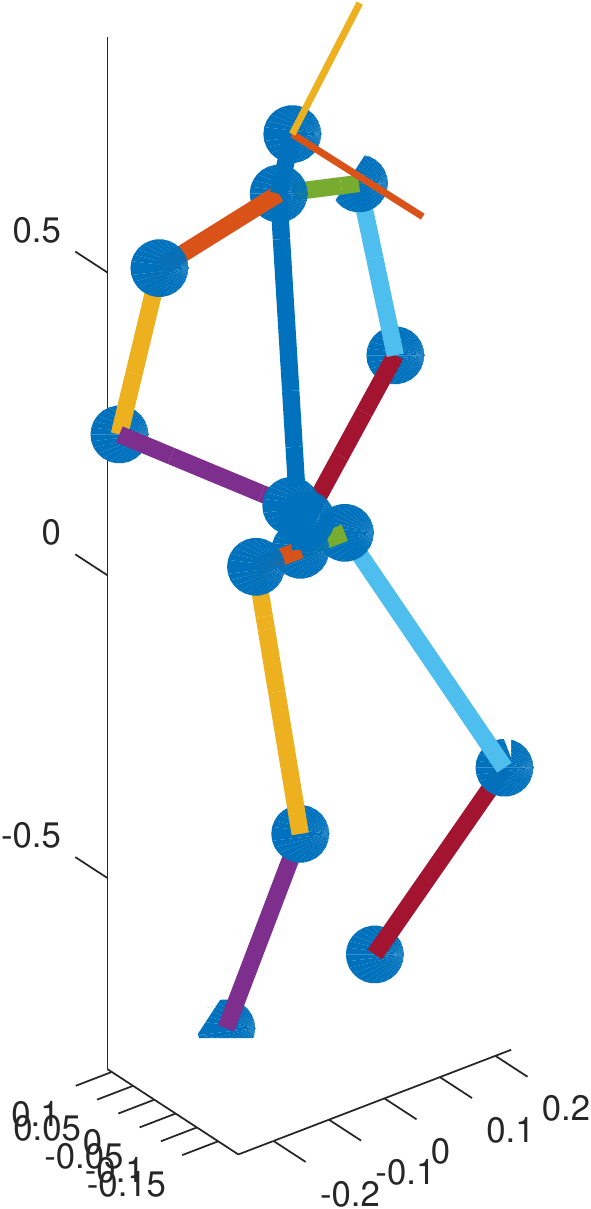}%
\includegraphics[width=0.05\linewidth, height=0.075\linewidth]{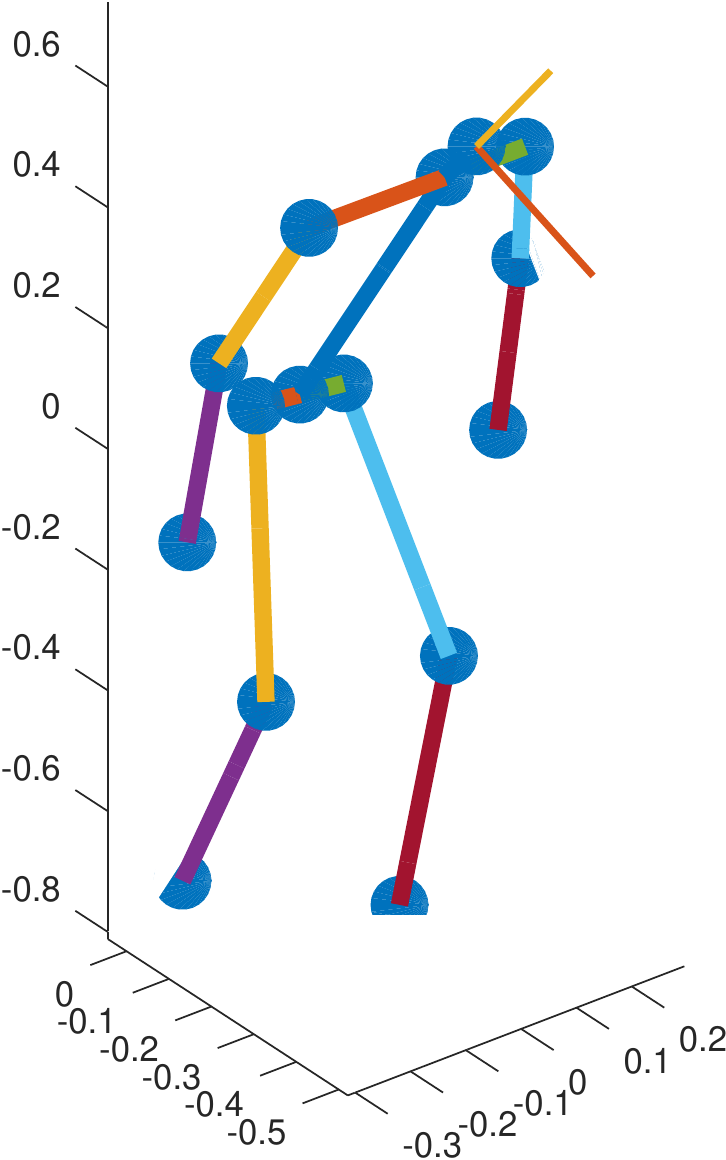}%
\includegraphics[width=0.05\linewidth, height=0.075\linewidth]{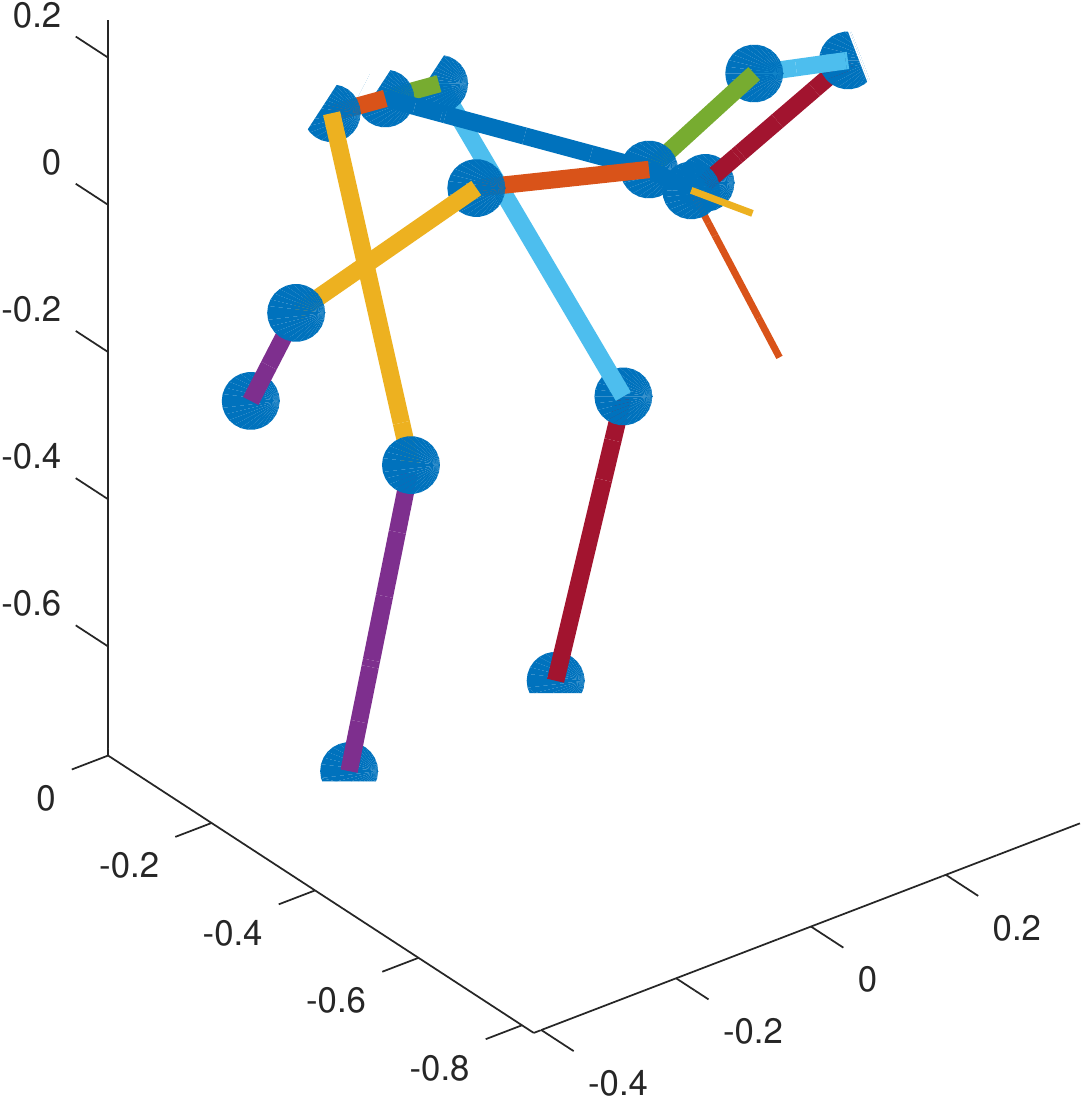}%
\includegraphics[width=0.05\linewidth, height=0.075\linewidth]{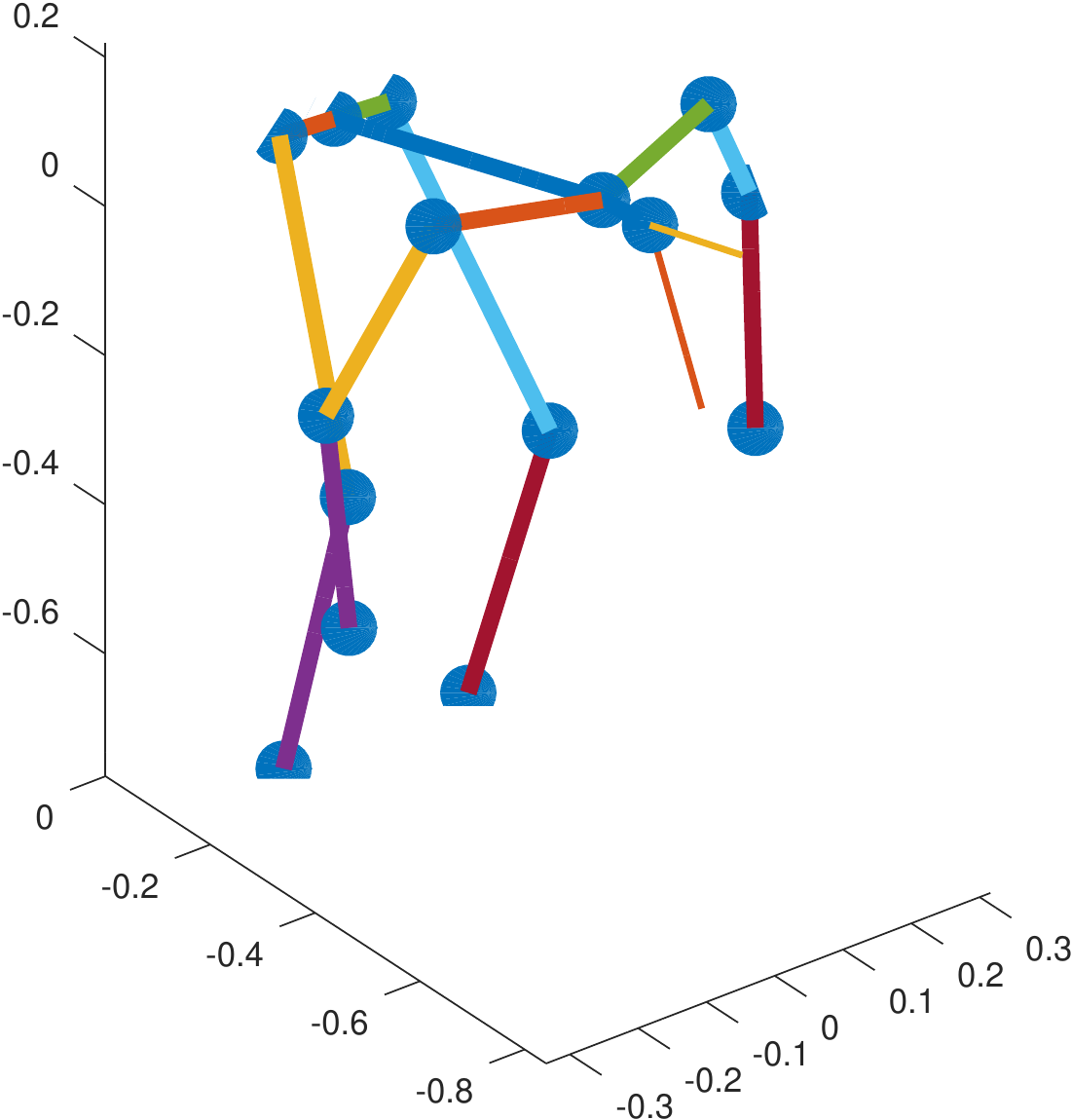}%
\includegraphics[width=0.05\linewidth, height=0.075\linewidth]{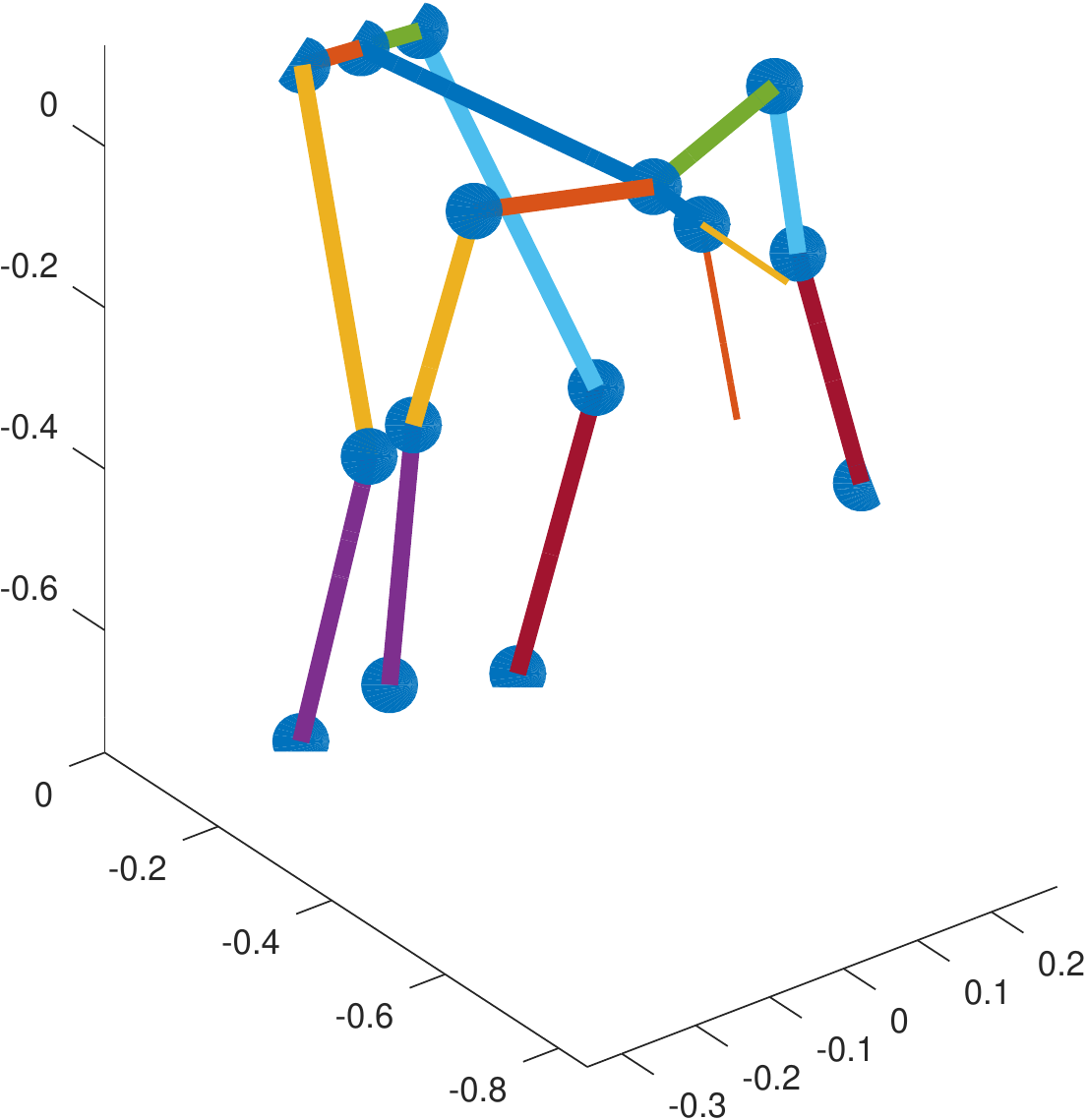}%
\includegraphics[width=0.05\linewidth, height=0.075\linewidth]{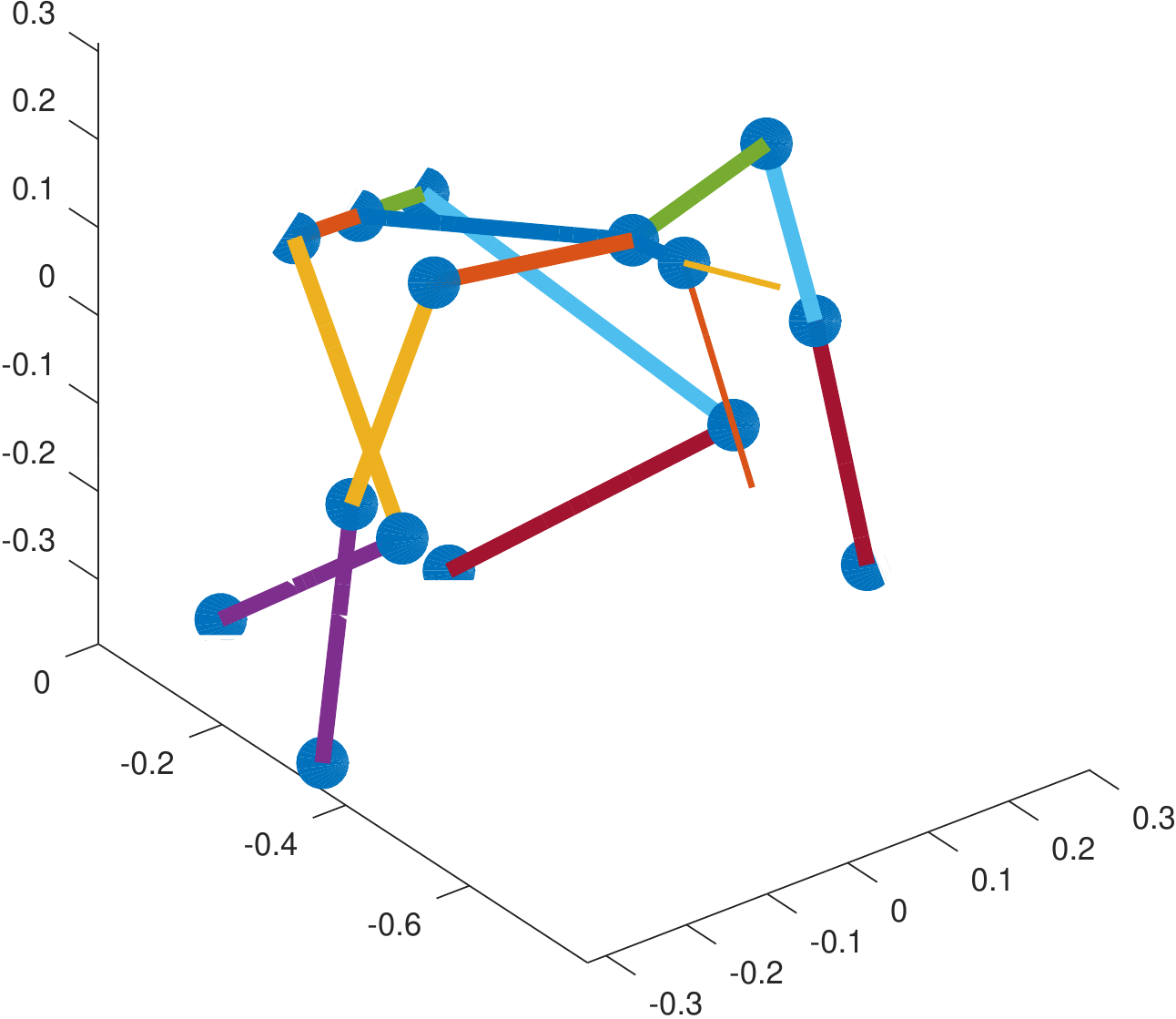}%
\includegraphics[width=0.05\linewidth, height=0.075\linewidth]{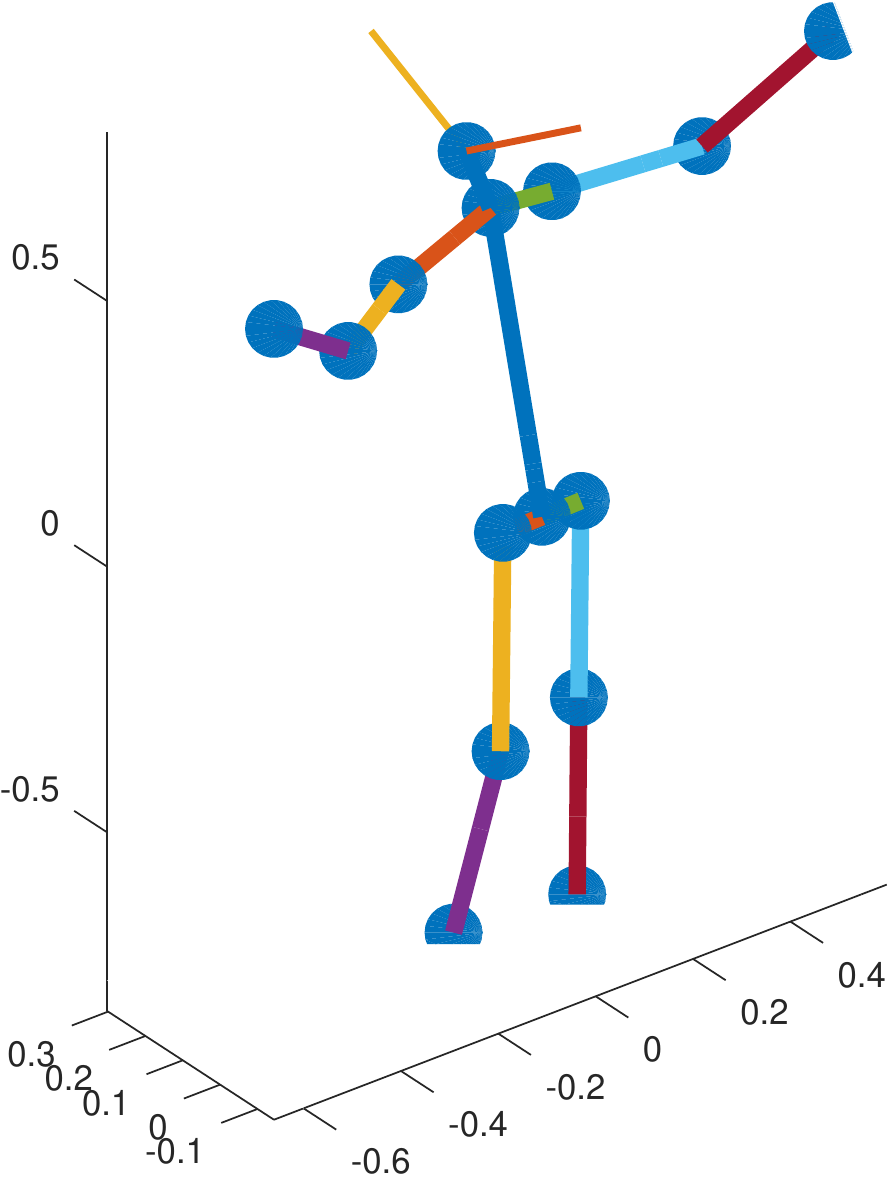}%
\includegraphics[width=0.05\linewidth, height=0.075\linewidth]{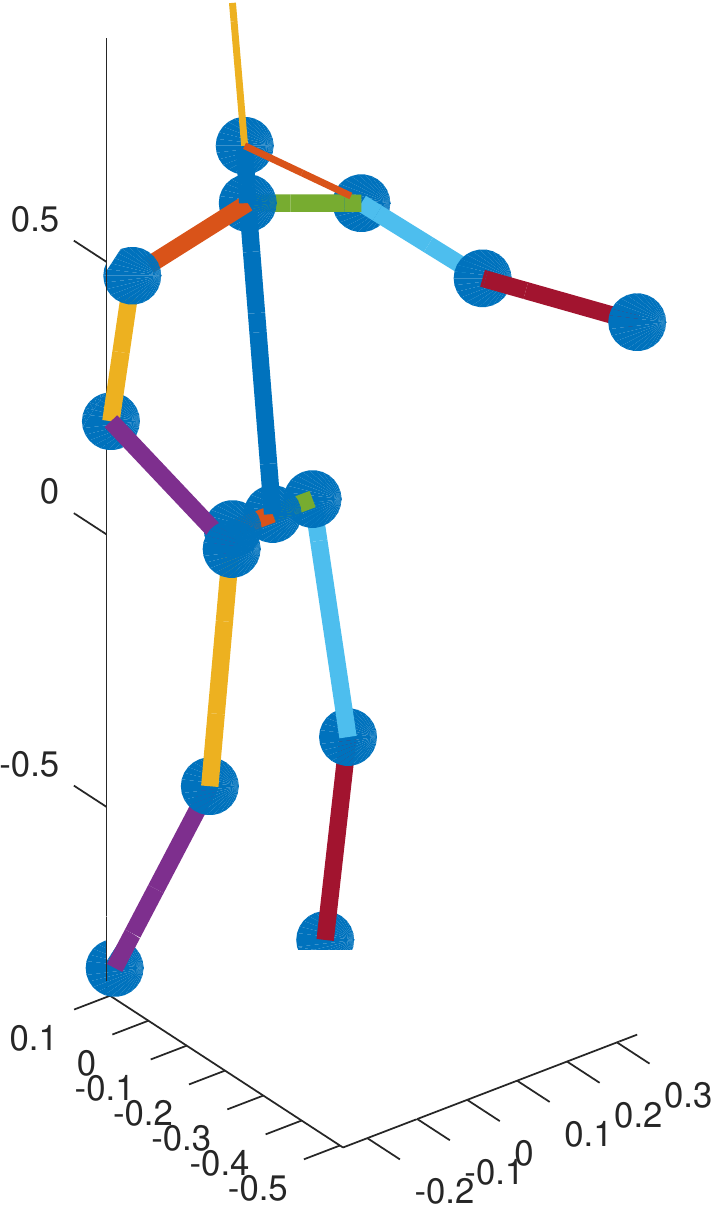}%
\includegraphics[width=0.05\linewidth, height=0.075\linewidth]{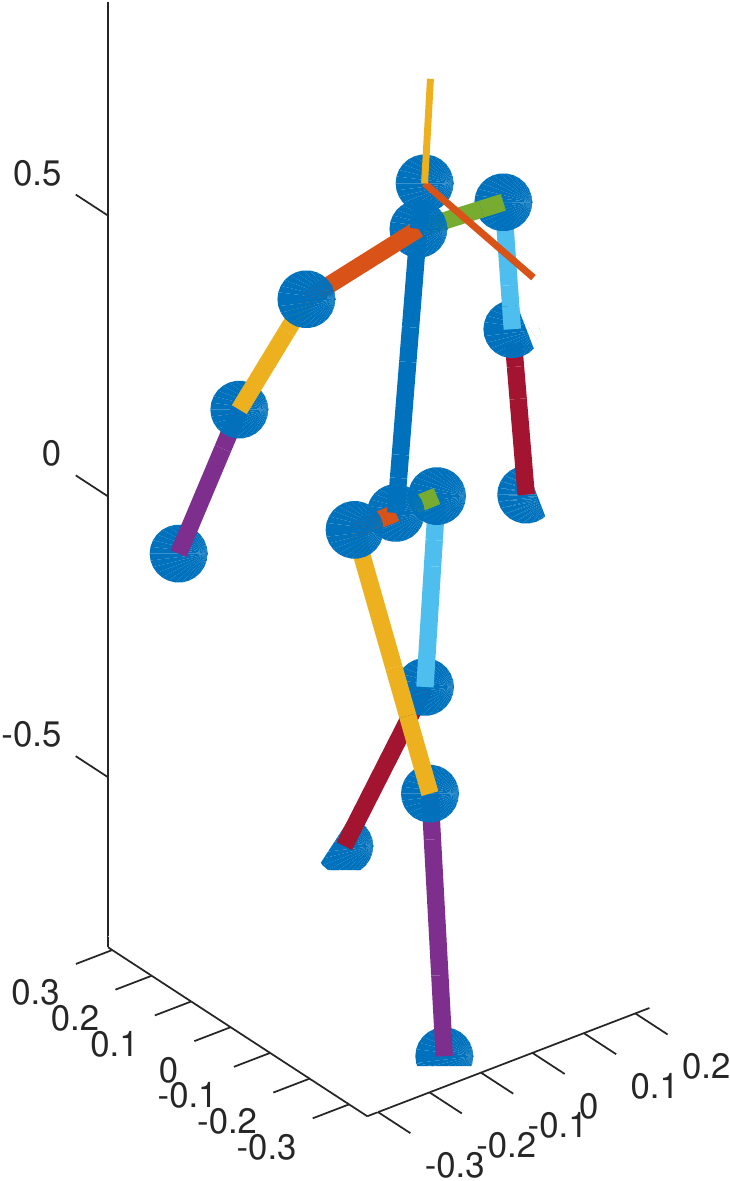}%
\includegraphics[width=0.05\linewidth, height=0.075\linewidth]{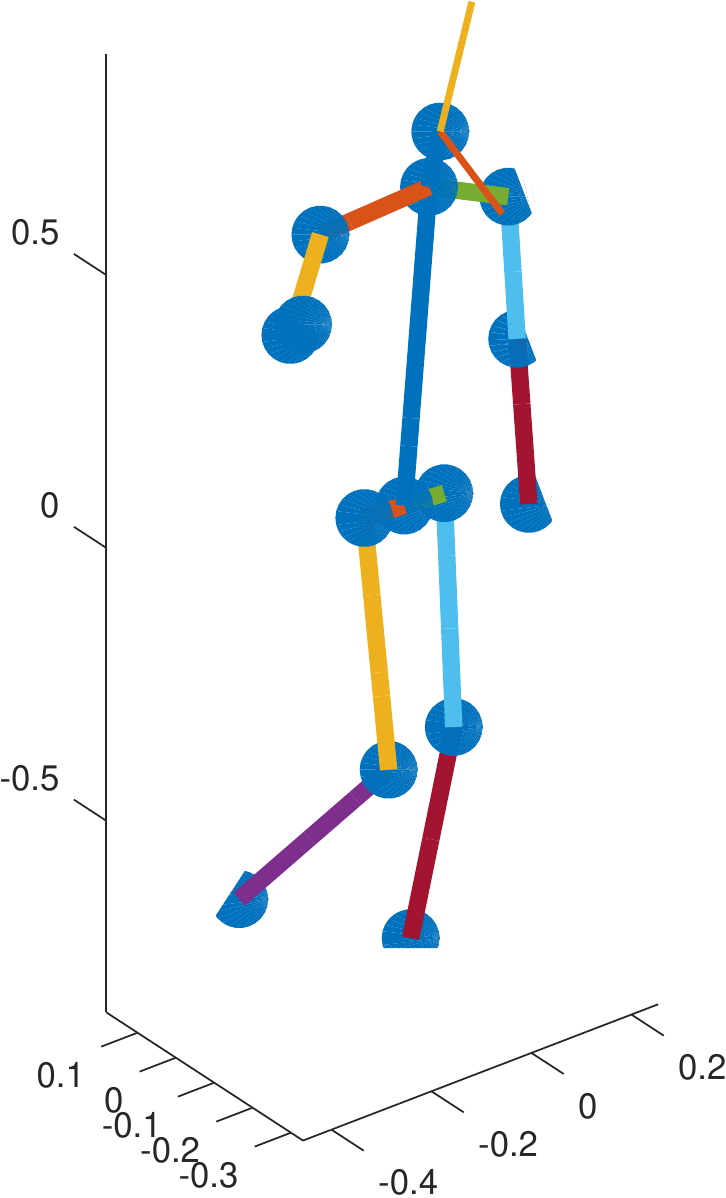}%
\\
 \rotatebox{90}{\hspace{15pt}{\tiny Ours}} &
\includegraphics[width=0.05\linewidth, height=0.075\linewidth]{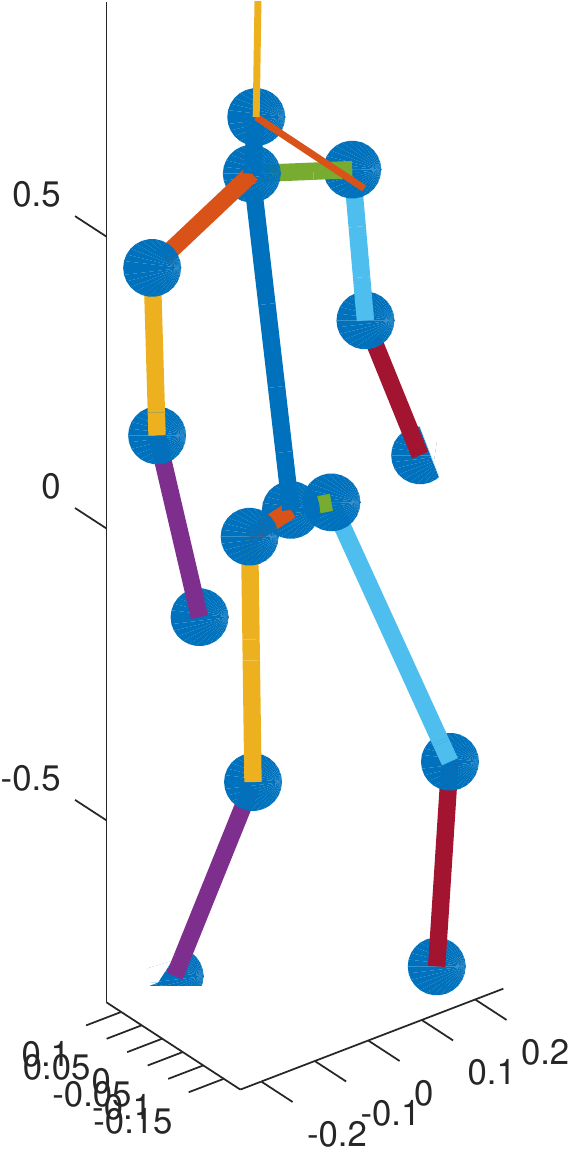}%
\includegraphics[width=0.05\linewidth, height=0.075\linewidth]{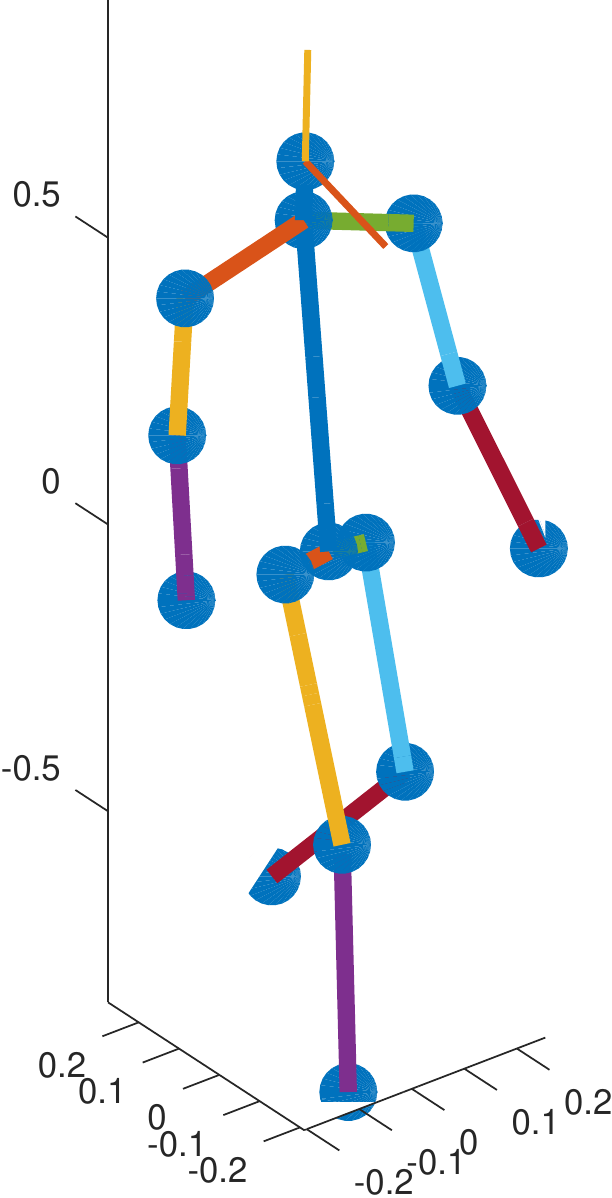}%
\includegraphics[width=0.05\linewidth, height=0.075\linewidth]{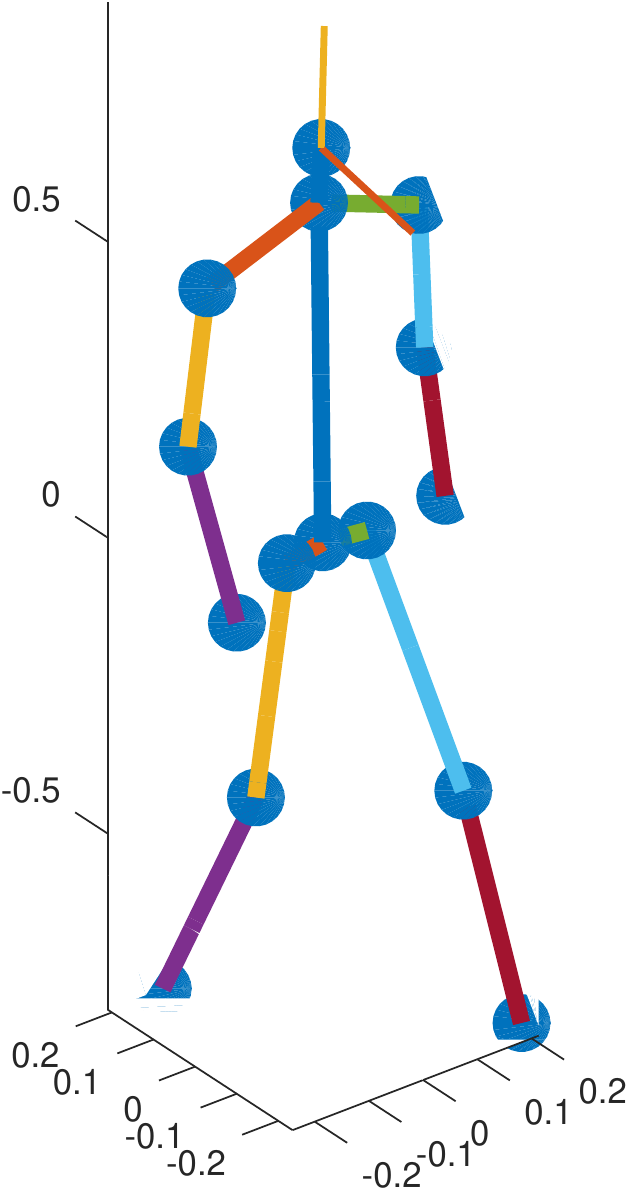}%
\includegraphics[width=0.05\linewidth, height=0.075\linewidth]{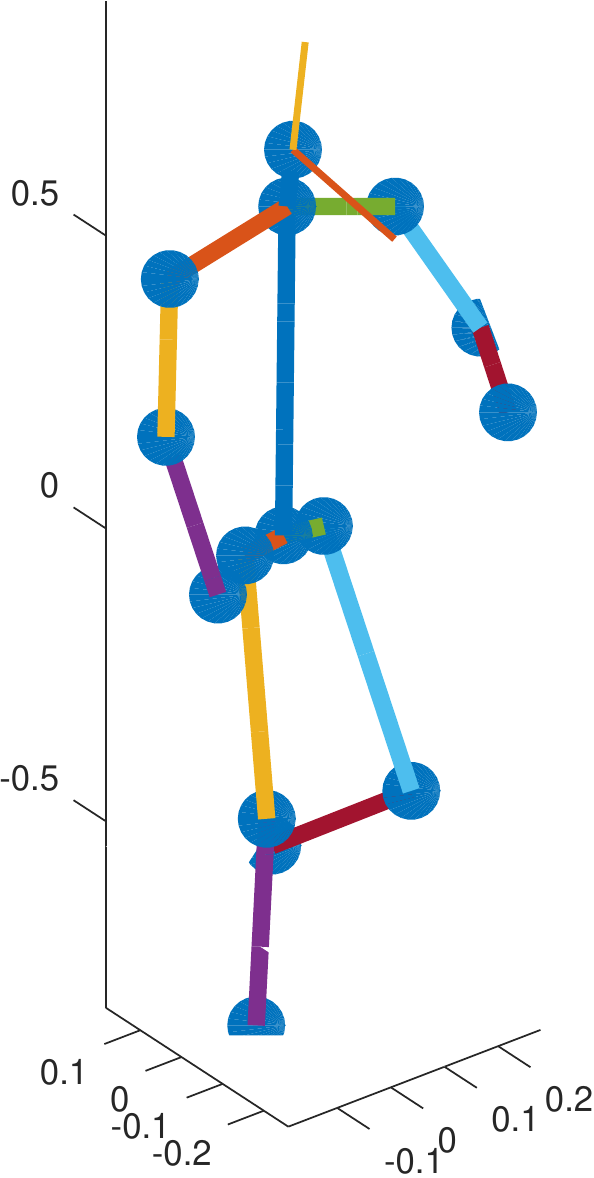}%
\includegraphics[width=0.05\linewidth, height=0.075\linewidth]{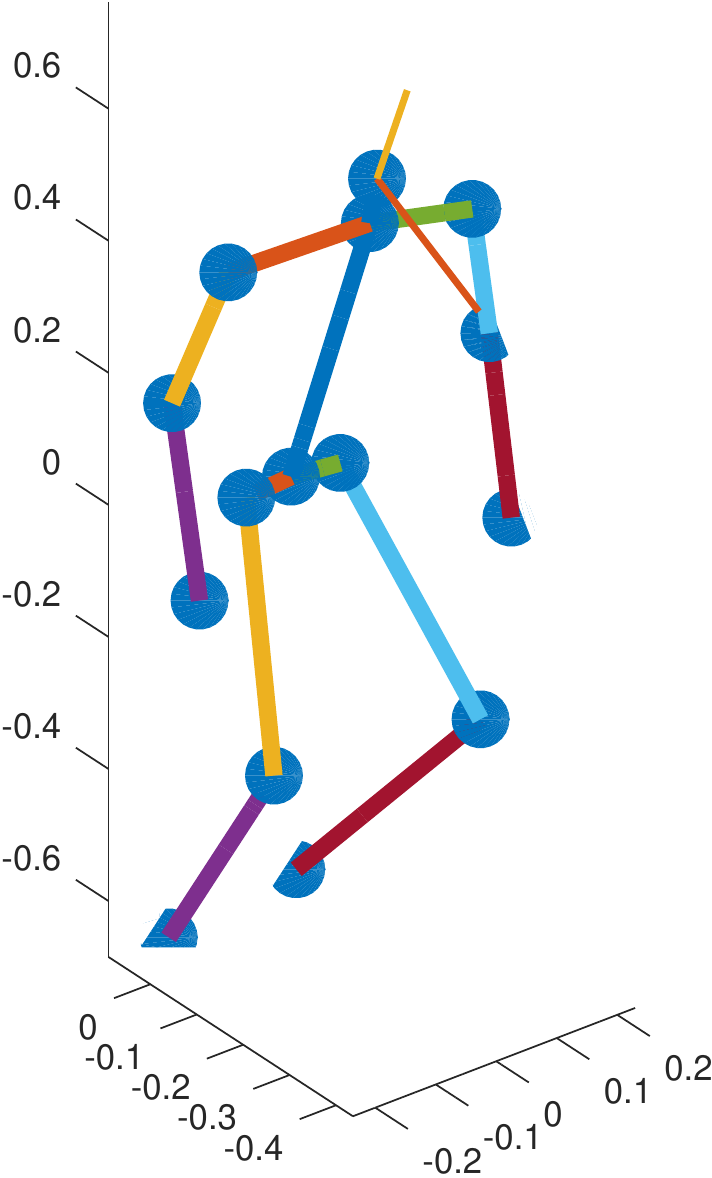}%
\includegraphics[width=0.05\linewidth, height=0.075\linewidth]{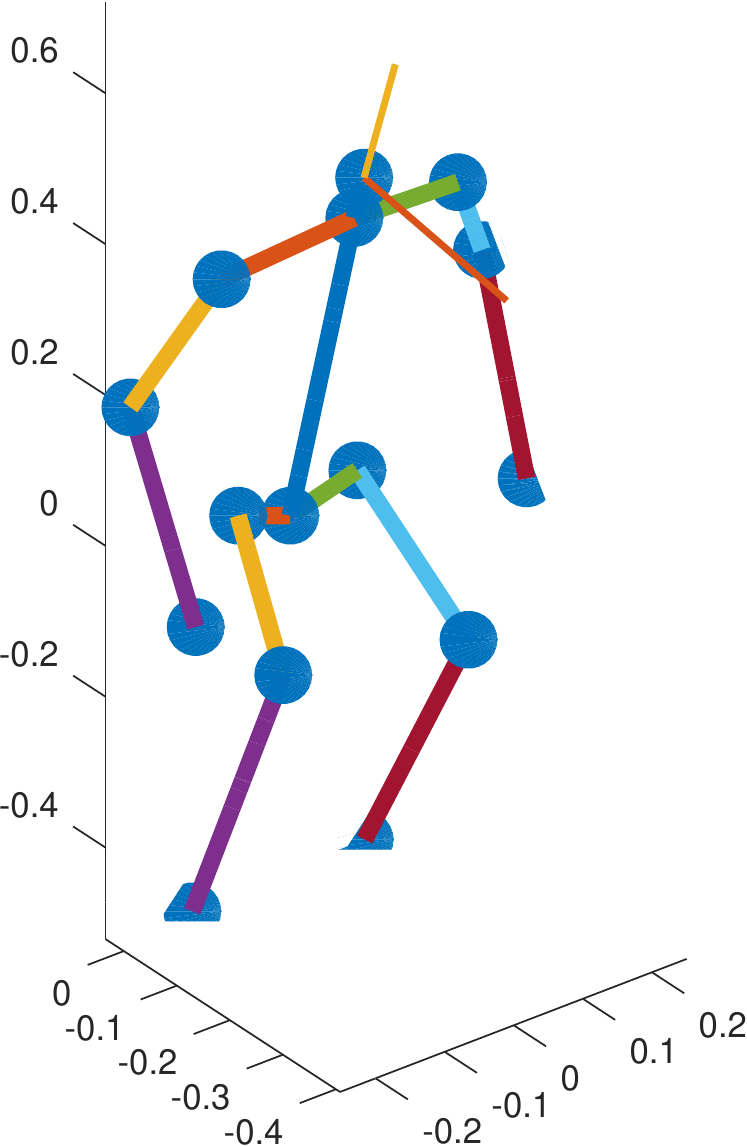}%
\includegraphics[width=0.05\linewidth, height=0.075\linewidth]{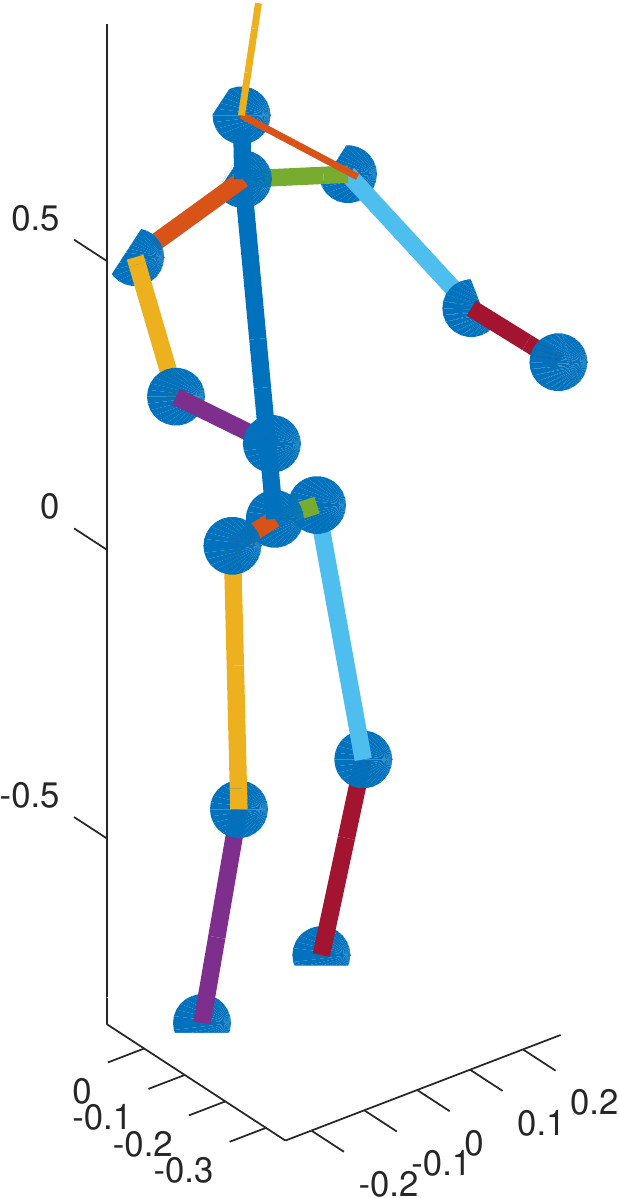}%
\includegraphics[width=0.05\linewidth, height=0.075\linewidth]{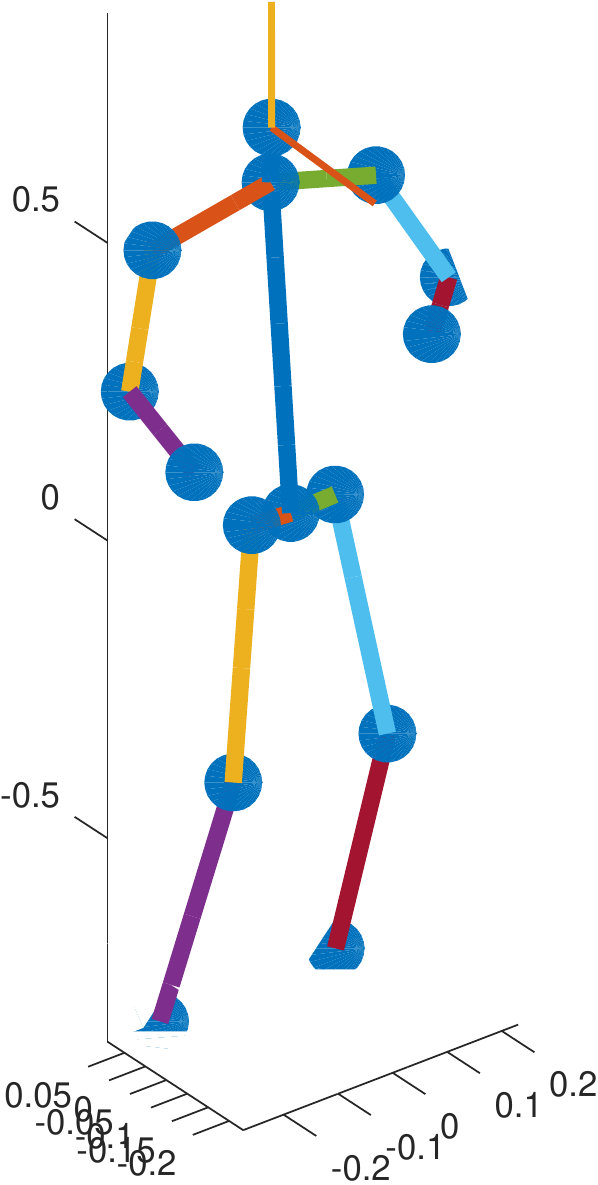}%
\includegraphics[width=0.05\linewidth, height=0.075\linewidth]{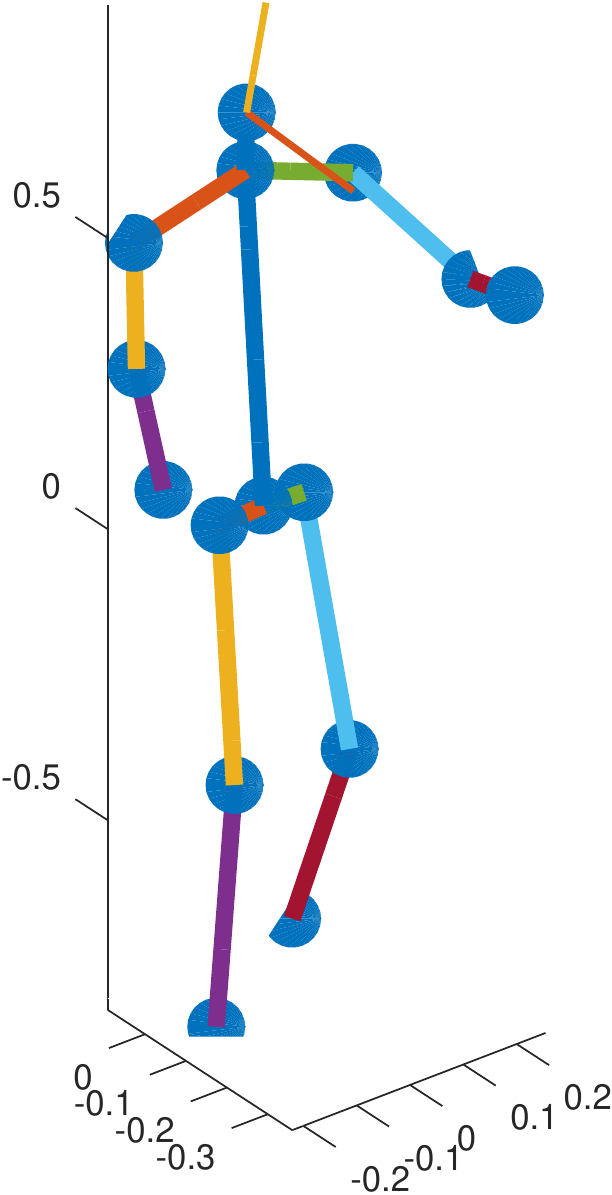}%
\includegraphics[width=0.05\linewidth, height=0.075\linewidth]{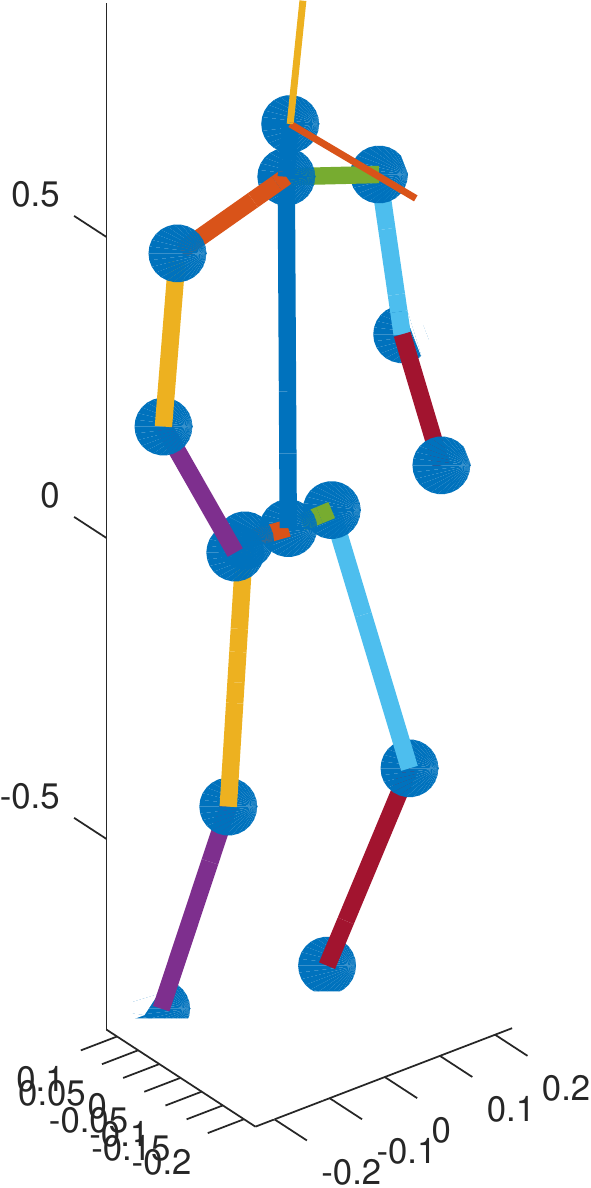}%
\includegraphics[width=0.05\linewidth, height=0.075\linewidth]{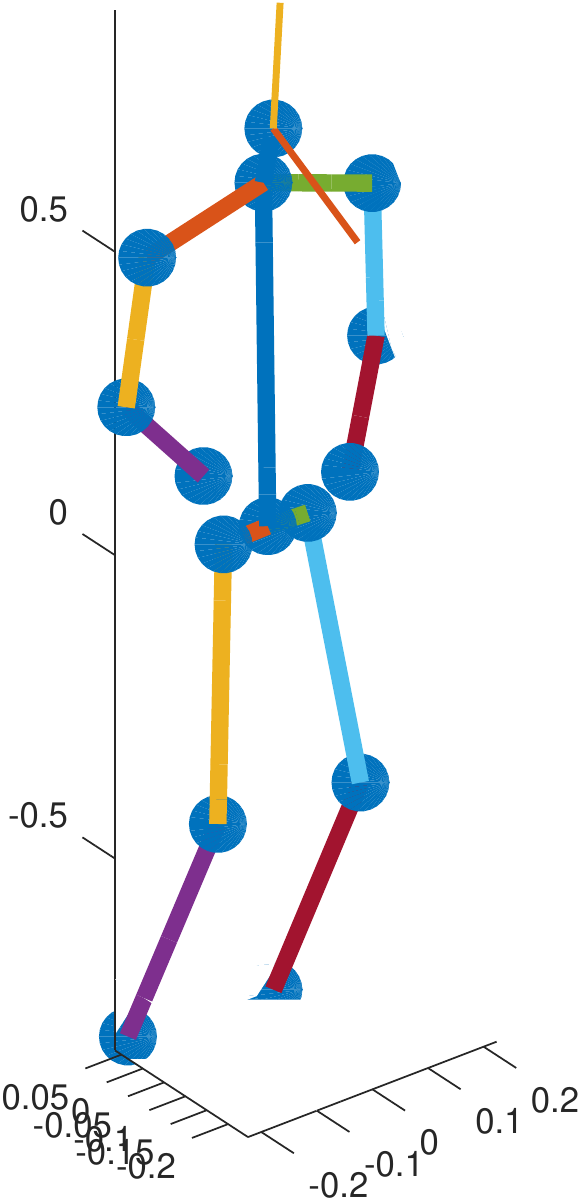}%
\includegraphics[width=0.05\linewidth, height=0.075\linewidth]{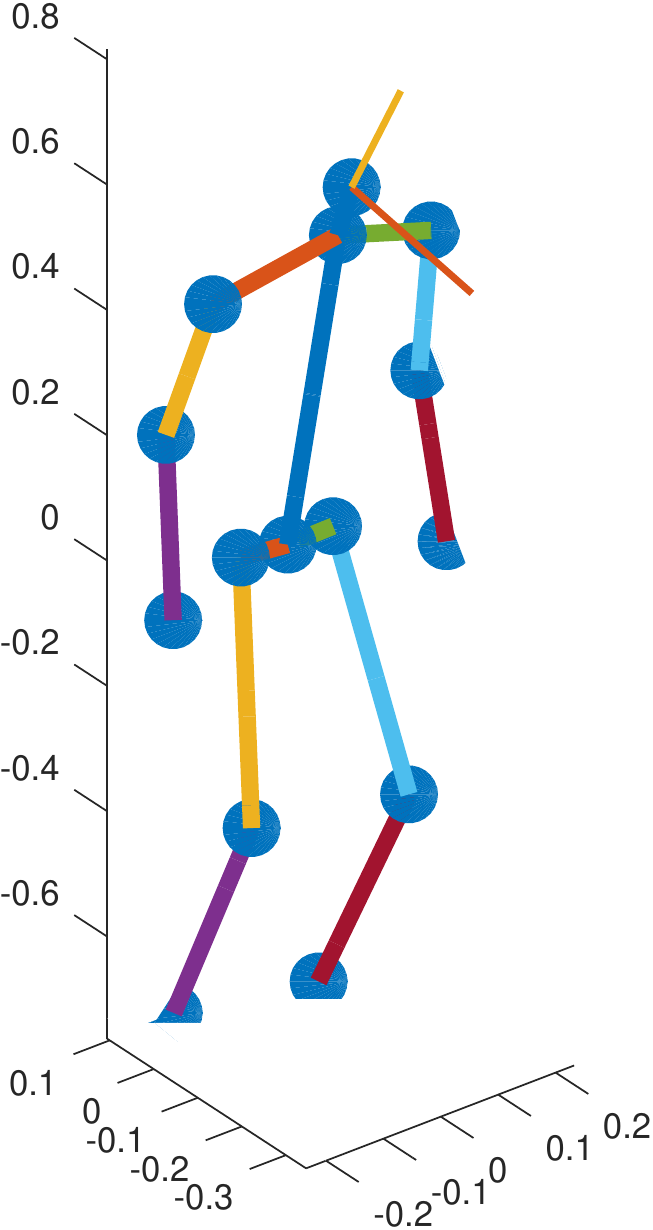}%
\includegraphics[width=0.05\linewidth, height=0.075\linewidth]{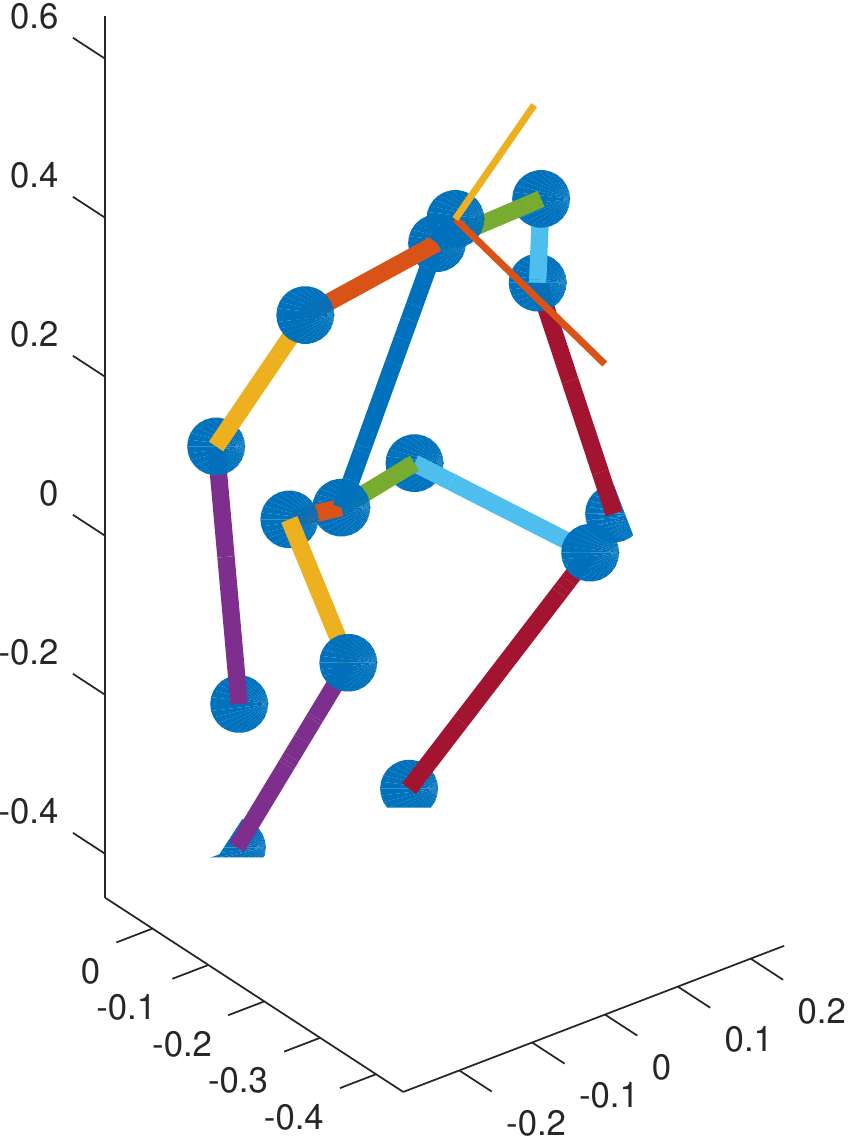}%
\includegraphics[width=0.05\linewidth, height=0.075\linewidth]{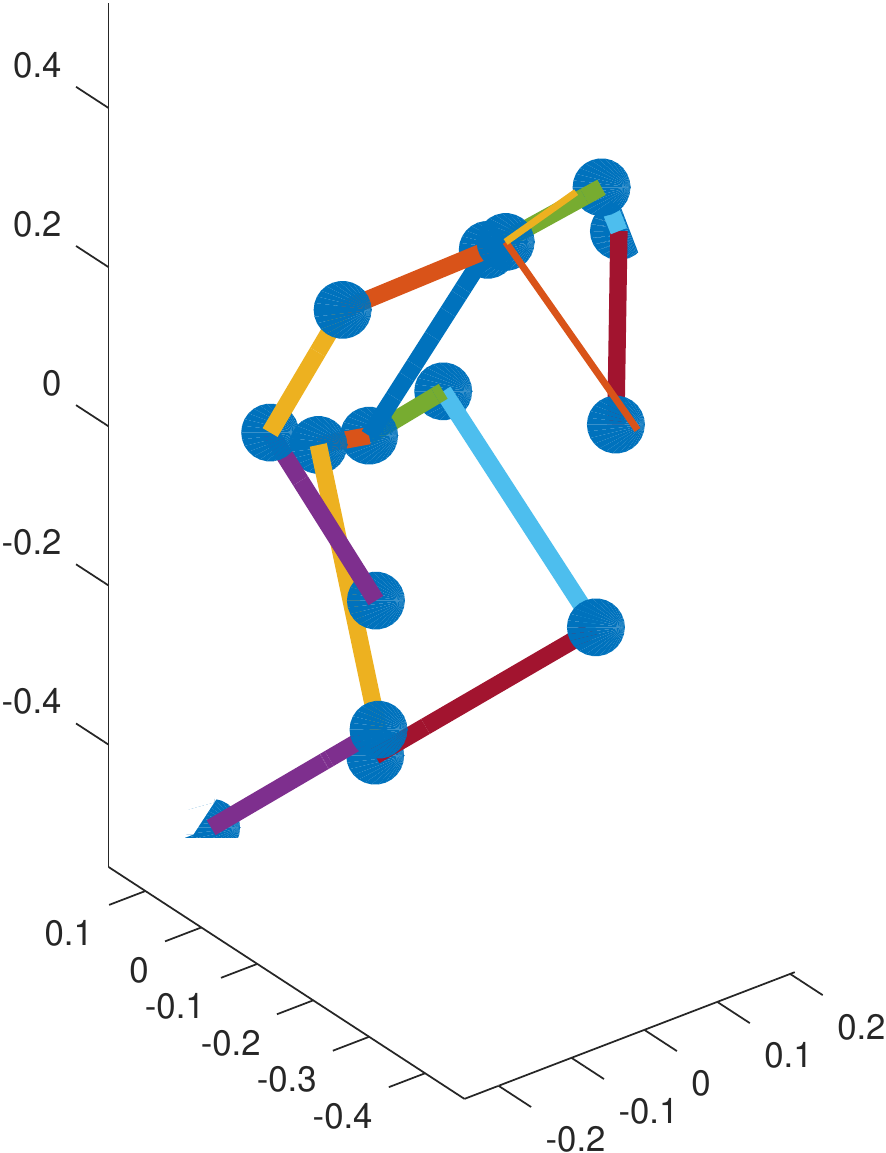}%
\includegraphics[width=0.05\linewidth, height=0.075\linewidth]{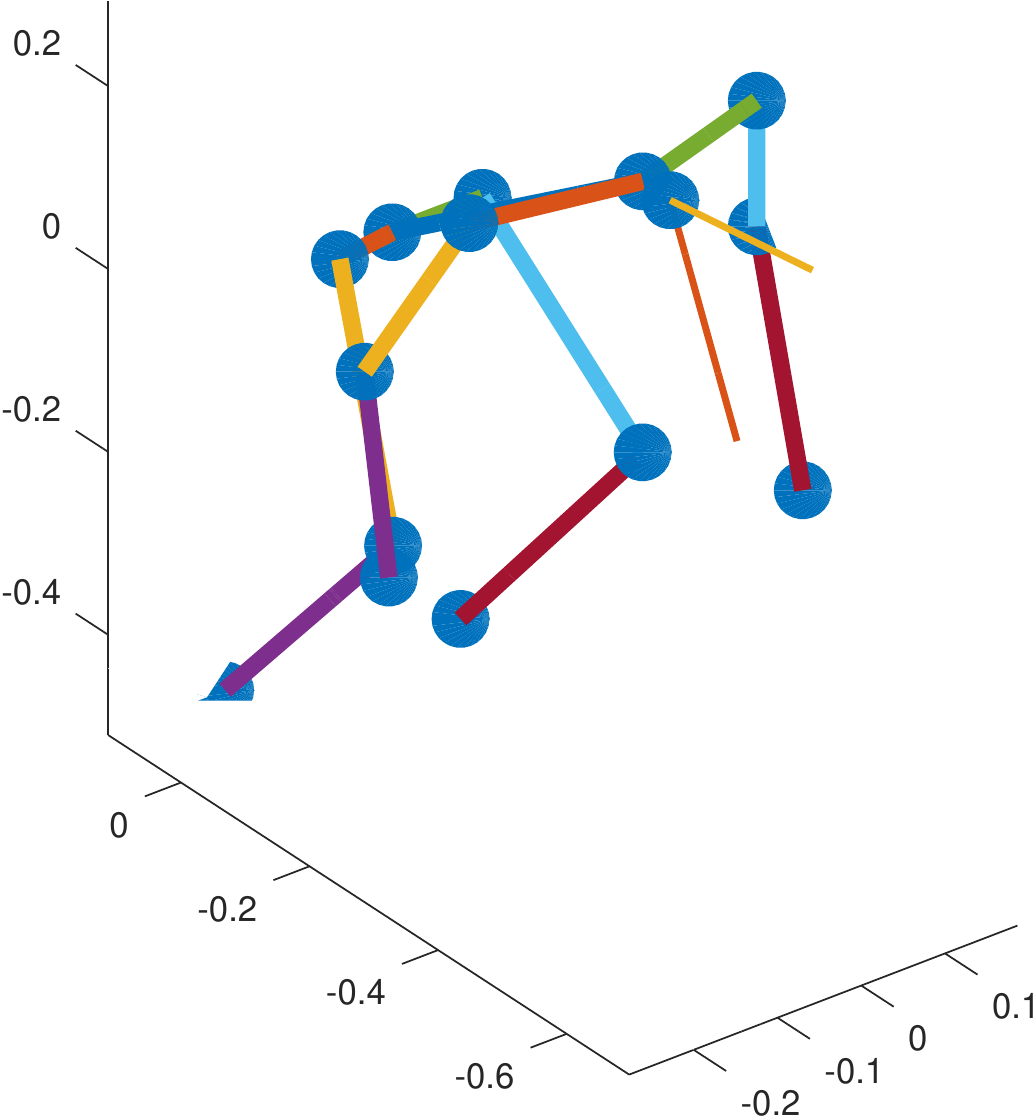}%
\includegraphics[width=0.05\linewidth, height=0.075\linewidth]{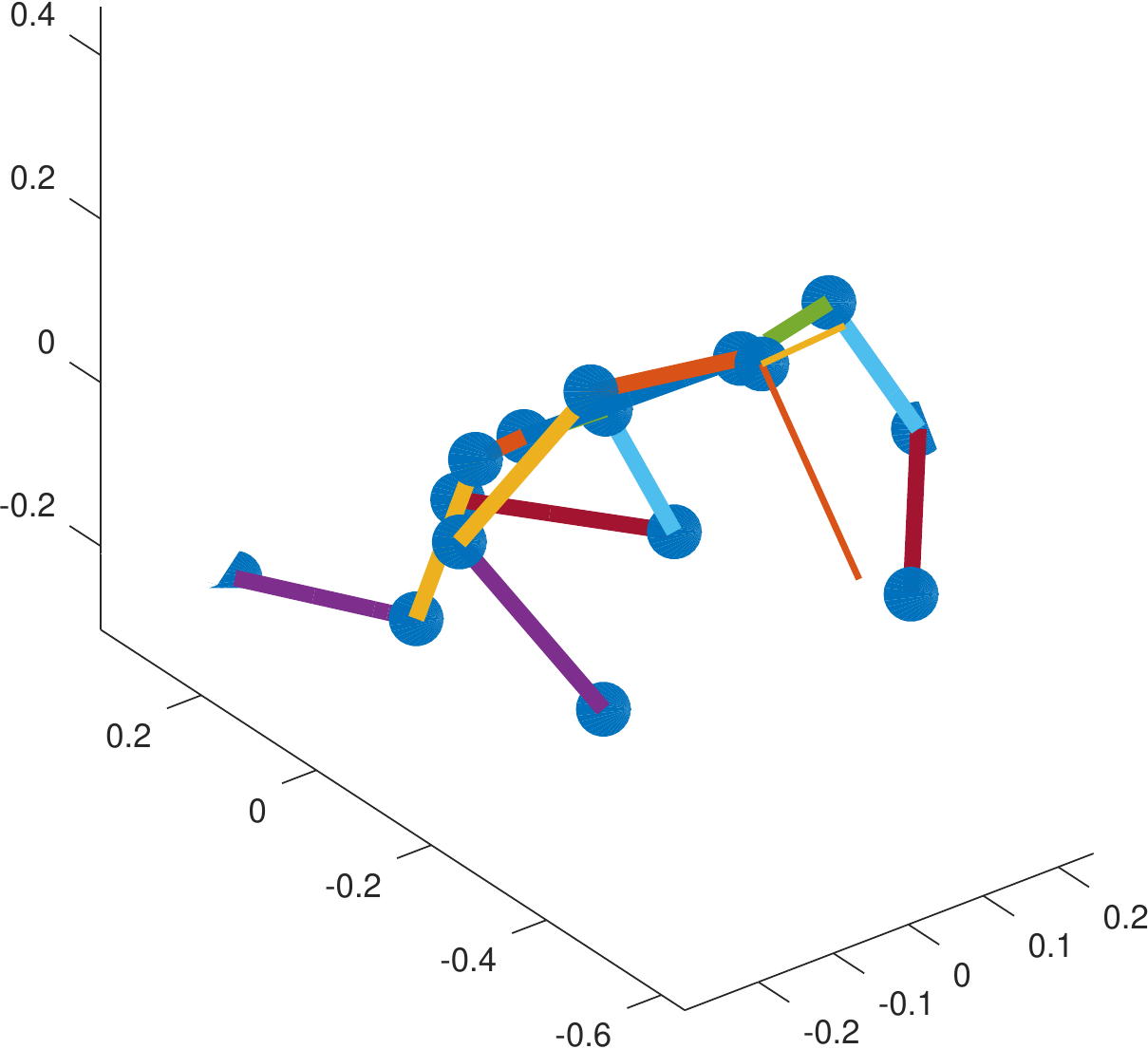}%
\includegraphics[width=0.05\linewidth, height=0.075\linewidth]{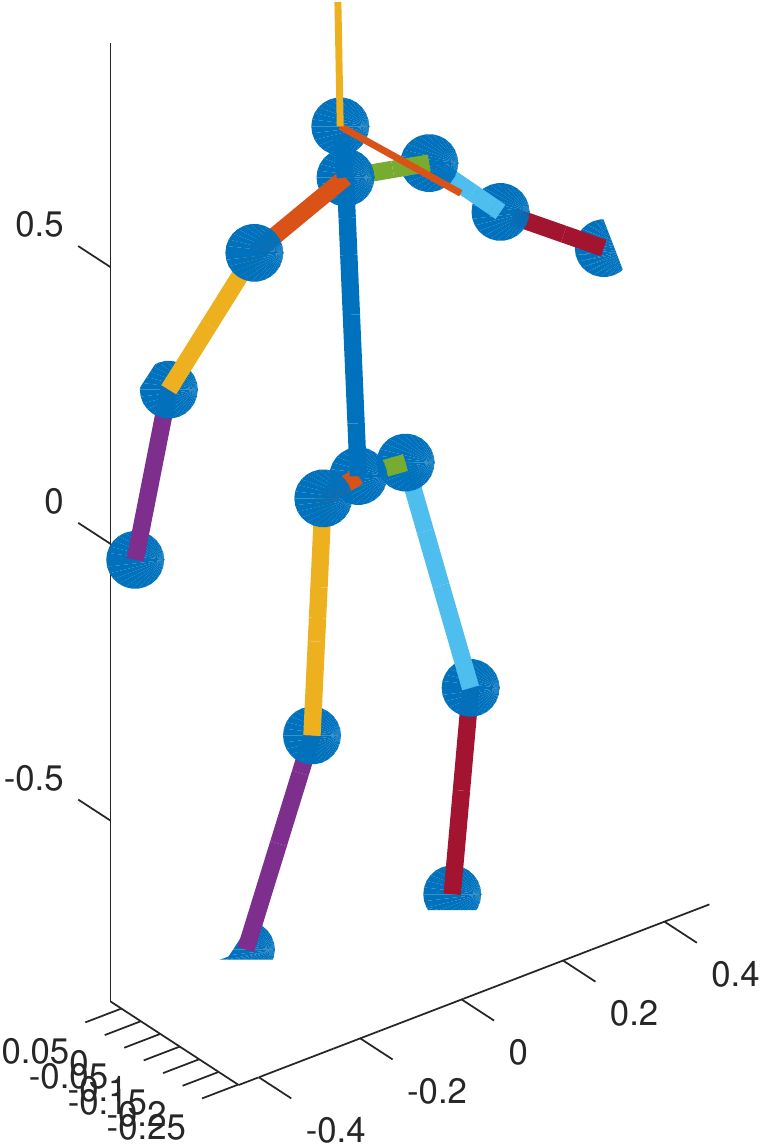}%
\includegraphics[width=0.05\linewidth, height=0.075\linewidth]{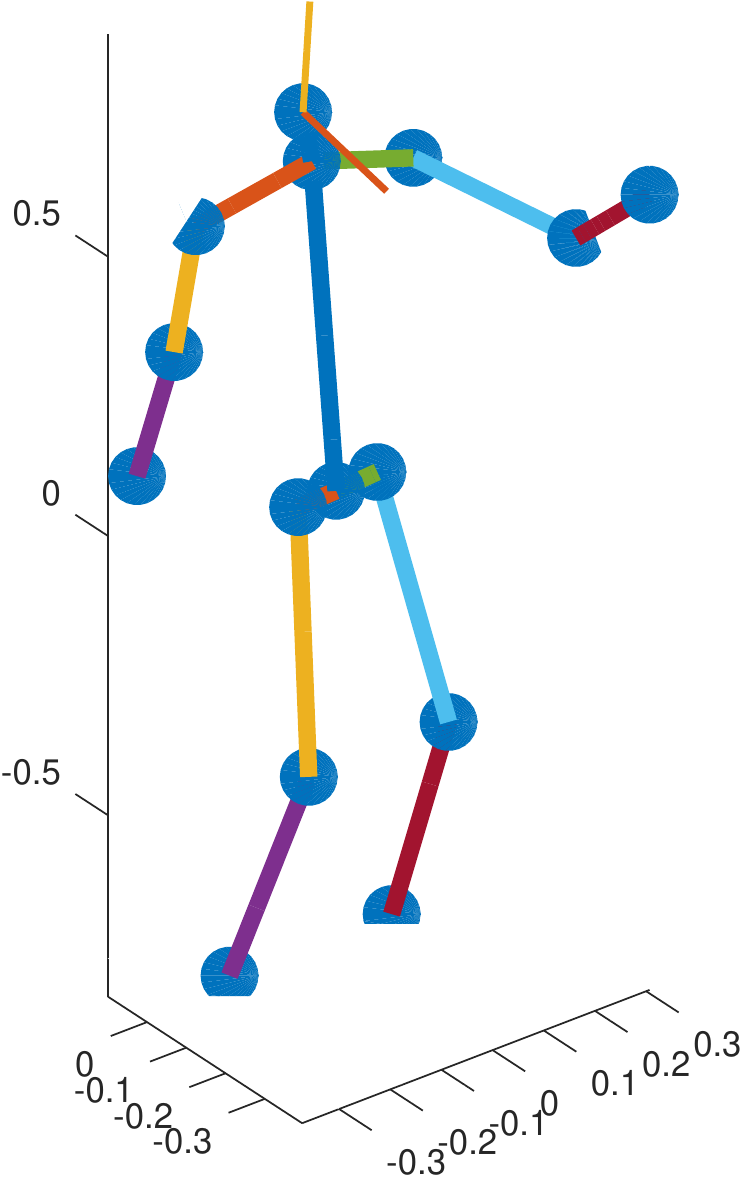}%
\includegraphics[width=0.05\linewidth, height=0.075\linewidth]{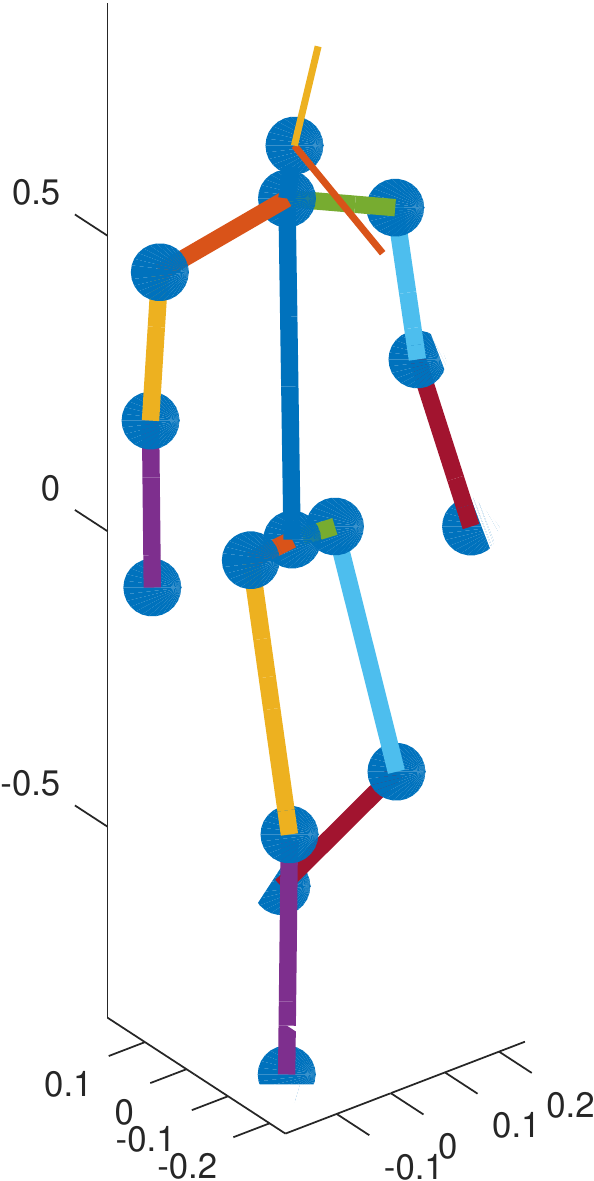}%
\includegraphics[width=0.05\linewidth, height=0.075\linewidth]{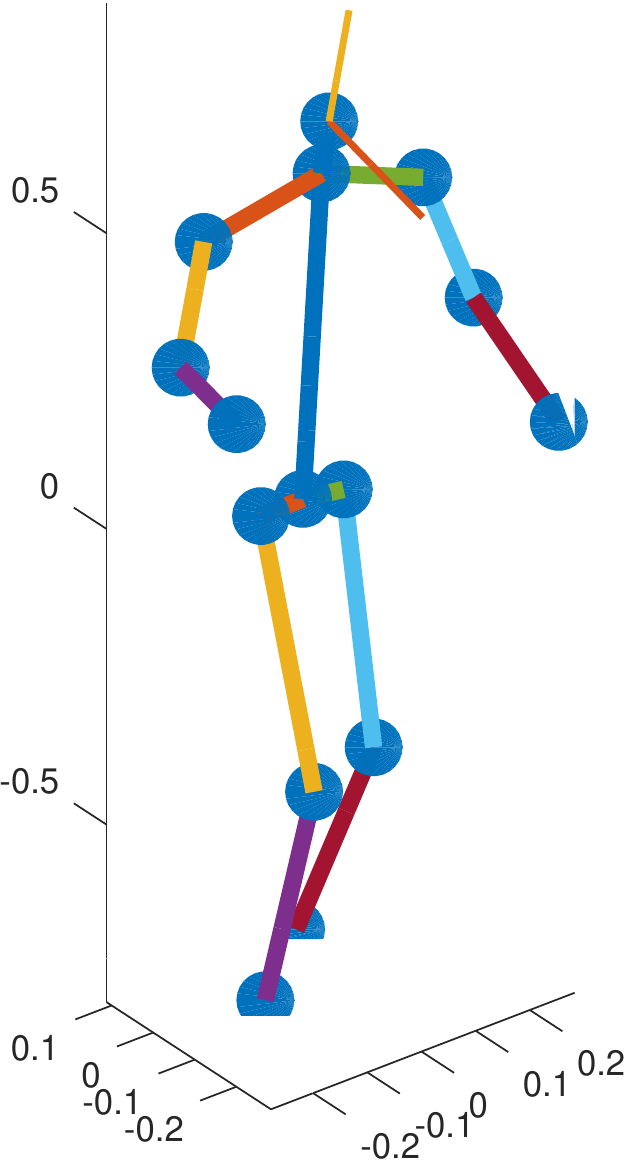}%
\\
 \rotatebox{90}{\hspace{0pt}{\tiny }} &
\includegraphics[width=0.05\linewidth, height=0.05\linewidth]{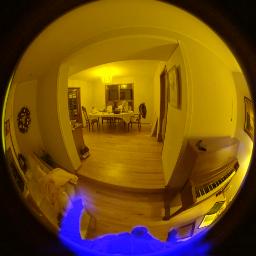}%
\includegraphics[width=0.05\linewidth, height=0.05\linewidth]{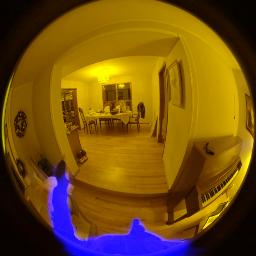}%
\includegraphics[width=0.05\linewidth, height=0.05\linewidth]{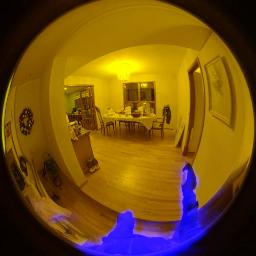}%
\includegraphics[width=0.05\linewidth, height=0.05\linewidth]{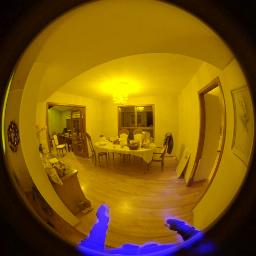}%
\includegraphics[width=0.05\linewidth, height=0.05\linewidth]{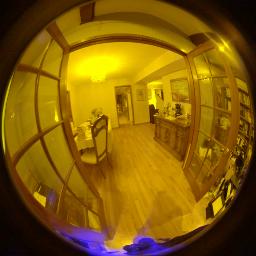}%
\includegraphics[width=0.05\linewidth, height=0.05\linewidth]{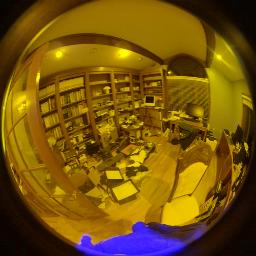}%
\includegraphics[width=0.05\linewidth, height=0.05\linewidth]{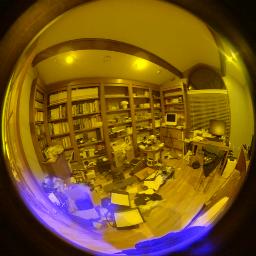}%
\includegraphics[width=0.05\linewidth, height=0.05\linewidth]{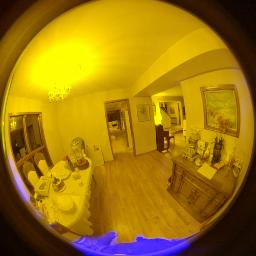}%
\includegraphics[width=0.05\linewidth, height=0.05\linewidth]{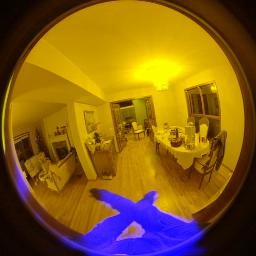}%
\includegraphics[width=0.05\linewidth, height=0.05\linewidth]{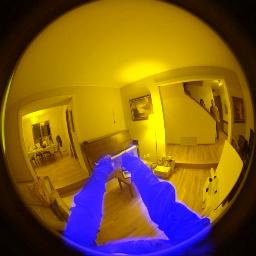}%
\includegraphics[width=0.05\linewidth, height=0.05\linewidth]{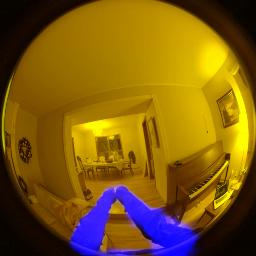}%
\includegraphics[width=0.05\linewidth, height=0.05\linewidth]{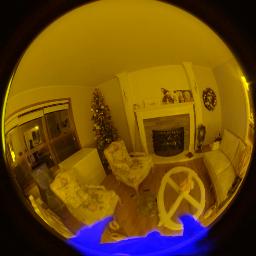}%
\includegraphics[width=0.05\linewidth, height=0.05\linewidth]{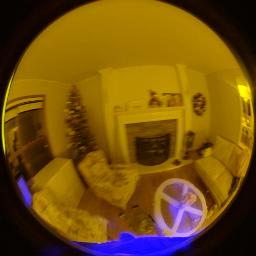}%
\includegraphics[width=0.05\linewidth, height=0.05\linewidth]{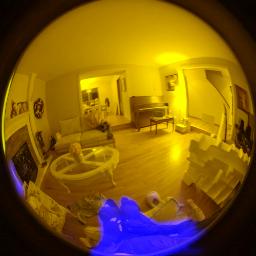}%
\includegraphics[width=0.05\linewidth, height=0.05\linewidth]{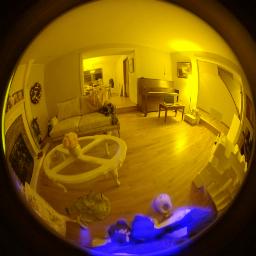}%
\includegraphics[width=0.05\linewidth, height=0.05\linewidth]{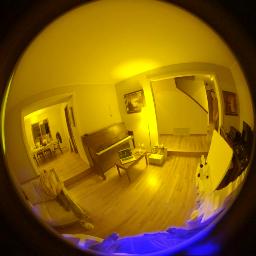}%
\includegraphics[width=0.05\linewidth, height=0.05\linewidth]{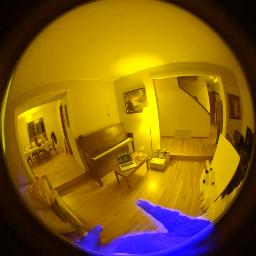}%
\includegraphics[width=0.05\linewidth, height=0.05\linewidth]{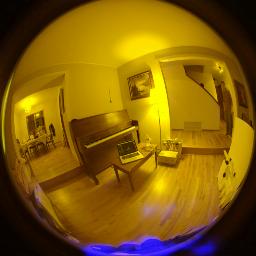}%
\includegraphics[width=0.05\linewidth, height=0.05\linewidth]{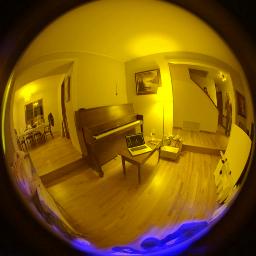}%
\includegraphics[width=0.05\linewidth, height=0.05\linewidth]{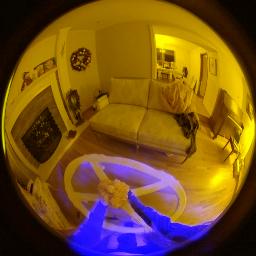}%
\\
 \rotatebox{90}{\hspace{4pt}{\tiny Ground Truth}} &
\includegraphics[width=0.05\linewidth, height=0.075\linewidth]{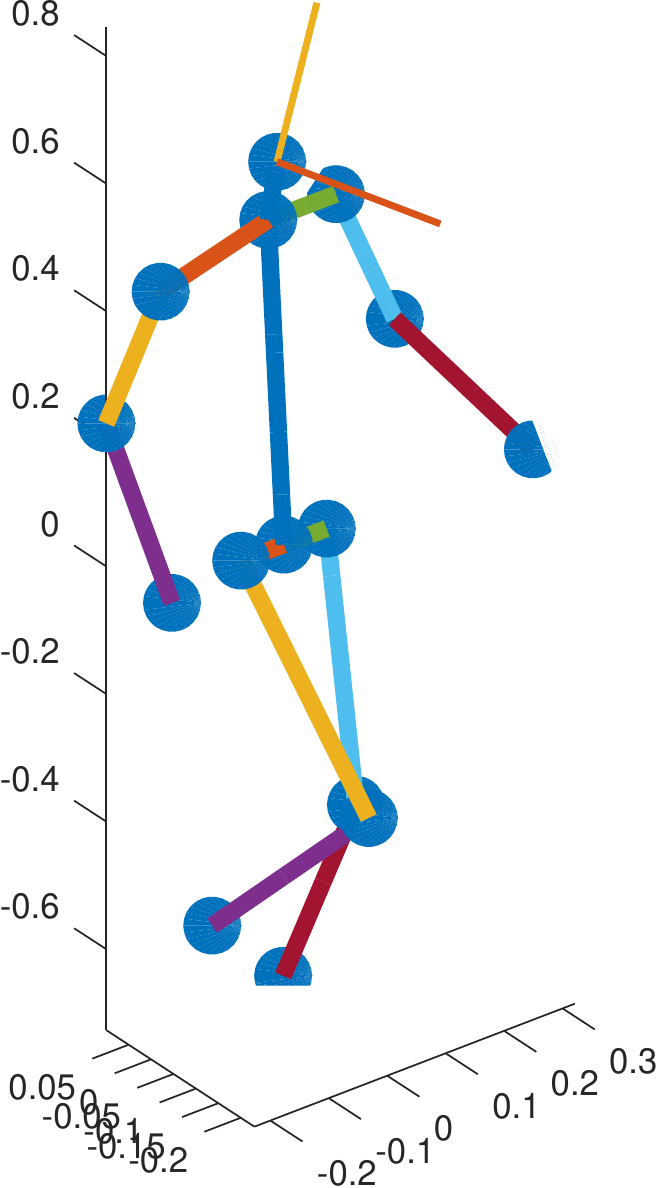}%
\includegraphics[width=0.05\linewidth, height=0.075\linewidth]{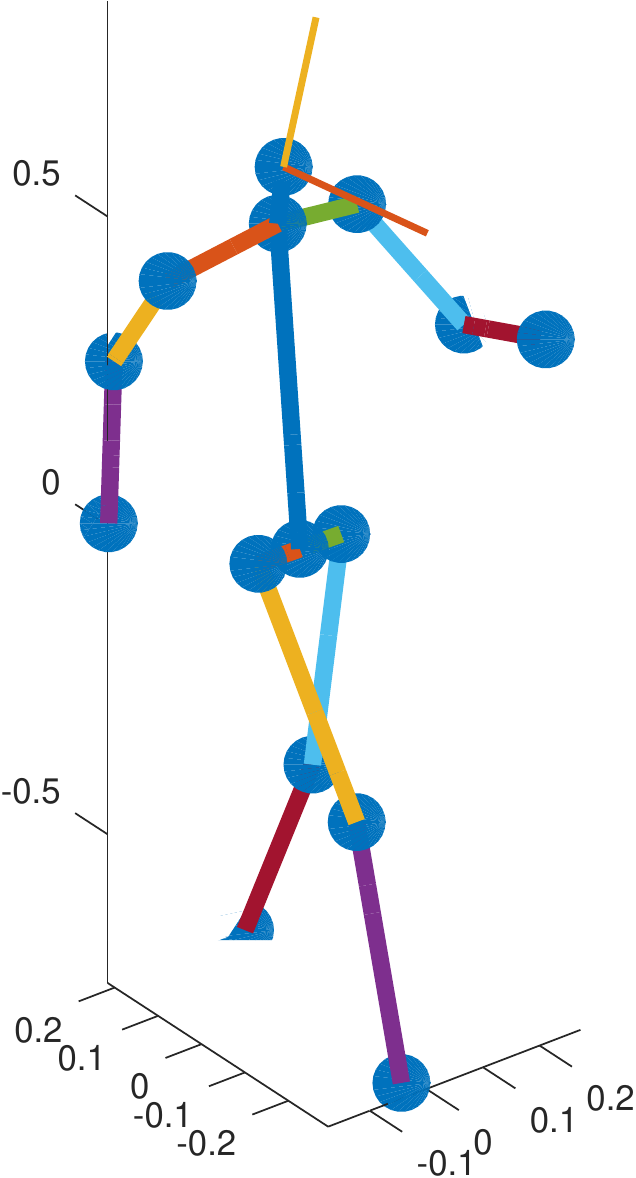}%
\includegraphics[width=0.05\linewidth, height=0.075\linewidth]{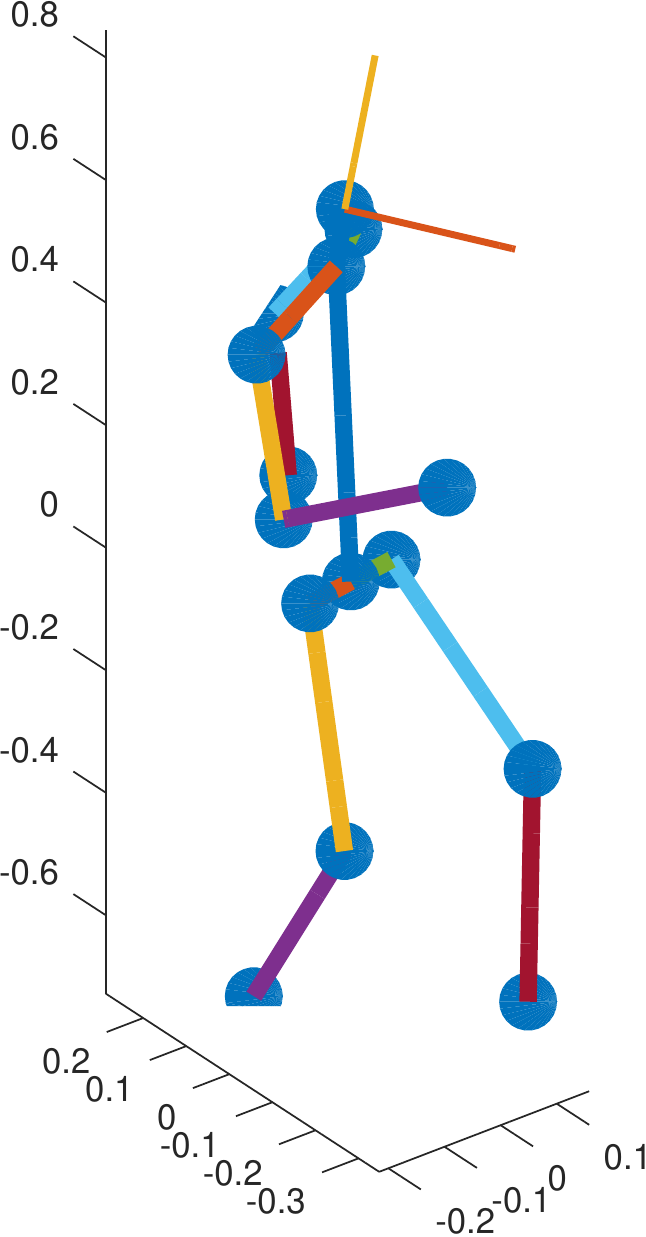}%
\includegraphics[width=0.05\linewidth, height=0.075\linewidth]{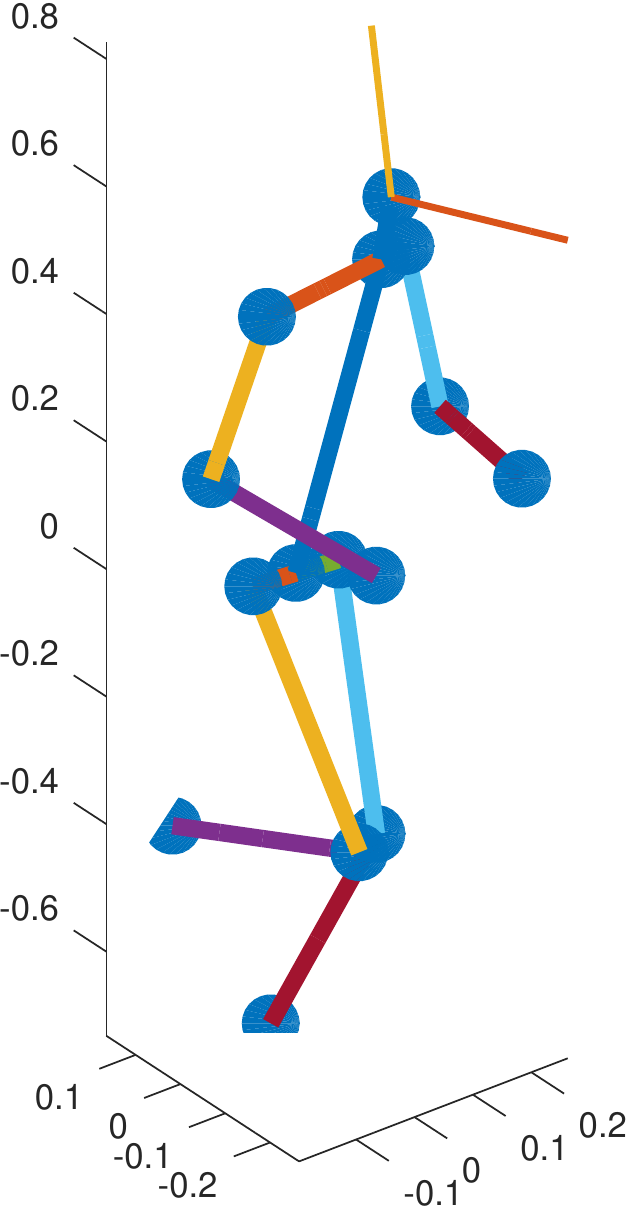}%
\includegraphics[width=0.05\linewidth, height=0.075\linewidth]{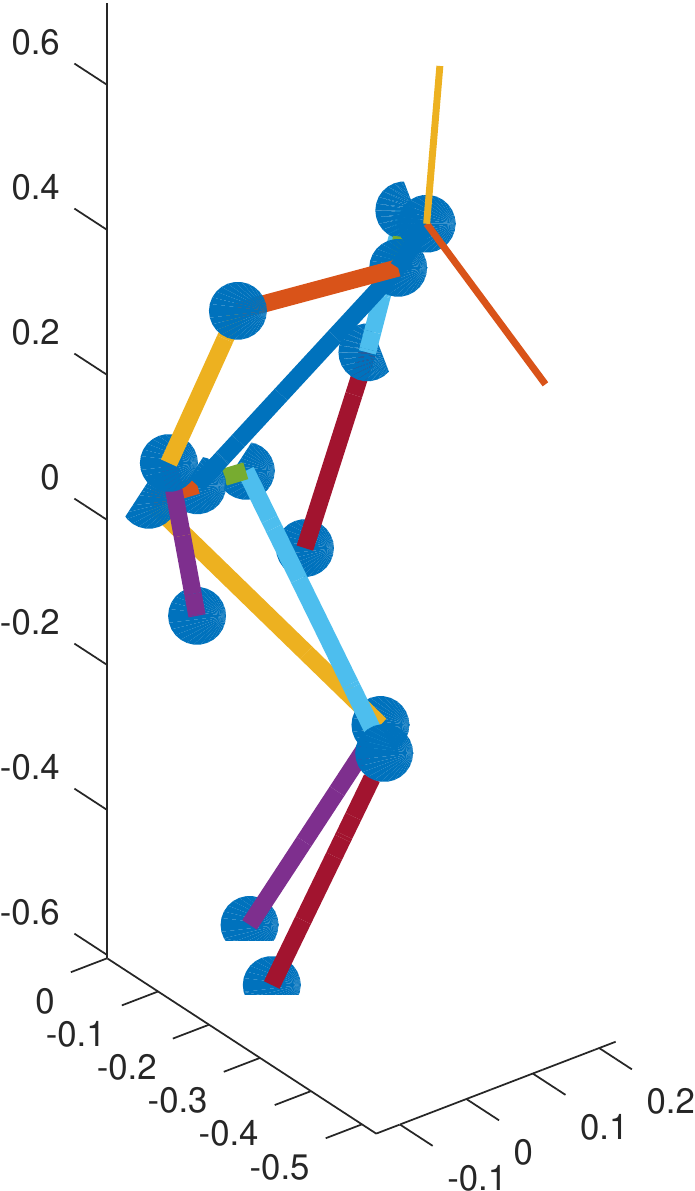}%
\includegraphics[width=0.05\linewidth, height=0.075\linewidth]{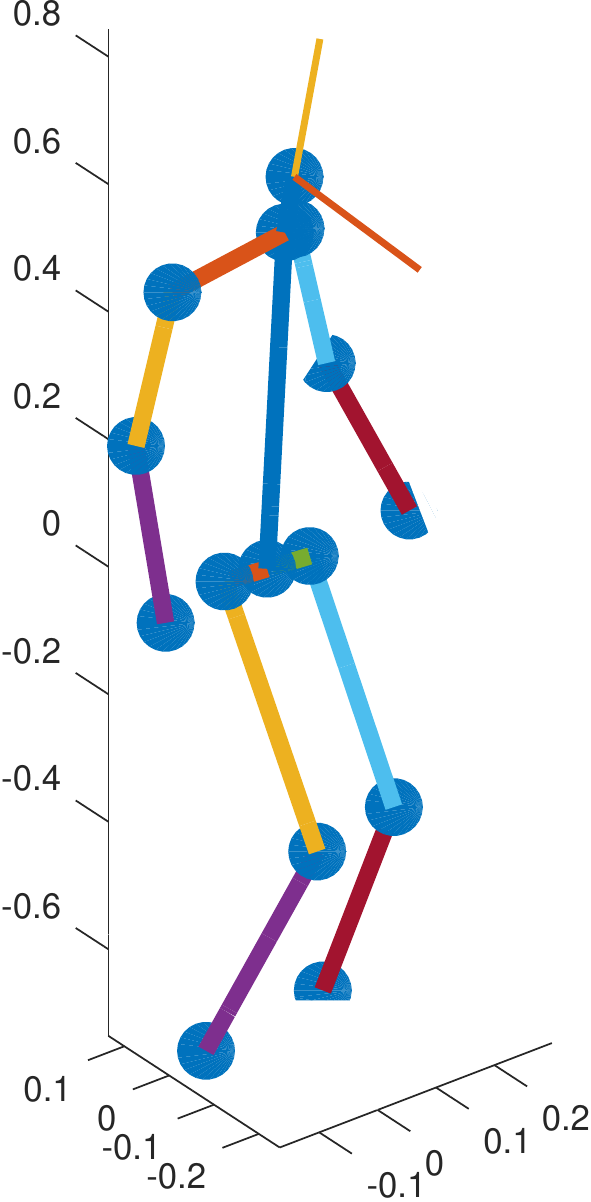}%
\includegraphics[width=0.05\linewidth, height=0.075\linewidth]{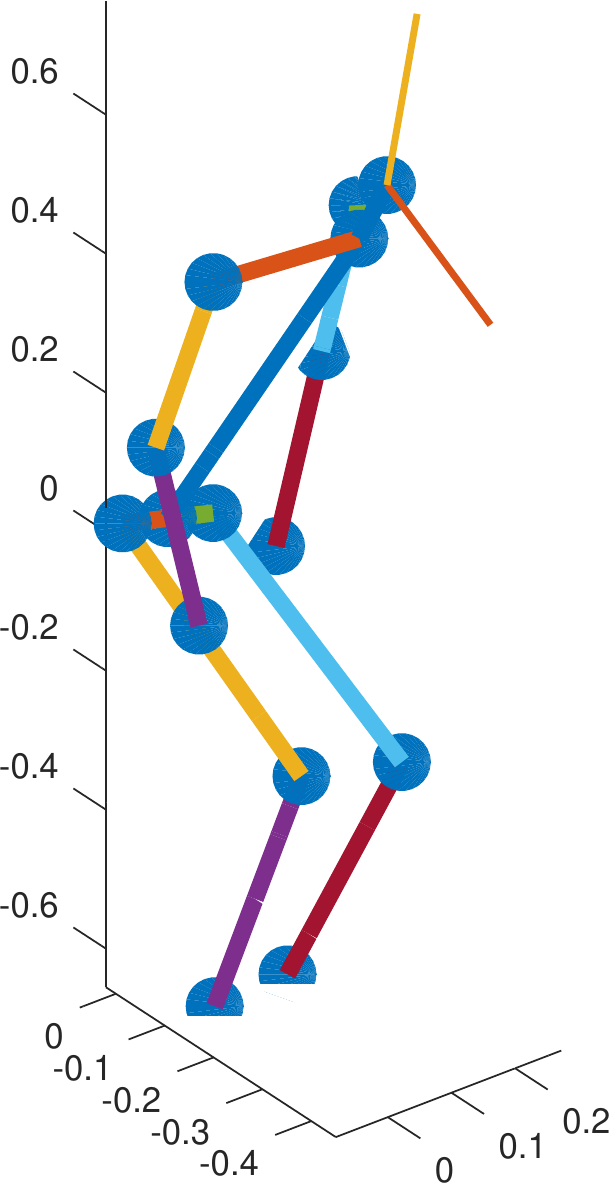}%
\includegraphics[width=0.05\linewidth, height=0.075\linewidth]{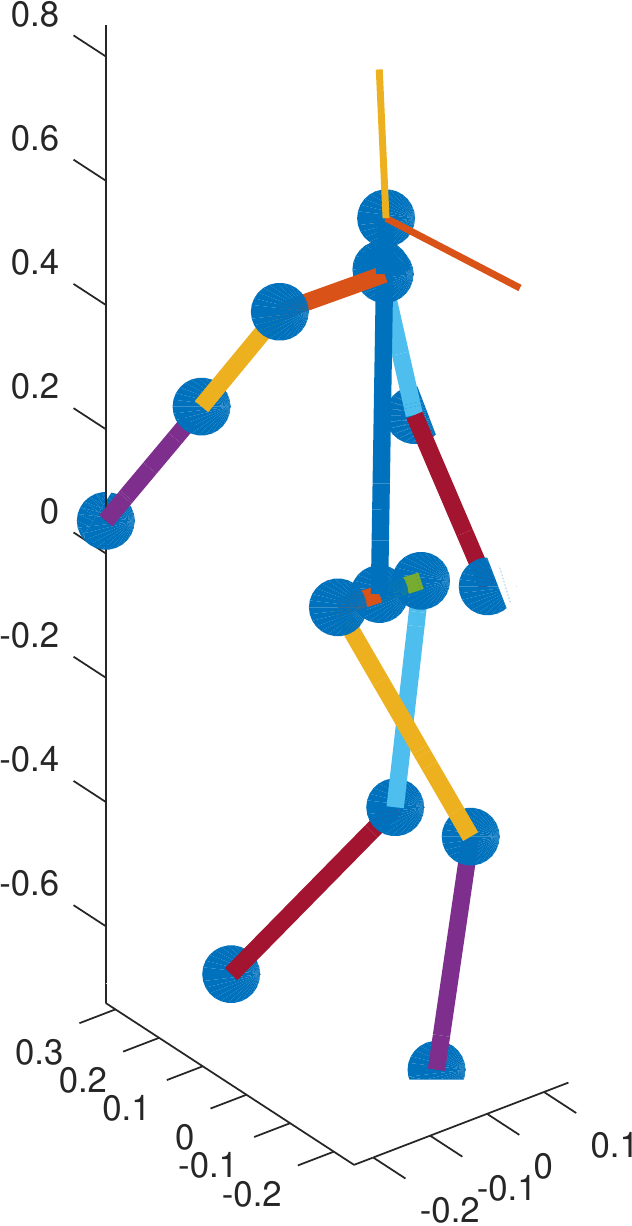}%
\includegraphics[width=0.05\linewidth, height=0.075\linewidth]{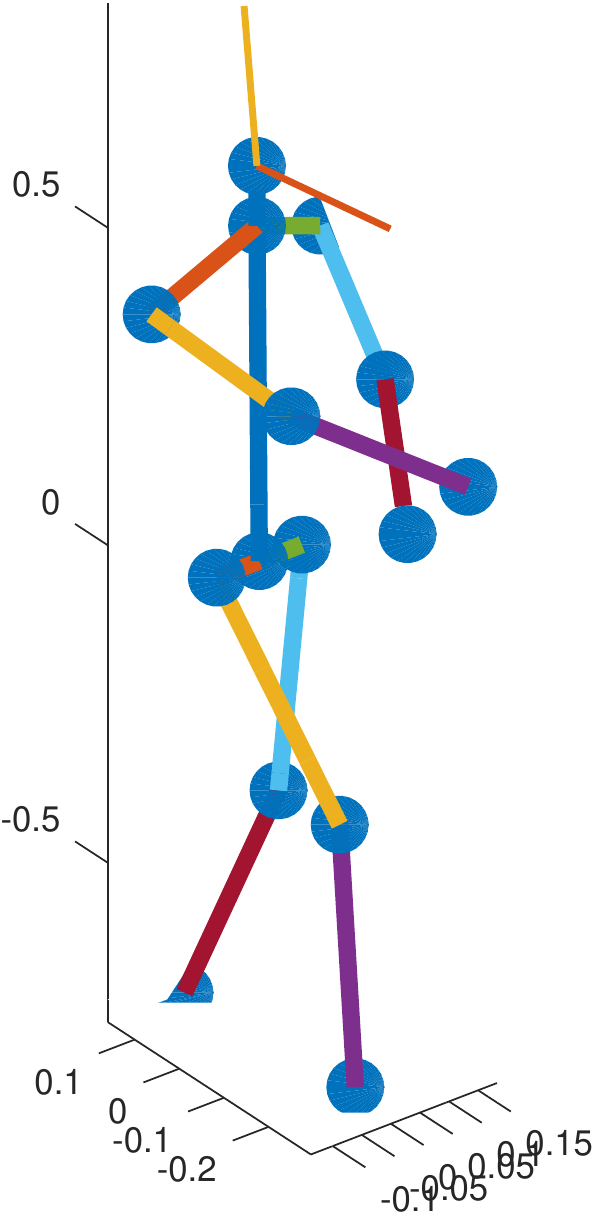}%
\includegraphics[width=0.05\linewidth, height=0.075\linewidth]{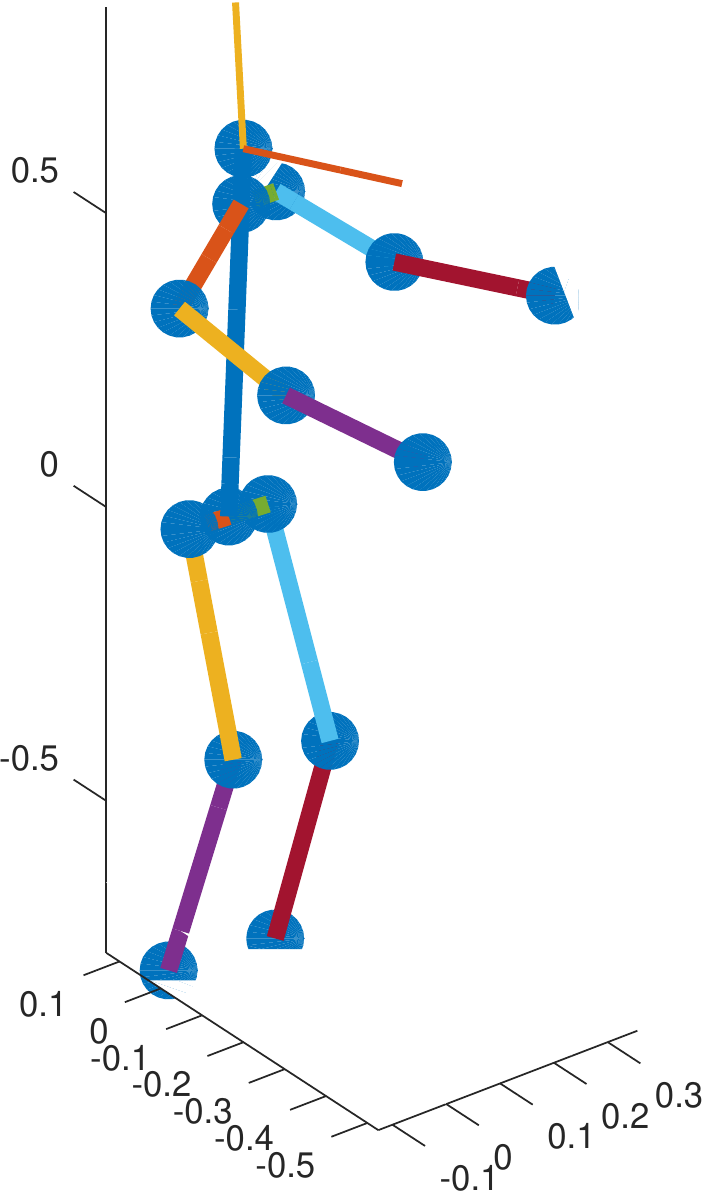}%
\includegraphics[width=0.05\linewidth, height=0.075\linewidth]{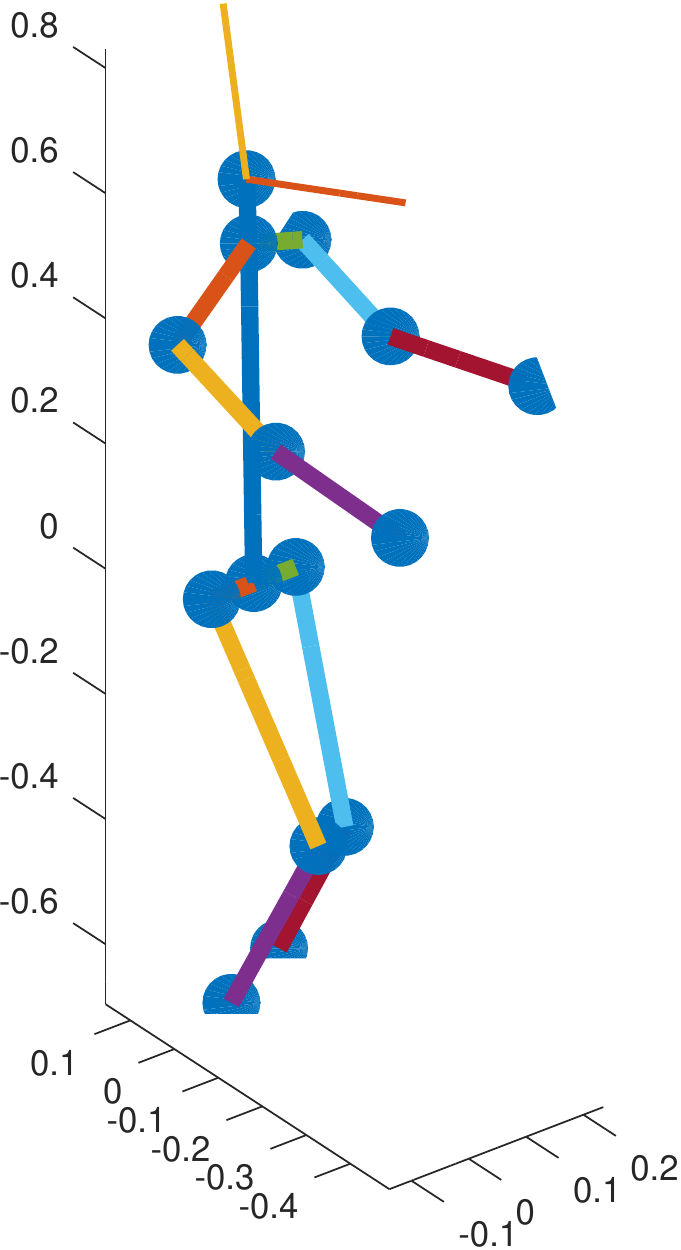}%
\includegraphics[width=0.05\linewidth, height=0.075\linewidth]{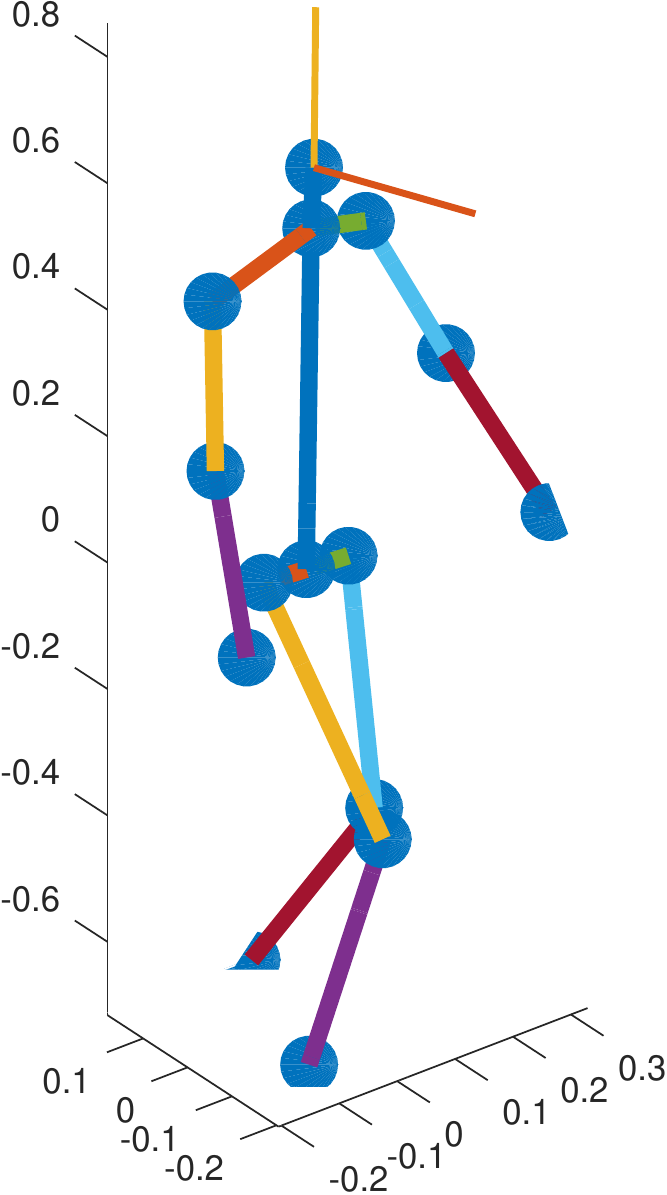}%
\includegraphics[width=0.05\linewidth, height=0.075\linewidth]{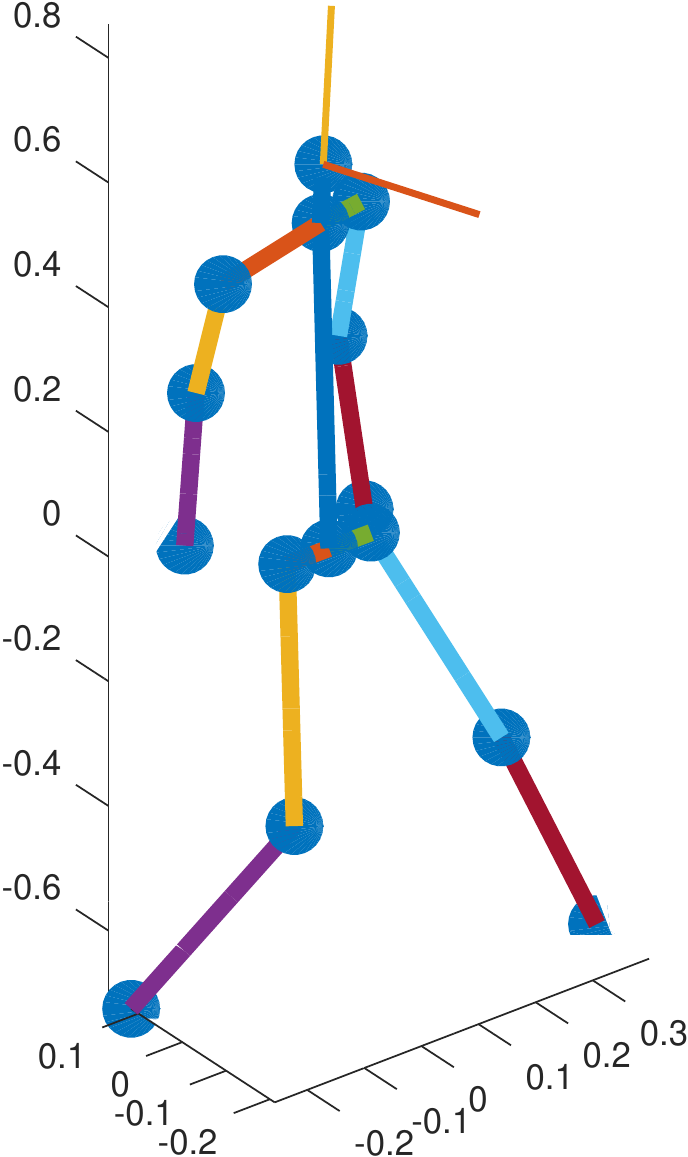}%
\includegraphics[width=0.05\linewidth, height=0.075\linewidth]{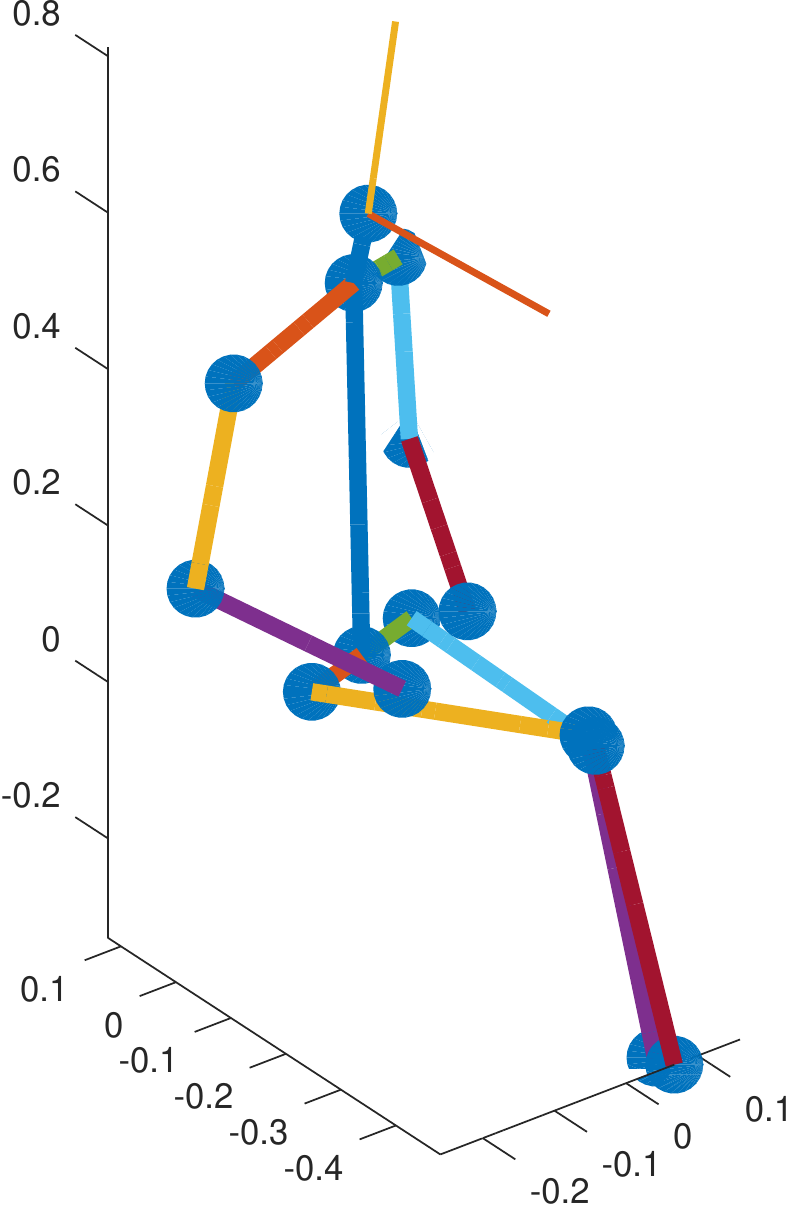}%
\includegraphics[width=0.05\linewidth, height=0.075\linewidth]{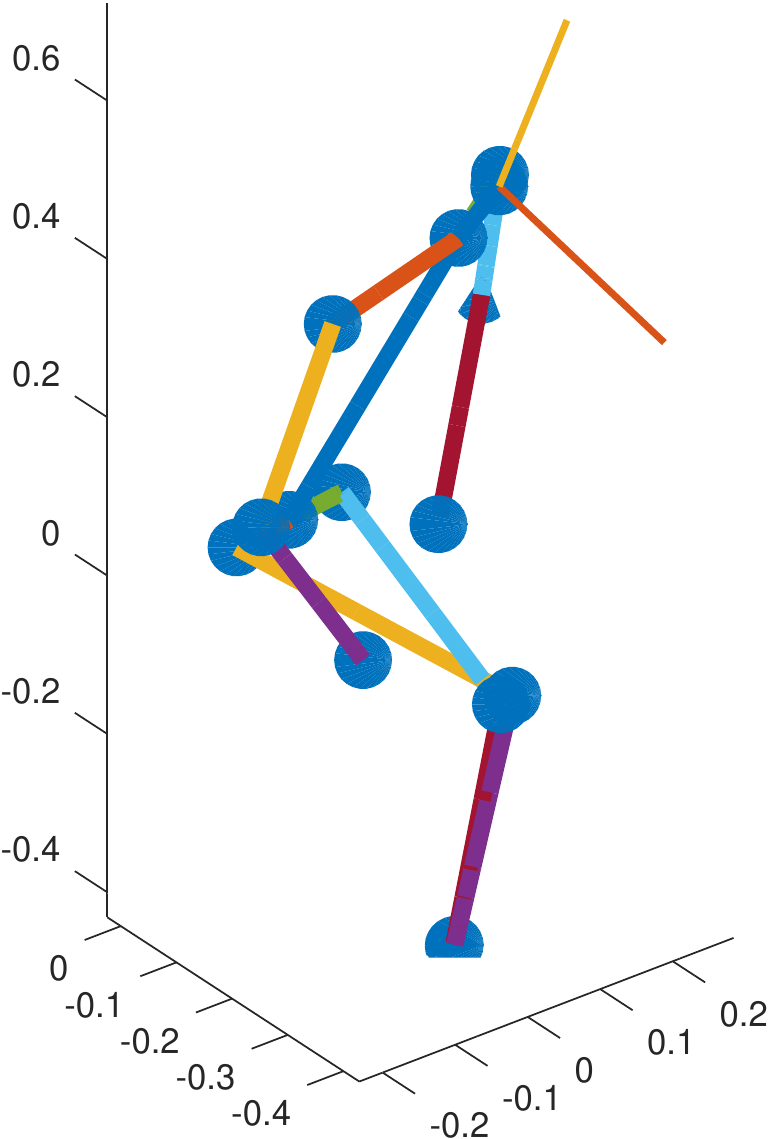}%
\includegraphics[width=0.05\linewidth, height=0.075\linewidth]{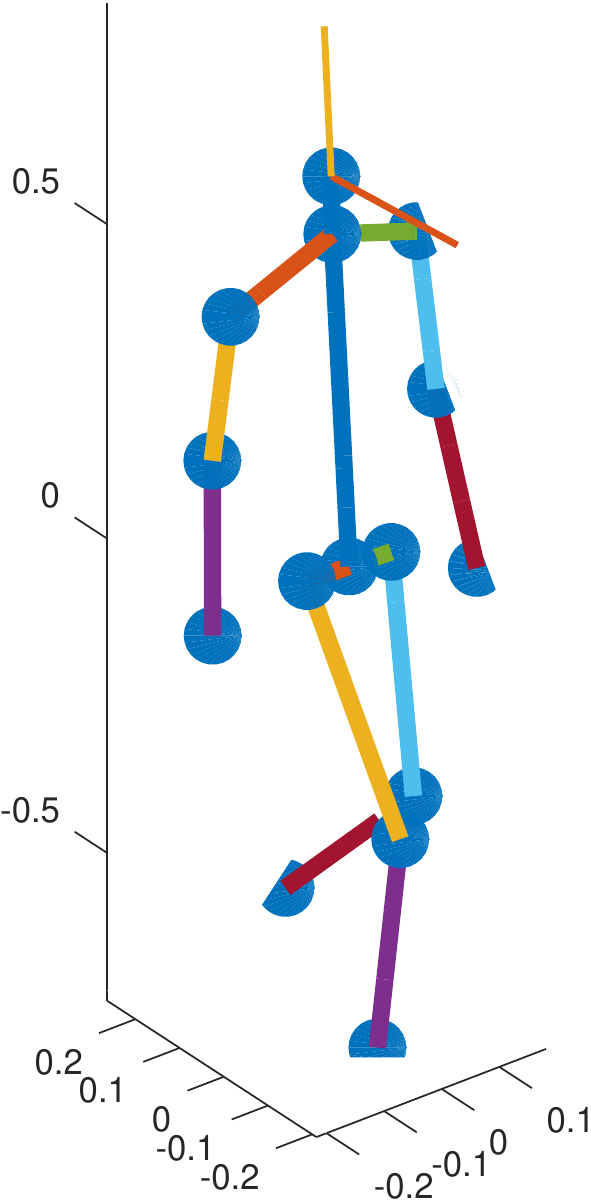}%
\includegraphics[width=0.05\linewidth, height=0.075\linewidth]{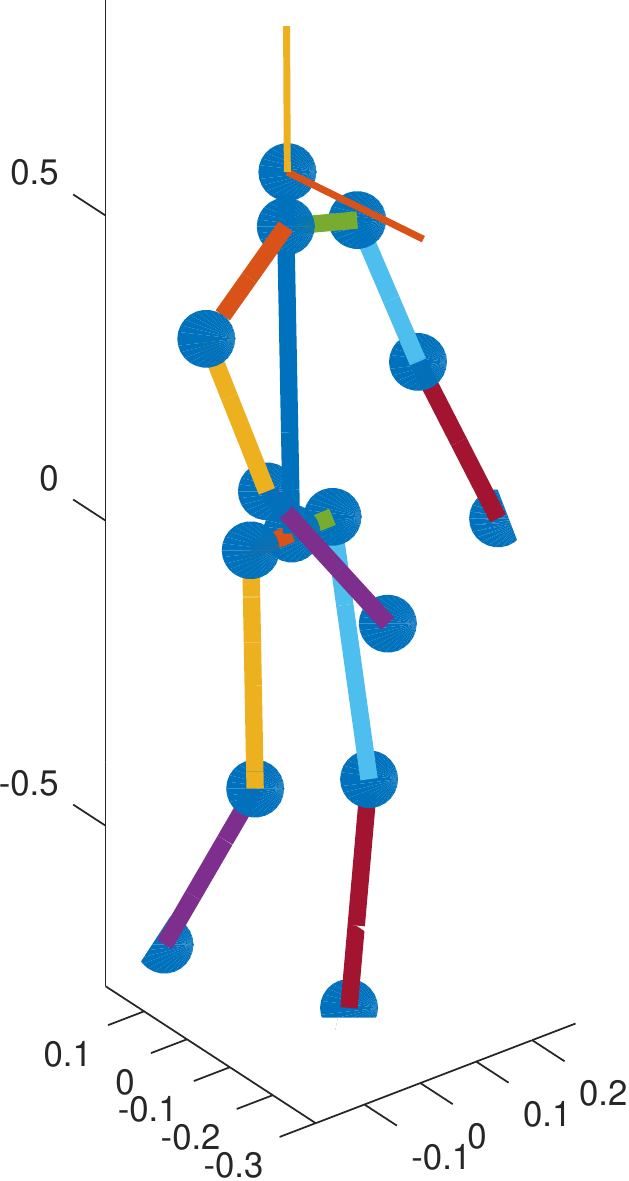}%
\includegraphics[width=0.05\linewidth, height=0.075\linewidth]{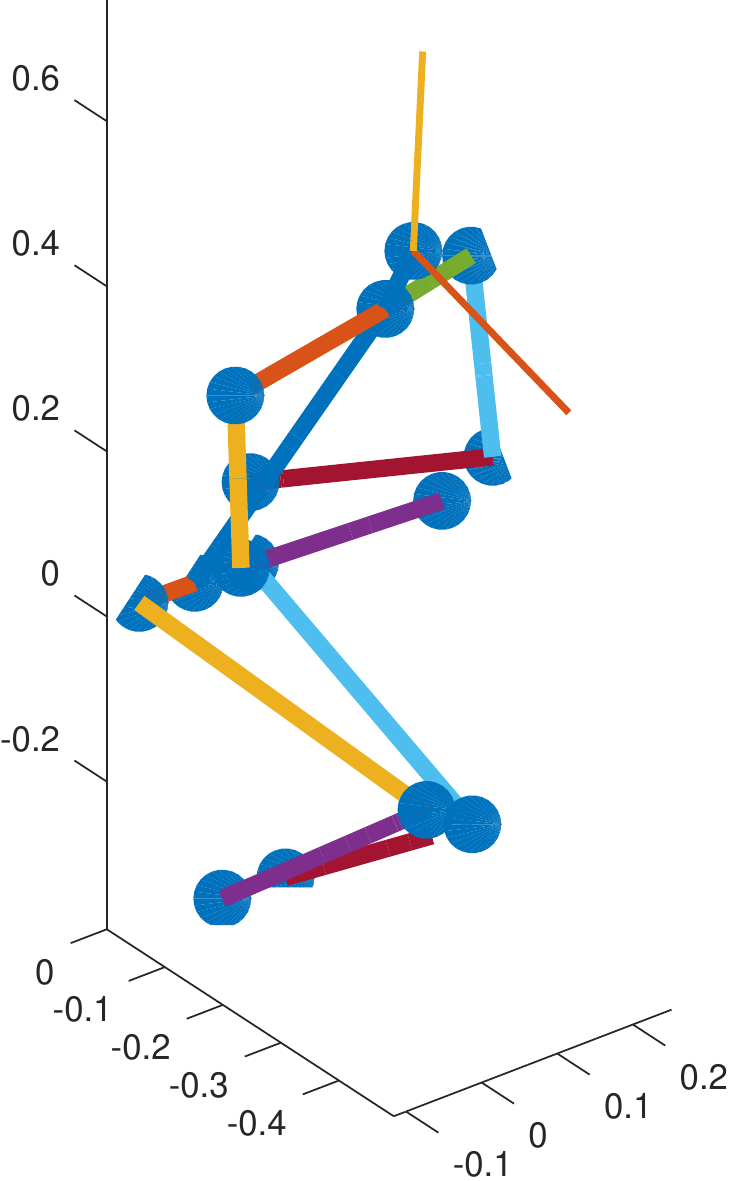}%
\includegraphics[width=0.05\linewidth, height=0.075\linewidth]{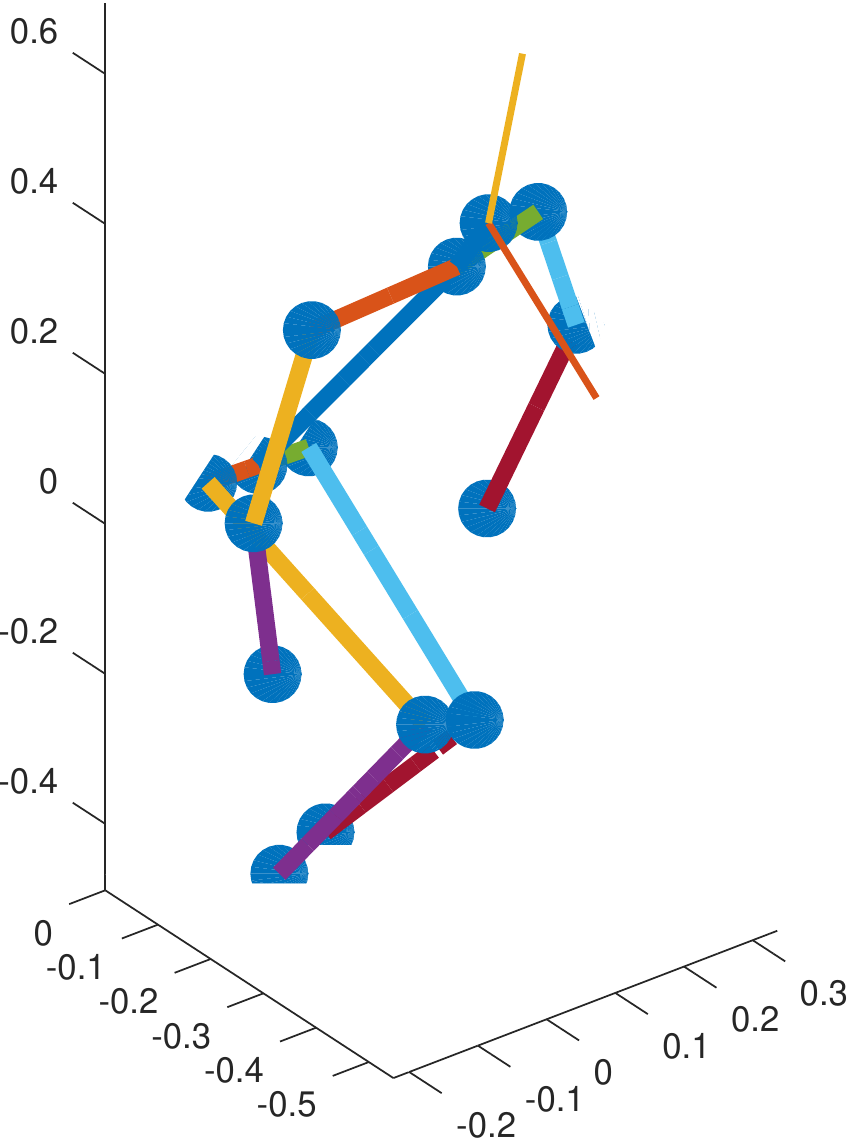}%
\includegraphics[width=0.05\linewidth, height=0.075\linewidth]{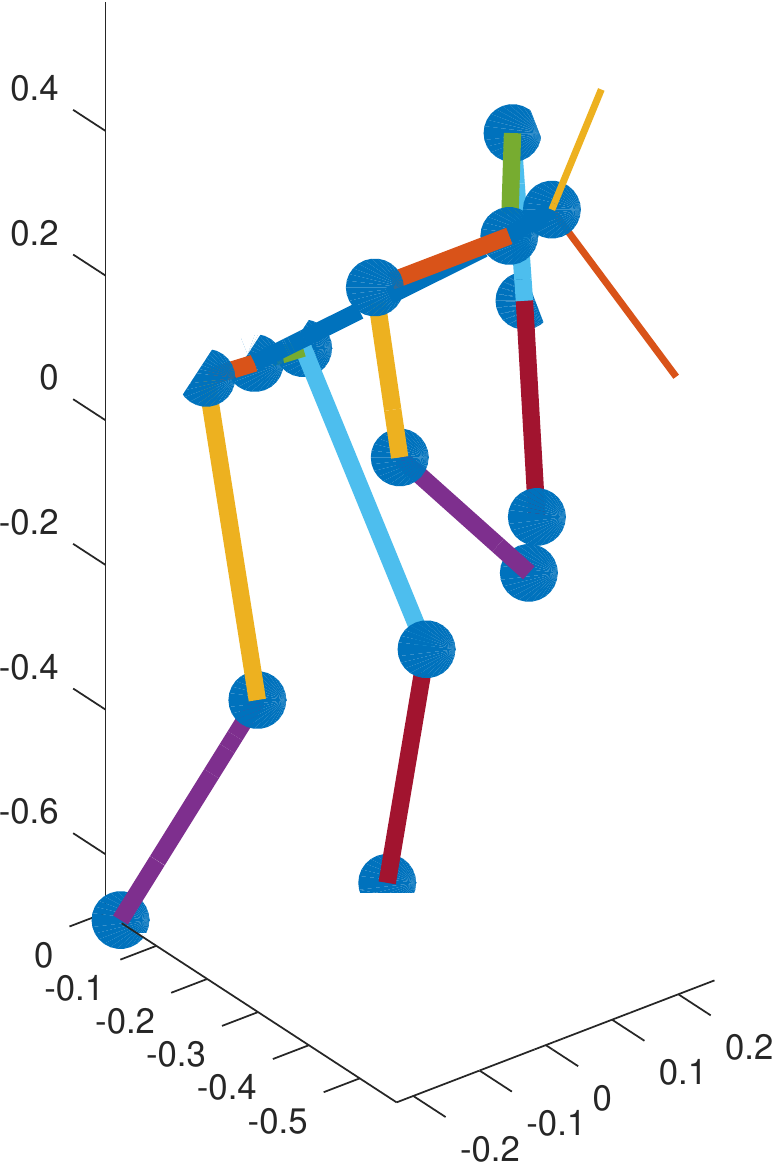}%
\\
 \rotatebox{90}{\hspace{15pt}{\tiny Ours}} &
\includegraphics[width=0.05\linewidth, height=0.075\linewidth]{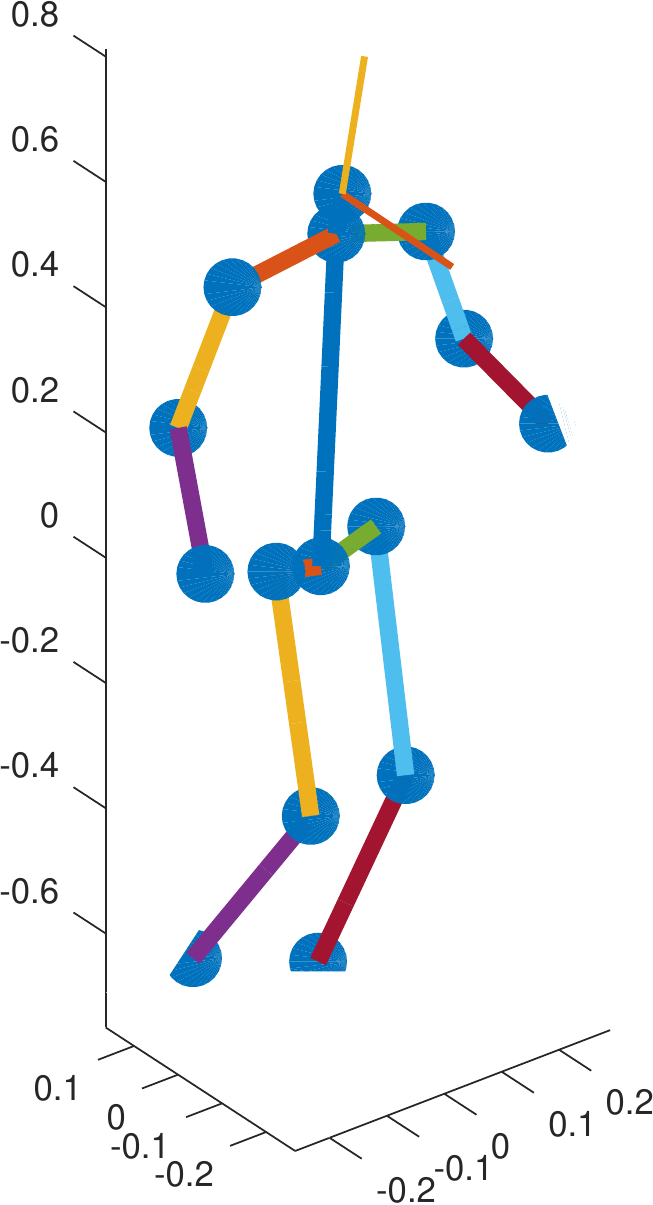}%
\includegraphics[width=0.05\linewidth, height=0.075\linewidth]{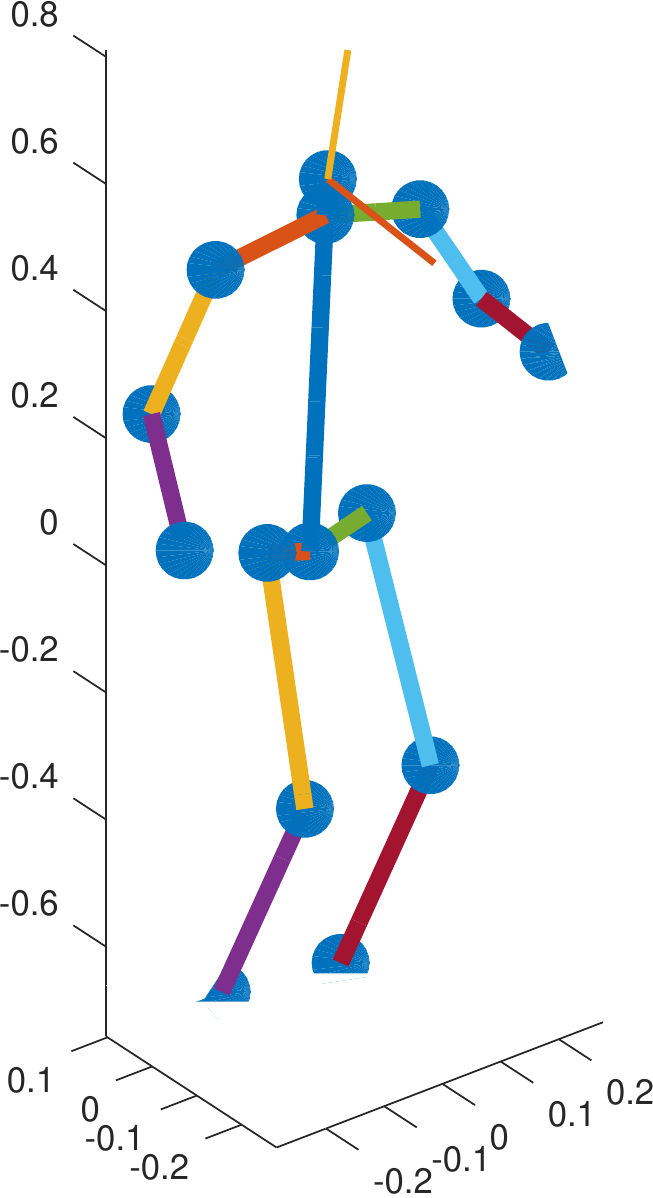}%
\includegraphics[width=0.05\linewidth, height=0.075\linewidth]{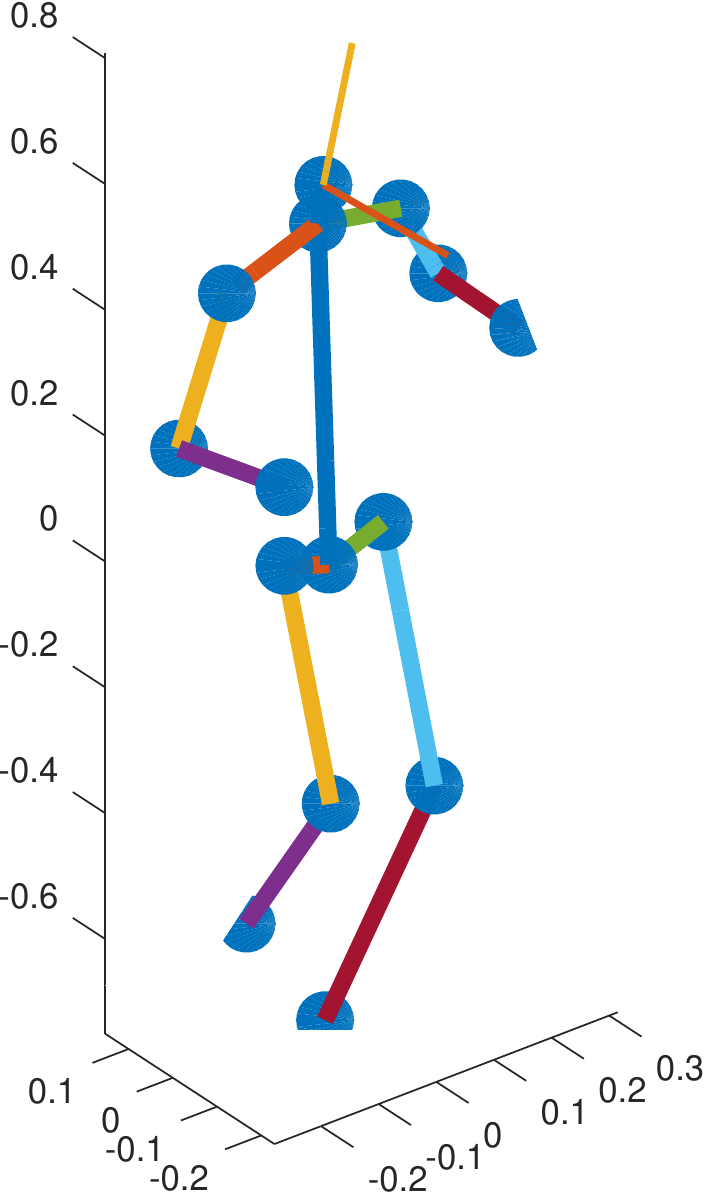}%
\includegraphics[width=0.05\linewidth, height=0.075\linewidth]{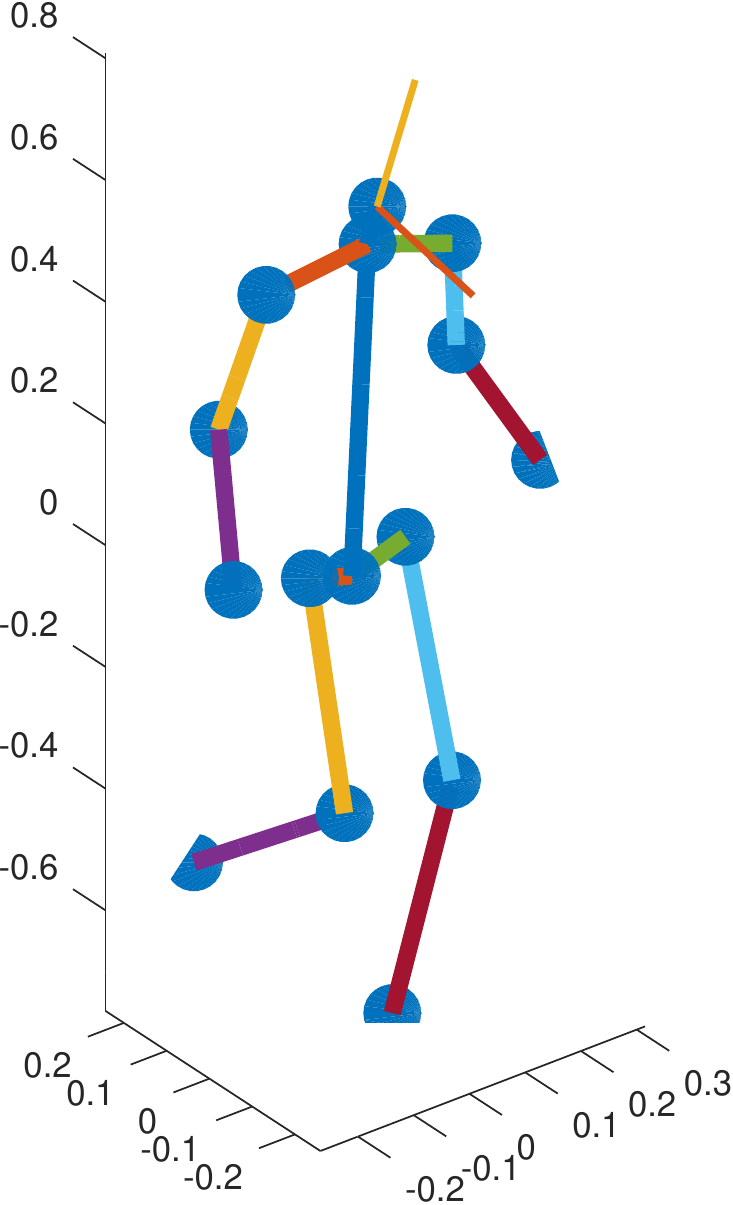}%
\includegraphics[width=0.05\linewidth, height=0.075\linewidth]{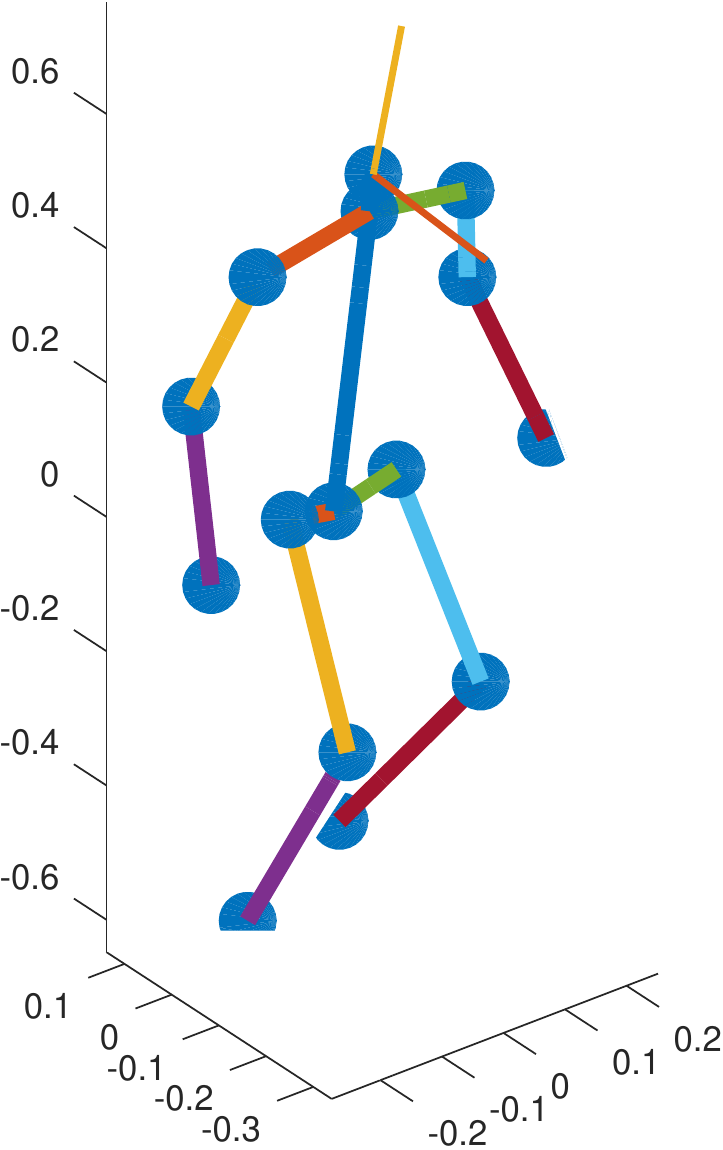}%
\includegraphics[width=0.05\linewidth, height=0.075\linewidth]{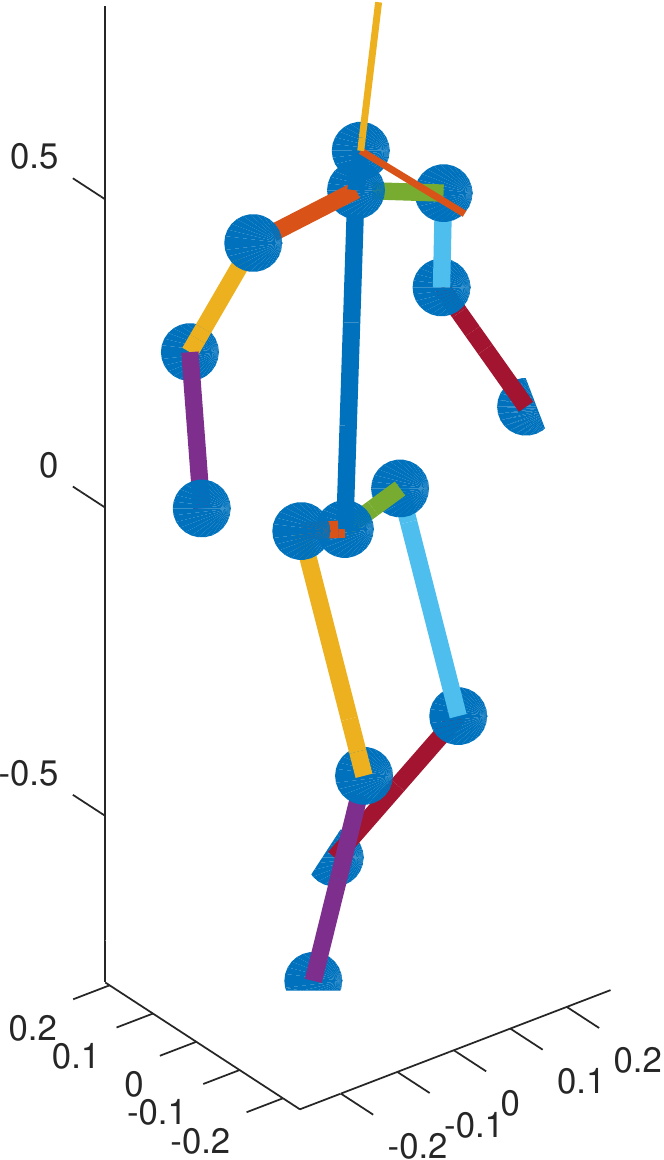}%
\includegraphics[width=0.05\linewidth, height=0.075\linewidth]{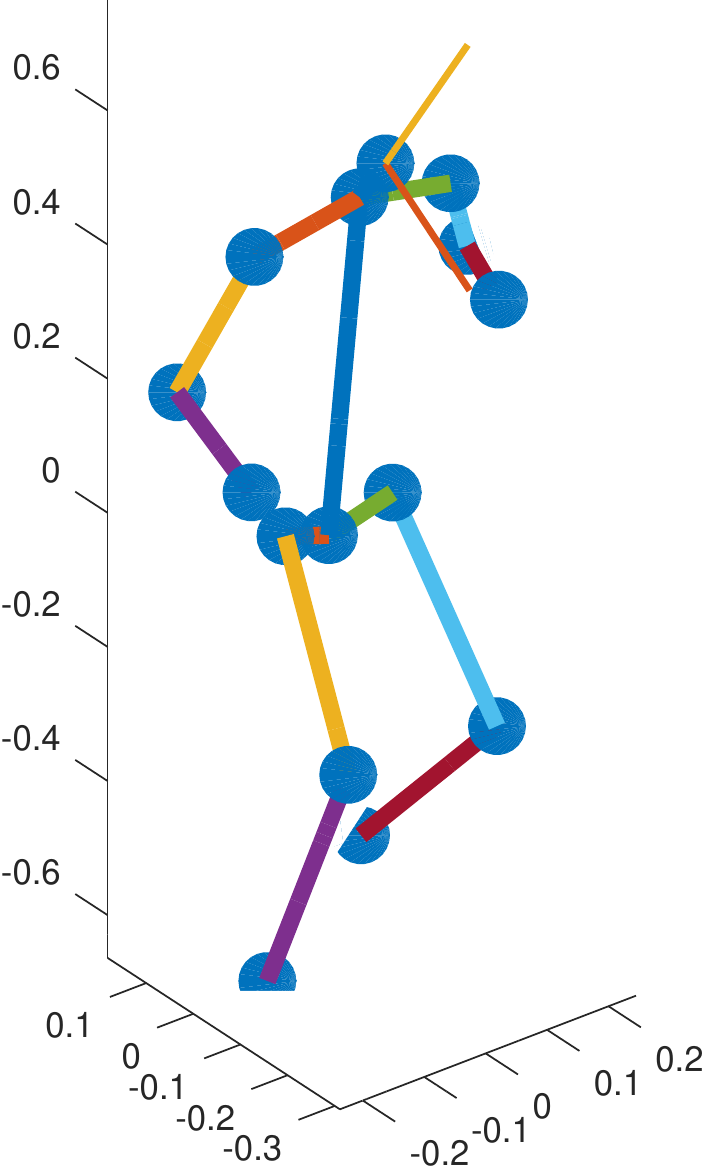}%
\includegraphics[width=0.05\linewidth, height=0.075\linewidth]{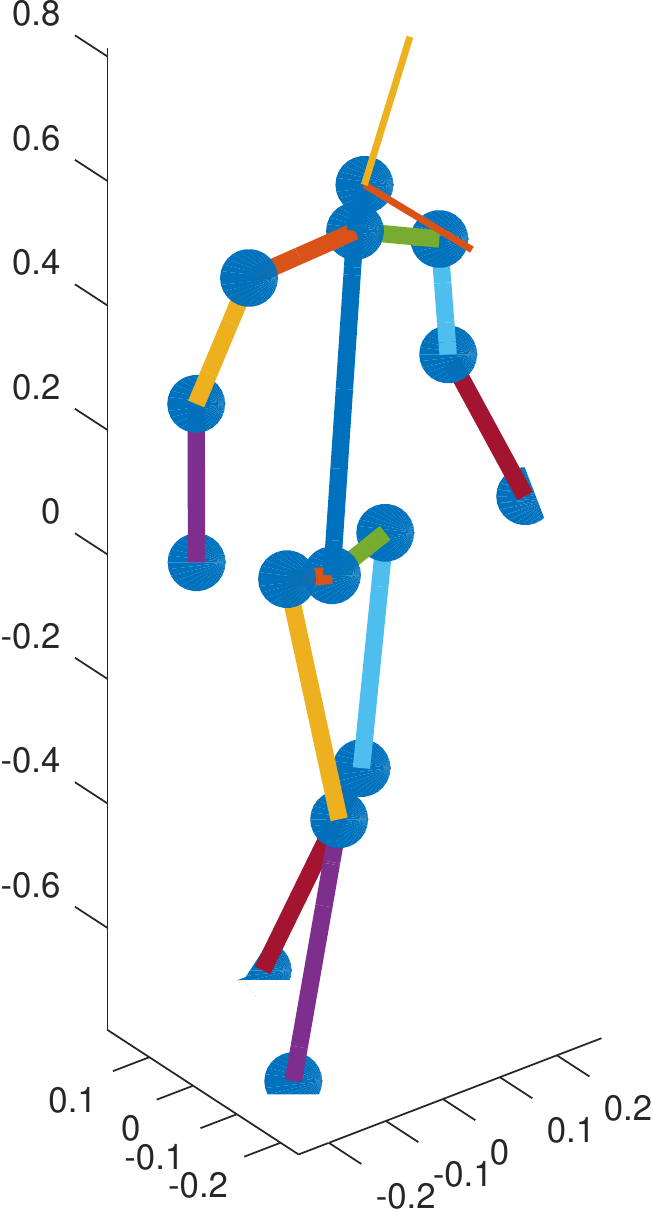}%
\includegraphics[width=0.05\linewidth, height=0.075\linewidth]{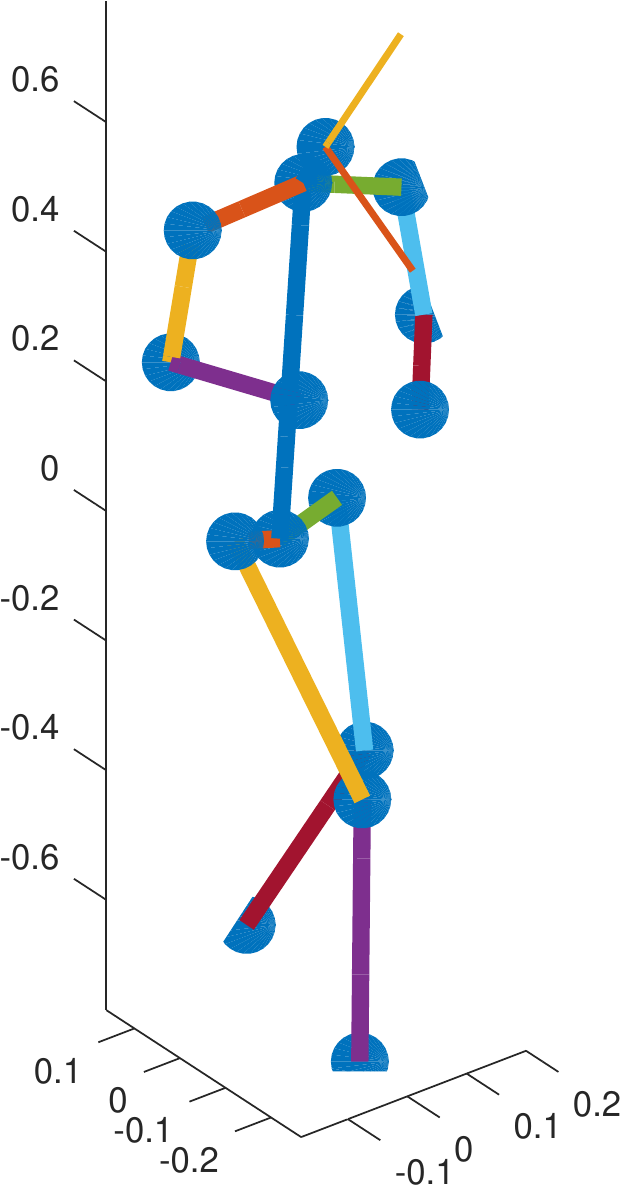}%
\includegraphics[width=0.05\linewidth, height=0.075\linewidth]{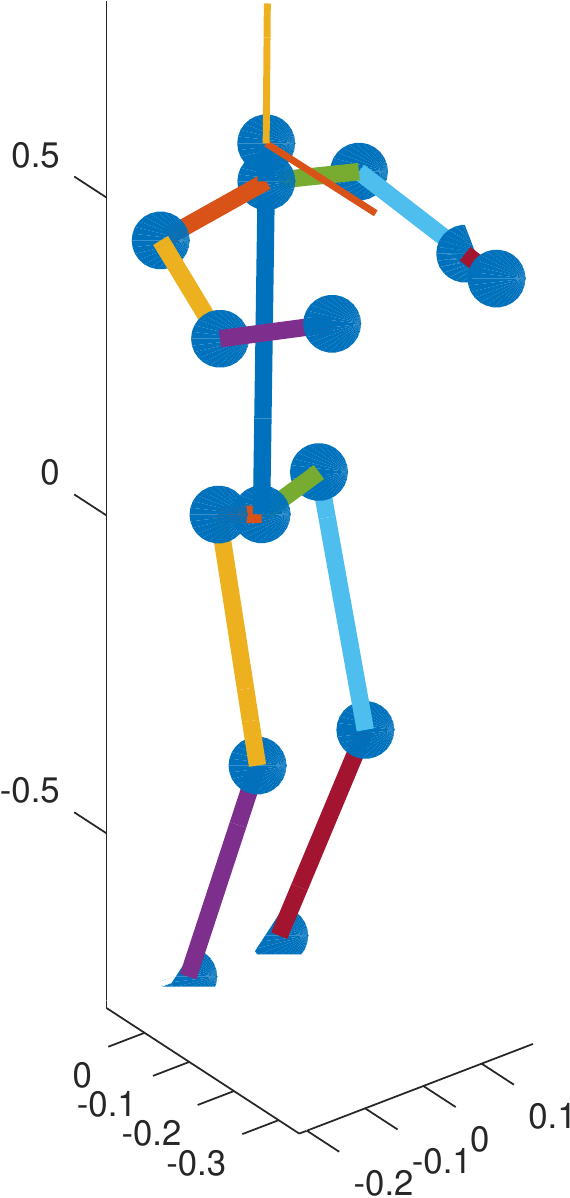}%
\includegraphics[width=0.05\linewidth, height=0.075\linewidth]{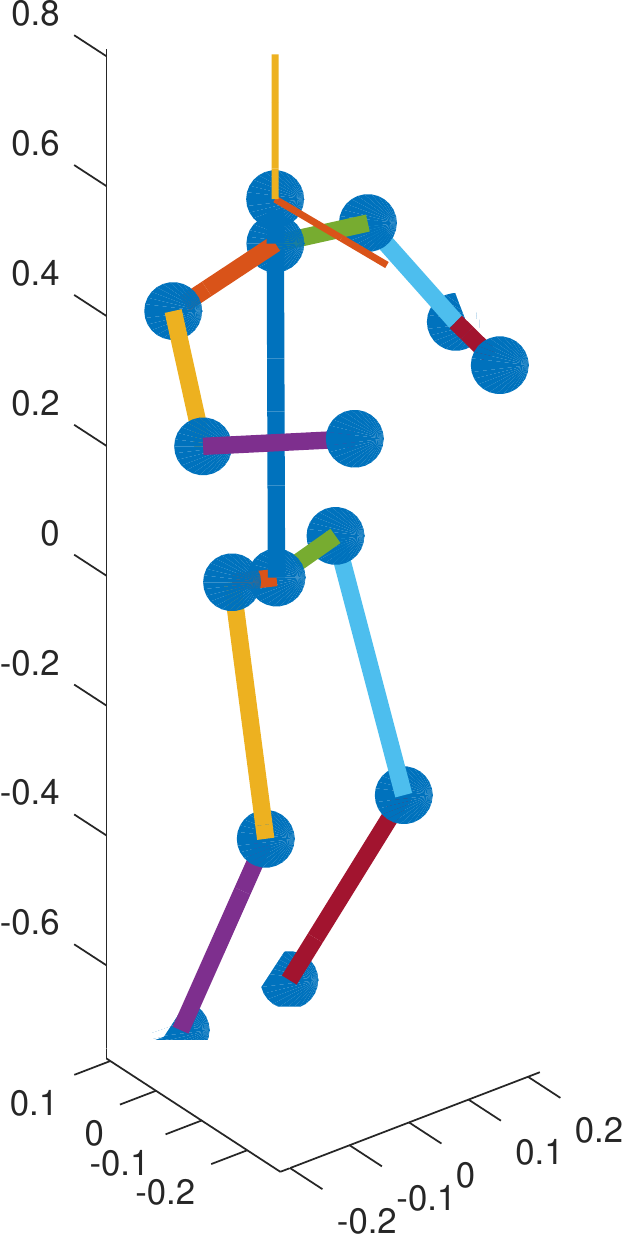}%
\includegraphics[width=0.05\linewidth, height=0.075\linewidth]{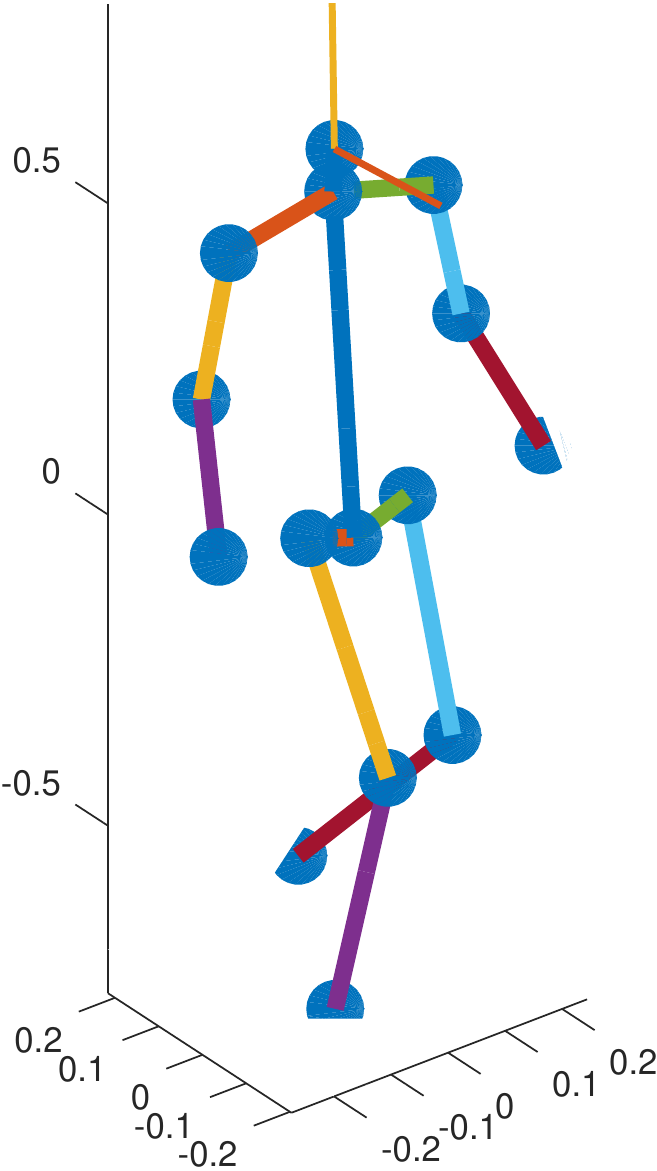}%
\includegraphics[width=0.05\linewidth, height=0.075\linewidth]{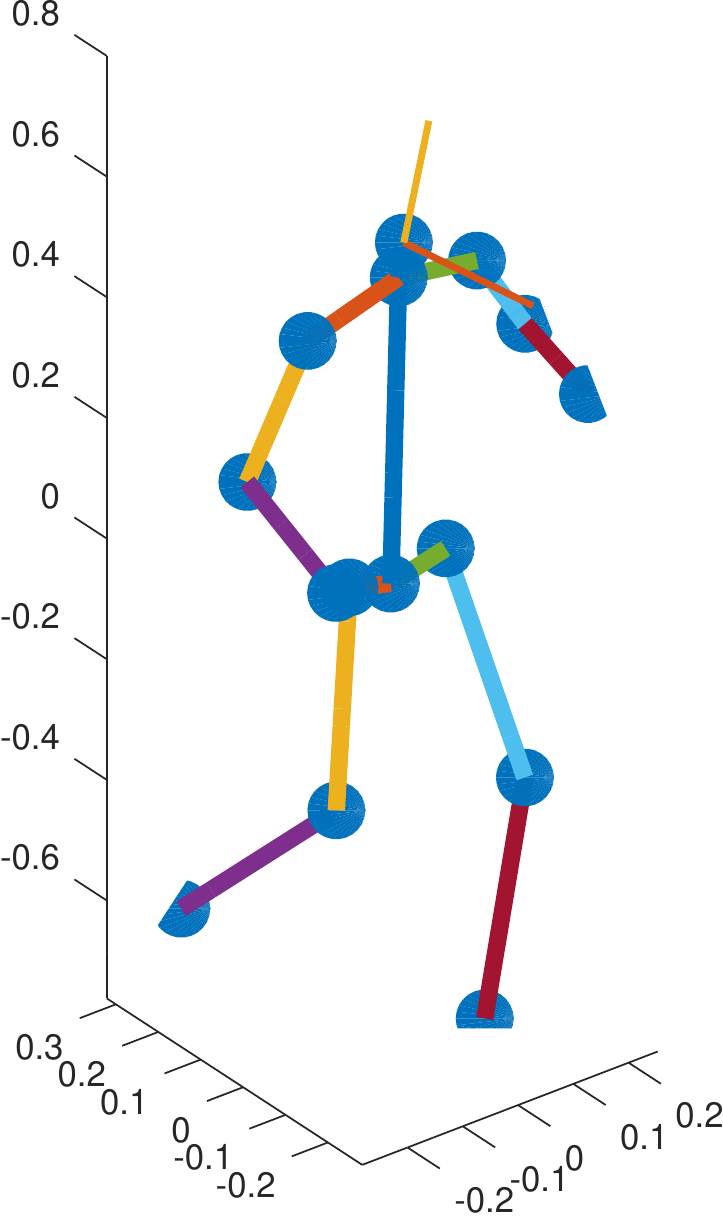}%
\includegraphics[width=0.05\linewidth, height=0.075\linewidth]{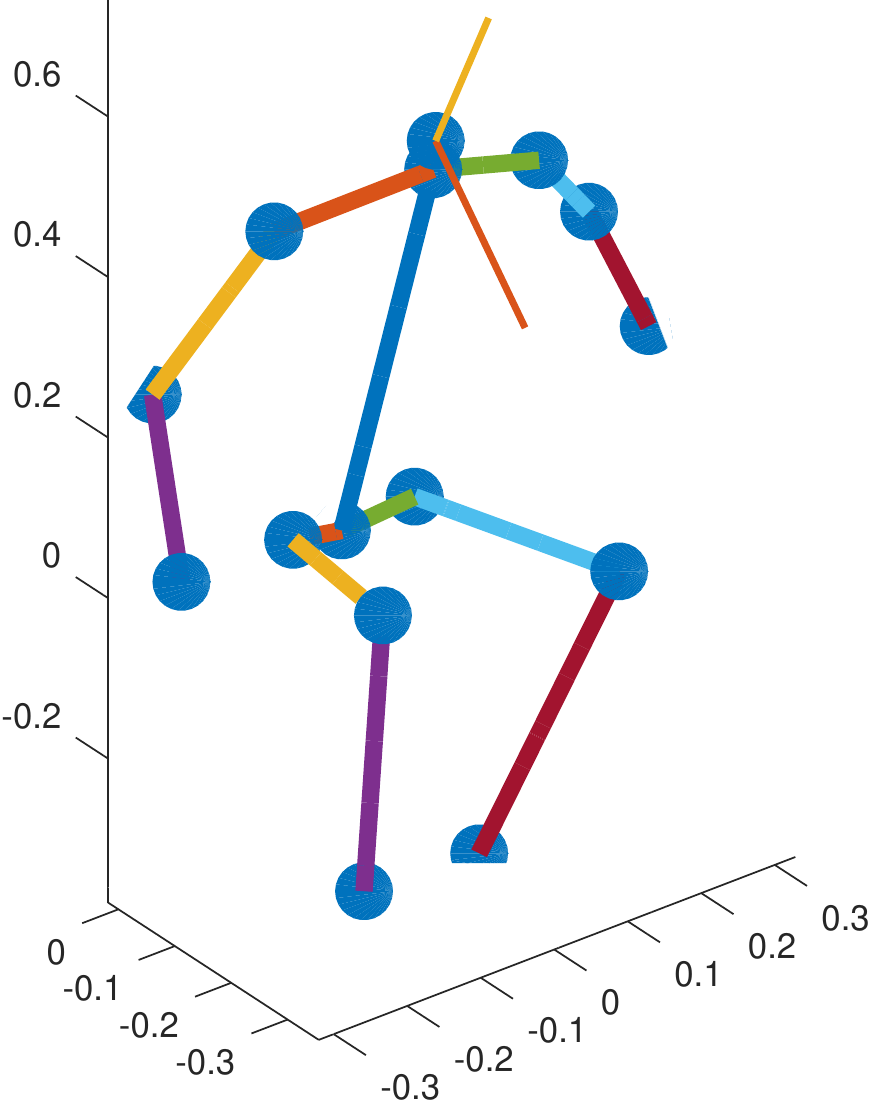}%
\includegraphics[width=0.05\linewidth, height=0.075\linewidth]{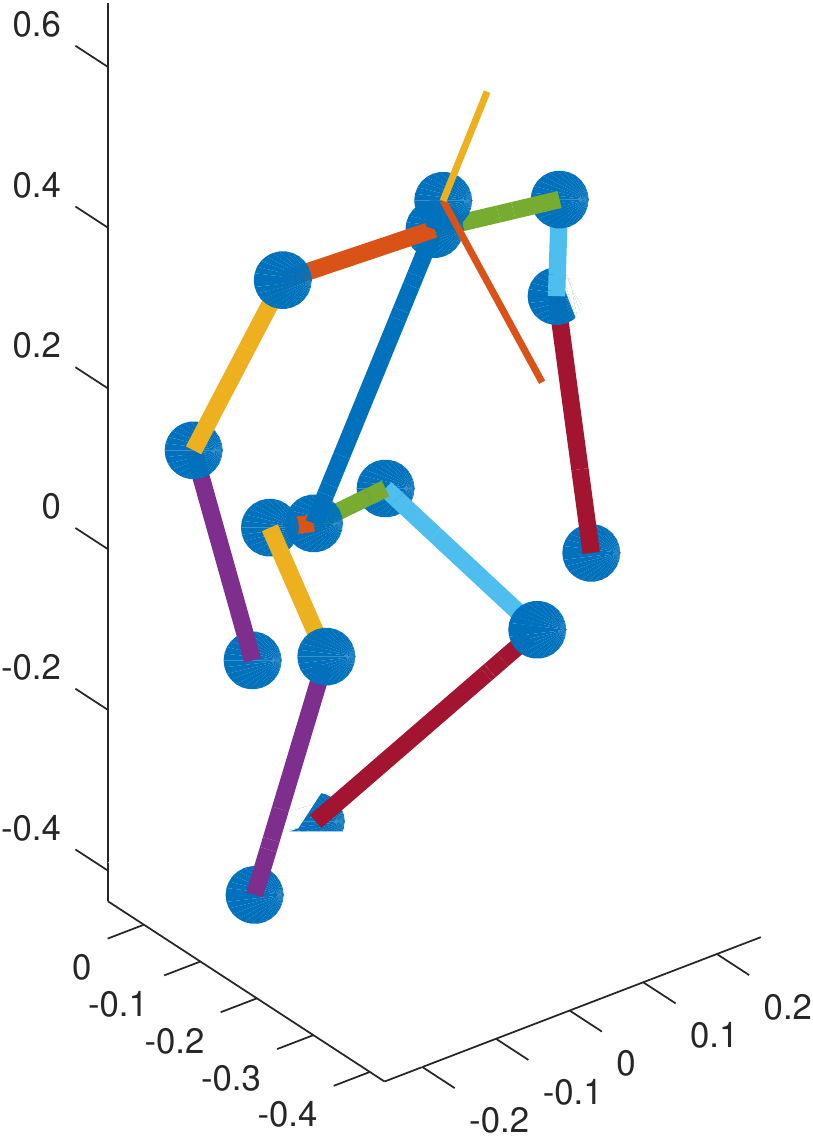}%
\includegraphics[width=0.05\linewidth, height=0.075\linewidth]{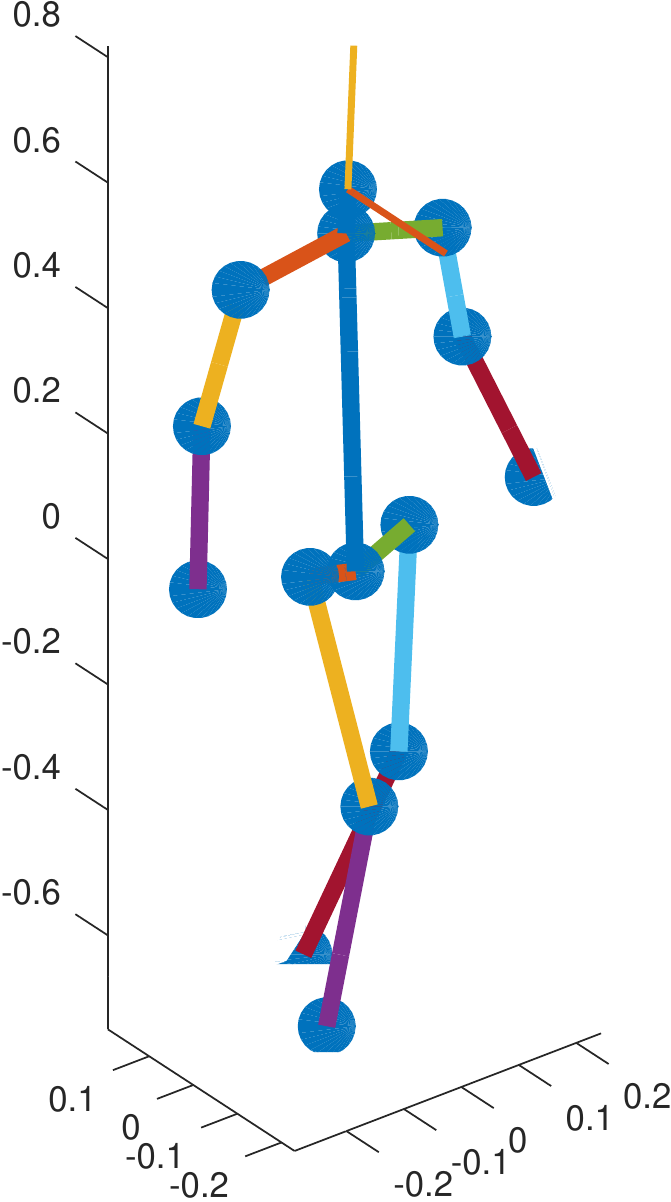}%
\includegraphics[width=0.05\linewidth, height=0.075\linewidth]{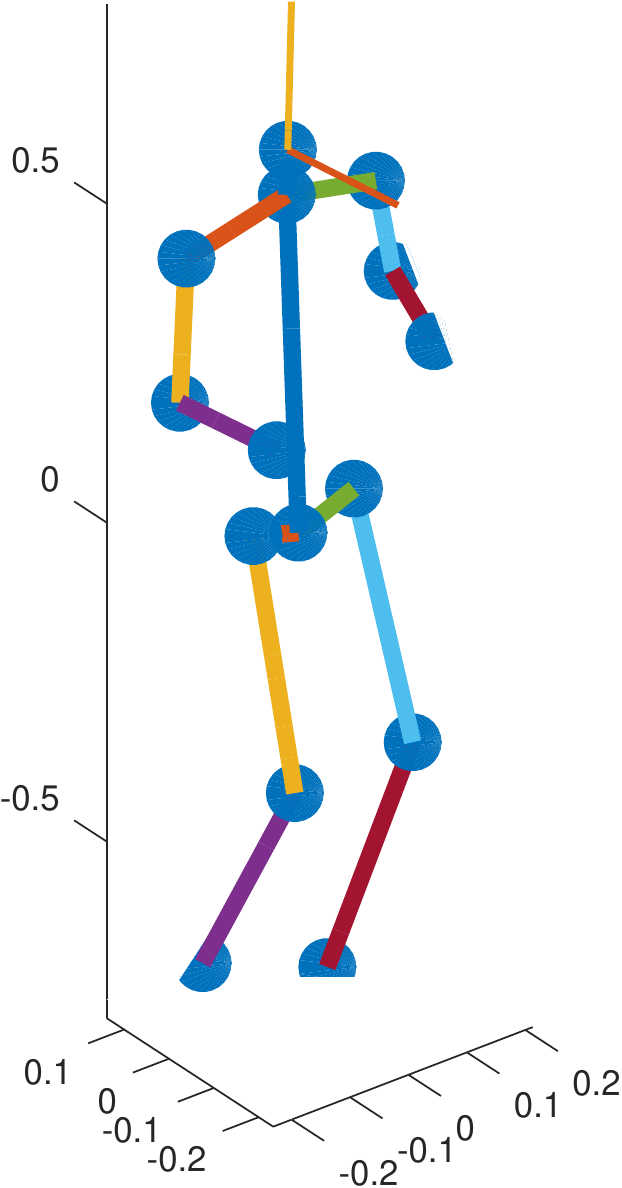}%
\includegraphics[width=0.05\linewidth, height=0.075\linewidth]{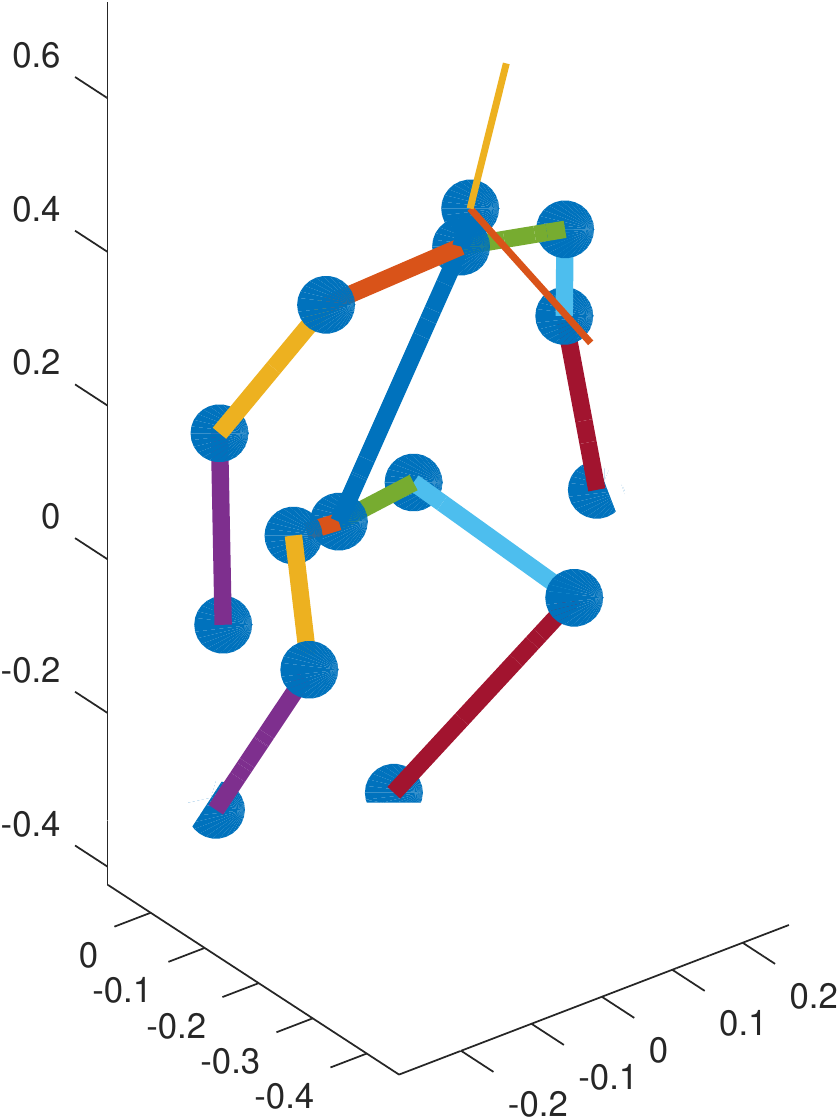}%
\includegraphics[width=0.05\linewidth, height=0.075\linewidth]{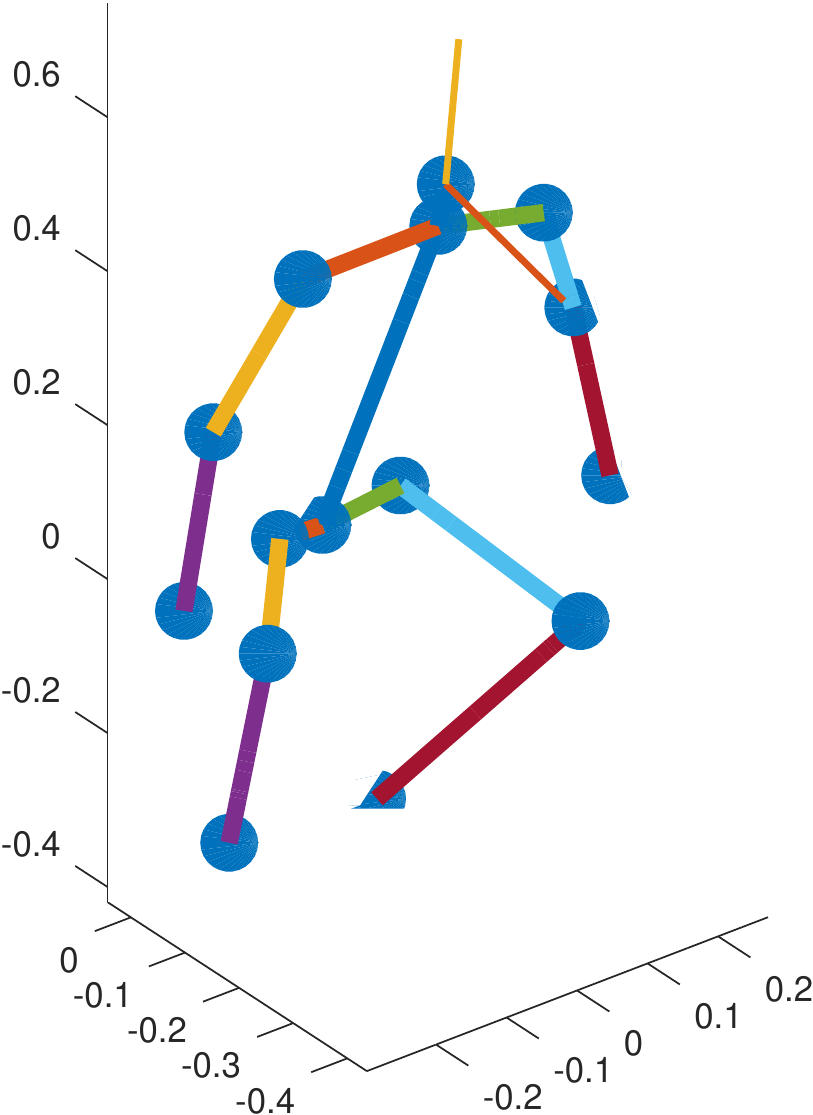}%
\includegraphics[width=0.05\linewidth, height=0.075\linewidth]{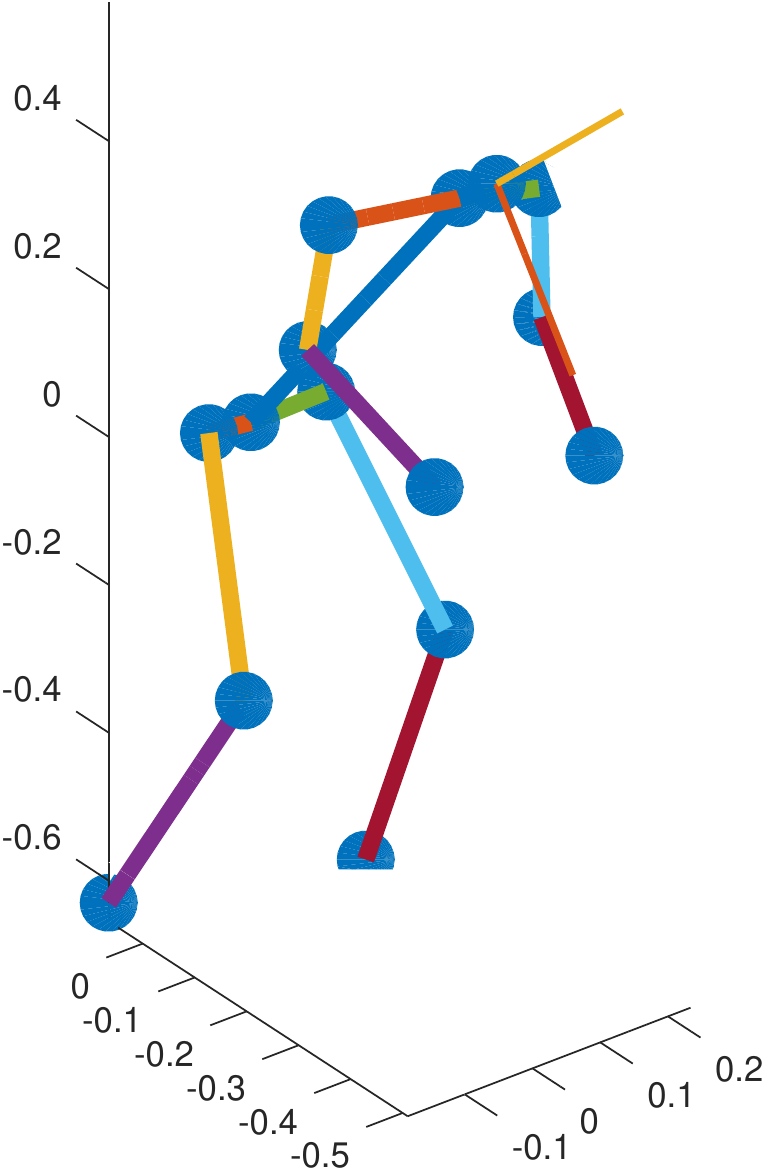}%
\end{tabularx}
}
\caption{\footnotesize
	Real video test. Rows 1,6,9: body shape segmentation overlaid on the egocentric video. 
	Rows 2,7,10: ground truth egoposes. Rows 3,8,11: the proposed method's (ours III) results.
	Row 4: \texttt{xr-egopose} II result. Row 5: \texttt{pd-egopose} result. (Best viewed in color).
}
\label{fig:real}
\vspace{-5pt}
\end{figure*}

\begin{table*}[tb]
\setlength{\tabcolsep}{5pt}	
\centering
\scriptsize
\begin{tabular}{|r|c|c|c|c|c|c|c|c|c|c|c|}
\hline
	& Stage1 & Stage1RNN & FullModel & ShapeOnly & MotionOnly & NoHeight & HandMap & xr-egopose\cite{fb1} & pd-egopose\cite{cmu2} & AllStand & AllSit\\
\hline \hline
	Keypoints (Avg) & 12.53  & 12.52 & \bf{11.76} & 13.89 & 14.55 & 14.52 & 26.10 & 13.26 & 14.46 & 23.23 & 29.81\\ 
	  (Std) & (21.00)   & (19.13) & (16.25) & (16.27) & (16.28) & (33.36) & (18.78) & (16.60) & (\bf{15.32}) & (19.00) & (20.78)\\
\hline
	Head V1 (Avg) & \bf{11.26} & 11.72 & \bf{11.26} & 13.03 & 15.36 & 13.01 & 16.26 & -- & -- & 67.84 & 76.04\\
	(Std) & (14.71)  & (13.96) & (14.71) & (17.25) & (\bf{13.14}) & (16.34) & (16.99) & -- & -- & (25.04) & (15.56)\\
\hline
	Head V2 (Avg) & \bf{13.04} & 14.00 & \bf{13.04} & 14.19 & 16.51 & 14.83 & 16.38 & -- & -- & 83.17 & 84.70\\ 
	(Std) & (13.57)  & (13.61) & (13.57) & (17.39) & (\bf{13.13}) & (18.51) & (14.12) & -- & --& (33.15) & (26.80) \\
\hline
\end{tabular}
\caption{\footnotesize Synthetic data comparison. The keypoints errors have the unit of centimeters and the head angle errors have 
	the unit of degrees.}
	\label{tab:syn}
	\vspace{-15pt}
\end{table*}

\begin{table}[tb]
\setlength{\tabcolsep}{1pt}	
\centering
\scriptsize
\begin{tabular}{|c|c|c|c|c|c|c|c|c|}
\hline
		 & Ous I    &   Ours II   & Ours III  & pdpose \cite{cmu2}  & xrpose I \cite{fb1}  & xrpose II \cite{fb1} & AllStand & AllSit\\
\hline \hline
KPs        & 15.70    &  16.13      & \bf{14.87}     & 17.59    &  16.86        &  17.29       &  20.61   & 27.87 \\ 
		 & (11.96)    &  (\bf{11.33})      & (11.42)     & (13.77)    &  (12.99)        &  (12.99)       &  (16.61)   & (18.44) \\ 
\hline
	HV1          & \bf{18.75}    &  21.01      & 19.31     & --       & --              &  --        & 71.83    & 78.75 \\
		 & (12.53)    &  (12.55)      & (\bf{11.73})     & --       & --              &  --        & (28.07)    & (17.51) \\  
\hline
	HV2          & 16.58    &  16.13      & \bf{15.81}     & --       & --              & --         & 73.96    & 77.53 \\ 
		 & (11.02)    &  (11.39)      & (\bf{9.60})      & --       & --              & --         & (31.60)    & (25.05) \\
\hline
\end{tabular}
	\caption{\footnotesize Comparison of errors in real video test. 
		Ours I, II, III: our method trained on real videos, on synthetic data only, and on a mixture of real and synthetic data.
	xrpose I, II: \texttt{xr-egopose} trained on real data only, and on the mixture of real and synthetic data. 
	pdpose: \texttt{pd-egopose} trained on real data. KPs: keypoints. HV1, 2: head orientation vector 1 and 2.        
	}
	\label{tab:real}
	\vspace{-10pt}
\end{table}

\subsection{Test on real videos} \label{sec:realeval}
\vspace{-3pt}
Building on top of the evaluations on synthetic data, we evaluate the proposed approach on real videos. 
We capture the body and head poses using a motion capture suit on three subjects. 
The synchronized fisheye video is captured by the GoPro fusion camera.
The training data is from one subject and we test on all three, with no overlap on the training and testing data.
The real video for training and each subject's test videos are about $15$-min long. 
In this test, the body shape model is trained on a dataset with $50$K images and the egopose model is trained using a mixture of the synthetic and real data. 
The synthetic training data include $900$ from the $2548$ sequences. Fig.~\ref{fig:real} and Table~\ref{tab:real} summarize the results. 

\begin{figure}[t]
\begin{center}
\includegraphics[width=0.5\linewidth, height=0.3\linewidth]{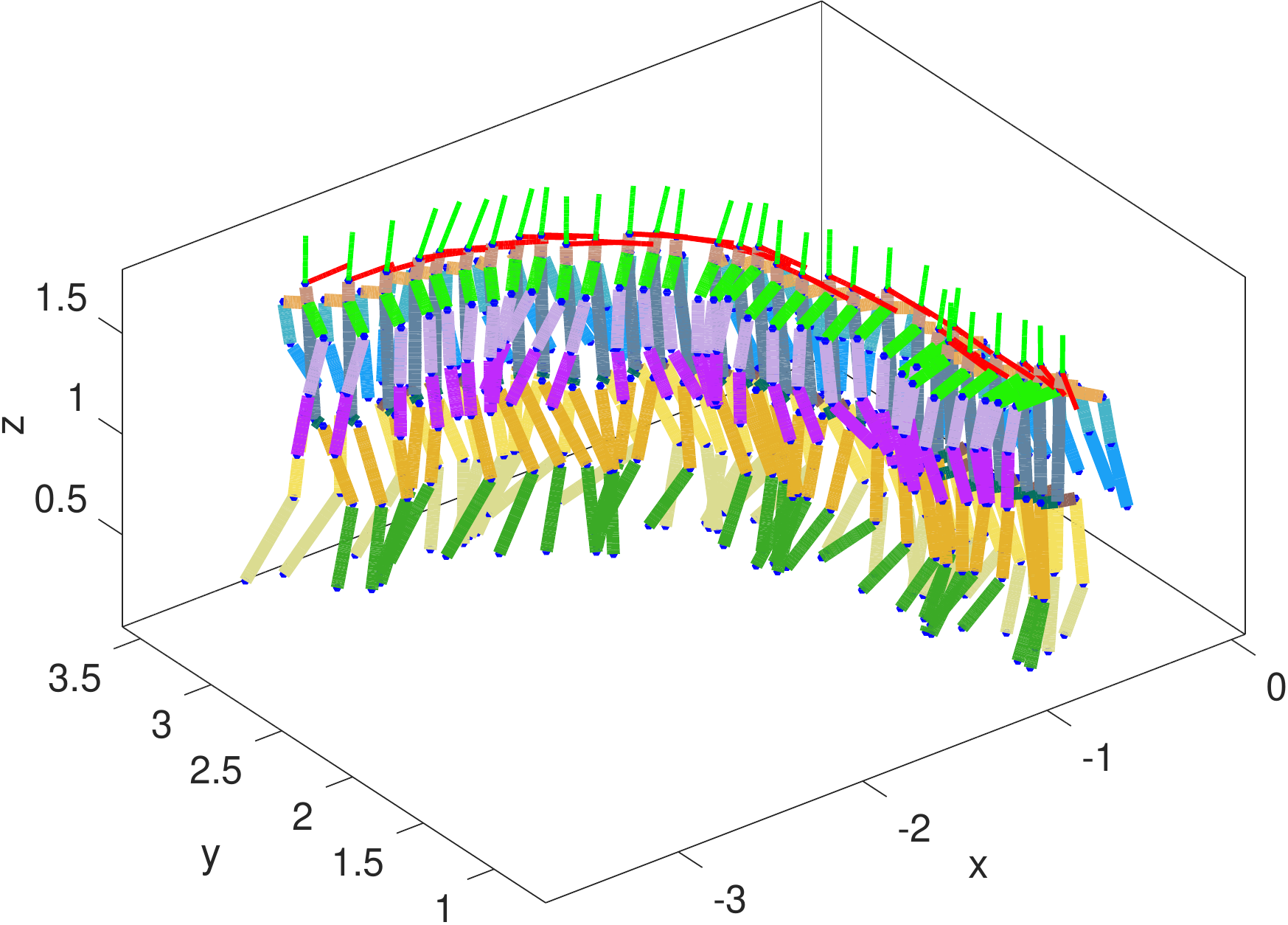}%
\includegraphics[width=0.5\linewidth, height=0.3\linewidth]{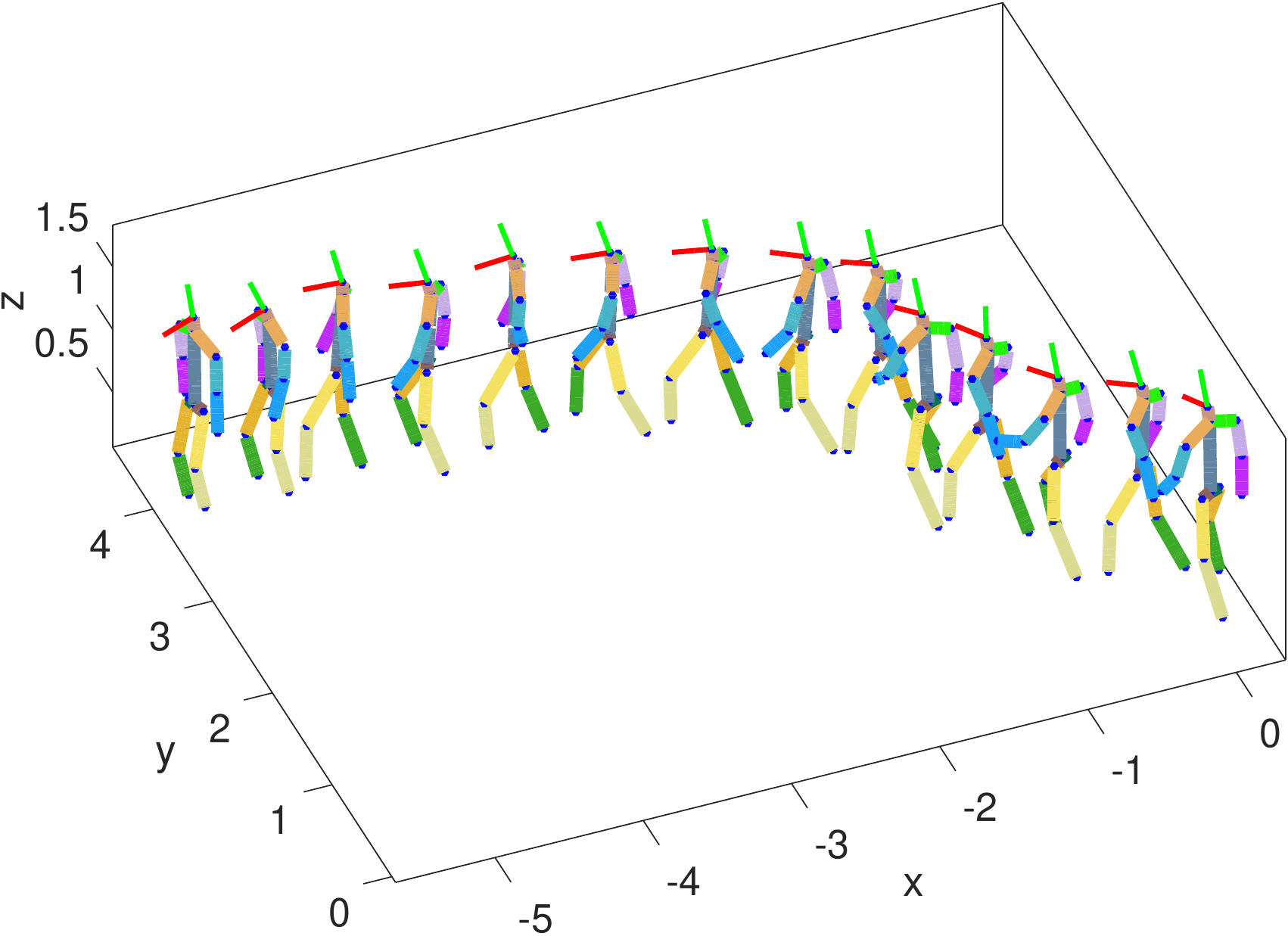}%
\end{center}
\caption{\footnotesize Repositioning the estimated egopose in a global coordinate system based on the estimated ego-head-pose and camera SLAM.
	Example results at $0.25$ (left) and $0.0625$ (right) of the original frame rates.}
\label{fig:global}
	\vspace{-10pt}
\end{figure}

As expected, the proposed method gives robust egopose estimation even though the camera only has a limited or no view of the wearer.
Table~\ref{tab:real} shows performance using different training schemes, and the proposed method clearly outperforms others in general.  
Using large synthetic data for training also boosts this superior performance. 
Although real video camera model is slightly different from the virtual cameras in synthetic dataset, we still recover estimates reasonably well. 
Note that we do not have the keypoint labeling for real video, and so we use \texttt{xr-egopose} network with foreground map instead of body keypoint map. 
Similar to the setup from Section \ref{sec:syneval}, we assume the initial foreground map estimation is fixed. 
The training thus optimizes the rest of the network.  
As Table~\ref{tab:real} shows, the proposed method gives much better result than both \texttt{xr-egopose} and \texttt{pd-egopose}.
Sample illustrations are shown in Fig.~\ref{fig:real}.
Further, we can reposition the estimated egopose in a global coordinate system using the head pose and the camera SLAM.
Here we rotate the egopose to align with the camera's pose only horizontally to ensure correct global pose even when the head pose estimation is imperfect. 
Fig.~\ref{fig:global} illustrates such egopose estimates in a global system. 
Lastly, as argued in Section \ref{sec:intro}, the proposed setup is real-time and efficient. 
With RTX2080Ti, Stages $1$ and $2$ take about $4$ms and $3$ms per frame respectively, and so the full system runs at $30$Hz per sec while taking up a fraction
of the GPU. 

\vspace{-5pt}
\section{Conclusion} \label{sec:conc}
\vspace{-6pt}
We introduce and tackle a new problem of estimating wearer's egopose from a human vision span. 
This is a challenging task, primarily due to the very limited view of the wearer with instances where the wearer is completely invisible in FOV. 
We propose a novel two-stage deep learning method which takes advantage of a new motion history image feature and the body shape feature. 
We estimate both the head and body pose at the same time while explicitly enforcing geometrical constraints.
Evaluations show good performance, robust to variation in camera settings while leveraging synthetic data sources thereby 
avoiding to re-collect large new datasets. 
The system is real-time and valuable for different egocentric experiences and applications in AR and VR. 


\end{document}